%% file: main.tex
\documentclass[10pt,journal,compsoc]{IEEEtran}

\usepackage{utils}
\usepackage[nocompress]{cite}
\usepackage{times}
\usepackage{graphicx}
\usepackage{amsmath}
\usepackage{amssymb}
\usepackage{diagbox}
\usepackage{color}
\usepackage{rotating}
\usepackage{siunitx}
\usepackage{changepage}
\usepackage[table]{xcolor}
\usepackage{multirow}
\usepackage{subcaption}
\graphicspath{{./Figures/}}

\usepackage{url}

\title{Simultaneous Surface Reflectance and Fluorescence Spectra Estimation}

\author{Henryk~Blasinski,~\IEEEmembership{Student~Member,~IEEE,} Joyce~Farrell and~Brian~Wandell%
\IEEEcompsocitemizethanks{\IEEEcompsocthanksitem All authors are with the Department of Electrical Engineering, Stanford University, CA 94305. E-mail: \{hblasins, jefarrell, wandell\}@stanford.edu}}

\begin{document}


\IEEEtitleabstractindextext{
\begin{abstract}

There is widespread interest in estimating the fluorescence properties of natural materials in an image. However, the separation between reflected and fluoresced components is difficult, because it is impossible to distinguish reflected and fluoresced photons without controlling the illuminant spectrum. We show how to jointly estimate the reflectance and fluorescence from a single set of images acquired under multiple illuminants. We present a framework based on a linear approximation to the physical equations describing image formation in terms of surface spectral reflectance and fluorescence due to multiple fluorophores. We relax the non-convex, inverse estimation problem in order to jointly estimate the reflectance and fluorescence properties in a single optimization step and we use the Alternating Direction Method of Multipliers (ADMM) approach to efficiently find a solution. We provide a software implementation of the solver for our method and prior methods. We evaluate the accuracy and reliability of the method using both simulations and experimental data. To acquire data to test the methods, we built a custom imaging system using a monochrome camera, a filter wheel with bandpass transmissive filters and a small number of light emitting diodes. We compared the system and algorithm performance with the ground truth as well as with prior methods. Our approach produces lower errors compared to earlier algorithms.

\end{abstract}
}

\maketitle

\section{Introduction}

\IEEEPARstart{M}{aterials} are commonly characterized by their surface reflectance spectra, which describe the fraction of incident photons that are reflected at each wavelength. In addition to reflectance, some materials absorb light in some wavelengths and then emit photons at longer wavelengths, a phenomenon called fluorescence. Fluorescent materials are common in nature \cite{Waggoner:06, Mitchell:88}, man-made objects such as paper, textiles, displays \cite{Grum:80,Kim:11,Tian:11}, or biological tissues  \cite{Georgakoudi:01,Monici:05,Ntziachristos:03}. In addition, the discovery of fluorophores that selectively bind to specific molecules has been extremely useful in biological and medical sciences \cite{Medintz:05,Dimitriadis:16}. 

To better characterize materials, it is useful to separate fluoresced photons from reflected ones, and there is widespread interest in natural and biological imaging methods that simultaneously estimate and disambiguate reflected and fluoresced photons \cite{Fu:13,Fuchs:01,Lam:13,Suo:14,Tominaga:15,Zhang:11,Zheng:14}. In this paper we describe an algorithm for simultaneously estimating reflectance and fluorescence. We design, implement, and evaluate a simple experimental system that performs this separation.

\begin{table*}
\renewcommand{\arraystretch}{1.3}
\centering
\caption{Comparison between existing reflectance and fluorescence estimation and separation algorithms.}
\label{tab:overview}
\small
\begin{tabular}{| l | l | l | l | l |}
\hline
\multicolumn{1}{| c |}{Author} & \multicolumn{1}{ c |}{Camera} & \multicolumn{1}{ c |}{Lights, quantity} & \multicolumn{1}{ c |}{Method} & \multicolumn{1}{ c |}{Estimate} \\
\hline
\hline
Tominaga \etal \cite{Tominaga:15} & $\sim 30$ bands  & Broadband, $2$ & Within-band ratios & Reflectance, one fluorophore, spectra \\
Fu \etal \cite{Fu:13} & $\sim 30$ bands & Sinusoidal, $2$ & Within-band ratios & Reflectance, one fluorophore, spectra \\
Zheng \etal \cite{Zheng:14} & $\sim 30$ bands & Broadband, $3$ & Sequence of optimizations & Reflectance, one fluorophore, spectra \\
Lam \etal \cite{Lam:13} & $\sim 30$ bands & Narrowband, $\sim 10$ & Multistep estimation & Reflectance, one fluorophore, spectra \\
Suo \etal \cite{Suo:14} & $\sim 10$ bands & Narrowband, $\sim 6$ & Biconvex optimization & Reflectance, multi fluorophore, intensity\\
Zhang \etal \cite{Zhang:11} & $3$ bands & Broadband, $2$ & Ind. Component Analysis & Chromaticity, intensity \\
Fu \etal \cite{Fu:14} & $3$ bands & Narrowband, $\sim 11$ & Sequence of optimizations & Reflectance, one fluorophore, spectra \\
\hline
\end{tabular}
\end{table*}

For the purpose of putting our work in context, it is useful to divide reflectance-fluorescence estimation algorithms (Table~\ref{tab:overview}) into two categories. \emph{Bispectral separation} methods use illuminated and radiated spectra sampled at high spectral resolution.  The approach combines complex instrumentation with relatively simple reconstruction algorithms. For example, Fuchs \cite{Fuchs:01} described a method that measures the sample radiance under ambient light with and without an additional, fluorescence exciting illuminant.  The fluorescence term is estimated by comparing the radiance ratios for wavelengths longer than the emission peak. A similar approach was presented by Tominaga, Horiuchi \etal \cite{Tominaga:13,Tominaga:15,Tominaga:11} who performed spectral separation using multispectral images of a scene acquired under two different, broadband light sources. These ideas were further extended by Fu, Lam \etal \cite{Fu:13} who used a multispectral camera together with light patterns with spectral power distributions sinusoidally modulated in the wavelength domain. They compared radiances for two illuminants that were phase-shifted in the wavelength domain. This approach requires multispectral acquisition and a spectrally controlled light source. A simpler method, requiring only a hyperspectral imager, was developed by Zheng, Sato \etal \cite{Zheng:14} who proposed a compact, four parameter fluorescent spectra parameterization based on Cauchy distribution and an estimation algorithm that solves a sequence of optimizations. 

\begin{figure*}
\includegraphics[width=\textwidth]{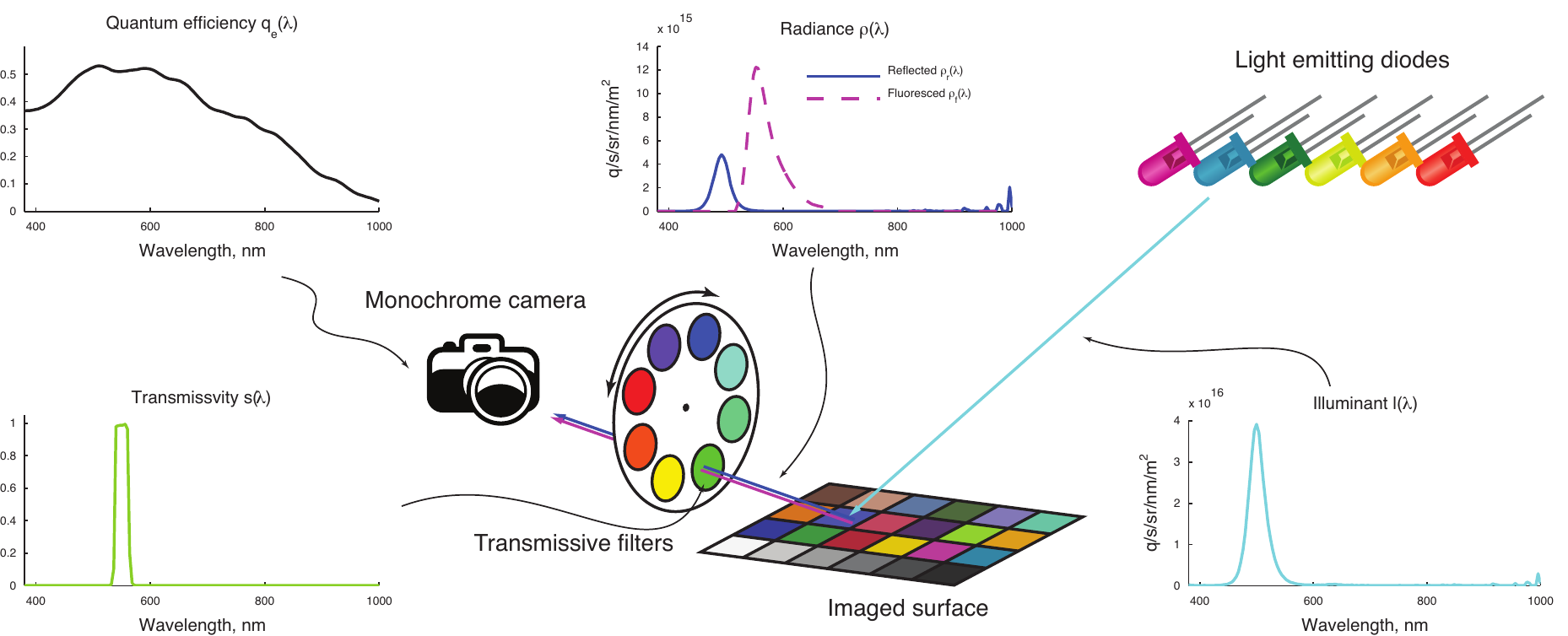}
 \caption{The reflectance and fluorescence imaging system includes a camera, narrowband light sources, and transmissive filters. Sub-figures present different spectral quantities. The camera's quantum efficiency, filter transmissivity and the illuminants are fixed and calibrated. The total scene radiance is estimated and separated into reflected and fluoresced components. The x-axis in all sub-figures represents wavelength in nanometers.}
 \label{fig:overview}
\end{figure*}

\emph{Computational separation} methods couple simple instrumentation with more complex algorithms that incorporate knowledge about the likely properties of the signals. Several algorithms use compressive sensing ideas and coarsely sampled measurements in the wavelength domain. For example, Zhang and Sato \cite{Zhang:11} use three-band (RGB) images to distinguish between reflected and fluoresced photons. Their Independent Component Analysis (ICA) based algorithm uses two images acquired under different illuminants and implements a simple camera responsivity model: Dirac delta functions in the wavelength domain. Lam and Sato \cite{Lam:13} and Fu, Lam \etal \cite{Fu:14,Fu:15}, extend this work to estimate reflectance and fluorescence spectra. Lam and Sato first estimate the reflectance component at about ten different wavelengths and then derive fluorescence components by subtracting the estimated reflectance contribution from the measurements. Fu, Lam \etal \cite{Fu:14,Fu:15} acquire a sequence of RGB images under nine different, narrowband illuminants. They develop a multi-step algorithm that estimates the chromaticity coordinates of the reflectance and fluorescence spectra and then searches a fluorophore database to determine the likely emission spectrum. Finally, Suo, Bian \etal \cite{Suo:14} estimate the reflectance spectrum and the Donaldson matrix, describing fluorescence properties of a surface, through nuclear norm minimization using Alternating Direction Minimization (ADM). 

We present a collection of computational separation methods to simultaneously estimate the reflectance, excitation and emission spectra of a surface. This framework builds on several prior approaches \cite{Fu:14,Fu:15,Suo:14,Zheng:14}. The methods also extend the active illumination reflection estimation methods \cite{Park:07,Parmar:12}, and the general case of fluorescence unmixing methods when surface spectral properties are unknown \cite{Alterman:10,Ikoma:14}. We formulate the reflectance and fluorescence spectra estimation as an inverse estimation problem, which we jointly solve in a single step for all unknown quantities. 

There are cases in which only a single fluorophore is present, and there are also cases in which multiple fluorophores co-exist.  For example, we are studying fluorescence in coral reefs which are known to contain multiple fluorophores \cite{Zawada:14,Treibitz:15}.  To assess the health of coral reefs it is useful to separate the fluorophores. Moreover, even if each substrate has only one fluorophore, the optics and pixel spacing frequently combine signals from adjacent scene spatial locations so that a pixel measures the signals from multiple substrates.  For example, in remote sensing applications the satellite pixel measures a region of the sea that includes more than one type of coral reef.  At the finer scale of microscopy, cells with different fluorophores may be interleaved within the tissue.  Hence, a single pixel within the digital microscope picks up the superposition of fluorescent signals from multiple cell types. An example is fluorescein angiography of the retina.  In this case, there is fluorescence in the blood vessels on the inner retina surface as well as auto-fluorescence from the cells within the retina, and the reflectance from the pigment epithelium layer \cite{Rodieck:98,Gray:06}. Although the fluorescent components are confined to distinct substrates, the image pixels contain an optically mixed signal.

The algorithms we describe are general enough to estimate fluorescence of a sample containing multiple fluorophores, and do not require assumptions regarding fluorescence emission properties, such as chromaticity invariance. We recognize that there are cases in which only a single fluorophore is present, and for those cases we derive a simpler and computationally efficient separation method.

The algorithms can be applied to data collected from a wide variety of imaging systems, including different types of sensors and light sources. The choice of sensor channels and light sources will influence the algorithm performance. We use a simulation environment to help  designers optimize system design given a specific choice of fluorophores. 

To evaluate the algorithms, we built a system (Figure~\ref{fig:overview}),  composed of off-the-shelf light emitting diodes (LED), transmissive filters and a monochrome camera. A target scene is illuminated with a single narrowband light source, from a small collection of available lights, and images of this scene are captured through a few narrowband filters. From the calibrated system characteristics, such as filter transmissivities or illuminant spectral power distributions, we estimate how accurately the algorithm reconstructs and separates reflected and fluoresced photons.

Section~\ref{sec:model} describes the image formation model for fluorescent surfaces. Section~\ref{sec:estMeth} presents the optimization framework and formulates spectral estimation and separation between reflectance and fluorescence radiance components as an inverse estimation problem. We analyze the limitations of the algorithms (Section \ref{sec:algoEval}) and  present the hardware system with experimental evaluations (Section \ref{sec:system}). We then discuss the experimental results (Section~\ref{sec:discussion}) and present our conclusions (Section~\ref{sec:concl}).
The source code and experimental results we present in the major figures are all accessible online.

\section{Fluorescent image formation model}
\label{sec:model}

\begin{figure}
\centering
\begin{subfigure}{0.33\columnwidth}
\includegraphics[width=\textwidth]{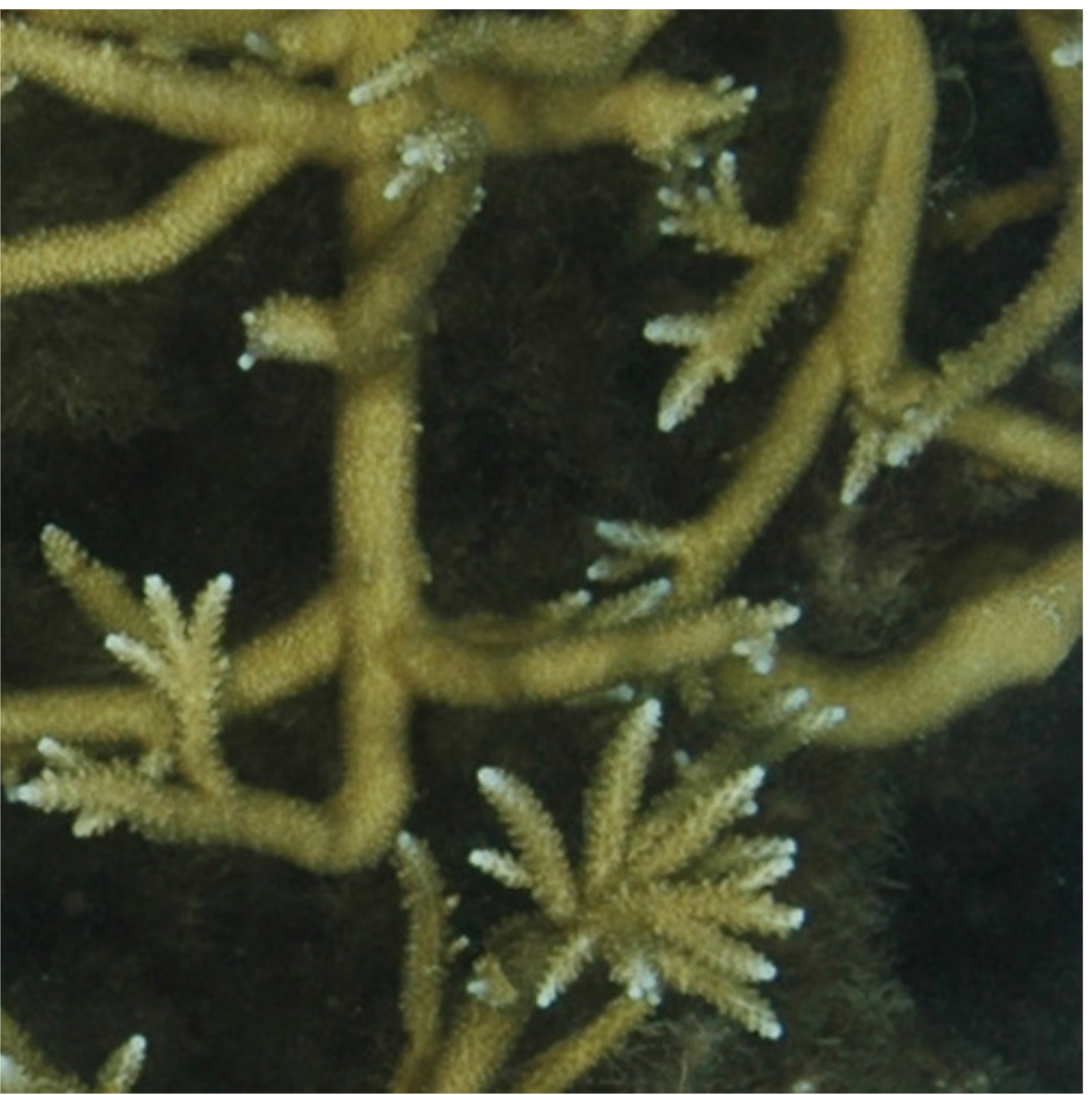}
\caption{Reflected}
\end{subfigure}%
\begin{subfigure}{0.33\columnwidth}
\includegraphics[width=\textwidth]{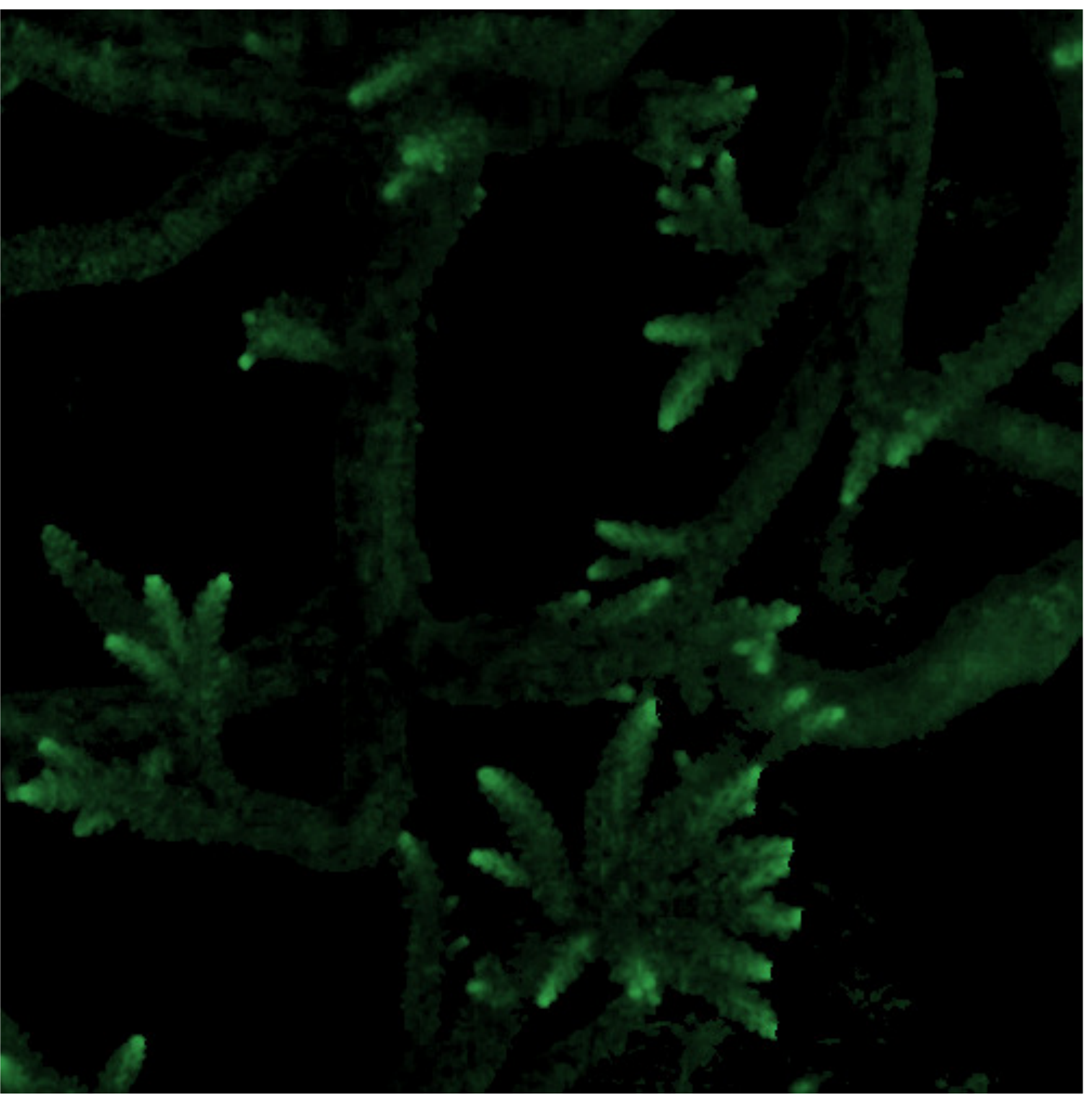}
\caption{Fluoresced}
\end{subfigure}%
\begin{subfigure}{0.33\columnwidth}
\includegraphics[width=\textwidth]{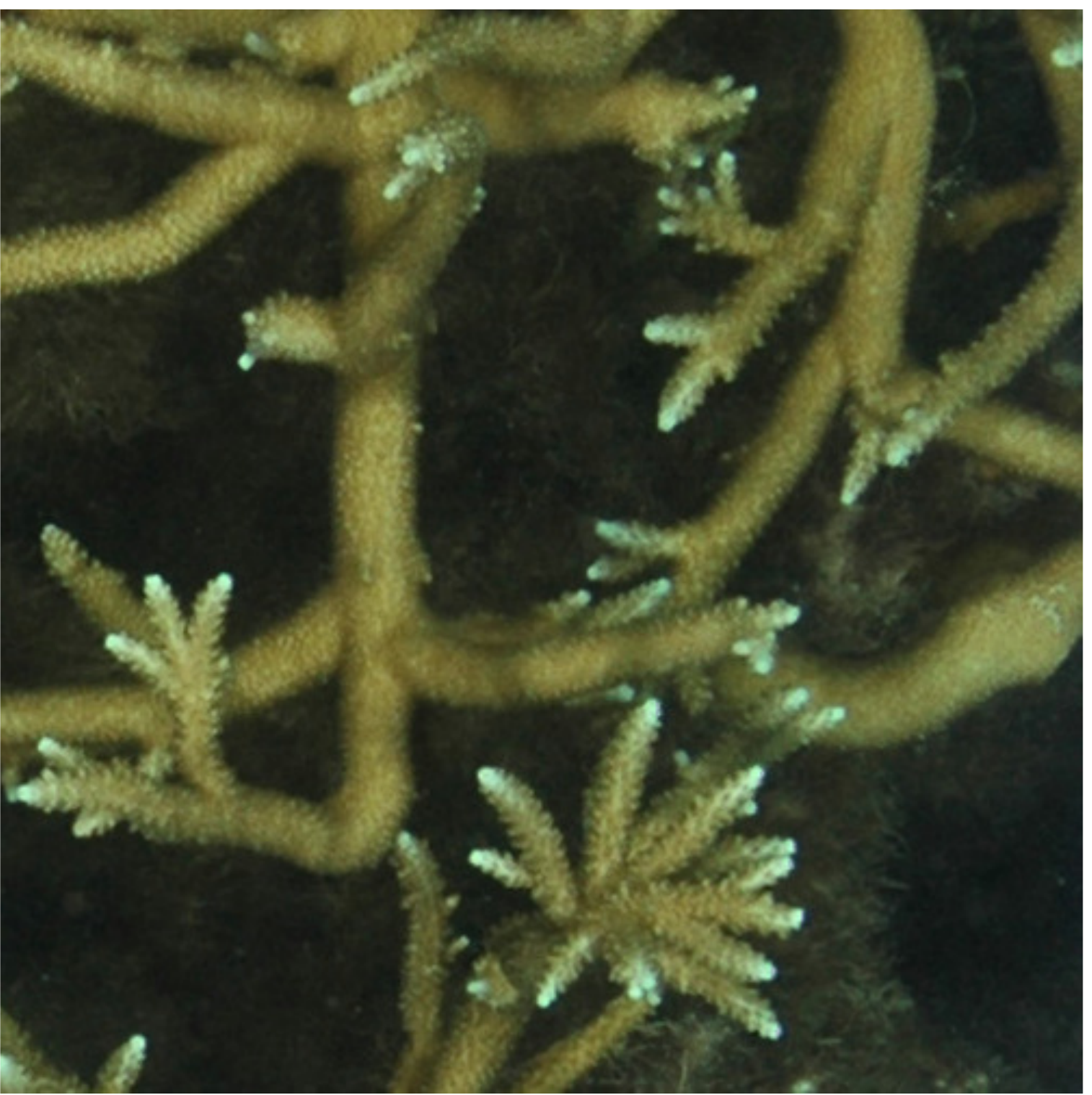}
\caption{Total}
\end{subfigure}%
\caption{Total scene radiance is a superposition of reflected and fluoresced light. Incident illumination is reflected from scene surfaces and also excites fluorescence. The image is \emph{Acropora} coral \cite{Beijbom:12} with reflected (a) and fluoresced (b) components synthesized using published data \cite{Hochberg:04,Treibitz:15}.}
\label{fig:reflFlSuperpos}
\end{figure}

Digital camera pixel response level $m$ is linearly related to the scene radiance $\rho$ \cite{Farrell:03}
\begin{align}
m = g \int q_e(\lambda) s(\lambda) \rho(\lambda)d\lambda,
\label{eq:imgFormModel}
\end{align}
where $q_e$ the photodetector quantum efficiency and  $s$ is the color filter transmissivity. The scalar $g$ combines a collection of camera parameters including the sensor gain, exposure duration and the aperture size. For any particular camera this combination of parameters in $g$ and the color filters $s$ can be fixed and calibrated.

The radiance, $\rho$, of any point in the scene is a superposition of radiances due to reflected $\rho_r$, and fluoresced light, $\rho_f$ (Fig.~\ref{fig:reflFlSuperpos}). Let $\rho(\lambda)$ denote the total radiance at some wavelength $\lambda$, then
\begin{align}
\rho(\lambda) = \rho_r(\lambda) + \rho_f(\lambda),
\label{eq:totalRadiance}
\end{align}
Assuming smooth and isotropic Lambertian surfaces, the reflected radiance at some wavelength $\lambda$ can be computed as a product of the illuminant $l(\lambda)$ and the surface reflectance $r(\lambda)$
\begin{align}
\rho_r(\lambda) = l(\lambda)r(\lambda).
\end{align}
Fluorescent radiance can be produced by a number of different fluorescent compounds present in the sample. Often their individual contributions $e_{f,z}(\lambda)$ are considered to be additive \cite{Alterman:10,Lakowicz:99}
\begin{align}
\rho_f(\lambda) = \sum_{z=1}^Z \rho_{f,z}(\lambda),
\end{align}
where $z$ indexes over different fluorophores. 

In the most general case the fluorescent radiance due to a single fluorophore $\rho_{f,z}(\lambda)$ is described by a two dimensional function $e_{xm,z}(\lambda,\lambda_x)$ \cite{Wyszecki:82}. This function expresses the number of emitted photons at a particular wavelength $\lambda$ as a fraction of incident monochromatic light of some other and different wavelength $\lambda_x$. If a broadband light source is used  it is necessary to consider fluorescence emissions arising from illumination at all spectral bands of the incident light
\begin{align}
\rho_{f,z}(\lambda) = \int e_{xm,z}(\lambda,\lambda_x)l(\lambda_x)d\lambda_x.
\label{eqn:fullFluorescence}
\end{align}

The vast majority of fluorophores exhibit physical properties that allow to simplify the emission model. First, it was observed by Stokes that the wavelength of the emitted photons is typically longer than that of exciting photons \cite{Stokes:52}. This implies that $e_{xm,z}(\lambda,\lambda_x) = 0$ if $\lambda \leq \lambda_x$ \cite{Wyszecki:82}. Second, per Kasha's rule, the shape of the fluorescence emission is constant and only its intensity varies with changes in the illumination wavelength \cite{Kasha:50}. This assumption, also called chromaticity invariance, implies that $e_{xm,z}(\lambda,\lambda_x)$ is a separable function and can be represented as a product of two univariate functions $e_{xm,z}(\lambda,\lambda_x) = e_{x,z}(\lambda)e_{m,z}(\lambda_x)$. The function $e_{m,z}(\lambda)$ is called the emission spectrum, which represents the spectral power distribution of the fluorescent light emitted by the surface. The second function $e_{x,z}(\lambda_x)$ is the excitation spectrum, sometimes referred to as the absorption spectrum \cite{Rendell:87}, which describes the efficiency with which incident photons of different wavelengths excite the fluorescence signal. Under these assumptions the fluorescence radiance due to a single fluorophore $z$ may be expressed as 
\begin{align}
\rho_{f,z}(\lambda) = e_{m,z}(\lambda)\int e_{x,z}(\lambda_x)l(\lambda_x)d\lambda_x.
\label{eq:flEmission}
\end{align}
The multiplicative relationship between the excitation and emission spectra implies that each of these spectra can be arbitrarily re-scaled and, as long as the reciprocal scaling is applied to the other quantity, the result will remain unchanged. Often both spectra are normalized so that their maximum intensities are equal to one \cite{McNamara:06}, or that the area under the curve is equal to one, i.e. $\int e(\lambda)d\lambda=1$ \cite{Zhang:11}. If this is the case an additional intensity scalar needs to be introduced into (\ref{eq:flEmission}) to reflect these normalizations.

\begin{figure}
\begin{minipage}[c]{.55\columnwidth}
  \vspace*{\fill}
  \centering
  \includegraphics[width=\textwidth]{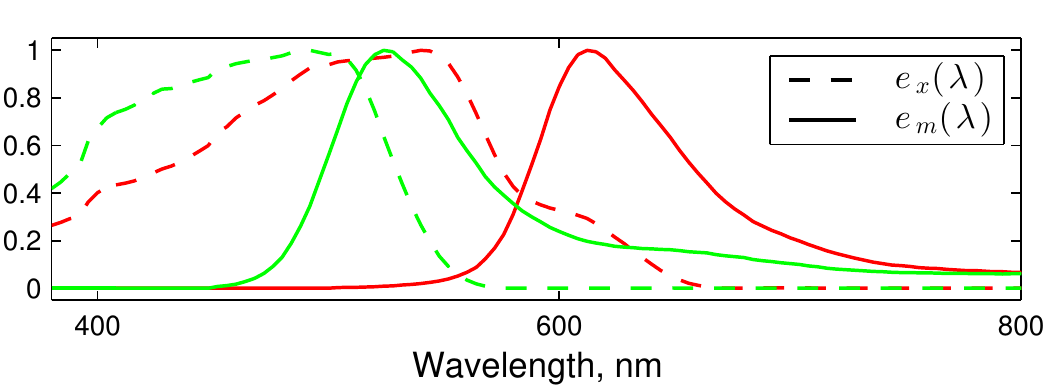}
    \subcaption{Fluorophores}
  \includegraphics[width=\textwidth]{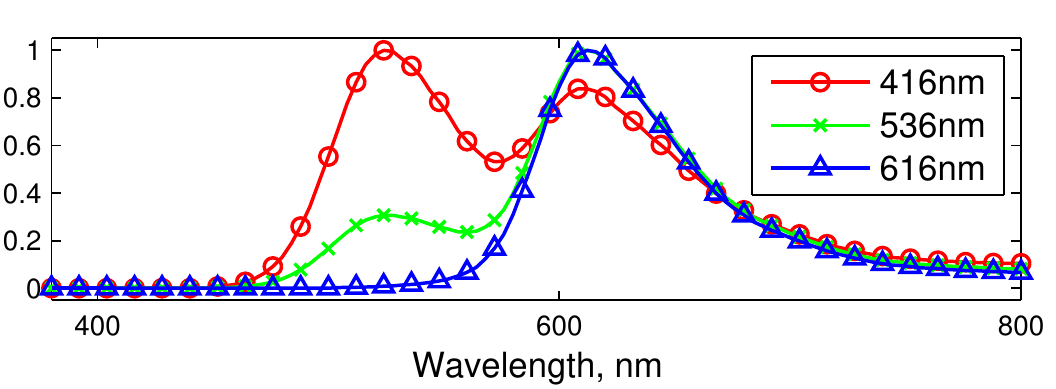}
  \subcaption{Emitted lights}
\end{minipage}%
\hfill%
\begin{minipage}[c]{.45\columnwidth}
  \vspace*{\fill}
  \centering
  \includegraphics[width=\textwidth]{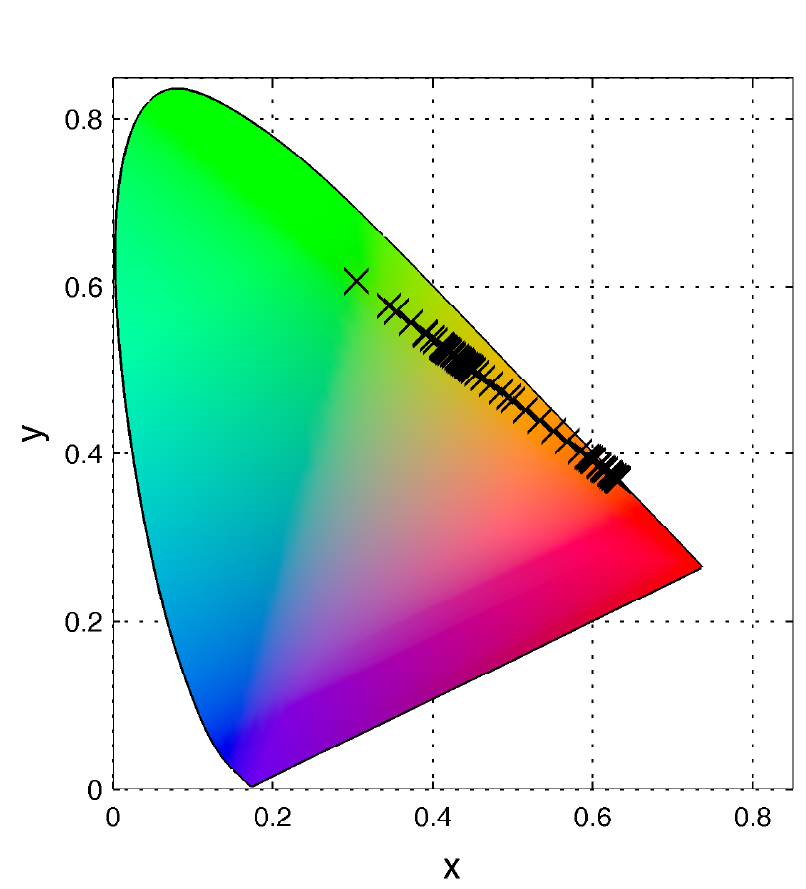}
  \subcaption{Chromaticity}
\end{minipage}%
\caption{Chromaticity invariance of the fluoresced signal is violated when the sample includes two or more fluorophores. (a) The excitation (dashed) and emission (solid) spectra of two fluorophores (green and blue). (b) Normalized spectral power distributions of  light emitted under three  monochromatic light sources. (c) The fluorescent emission chromaticity changes substantially as the light source varies from $400$nm to $800$nm.}
\label{fig:chrInv}
\end{figure}

Note that Kasha's rule holds only when the excitation and emission spectra do not overlap. When they do overlap, Stokes shift implies that the emission spectrum will vary with the illumination. Furthermore, the emission spectrum is not invariant when two or more fluorophores are present in a sample; each of the fluorophores will contribute different amounts depending on the illumination. This effect is illustrated in Fig.~\ref{fig:chrInv}, which shows normalized emission spectra of a two fluorophore sample under different monochromatic lights.

\subsection{Discretized image formation model}
We represent spectral functions using vectors and matrices quantized to $d$ narrow spectral bins. When a particular surface  with $n$ fluorophores is observed using $i$ different camera filters and under $j$ different illuminants the discrete image formation model may be written as
\begin{align}
M = G \circ C^T \left( \mathbf{diag}(r) + T \circ \sum_{z=1}^n e_{m,z}e_{x,z}^T\right)L, 
\label{eq:imageFormation}
\end{align}
with
\begin{align}
T = \left[ \begin{array}{ccccc}
			0 & 0 & & & 0\\
       		1 & 0 & 0 & &  \\
            1 & 1 & 0 & & \\
  				& & & \ddots &  \\
             1 &  &  & 1 &  0  \end{array}  \right].
\end{align}
The $\mathbf{diag}(r)$ operator places the entries of the reflectance vector $r \in \mathbf{R}^d$ along the diagonal of a matrix. The matrix $\sum_{z=1}^n e_{m,z}e_{x,z}^T$ with components $e_{m,z},e_{x,z} \in \mathbf{R}^d$, sometimes called the Donaldson matrix, is a discrete representation of the $e_{xm}(\lambda,\lambda_x)$ function \cite{Resch:08}. The Donaldson matrix is element-wise multiplied (Hadamard product denoted with $\circ$) with $T \in \mathbf{R}^{d\times d}$, forcing this matrix into a lower triangular form, as predicted by the Stokes rule. The columns of matrix $C=\mathbf{diag}(q_e)\left[s_1,\ldots, s_i\right], C \in \mathbf{R}^{d\times i}$ are formed by filter transmissivities scaled by the sensor quantum efficiency, similarly $L=\left[l_1,\ldots,l_j\right], L \in \mathbf{R}^{d\times j}$ is a matrix whose columns are the illuminant spectral power distributions. The $op$th entry of $G\in \mathbf{R}^{i \times j}$ represents the camera gain parameter associated with the $o$th filter and $p$th illuminant. Finally, the $op$th entry of $M \in \mathbf{R}^{i \times j}$ is the pixel value observed though the $o$th filter and under $p$th illuminant.  

Reflectance and fluorescence spectra are typically smooth functions that fall within a low-dimensional subspace spanned by a small number of basis functions \cite{Fu:14,Maloney:86}. As a consequence any reflectance or fluorescence spectrum can be compactly represented using low dimensional linear models 
\begin{align}
r & = B_rw_r, \label{eq:basisFcnRefl}\\
e_{x,z} &  = B_x w_{x,z}, \label{eq:basisFcnEx}\\
e_{m,z} & =  B_m w_{m,z},  \label{eq:basisFcnEm}
\end{align}
where $B_r \in \mathbf{R}^{d \times n_r}$, $B_x \in \mathbf{R}^{d \times n_x}$ and $B_e \in \mathbf{R}^{d \times n_e}$ are matrices whose columns are basis functions for reflectance, excitation and emission spectra respectively. Similarly $w_r,w_{x,z},w_{e,z}$ are the corresponding weight coefficient vectors. This modeling approach permits to reduce the number of parameters in the image formation model.

Finally, the compact image formation model, with linear approximations for reflectance, excitation and emission spectra (\ref{eq:basisFcnRefl}) -- (\ref{eq:basisFcnEm}), expresses measured pixel intensities in terms of basis function weights 
\begin{align}
M = G \circ C^T \left( \mathbf{diag}(B_rw_r) + T \circ B_mWB_x^T\right)L,
\end{align}
where 
\begin{align}
W = \sum_{z=1}^n w_{m,z}w_{x,z}.
\end{align}
Note that $\mathbf{rank}(W) = n$ and $W \in \mathbf{R}^{n_e \times n_x}$.

\section{Estimation methods}
\label{sec:estMeth}

Using the image formation model and data, we estimate the reflectance and fluorescence spectra that minimize the Euclidean error between model predictions and measurements $M$. We propose three estimation algorithms; first, a general method applicable when multiple fluorophores are present in a sample. Second, we simplify the general model for the case when the sample is known to contain a single fluorophore. Third, we show how the single fluorophore model can be simplified even further when only the emission spectrum needs to be estimated.  

\subsection{Multi-fluorophore model}

The goal of multi-fluorophore estimation is to find such reflectance basis function weights $w_r$ and a matrix $W$ that minimize the Euclidean error in the measurements $M$ subject to physics motivated constraints
\begin{align}
\text{minimize}\; & \; \|M - G \circ C^T \left( \mathbf{diag}(B_rw_r) + T \circ B_mWB_x^T\right)L\|_F^2 \nonumber\\
&\; + \alpha\|\nabla B_rw_r\|_2^2 +  \beta  \|\nabla \left(T \circ B_mWB_x^T\right)\|_F^2 \nonumber \\
&\; + \beta \|\left(T \circ B_mWB_x^T \right) \nabla^T\|_F^2 \nonumber\\ 
\text{subject to}\; & \; 0 \leq B_rw_r \leq 1, \nonumber\\ 
& \; 0 \leq T \circ B_mWB_x^T, \nonumber \\
& \; \mathbf{rank}(W) = n,
\end{align}
and
\begin{align}
\nabla = \left[ \begin{array}{cccc}
			1 & -1 & & 0\\
       		0 & 1 & -1 &  \\
  				& & \ddots &  \\
             0 &  & 1 & -1  \end{array}  \right],
\end{align}
where $\|\cdot\|_F$ is the Frobenius norm of a matrix and the $\nabla$ operator computes differences between adjacent entries in a vector. 

The objective function is composed of three terms. The first is a data fidelity term that measures the difference between the model and data. Two additional terms, scaled by tuning parameters $\alpha$ and $\beta$, encourage smooth solutions by penalizing the objective if neighboring entries in the estimates of the reflectance $B_rw_r$ or the Donaldson matrix $B_eWB_x^T$ vary. In the case of the Donaldson matrix the roughness penalty is imposed on both the rows and columns. 

The solution space is further restricted by three constraints. The first constraint follows from the fact that reflectance is a passive process, which does not create new photons. The second constraint is a consequence of nonnegativity of light. Note however that the nonnegativity is applied to the entire Donaldson matrix estimate, not the contributing fluorophore excitation and emission spectra. The third constraint enforces a solution with a specific number of fluorophores.  The last constraint is cumbersome for two reasons. In general, we do not know in advance how many fluorophores are present in a given sample. Additionally, the rank equality constraint makes the optimization problem non-convex and hard to solve globally.

Instead, we can approximate the original problem by replacing the non-convex constraint with a convex penalty. The rank of a matrix is equal to the number of its nonzero singular values. We can impose a less stringent constraint by penalizing the sum of all the singular values, \ie matrix nuclear norm, which is a convex function \cite{Suo:14}. This penalty is analogous to an $l_1$ penalty which is typically used to enforce solution sparsity \cite{Donoho:09,Hastie:09}. The substitution of nuclear norm penalty for rank constraint produces the following convex relaxation of the original problem 
\begin{align}
\text{minimize}\; & \; \|M - G \circ C^T \left( \mathbf{diag}(B_rw_r) + T \circ B_mWB_x^T\right)L\|_F^2 \nonumber\\
&\; + \alpha\|\nabla B_rw_r\|_2^2 +  \beta  \|\nabla \left(T \circ B_mWB_x^T\right)\|_F^2 \nonumber \\
&\; + \beta \|\left(T \circ B_mWB_x^T \right) \nabla^T\|_F^2 + \eta \|W\|_\star \nonumber\\ 
\text{subject to}\; & \; 0 \leq B_rw_r \leq 1, \nonumber\\ 
& \; 0 \leq T \circ B_mWB_x, 
\label{eq:relaxedProblem}
\end{align}
where $\|W\|_\star$ denotes the nuclear norm of $W$ and $\eta$ is the penalty tuning parameter. This convex optimization problem, which we will refer to as \emph{multi-fluorophore}, can be efficiently solved using the Alternating Direction Method of Multipliers (ADMM). Implementation details are available in the Supplemental Material (Appendix~A), where we also show how to explicitly enforce the matrix $B_mWB_x$ estimate to have rank $n$.

\subsection{Single fluorophore model}
When the sample contains only one fluorophore, the estimation problem is substantially simplified: The problem becomes biconvex in the unknown parameters $w_r, w_x$ and $w_e$. It is possible to strictly enforce the rank constraint $\mathbf{rank}(W) = 1$ by alternating minimization over subsets of parameters in which the objective is convex. Even though the solution algorithm is still iterative, it is easier to solve because the nuclear norm penalty is eliminated from the objective. In addition, the optimization is performed over a single excitation and emission spectrum, which allows to impose nonnegativity directly on those spectra. The \emph{single fluorophore} optimization problem becomes
\begin{adjustwidth}{-0.5cm}{0cm}
\begin{align}
\text{minimize}\; & \; \|M - G \circ C^T \left( \mathbf{diag}(B_rw_r) + T \circ B_mw_mw_x^TB_x^T\right)L\|_F^2 \nonumber\\
&\; + \beta \left(\|\nabla B_xw_x\|_2^2 + \|\nabla B_mw_m\|_2^2\right) \nonumber\\ 
&\; +  \alpha\|\nabla B_rw_r\|_2^2 \nonumber \\
\text{subject to}\; & \; 0 \leq B_rw_r \leq 1, \nonumber\\ 
& \; 0 \leq B_xw_x, \nonumber \\
& \; 0 \leq B_mw_m,
\label{eq:biconvex}
\end{align}
\end{adjustwidth}
The optimization is quadratic in $w_r,w_m$ and $w_r,w_x$. First, a quadratic  problem (QP) is solved over the variables $w_r,w_m$ holding $w_x$'s fixed. Next, a QP is solved over $w_r$ and $w_x$ while $w_m$ is fixed. These steps are repeated until no improvement in the objective is observed. 

In general there is always a scaling ambiguity in specifying the excitation and emission spectra, which are estimated up to a free multiplicative scale $\Delta w_x$ and $(1 / \Delta) w_m$. Despite the scaling uncertainty, the algorithm can still correctly recover the total number of fluoresced photons and relative spectra shapes.

\subsection{Chromaticity invariant model}
In some cases a fluorophore emits photons within the one wavelength range, but is excited only by wavelengths below the emission range.  This case permits a further simplification because the emission spectrum has the same chromaticity, independent of the light source.  This case can be modeled by optimizing over $p = L^TB_xw_x$, $p \in \mathbf{R}^j$ rather than $w_x$
\begin{align}
\text{minimize}\; & \; \|M - G \circ C^T \left( \mathbf{diag}(B_rw_r)L + B_mw_mp^T\right)\|_F^2 \nonumber\\
&\; + \beta \|\nabla B_xw_x\|_2^2 +  \alpha\|\nabla B_rw_r\|_2^2 \nonumber \\
\text{subject to}\; & \; 0 \leq B_rw_r \leq 1 \nonumber\\ 
& \; 0 \leq B_xw_x \nonumber \\
& \; 0 \leq p.
\label{eq:cim}
\end{align}
In this chromaticity invariant model (\emph{CIM}) only the shape of the fluorescence emission and an intensity scaling factor are estimated. The scaling factor $p_j$ compactly represents all excitation phenomena for a given illuminant $j$. The wavelength dependency of the excitation spectrum is not included in the image formation model,  it is no longer meaningful to impose the Stokes rule, and the matrix $T$ can be dropped.

\section{Algorithm evaluation}
\label{sec:algoEval}

The proposed estimation algorithms are implemented in Matlab\footnote{The source code is available for download at \texttt{https://github.com/hblasins/fiToolbox}}. The multi-fluorophore, ADMM solver uses standard matrix operators, and the single fluorophore biconvex solver uses the \texttt{cvx} convex optimization toolbox \cite{cvx:14,Nocedal:06}. We use simulations to understand the effect of key system parameters including (a) the number of basis excitation and emission basis functions, (b) the number of illuminants and filters, (c) the robustness to noise, and (d) algorithm convergence rates.

To validate the estimation algorithms, we created synthetic data that comply with the image formation model (\ref{eq:imageFormation}). We used Macbeth chart reflectance spectra and Donaldson matrices from the McNamara-Boswell data set \cite{McNamara:06}, restricted to samples with peak excitation and emission within the $400$ to $980$nm range. We choose this interval, slightly smaller  than the camera spectral range ($380$ to $1000$nm sampled at $4$nm, $d = 156$) to eliminate edge cases that may be difficult to analyze. In all evaluation experiments we used $24$ test patches, each of which had distinct surface spectral reflectance properties and fluorescence excitation--emission properties of a single fluorescent compound.

We evaluate the accuracy by computing the root-mean-square error, RMSE between estimated and ground truth spectral reflectance curves, Donaldson matrices and excitation and emission spectra. We report the average RMSE and standard deviation over estimates for all test patches in a given experiment. These RMSE quantities typically occupy different ranges. The reflectance values are often around $0.5$, while the Donaldson matrix entries rarely exceed $10^{-2}$. Consequently, the low absolute values of the RMSE for the Donaldson matrix do not imply superior accuracy but simply capture the level of the fluorescence signal. We refer to all comparisons that preserve the absolute reflectance or fluorescence scales as the \emph{absolute} comparisons. To better match the RMSE scales, we also compute the RMSE for normalized  quantities which we will refer to as \emph{normalized} comparisons. In this case the estimate and ground truth are divided by their maximal values before computing the RMSE. 

For the case we analyze below we found that (a) $12$ fluorescence excitation and emission basis functions provide good approximations for the Donaldson matrix, (b) a system composed of about $20$ channels and illuminants performs reliable reflectance and fluorescence separation. In addition, (c) the algorithm is robust against noise and (d) converges to a solution typically in a few hundred iterations. 

\subsection{Linear model dimensionality} 
First, we investigated the model accuracy by varying the number of linear model basis functions. We analyze the approximation for both the excitation and emission spectra.  The basis functions were derived from the McNamara-Boswell dataset. We chose five basis to approximate Macbeth chart reflectances, which corresponds to the typically reported dimensionality of that set \cite{Wandell:95}. In this experiment we used a bispectral system, where  the camera samples spectral bands and the light source generates narrowband light, i.e., $C=L=I$, where $I \in \mathbf{R}^{d \times d}$ is the identity matrix. Camera gain $G$ was adjusted to a maximal pixel intensity of one, and tuning parameters were set to $\alpha=\beta=\eta=0.001$. 

\begin{figure}
\centering
\begin{subfigure}{0.49\columnwidth}
\includegraphics[width=\textwidth]{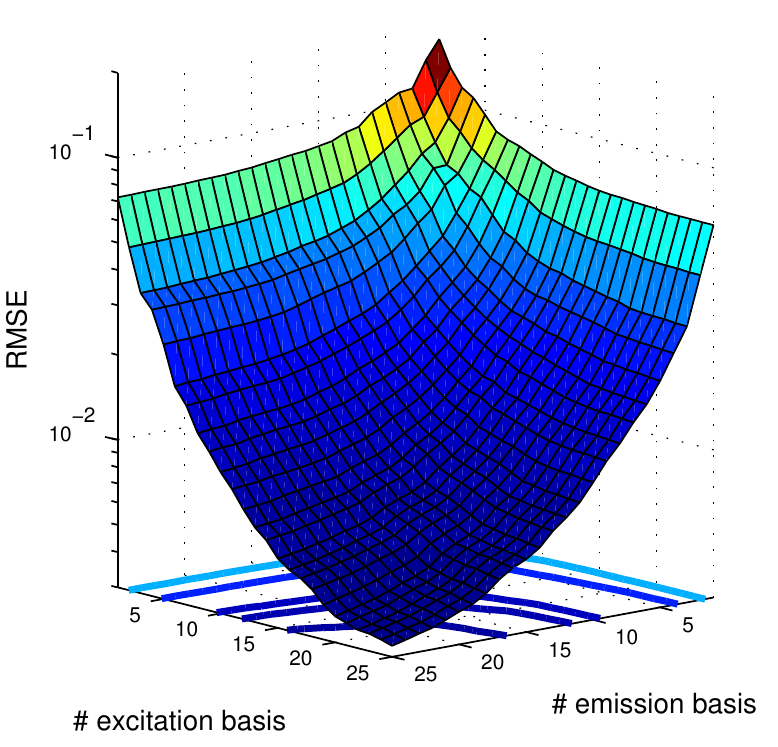}
\caption{Multi-fluorophore}
\end{subfigure}%
\begin{subfigure}{0.49\columnwidth}
\includegraphics[width=\textwidth]{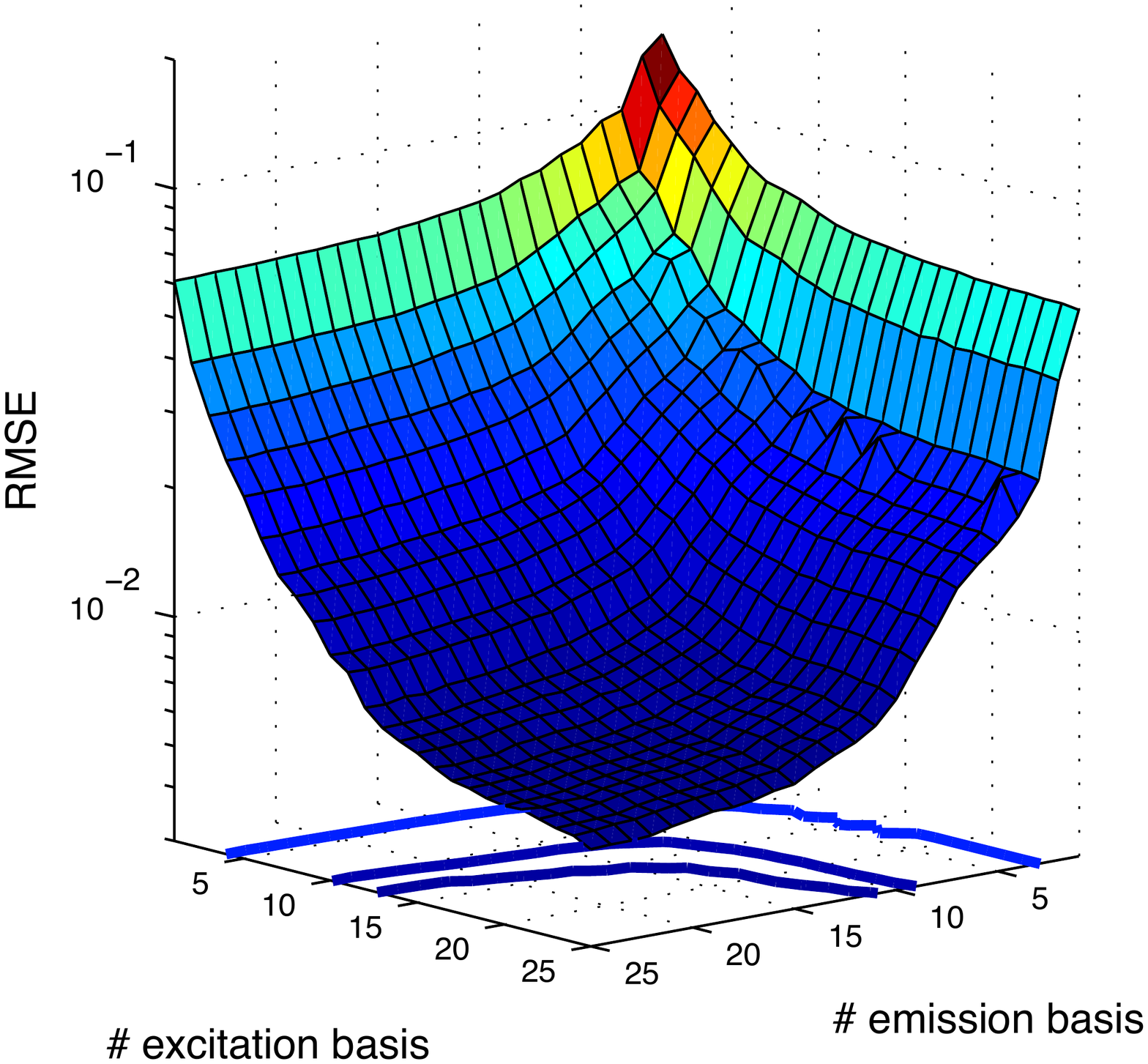}
\caption{Single fluorophore}
\end{subfigure}%
\caption{Normalized Donaldson matrix estimation error (RMSE) as a function of the number of linear basis functions approximating excitation and emission spectra. Contour plots show iso-RMSE lines at $0.025$, $0.01$ and $0.0075$.}
\label{fig:nBases}
\end{figure}

\begin{figure}
\centering
\begin{subfigure}{0.49\columnwidth}
\includegraphics[width=\textwidth]{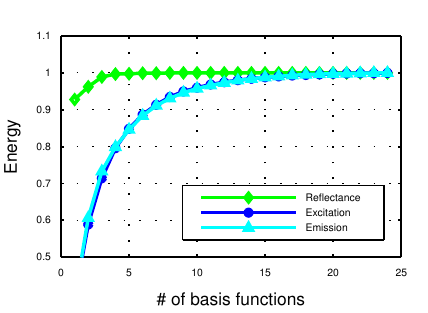}
\caption{Approximation error}
\end{subfigure}%
\begin{subfigure}{0.49\columnwidth}
\includegraphics[width=\textwidth]{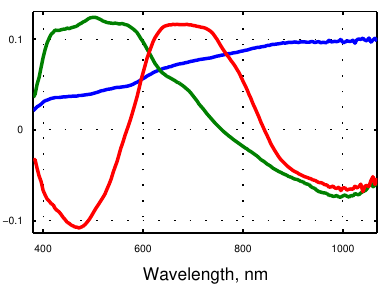}
\caption{Reflectance basis}
\end{subfigure}\\%
\begin{subfigure}{0.49\columnwidth}
\includegraphics[width=\textwidth]{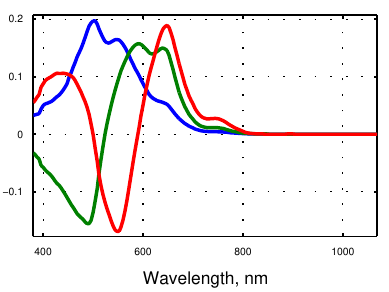}
\caption{Excitation basis}
\end{subfigure}%
\begin{subfigure}{0.49\columnwidth}
\includegraphics[width=\textwidth]{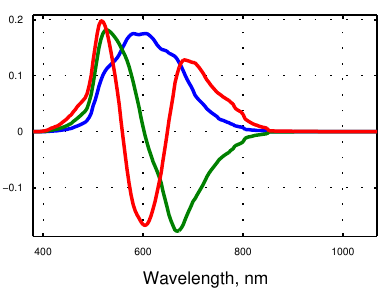}
\caption{Emission basis}
\end{subfigure}%
\caption{Linear basis function approximation of spectral quantities. Five reflectance basis functions explain $99\%$ of variance of Macbeth chart reflectances. Similarly, $12$ excitation and emission basis functions explain $97\%$ of variance in the McNamara-Boswell data set (a). Remaining panels show three most dominant basis functions for reflectance (b), excitation (c), and emission (d) spectra.}
\label{fig:basis}
\end{figure}

Figure~\ref{fig:nBases} presents the root-mean-square error (RMSE) of the scaled Donaldson matrix estimates using the multi-fluorophore and single fluorophore models averaged over estimates for all test patches. About $12$ basis functions provide normalized Donaldson matrix estimate RMSE of the order of $0.01$. Furthermore, $12$ excitation and emission basis functions account for $97\%$ of variance in the McNamara-Boswell data set (Fig.~\ref{fig:basis}). Based on these calculations, we used $12$ excitation and emission bases for subsequent experiments.

\subsection{Number of camera and illuminant channels}
\begin{figure}
\centering
\begin{subfigure}{0.49\columnwidth}
\includegraphics[width=\textwidth]{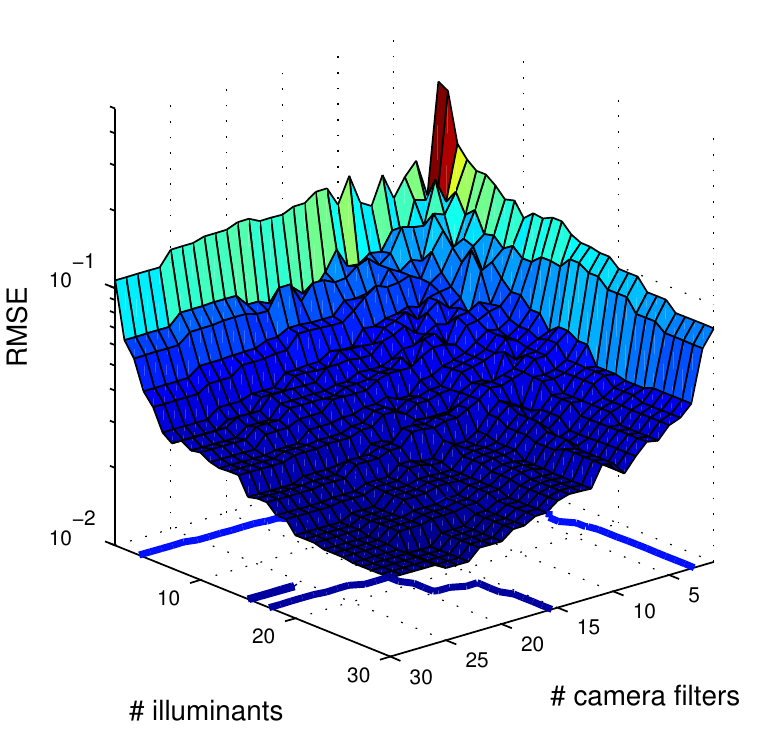}
\caption{Multi-fluorophore}
\end{subfigure}%
\begin{subfigure}{0.49\columnwidth}
\includegraphics[width=\textwidth]{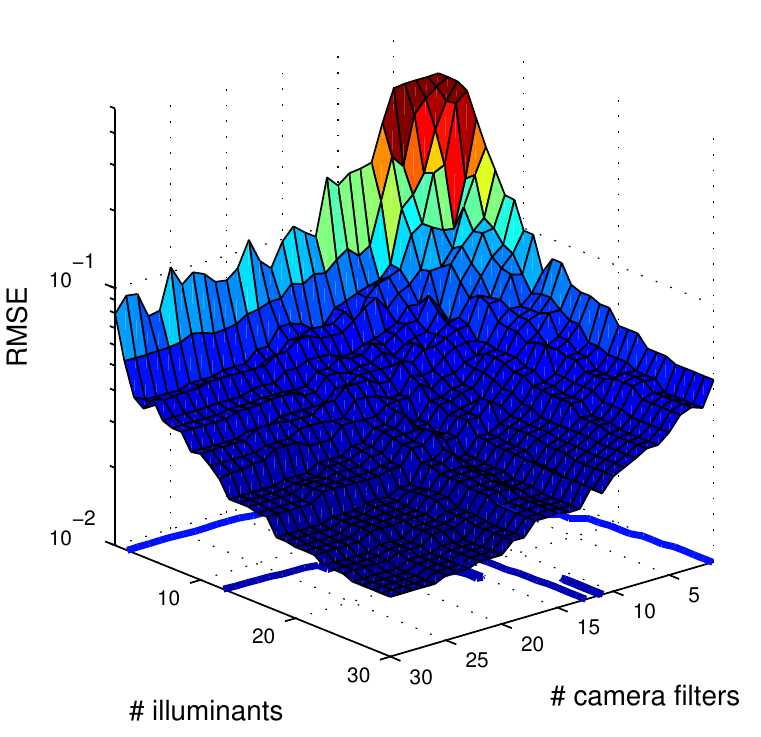}
\caption{Single fluorophore}
\end{subfigure}%
\caption{Normalized Donaldson matrix estimation error (RMSE) as a function of the number of camera filters and illuminants. The number of linear bases used to approximate the  excitation and emission spectra was set to $12$. Contour plots show iso-RMSE lines at $0.05$ and $0.02$.}
\label{fig:nChannels}
\end{figure}

Second, we varied the number of camera filters and illuminant channels. Both camera filters and illuminant spectral profiles were rectangular, and their widths were adjusted so that the sum of all channels produced a flat response over the entire spectral range. Just as before we set the tuning parameters to $\alpha = \beta = \eta = 0.001$. Figure~\ref{fig:nChannels} presents the RMSE of the Donaldson matrix estimates averaged over $24$ different fluorophores. The accuracy of fluorescence detection depends also on spectral shapes and pass-band positions of camera filters and illuminants. We did not change these parameters. This is why the error surfaces in Fig.~\ref{fig:nChannels} are less smooth, compared to those obtained by varying the number of basis functions. Using about $20$ filters and illuminants produced normalized RMSE on the order of $0.02$.

\subsection{Noise performance}
\begin{figure}
\centering
\begin{subfigure}{0.49\columnwidth}
\includegraphics[width=\textwidth]{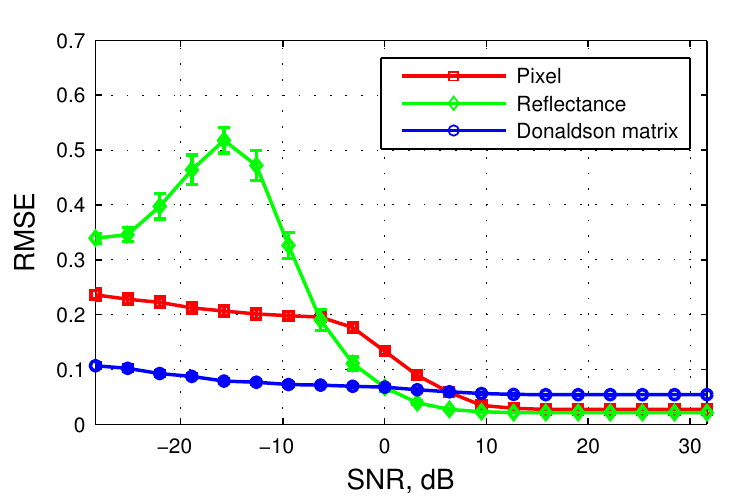}
\caption{Multi-fluorophore}
\end{subfigure}%
\begin{subfigure}{0.49\columnwidth}
\includegraphics[width=\textwidth]{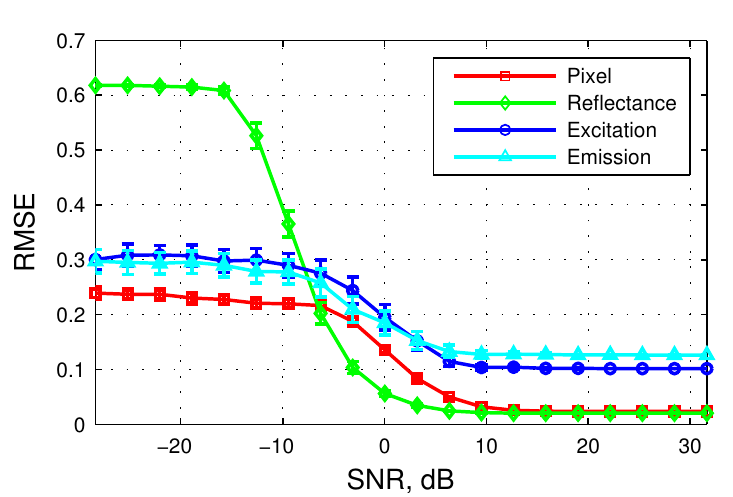}
\caption{Single fluorophore}
\end{subfigure}%
\caption{Estimation accuracy in the presence of noise.  Each curve shows the reduction in RMSE as the SNR increases for multi (a) and single (b) fluorophore estimation algorithms.  In both cases the estimation accuracy asymptotes as SNR approaches $10$dB. }
\label{fig:noise}
\end{figure}

Third, we analyzed the estimation accuracy in the presence of noise. We fixed tuning parameters to $\alpha=\beta=\eta=0.01$, and we added different amounts of Gaussian noise to the simulated pixel intensities $M$. At each noise level and for each sample we used $10$ different instances of noise patterns, producing $240$ estimates per noise level. Figure~\ref{fig:noise} shows the average RMSE of the Donaldson matrix, reflectance, and pixel values estimates as a function of the signal to noise ratio (SNR). The error bars represent standard errors computed for the $10$ noise instances and averaged over $24$ samples. The accuracy asymptotes with the SNR reaching $10$dB.

\subsection{Algorithm convergence} 
\begin{figure}
\centering
\begin{subfigure}{0.49\columnwidth}
\includegraphics[width=\textwidth]{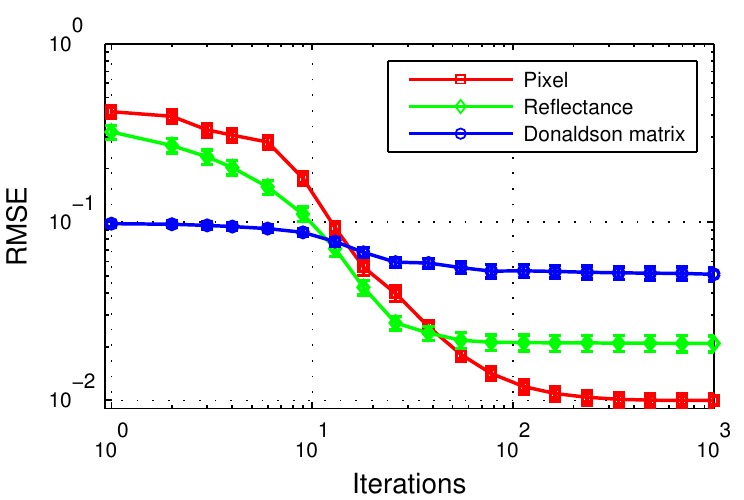}
\caption{Multi-fluorophore}
\end{subfigure}%
\begin{subfigure}{0.49\columnwidth}
\includegraphics[width=\textwidth]{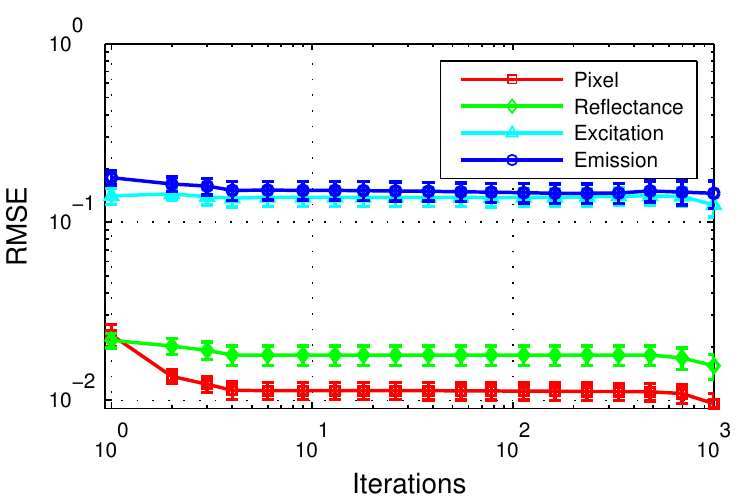}
\caption{Single fluorophore}
\end{subfigure}%
\caption{Estimation accuracy with increasing number of algorithm iterations. The multi-fluorophore method (a) converges to a solution more slowly than the single fluorophore algorithm (b).}
\label{fig:conv}
\end{figure}

Finally, we tested the estimation accuracy for different number of algorithm iterations. Figure~\ref{fig:conv} presents the multi-fluorophore estimates RMSE as a function of the number of ADMM iterations as well as the single fluorophore estimates RMSE as a function of the number of biconvex iterations. Algorithm parameters were set to $\alpha=\beta=\eta=0.01$. All curves are averaged over estimates for $24$ different fluorophores and reflectance spectra. The multi and single fluorophore estimates of the reflectance and pixel values converge to approximately the same RMSE values. The multi-fluorophore ADMM method converges much more slowly than the single fluorophore, biconvex solver.

\begin{figure}
\begin{minipage}[c]{.5\columnwidth}
  \vspace*{\fill}
  \centering
  \includegraphics[width=\textwidth]{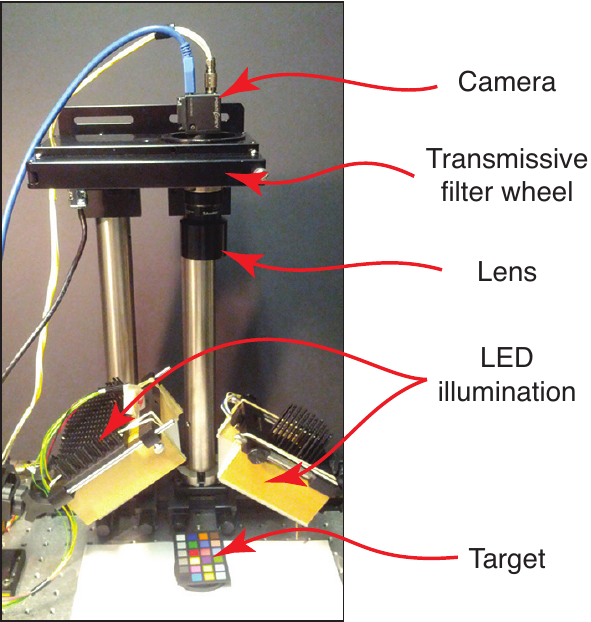}
  \subcaption{Overview}
  \label{fig:test1}
\end{minipage}%
\hfill
\begin{minipage}[c]{.5\columnwidth}
  \vspace*{\fill}
  \centering
  \includegraphics[width=\textwidth]{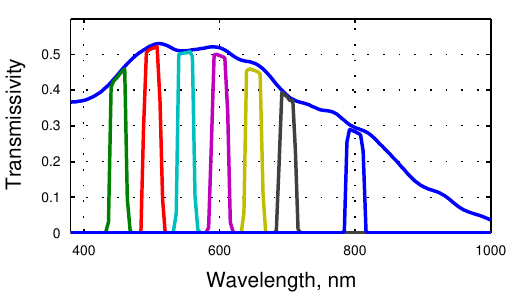}
  \subcaption{Filters}
  \label{fig:filters}
  \includegraphics[width=\textwidth]{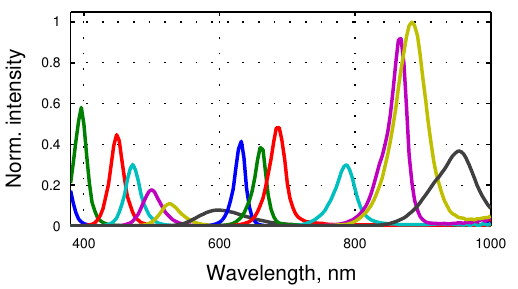}
  \subcaption{Illuminants}
  \label{fig:lights}
\end{minipage}
\caption{The experimental image acquisition system. (a) A monochrome, PointGrey Flea3 FL3-U3-13Y3M-C, 1.3MP camera with a Schneider Optics Tele-Xenar $70$mm lens. (b) The system acquires images through one broadband and seven $25$mm bandpass filters centered at $450$, $500$, $550$, $600$, $650$, $700$ and $800$nm. The filters are housed inside an Edmund Optics motorized filter wheel placed between the camera and the lens. (c) The scenes are illuminated by one of $14$ narrowband, LED illuminants. The LEDs with $350-400$nm peaks were manufactured by International Light Technologies, the $400-700$nm LEDs were from Luxeon, and the $700-950$nm LEDs from Epitex. Filter wheel positions and illumination times are synchronized using an Arduino Mega2560 controller. Non-uniform illumination was corrected using an image of a block of magnesium oxide (chalk).}
\label{fig:acqSystem}
\end{figure} 

\section{System evaluation}
\label{sec:system}

\begin{figure*}
\centering
\begin{subfigure}{0.02\textwidth}
\rotcaption*{Multi-fluorophore}
\end{subfigure}\hfill%
\begin{subfigure}{0.22\textwidth}
\includegraphics[width = \textwidth]{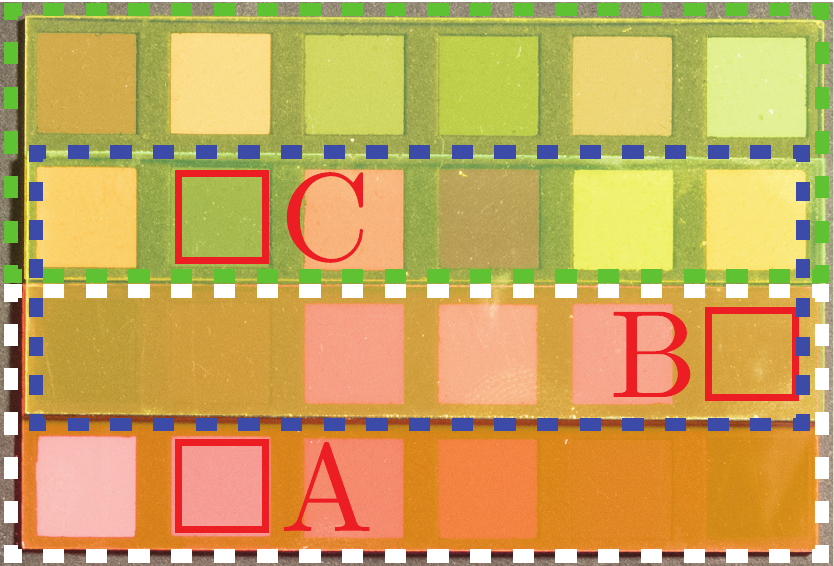}
\end{subfigure}\hfill%
\begin{subfigure}{0.22\textwidth}
\includegraphics[width = \textwidth]{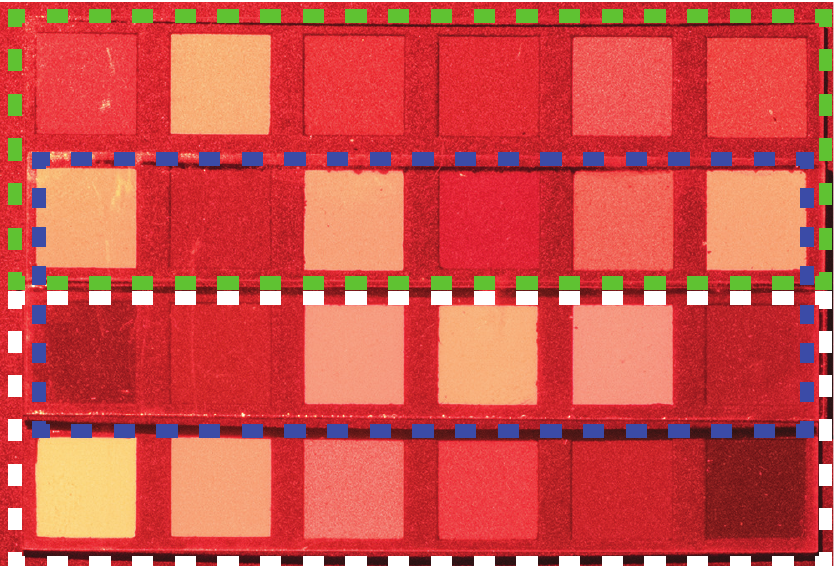}
\end{subfigure}\hfill%
\begin{subfigure}{0.22\textwidth}
\includegraphics[width = \textwidth]{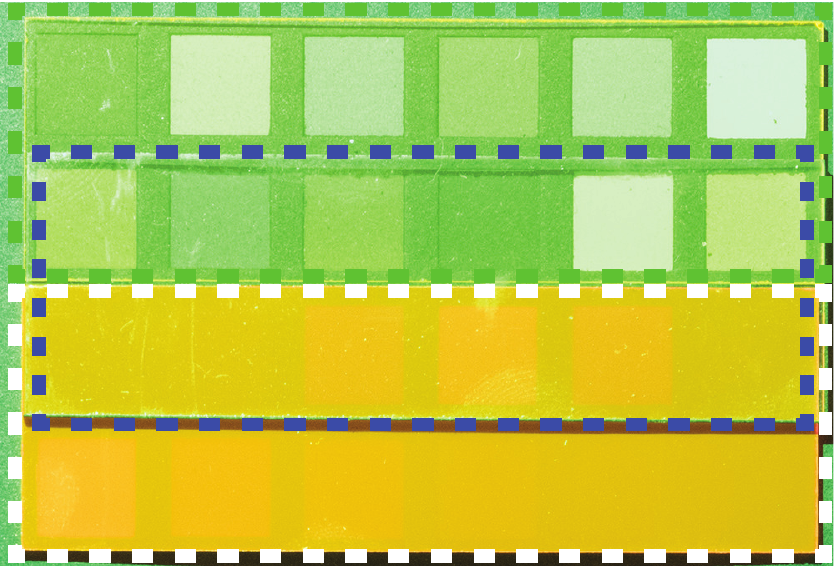}
\end{subfigure}\hfill%
\begin{subfigure}{0.22\textwidth}
\includegraphics[width = \textwidth]{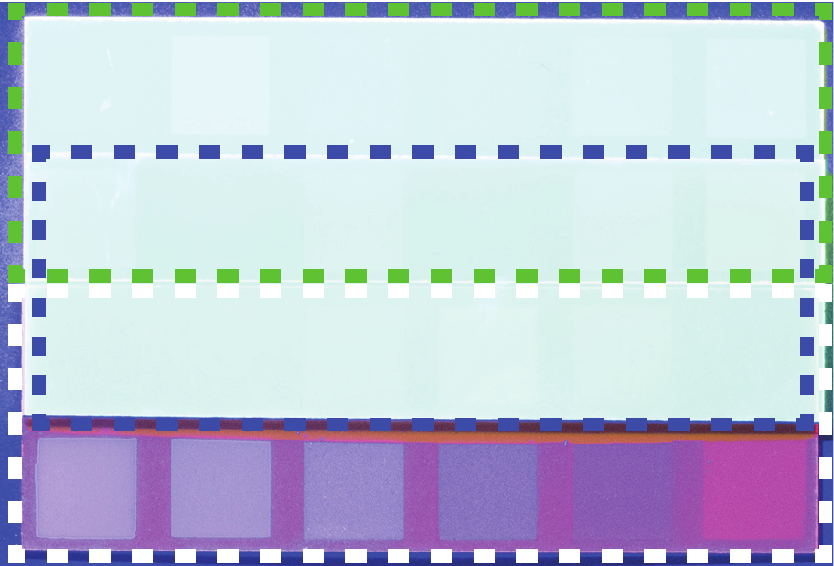}
\end{subfigure}\\
\vspace{0.25cm}
\begin{subfigure}{0.02\textwidth}
\rotcaption*{Single fluorophore}
\caption*{}
\end{subfigure}\hfill%
\begin{subfigure}{0.22\textwidth}
\includegraphics[width = \textwidth]{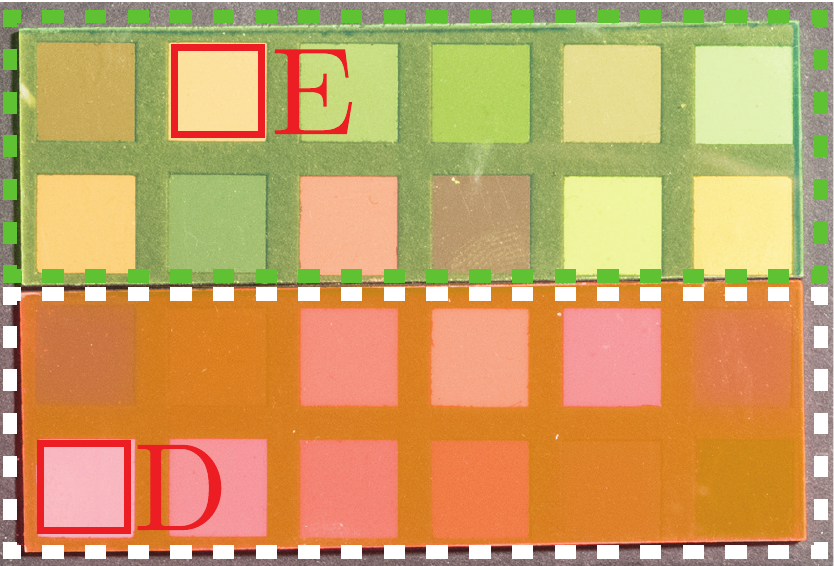}
\caption{Tungsten}
\end{subfigure}\hfill%
\begin{subfigure}{0.22\textwidth}
\includegraphics[width = \textwidth]{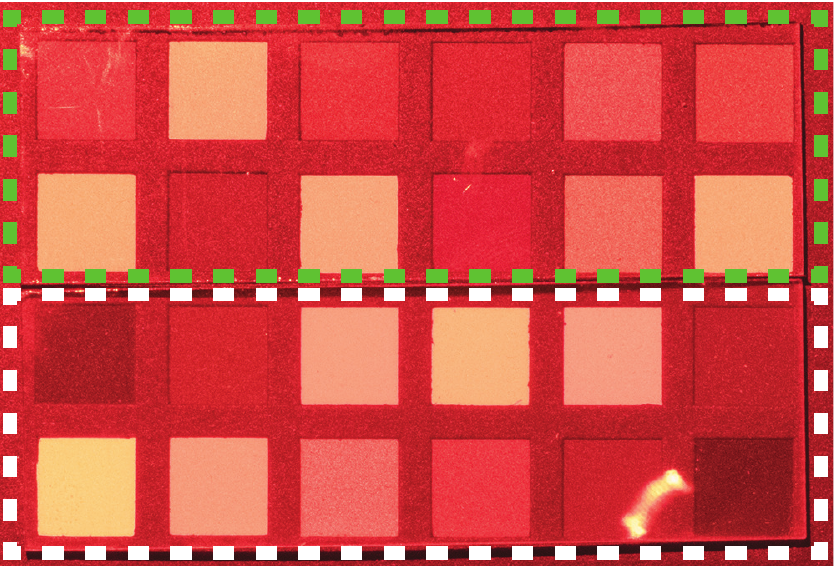}
\caption{Red LED}
\end{subfigure}\hfill%
\begin{subfigure}{0.22\textwidth}
\includegraphics[width = \textwidth]{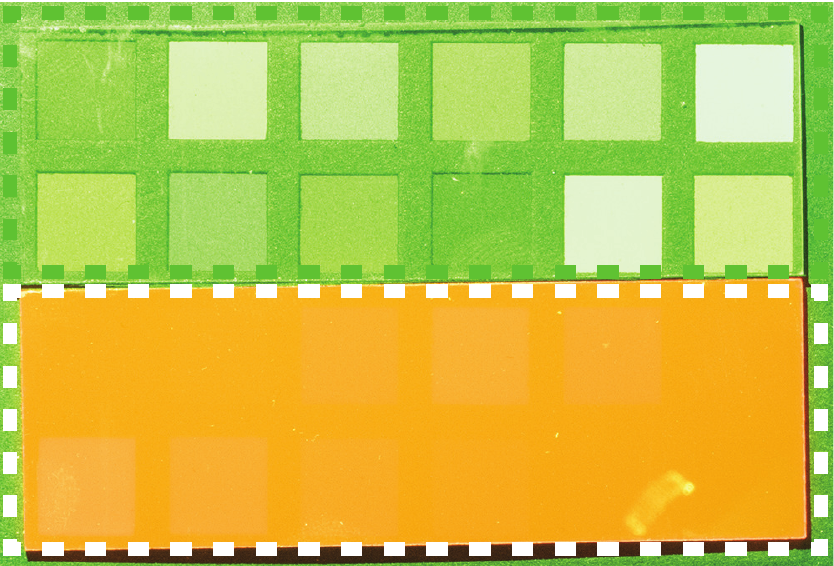}
\caption{Green LED}
\end{subfigure}\hfill%
\begin{subfigure}{0.22\textwidth}
\includegraphics[width = \textwidth]{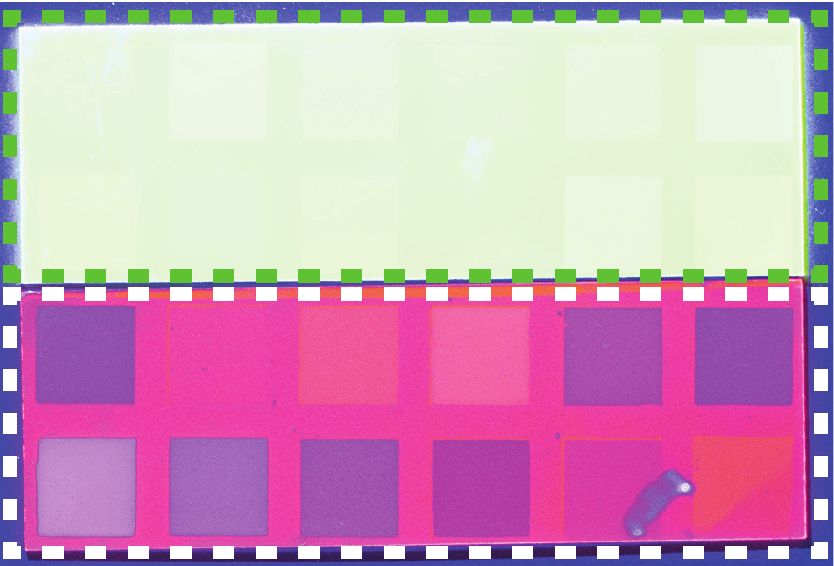}
\caption{Blue LED}
\end{subfigure}
\caption{Conventional color camera images of the test target composed of fluorescent slides placed on top of a reflective Macbeth chart. Different slide boundaries are marked with dashed lines. The multi-fluorphore configuration (top) has two layers of slides placed over middle rows of the Macbeth chart (blue dashed line). In the single fluorophore configuration (bottom) fluorescent slides form a single layer. Columns present the appearance of the targets under different light sources (a) tungsten, (b) red, (c) green, and (d) blue LEDs. The fluorescence and reflectance of patches labeled A--E in (a) is further analyzed in Fig.~\ref{fig:multiExExp}~and~\ref{fig:singleExExp}. The fluorescent slides are transparent under red light (b), but emit photons in the orange and green bands, when illuminated with green and blue light respectively (c,d). Note that the patch B changes color from green to orange when the illuminant changes from blue to green, and thus violates the chromaticity invariance principle.}
\label{fig:sampleTargetRGB}
\end{figure*}
\input{SampleFluorescenceImage.tex}

We describe a system to test the multiple fluorophore estimation algorithms in a practical setting.  The system evaluates the improvements our methods might offer in practical settings, where noise, calibration accuracy and other sources of uncertainty are non-negligible, we implemented a custom fluorescence imaging system and applied different algorithms to the measurements. 

In all computations we use five basis functions derived from the set of Macbeth chart reflectances \cite{Wandell:95} and $12$ basis for excitation and emission spectra, derived from the McNamara-Boswell data set \cite{McNamara:06}. The choice of tuning parameters ($\alpha$, $\beta$ and $\eta$) had little influence on the algorithm accuracy over a broad range of values; therefore, we adjusted them manually, rather than through rigorous cross-validation (see Supplemental Material, Appendix~B).

\subsection{System architecture}

We built a system with eight different camera filters and $14$ LED illuminants with peak emission spectra in the $350$ to $950$nm range Fig.~\ref{fig:acqSystem}. This is a practical system, but it has fewer than the optimal number of channels (20 filters and illuminants, Fig.~\ref{fig:nChannels}).

\input{ResultImage.tex}

\subsection{Targets}
The first experimental test target was composed of two building blocks; a purely reflective Macbeth color test chart and semi-transparent fluorescent microscopy slides from Chroma Technology Corporation\footnote{https://www.chroma.com/products/filter-accessories/diagnostic-slides}. The slides were placed on top of the Macbeth chart to create targets with a range of reflectance and fluorescence properties. We estimated the Macbeth reflectances using standard procedures of illuminating with a broadband light and measuring the returned radiance.  We estimated the fluorescence and transmissivity of the slides in the $380$nm to $1000$nm range in $4$nm bands using gold-standard bispectral methods.  We illuminated the slide with monochromatic light (Oriel Cornerstone $130$ monochromator) and measured the radiance (SpectraScan PR715 spectrophotometer). We computed the fluorescent emission spectrum by illuminating with a  short wavelength light ($360$nm) and using the radiance as the fluorescent emission. Once this is known, the transmittance and excitation was estimated using our multi-fluorophore algorithm  (\ref{eq:relaxedProblem}). 

The number of layers of fluorophore slides placed on top of the Macbeth chart defined the number of fluorophores. We used the chart plus one slide, or the chart and the superposition of two slides to evaluate the performance of the multi-fluorophore method.  We used the chart and one slide to evaluate single fluorophore algorithm (Fig.~\ref{fig:sampleTargetRGB}).

A second multi-fluorophore test target was created by coloring different shapes with fluorescent paints and using a sheet of traditional white office paper as a substrate. The paints contained one type of fluorophore, which combined with the intrinsic paper fluorescence \cite{Tian:11} to produce a multi-fluorophore target.

\input{multiFlAccuracy.tex}

Figure~\ref{fig:sampleTarget} presents an $8 \times 14$ matrix of images of the single fluorophore target. We acquired each image under a specific illuminant (columns) with a particular filter (rows). The top shows the broadband, monochromatic images.   The images near the diagonal are dominated by reflectance, and the images below the diagonal measure fluorescence. The data using $395$nm through $530$nm illuminants produces clearly visible fluorescent responses. 

\subsection{Estimation}

We use bootstrapping to calculate the $95\%$ confidence intervals on the estimated curves \cite{Efron:86}. Given a particular test patch we run the estimation algorithms $100$ times using pixel values randomly selected from the image area representing that patch. Confidence interval boundaries are given by the $2.5$th and $97.5$th percentiles of estimate distributions at a particular wavelength. 

\input{RelightImagesV2.tex}

\subsubsection{Multi-fluorophore estimation}

We evaluate the multi-fluorophore estimation algorithm using $24$ test patches containing one or two superimposed fluorophores (Table~\ref{tab:multiAccExp}). Figure~\ref{fig:multiExExp} shows spectral estimates and ground truth data for three patches outlined on the test chart presented in Fig.~\ref{fig:sampleTargetRGB}. Patch~A contains single 'orange' fluorophore, patch~B is a mixture of 'orange' and 'green' fluorophores and patch C is a mixture of 'green' and 'amber' fluorophores. 

We used the multi-fluorophore method with tuning parameters $\alpha=0.1$, $\beta=5$, $\eta=0.01$ to perform spectral estimation. Our algorithm correctly determines that patch A contains a single fluorophore with the emission peak around $600$nm and the excitation peak above $500$nm. The the Donaldson matrix estimate for patch B is bimodal. The estimate contains the orange fluorophore peak at $600$nm emission wavelength, just like the estimate for patch~A, and another peak representing the green fluorophore, with the emission around $500$nm. In case of patch C the spectral distinction between green and amber fluorophores is small (Fig.~\ref{fig:sampleTargetRGB}d). For this reason the Donaldson matrix estimate is unimodal. 

\begin{figure*}
\centering
\begin{subfigure}{0.04\textwidth}
\rotcaption*{Patch D}
\end{subfigure}%
\begin{subfigure}{0.19\textwidth}
\includegraphics[width = \textwidth]{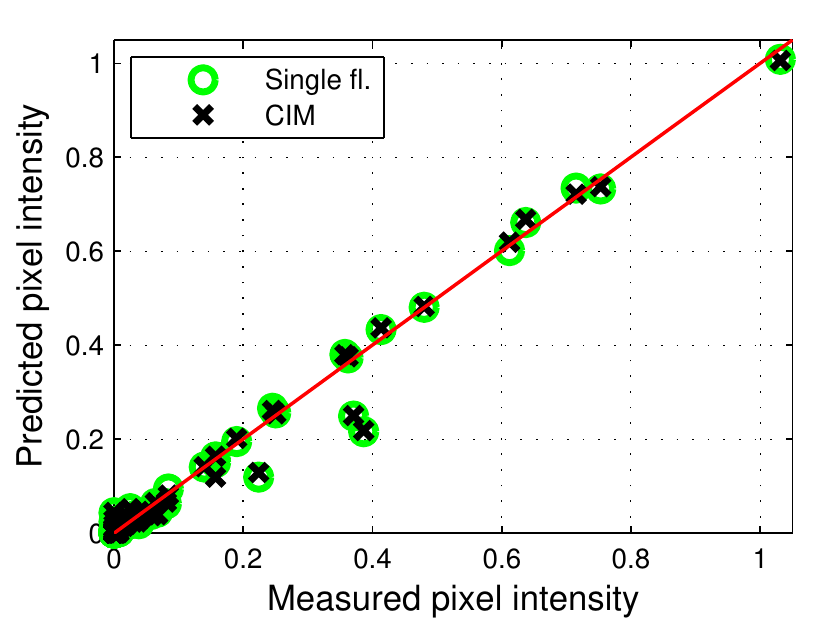}
\end{subfigure}%
\begin{subfigure}{0.19\textwidth}
\includegraphics[width = \textwidth]{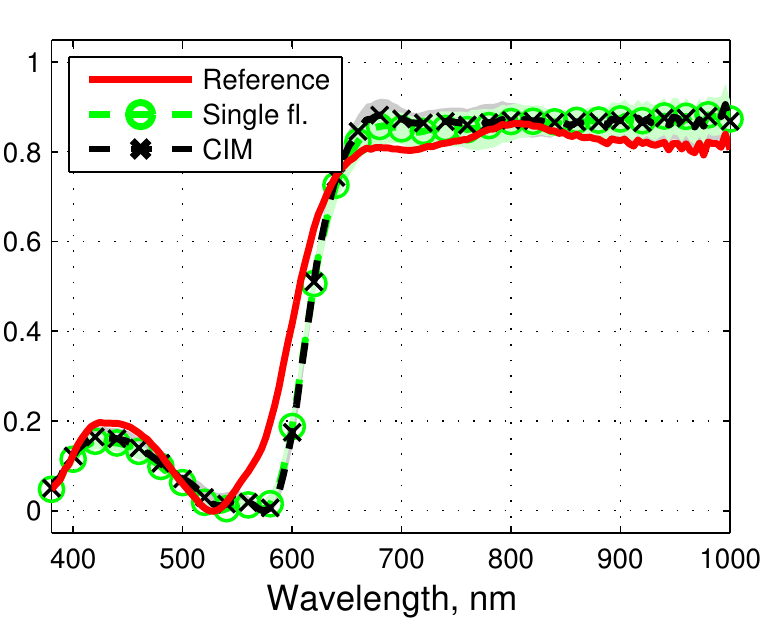}
\end{subfigure}%
\begin{subfigure}{0.19\textwidth}
\includegraphics[width = \textwidth]{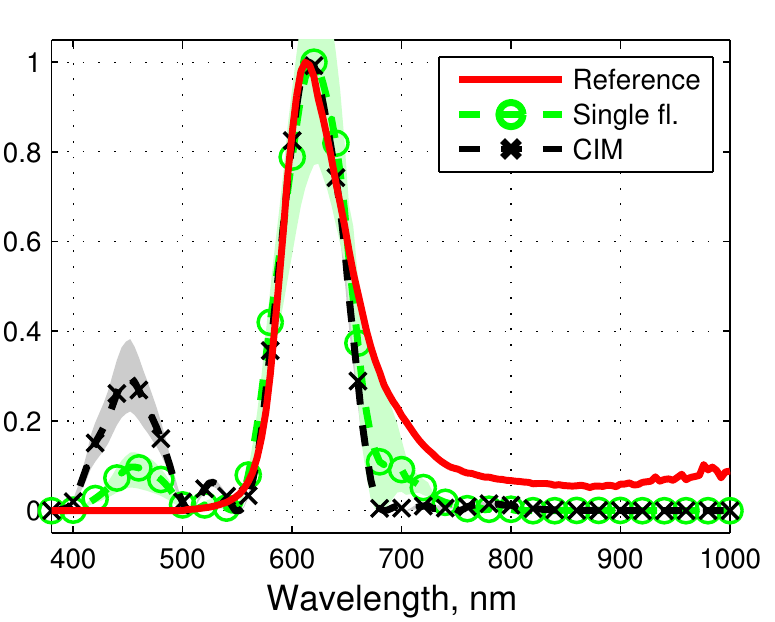}
\end{subfigure}%
\begin{subfigure}{0.19\textwidth}
\includegraphics[width = \textwidth]{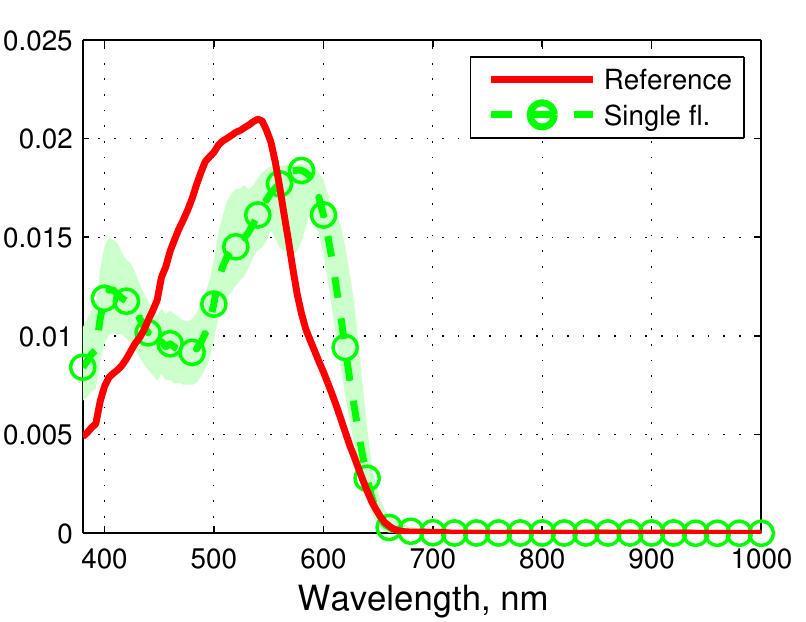}
\end{subfigure}%
\begin{subfigure}{0.19\textwidth}
\includegraphics[width = \textwidth]{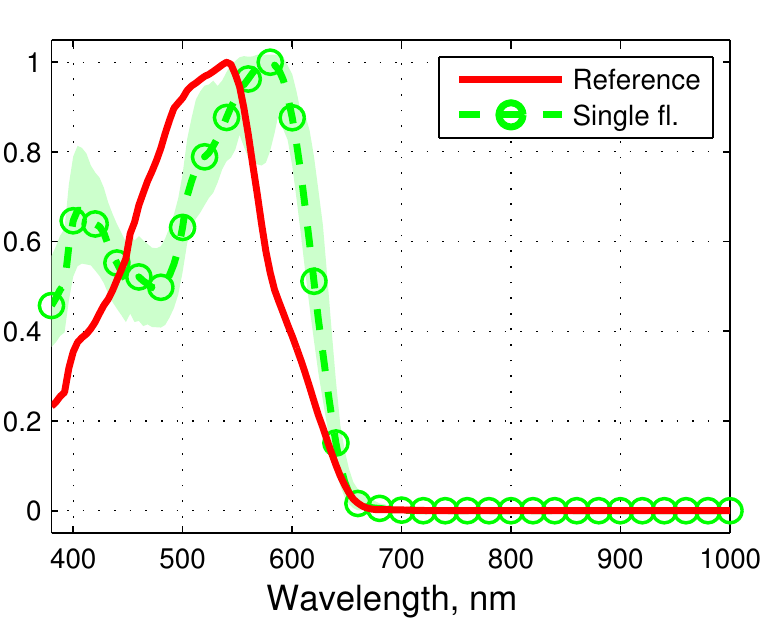}
\end{subfigure}\\
\begin{subfigure}{0.04\textwidth}
\rotcaption*{Patch E}
\end{subfigure}%
\begin{subfigure}{0.19\textwidth}
\includegraphics[width = \textwidth]{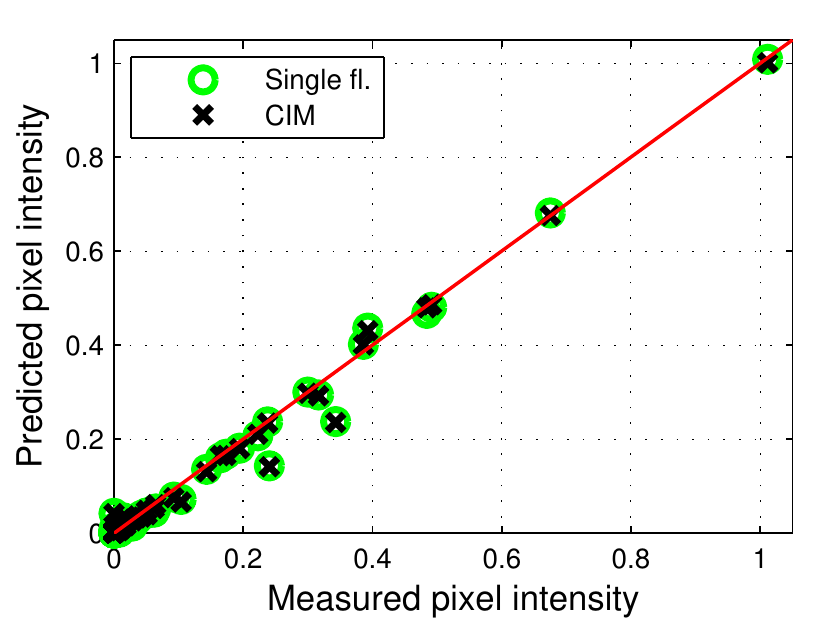}
\end{subfigure}%
\begin{subfigure}{0.19\textwidth}
\includegraphics[width = \textwidth]{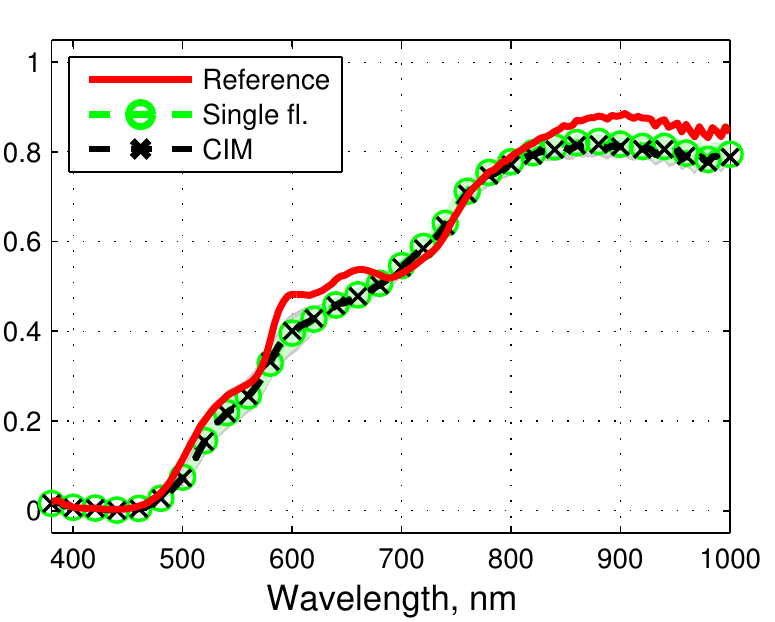}
\end{subfigure}%
\begin{subfigure}{0.19\textwidth}
\includegraphics[width = \textwidth]{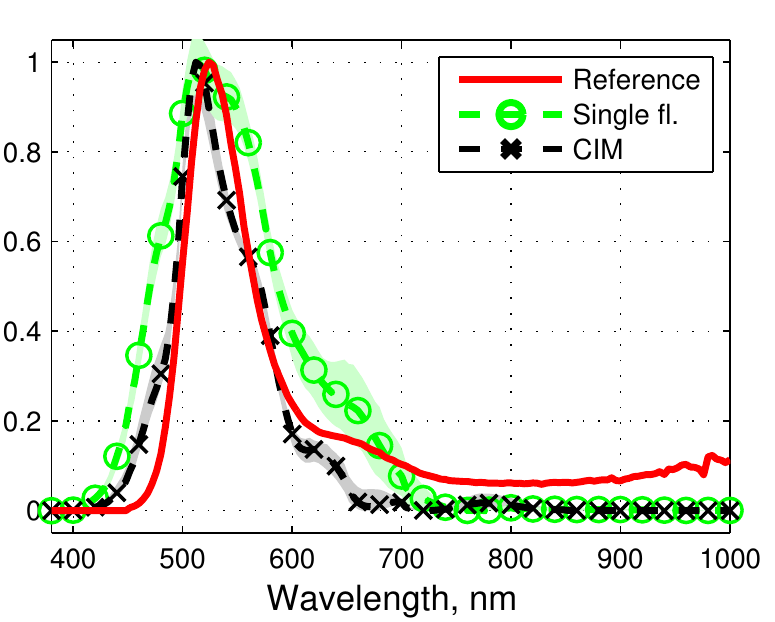}
\end{subfigure}%
\begin{subfigure}{0.19\textwidth}
\includegraphics[width = \textwidth]{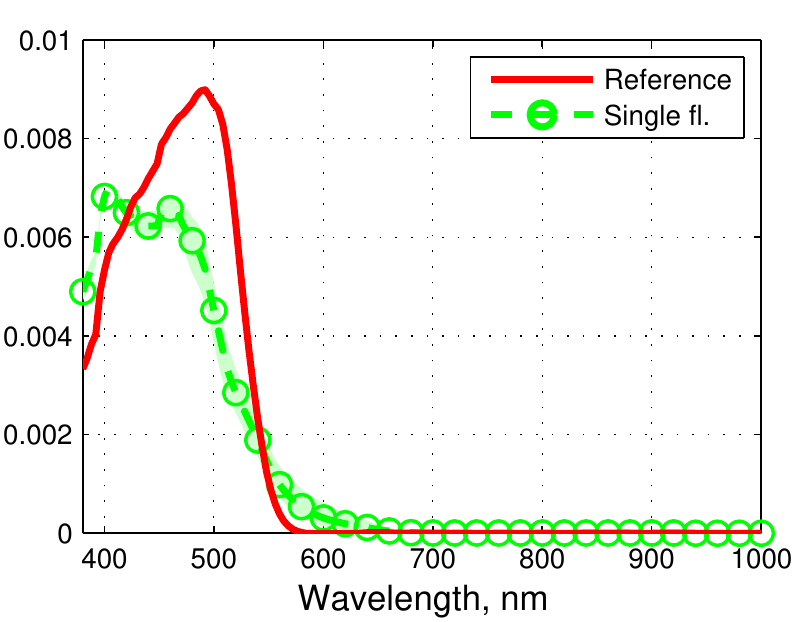}
\end{subfigure}%
\begin{subfigure}{0.19\textwidth}
\includegraphics[width = \textwidth]{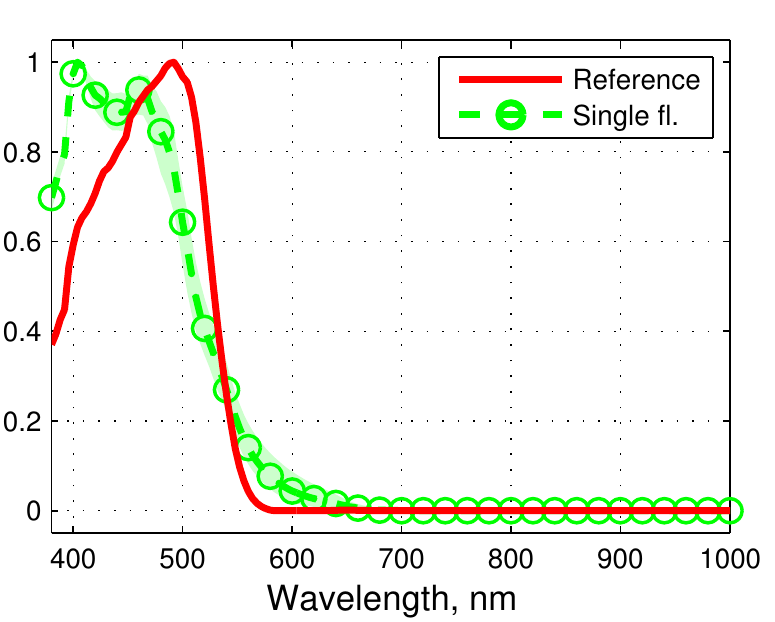}
\end{subfigure}\\
\begin{subfigure}{0.04\textwidth}
\caption*{}
\end{subfigure}%
\begin{subfigure}{0.19\textwidth}
\caption{Pixels}
\end{subfigure}%
\begin{subfigure}{0.19\textwidth}
\caption{Reflectance}
\end{subfigure}%
\begin{subfigure}{0.19\textwidth}
\caption{Emission}
\end{subfigure}%
\begin{subfigure}{0.19\textwidth}
\caption{Excitation (absolute)}
\end{subfigure}%
\begin{subfigure}{0.19\textwidth}
\caption{Excitation (normalized)}
\end{subfigure}
\caption{Single fluorophore estimates. A row of panels represents estimates for a single test patch and each panel compares  true (solid) and estimated (dashed) pixel intensities (a), reflectance (b), emission (c), absolute  excitation (d), and normalized excitation (e). The shaded areas represent $95\%$ confidence intervals around the estimates.}
\label{fig:singleExExp}
\end{figure*}
\input{singleFlAccuracy.tex}

Once the image is separated into reflected and fluoresced components, we can predict the returned radiance when the object is illuminated by arbitrary lights (scene relighting). Figure~\ref{fig:relighting} shows a scene composed of two symbols on a  sheet of white paper. Each symbol was painted with a different fluorescent paint. We imaged the target with our apparatus and used the multi-fluorophore analyses to calculate reflectance and flourescence. We also captured images of the target with a Canon G7 X consumer camera. 

We then tested the ability to 'relight' the image. Specifically, we used the estimated Donaldson matrices and reflectance spectra to predict the spectral image for a set of illuminants generated by a Thouslite LEDCube\footnote{http://www.thouslite.com/show.asp?id=16}.  The rendering was based on a model of the Canon camera, which we created from a set of images of a Macbeth chart captured under different illuminants with known spectral power distributions. We compare the simulated and captured camera pixel intensities using RGB root-mean-squared error maps (Figure~\ref{fig:relighting}e). To simplify the comparison, the captured images were cropped, downsampled and aligned with the simulated images. The relighting is generally accurate to about 5\% error, with some outliers due to illuminant nonuniformities.

\subsubsection{Single fluorophore estimation}

We use the single fluorophore and CIM models to analyze the target with one fluorophore ( $\alpha=0.01$ and $\beta=0.1$). The $\beta$ value is reduced compared to the multi-fluorophore setting because the two methods compute smoothness of excitation end emission spectra only, rather than every row and column of the Donaldson matrix. In our case the Donaldson matrix contains $156$ rows and columns, which explains the two order of magnitude difference in the value of $\beta$.

Figure~\ref{fig:singleExExp} shows the estimated pixel values, reflectance and fluorescence excitation and emission spectra of patches D and E (Fig.~\ref{fig:sampleTargetRGB}a). Both algorithms accurately model the measured pixel intensities and provide good estimates of the reflectance, excitation and emission spectra shapes as well as intensity scales. The estimates are reliable and repeatable, as indicated by the $95\%$ confidence intervals. Table~\ref{tab:singleAccExp} summarizes the average error (RMSE) over $24$ test patches. The single fluorophore and CIM algorithms achieve similar RMSE scores. The CIM approach does not recover the excitation spectra, but it more accurately estimates the fluorescence emission.

\section{Discussion}
\label{sec:discussion}

We present a simultaneous reflectance and fluorescence estimation framework that unifies multi-fluorophore and single fluorophore applications. Simulations show that for typical data sets, the algorithms reach asymptotic performance with $12$ excitation and emission basis functions and $20$ camera channels and $20$ illuminants. The iterative algorithms converge to a solution in a few hundred iterations.

We experimentally evaluate the algorithms using a slightly simplified imaging system with $14$ illuminants and $8$ channels. The algorithms were applied to data captured with this system and we report on the accuracy of the estimated surface spectral reflectance (Table~\ref{tab:multiAccExp}, \ref{tab:singleAccExp}), Donaldson matrix peak positions, shapes, and the overall amount of fluorescence (Fig.~\ref{fig:multiExExp}), as well as single fluorophore excitation and emission spectra (Fig.~\ref{fig:singleExExp}). We compare our algorithm performance with our implementations of prior work.

\subsection{Algorithm performance}

The CIM algorithm produces the most accurate emission spectra estimates, though it does not return the excitation spectrum estimate. CIM achieves this accuracy by solving an optimization problem that allows independent scaling of the fluorescence emission under every illuminant. The other multi- and single fluorophore approaches couple the fluorescent scaling through the excitation spectrum, which reduces the accuracy slightly (Table~\ref{tab:singleAccExp}).

When two or more fluorophores with distinct emission spectra are present, the proposed imaging system, combined with the multi-fluorophore approach, correctly identifies the multimodal character of the Donaldson matrix (Fig.~\ref{fig:multiExExp}, patches A, B). When the fluorescent emission spectra overlap, the multi-fluorophore approach produces a unimodal estimate, which can be confused with a single fluorophore case (patch C). The experimental accuracy is limited by the imaging system, rather than the algorithms; the simulations demonstrate that resolution can be increased with systems that include a larger number of imaging channels and/or illuminants (Fig.~\ref{fig:nChannels}). 

Knowledge of the reflectance and fluorescence properties allow us to predict the spectral radiance under different illuminants (Fig.~\ref{fig:relighting}). We evaluated the accuracy of this calculation by capturing images of the same object under the simulated illumination conditions. The measured and predicted camera images are accurate to about 5\%, with much of the error being due to surface unevenness and spatial light non-uniformity. We note that the present algorithms operate independently on each pixel, and it is likely that additional spatial constraints, such as a total variation prior, may improve performance.  Although this greatly increases the size of the calculation, early tests show that the spatial constraints can be incorporated and solved with the Alternating Direction Method of Multipliers \cite{Blasinski:15}.

We note some practical issues relating to system performance. To determine the correct scales of fluoresced and reflected radiances the camera has to be accurately calibrated over all gains (ISO), photo response non-uniformity, shutter speed and aperture settings. We also observed that it is important to assure proper thermal management of the LED light sources. High power LEDs produce significant amounts of heat which, if not dissipated, affects light output and causes wavelength shifts in the illuminant spectrum. These calibration errors have smaller impact on estimating the shapes of excitation and emission spectra, but they greatly influence the measurements of absolute spectral levels.

\subsection{Comparison with prior work}

We implemented and compared prior methods with the proposed method. First, we adapted the nuclear norm minimization approach of Suo \etal \cite{Suo:14}, (Supplemental Material, Appendix~D). Second we implemented the multi-step algorithm of Fu \etal \cite{Fu:14}. These implementations, along with our algorithm, are available in our code repository.

Our methods have smaller error (RMSE) compared to these algorithms. Fu \etal \cite{Fu:14,Fu:15} (Table~\ref{tab:singleAccExp}) use a sequence of optimizations while the single fluorophore and chromaticity invariant (CIM) methods solve with a single optimization step. Perhaps the performance improvement is because the single step avoids accumulating errors across different stages. 

Suo \etal \cite{Suo:14} include a tuning parameter that sets a bound on the pixel prediction error. We adjusted this parameter so that the error in measured pixel intensities is the same as in our multi-fluorophore approach.  The high accuracy in pixel value predictions does not translate to accurate reflectance and Donaldson matrix estimates. For the same pixel error, our method produces more accurate Donaldson matrix and reflectance estimates. 



\section{Conclusions}
\label{sec:concl}

Fluorescent materials are common in our environment, and fluorescent signals are particularly important in biology and medicine. The separation and estimation of reflected and fluoresced radiances is a complex problem, because photons radiated in the two phenomena are indistinguishable from one another. Algorithms that can separate reflected and fluoresced components can provide useful information about substrates (coral reefs, biological tissues) that can be used in diagnostics, analysis or classification.

We present a unified framework for simultaneous estimation of surface reflectance and fluorescence properties. We show how to derive these properties from a small number of images taken with different narrowband filters and under narrowband illuminants. Our image formation model makes few assumptions regarding the properties of fluorescence emission and can account for multiple fluorphores present in the sample. We show how the general, multi-fluorophore estimation algorithm can be further simplified when it is known that only one fluorescent compound is present in the sample. The simplified single fluorophore and CIM models are more computationally efficient. 

We evaluated the algorithms using data from a simple imaging system we built. The system uses standard, off-the-shelf components: bandpass filters, LEDs and a CMOS sensor that can be easily integrated into other imaging devices operating at micro and macro scales. We showed that our approaches produce lower errors compared to earlier algorithms.

We frame reflectance and fluorescence estimation as inverse estimation problems and use convex optimization techniques to search for solutions. Such formulations allow easy algorithmic modifications, when for example, fluorescence emission properties are known and only their amounts need to be quantified. We provide an implementation of the algorithms as well as critical data to help readers reproduce and improve upon our results.





\bibliographystyle{./bibtex/ieee}
\bibliography{./bibtex/Biblio,./bibtex/Bibliography,./bibtex/Coral}

\begin{IEEEbiography}[{\includegraphics[width=1in,height=1.25in,clip,keepaspectratio]{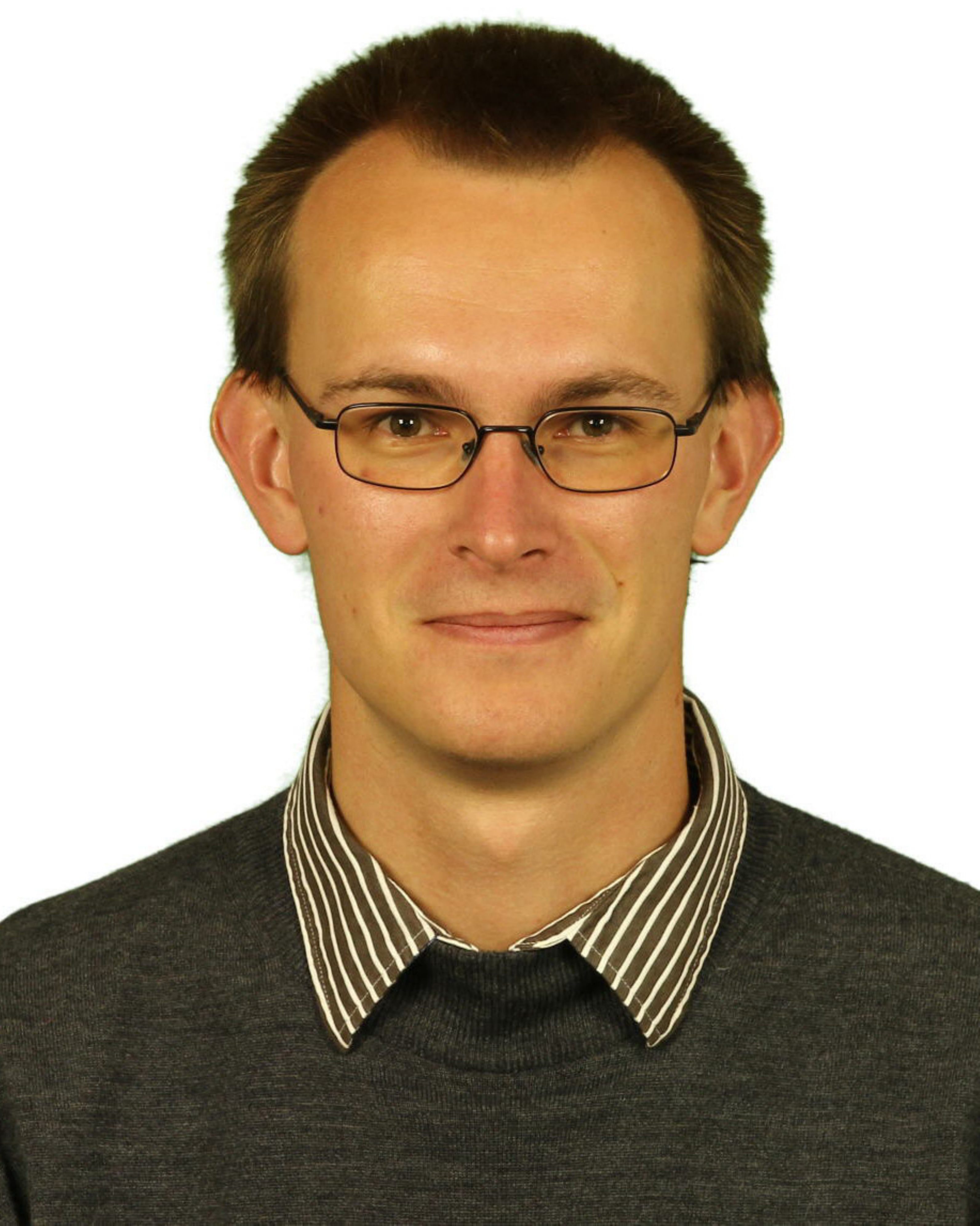}}]{Henryk Blasinski} Henryk Blasinski (S'10) received the M.S. degree (Hons.) in telecommunications and computer science from the Lodz University of Technology, Lodz, Poland, and the Diplome d'Ingeneiur degree from the Institut Superieur d'Electronique de Paris, France, in 2008 and 2009, respectively. He was a Fulbright Scholar with the Department of Electrical and Computer Engineering, University of Rochester, Rochester, NY, from 2010 to 2011. At present he is pursuing a Ph.D. degree at the Department of Electrical Engineering, Stanford University, CA. Henryk's research interests include image processing, human and computer vision and machine learning. Henryk is a recipient of several awards, including the Fulbright Fellowship, the Fellowship from the Minister of Higher Education of the Republic of Poland, the Polish Talents Award, the DP Systems Award, the Fellowship of the Lodz Region Marshall, the Crawford Prize for the best M. Sc. project and the 2014 SPIE Digital Photography X Best Paper Award.
\end{IEEEbiography}

\begin{IEEEbiography}[{\includegraphics[width=1in,height=1.25in,clip,keepaspectratio]{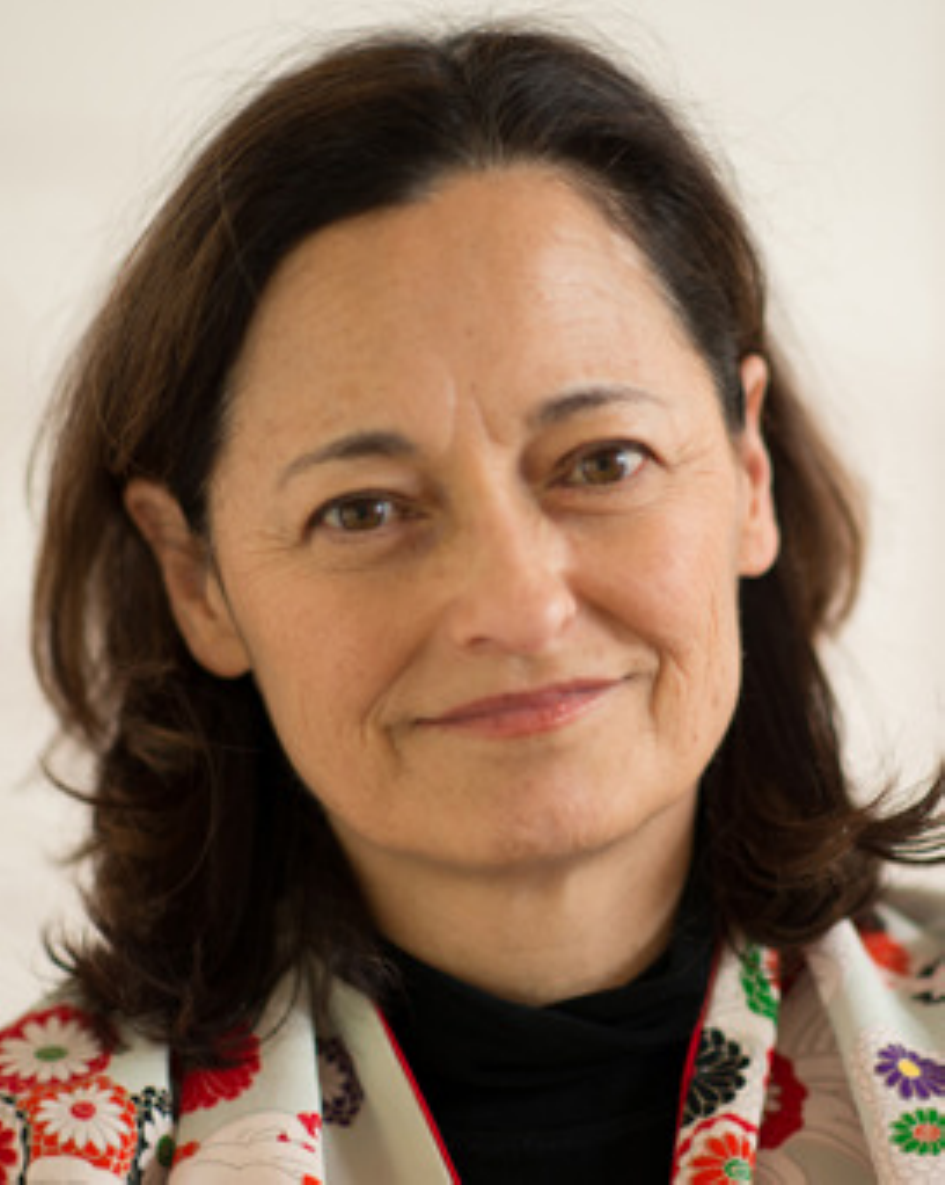}}]{Joyce Farrell} is the Executive Director of the Stanford Center for Image Systems Engineering and a senior research associate in the Department of Electrical Engineering at Stanford University. She has a doctorate degree from Stanford University and more than 20 years of research and professional experience  working at a variety of companies and institutions, including the NASA Ames Research Center, New York University, the Xerox Palo Alto Research Center, Hewlett Packard Laboratories and Shutterfly. She is also the CEO and founder of ImagEval Consulting, LLC.
\end{IEEEbiography}

\begin{IEEEbiography}[{\includegraphics[width=1in,height=1.25in,clip,keepaspectratio]{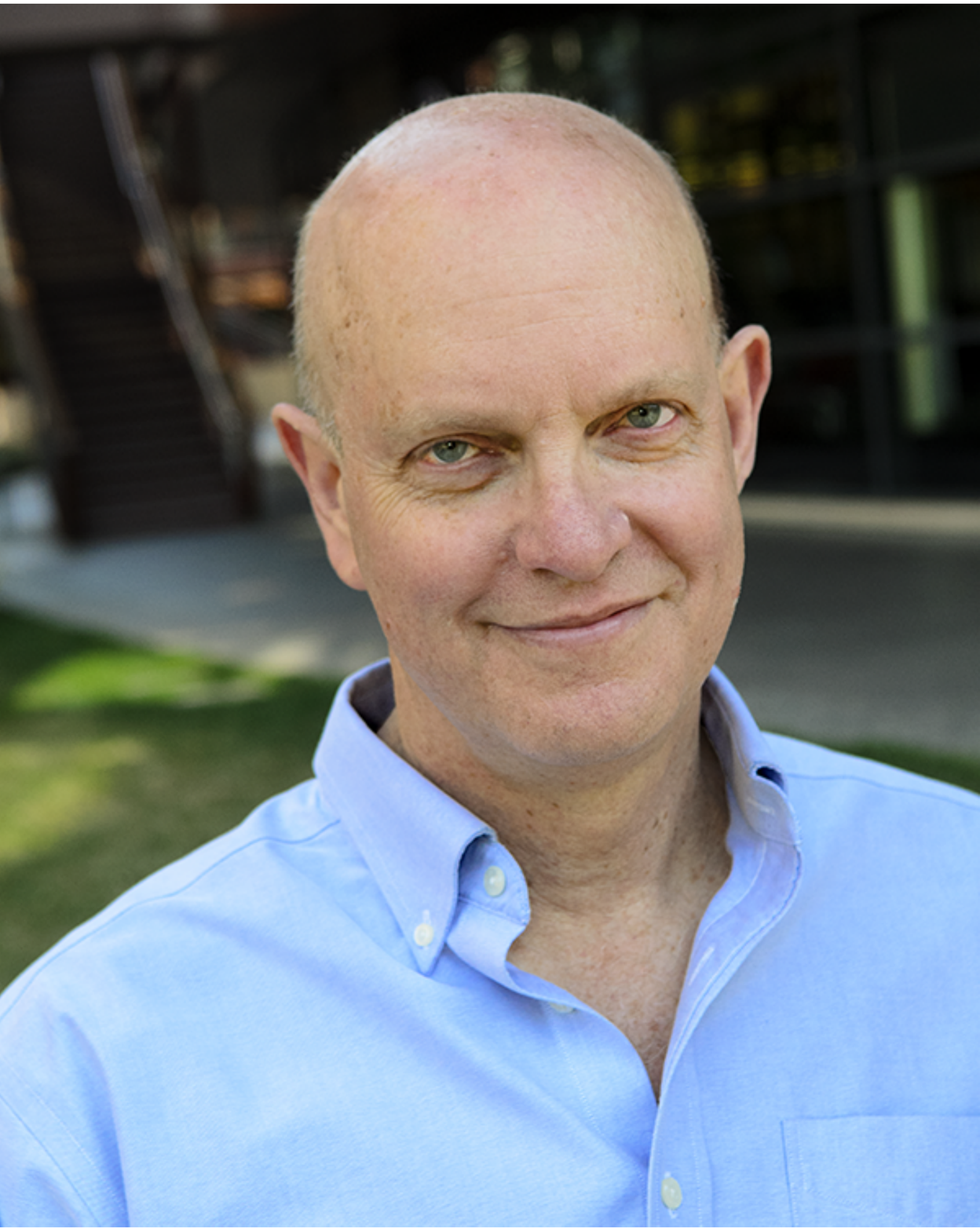}}]{Brian Wandell} is the first Isaac and Madeline Stein Family Professor. He joined the Stanford Psychology faculty in 1979 and is a member, by courtesy, of Electrical Engineering and Ophthalmology. He is Director of Stanford’s Center for Cognitive and Neurobiological Imaging, and Deputy Director of Stanford’s Neuroscience Institute. Wandell’s research centers on vision science, spanning topics from visual disorders, reading development in children, to digital imaging devices and algorithms for both magnetic resonance imaging and digital imaging.

Brian Wandell graduated from the University of Michigan in 1973 with a B.S. in mathematics and psychology. In 1977, he earned a Ph.D. in social science from the University of California at Irvine. After a year as a postdoctoral fellow at the University of Pennsylvania, he joined the faculty of Stanford University in 1979. 

In 1986, Wandell won the Troland Research Award from the National Academy of Sciences for his work in color vision. He was made a fellow of the Optical Society of America in 1990; in 1997 he became a McKnight Senior Investigator and received the Edridge Green Medal in Ophthalmology for work in visual neuroscience. In 2000, he was awarded the Macbeth Prize from the Inter-Society Color Council, and in 2007 he was named Electronic Imaging Scientist of the Year by the SPIE/IS\&T, and he was awarded the Tillyer Prize from the Optical Society of America in 2008. He was elected to the American Academy of Arts and Sciences in 2011. Oberdorfer Award from the Association for Research in Vision and Ophthalmology, 2012. In 2014 he was awarded the highest honor of the Society for Imaging Science and Technology. Wandell was elected to the US National Academy of Sciences in 2003.
\end{IEEEbiography}

\end{document}

%% file: SampleFluorescenceImage.tex
\begin{figure*}
\centering
\begin{subfigure}{0.06\textwidth}
\captionsetup{justification=raggedright,font=scriptsize}
\caption*{MC}
\end{subfigure}%
\begin{subfigure}{0.06\textwidth}
\centering
\includegraphics[width=\textwidth]{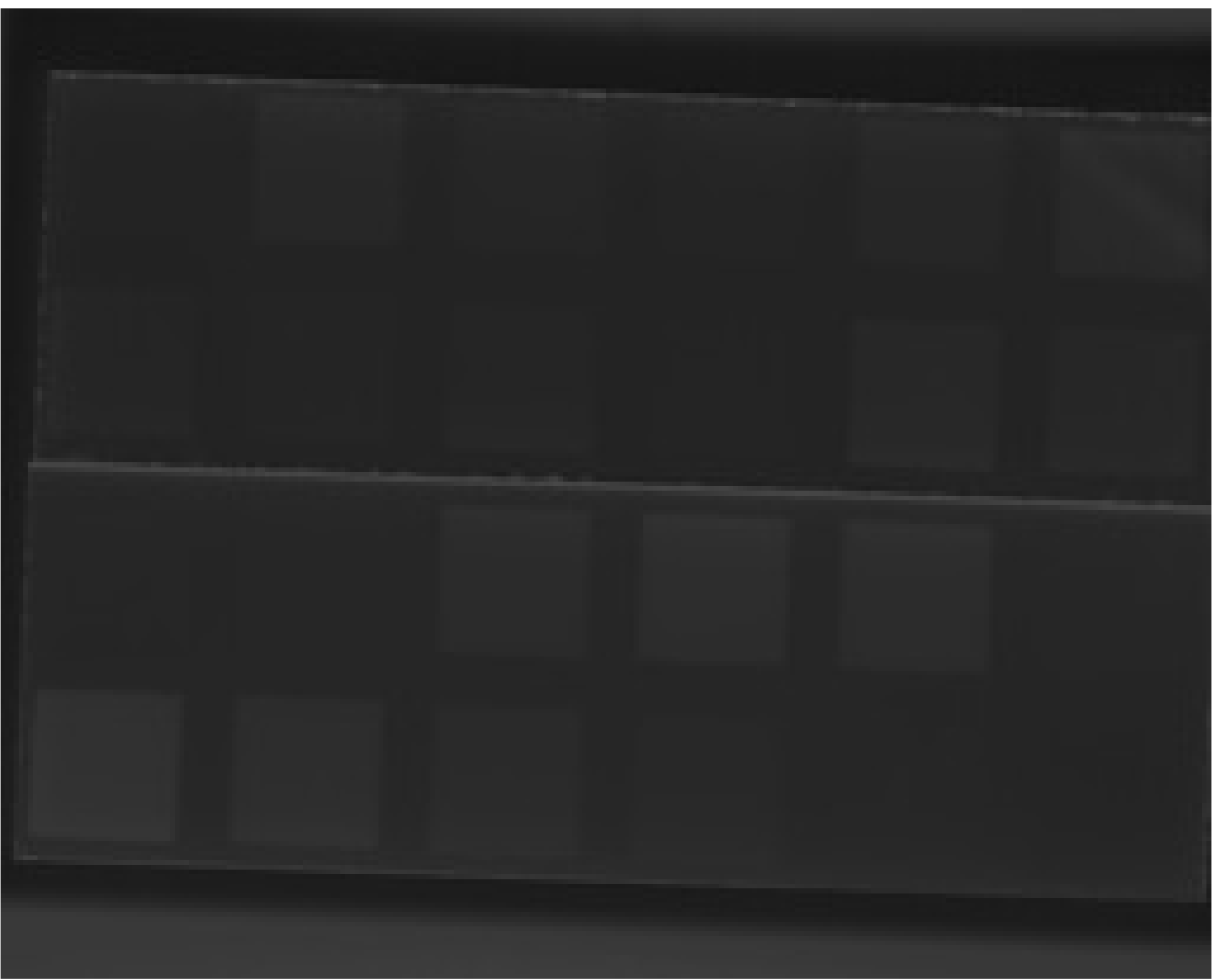}
\end{subfigure}%
\begin{subfigure}{0.06\textwidth}
\centering
\includegraphics[width=\textwidth]{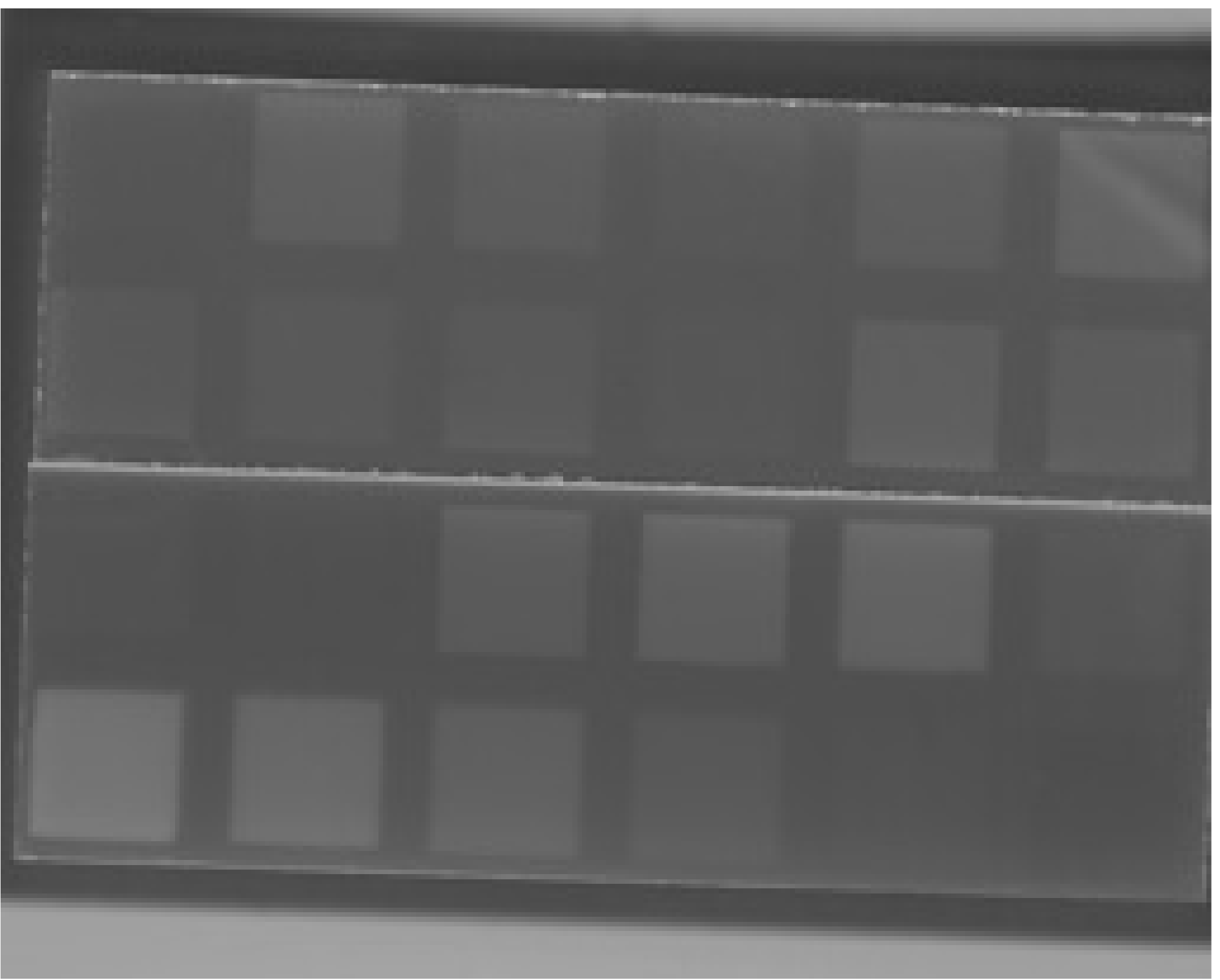}
\end{subfigure}%
\begin{subfigure}{0.06\textwidth}
\centering
\includegraphics[width=\textwidth]{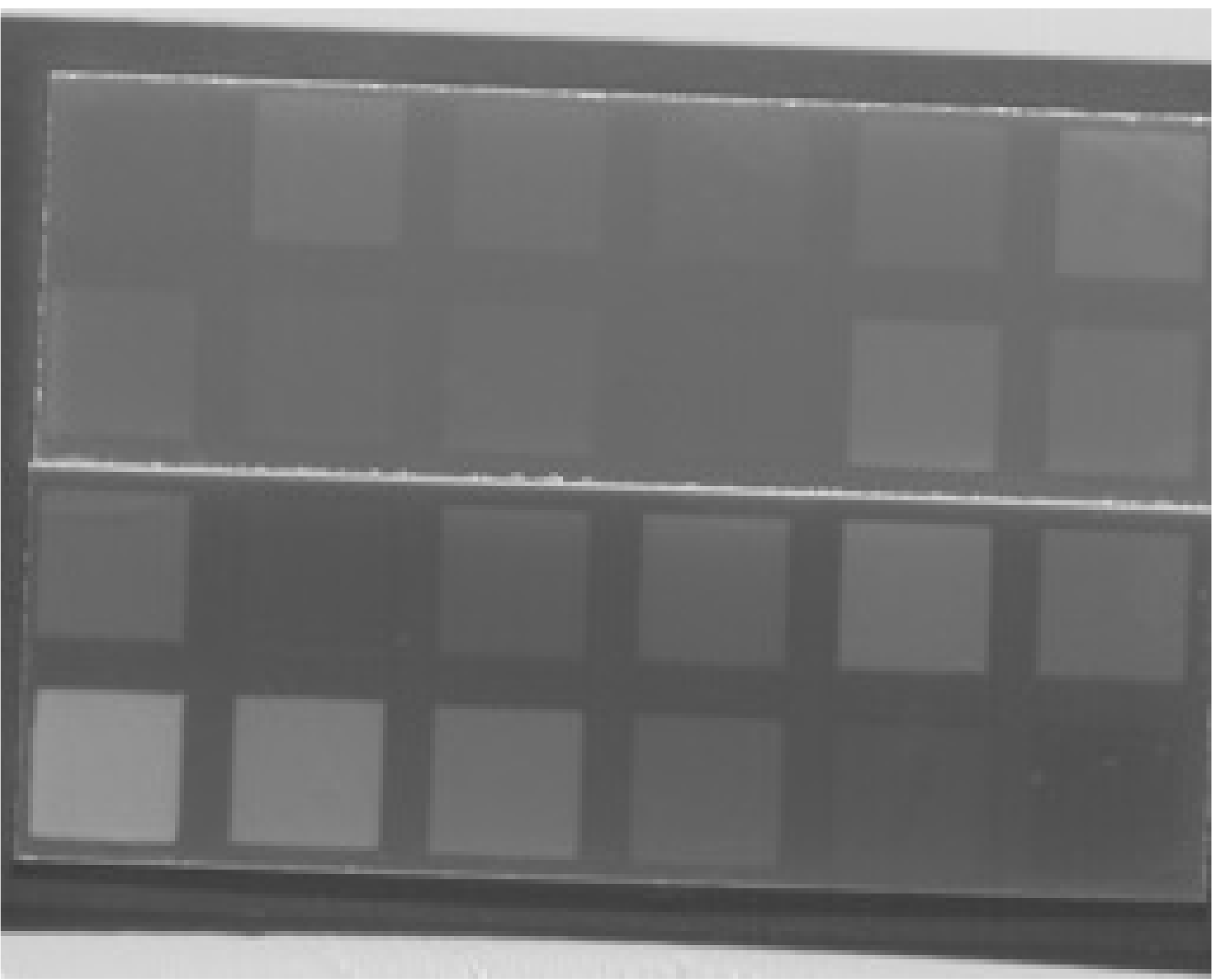}
\end{subfigure}%
\begin{subfigure}{0.06\textwidth}
\centering
\includegraphics[width=\textwidth]{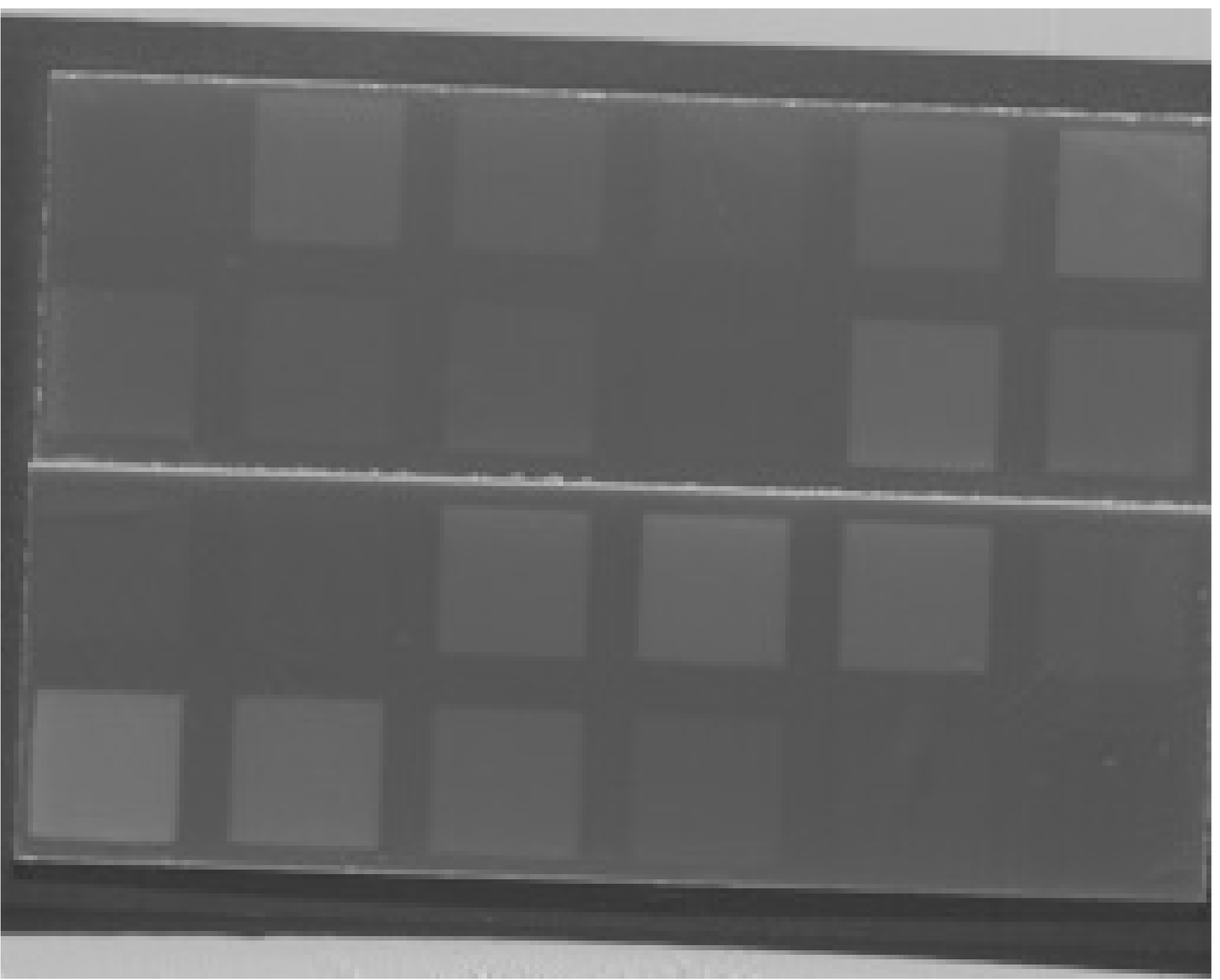}
\end{subfigure}%
\begin{subfigure}{0.06\textwidth}
\centering
\includegraphics[width=\textwidth]{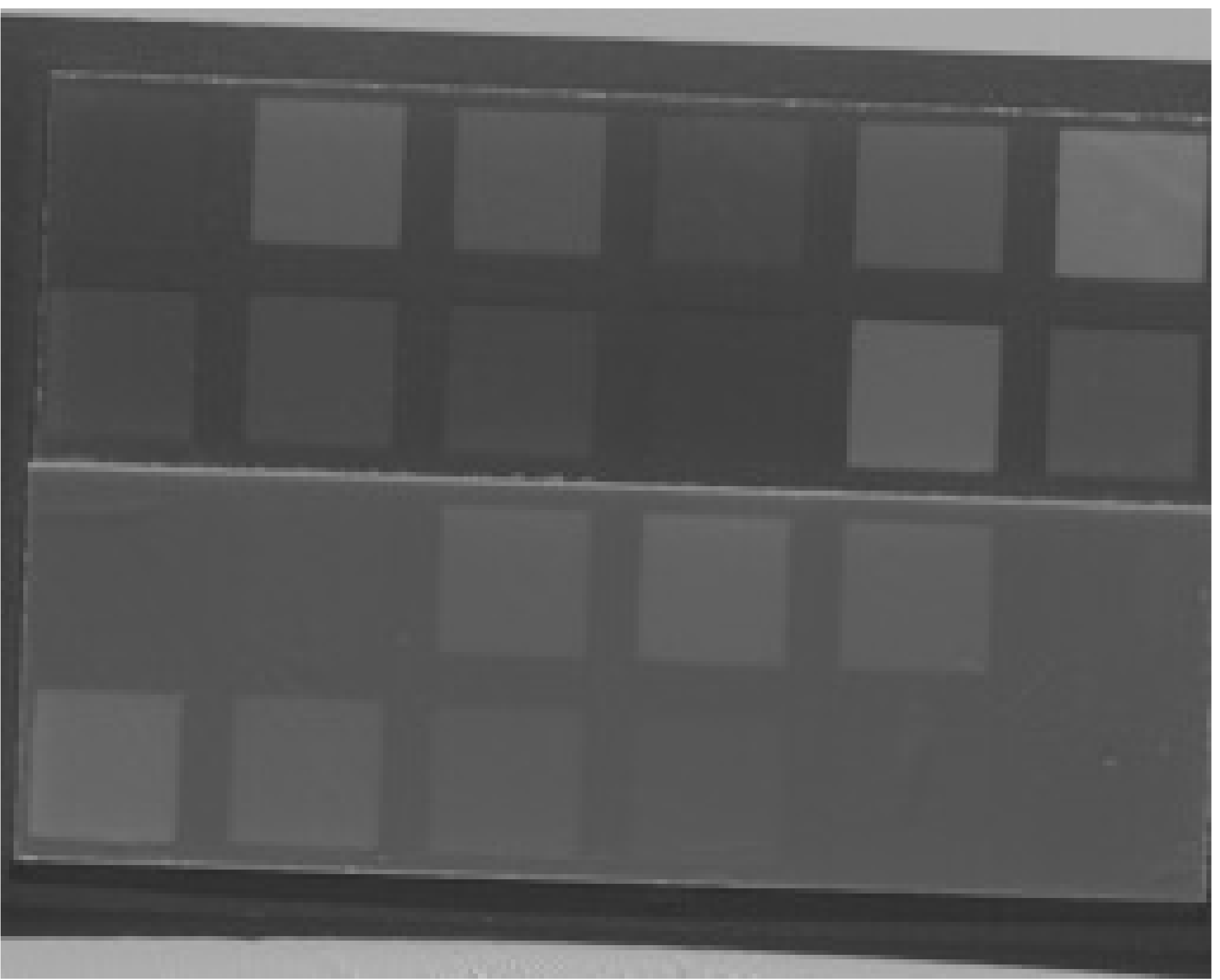}
\end{subfigure}%
\begin{subfigure}{0.06\textwidth}
\centering
\includegraphics[width=\textwidth]{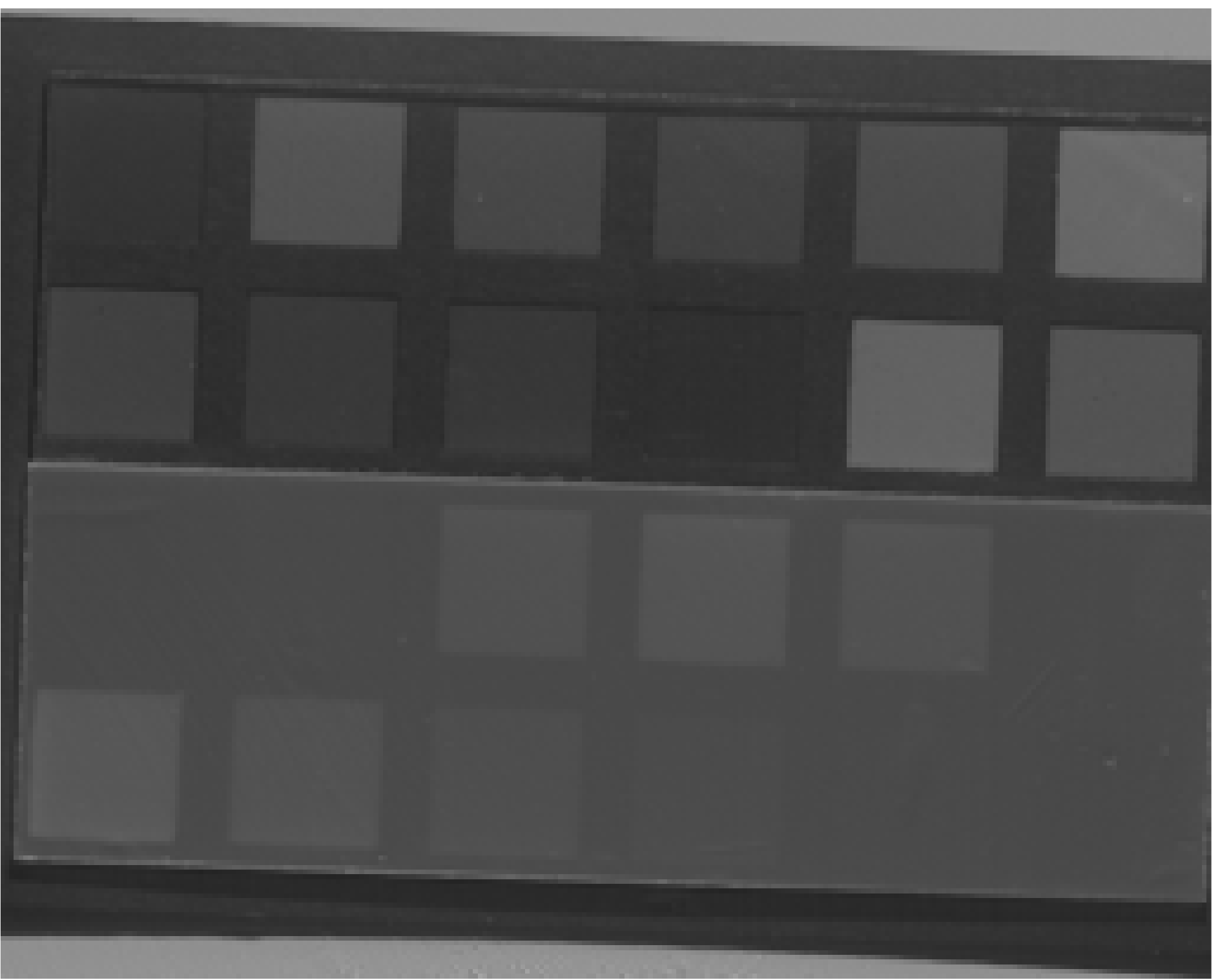}
\end{subfigure}%
\begin{subfigure}{0.06\textwidth}
\centering
\includegraphics[width=\textwidth]{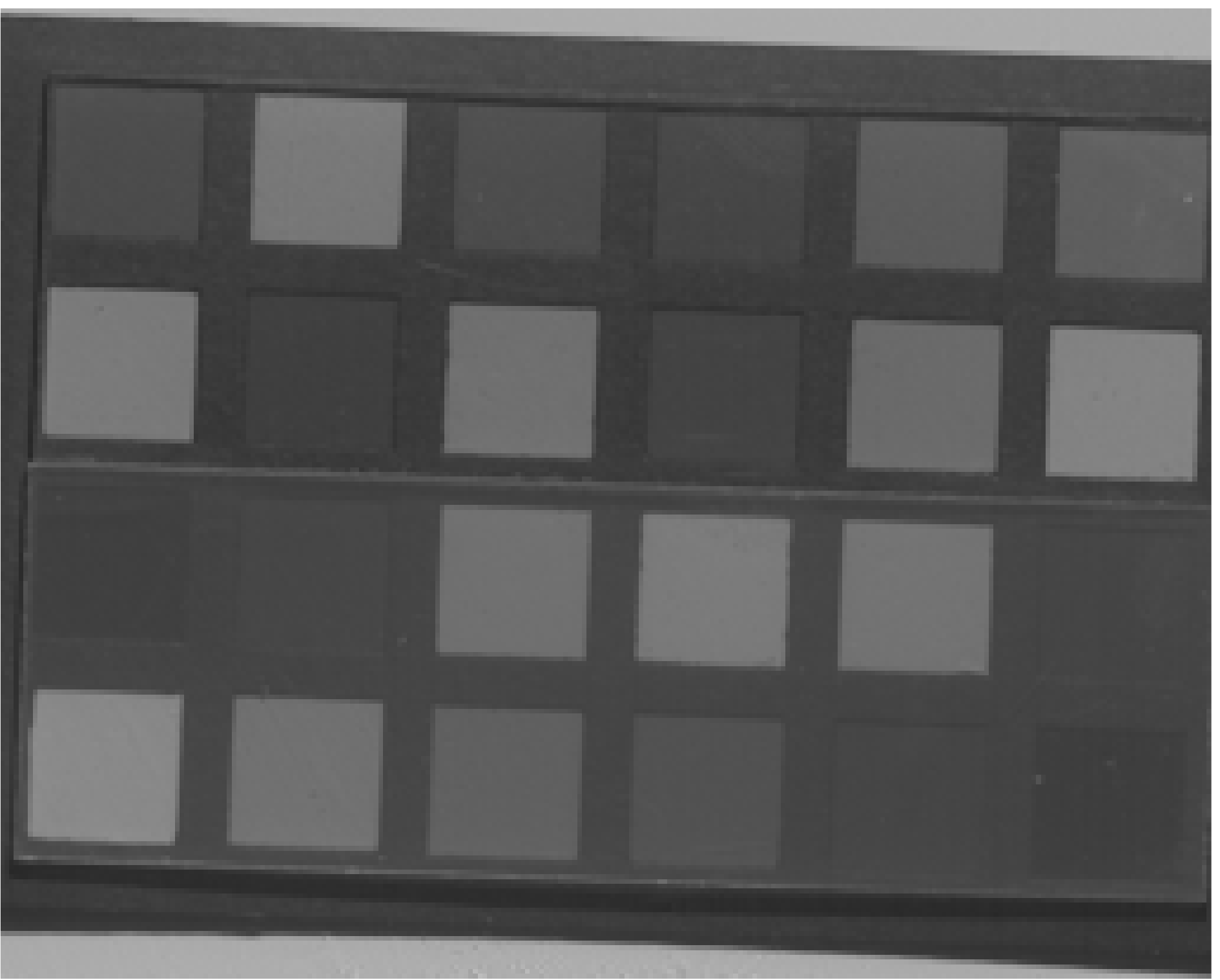}
\end{subfigure}%
\begin{subfigure}{0.06\textwidth}
\centering
\includegraphics[width=\textwidth]{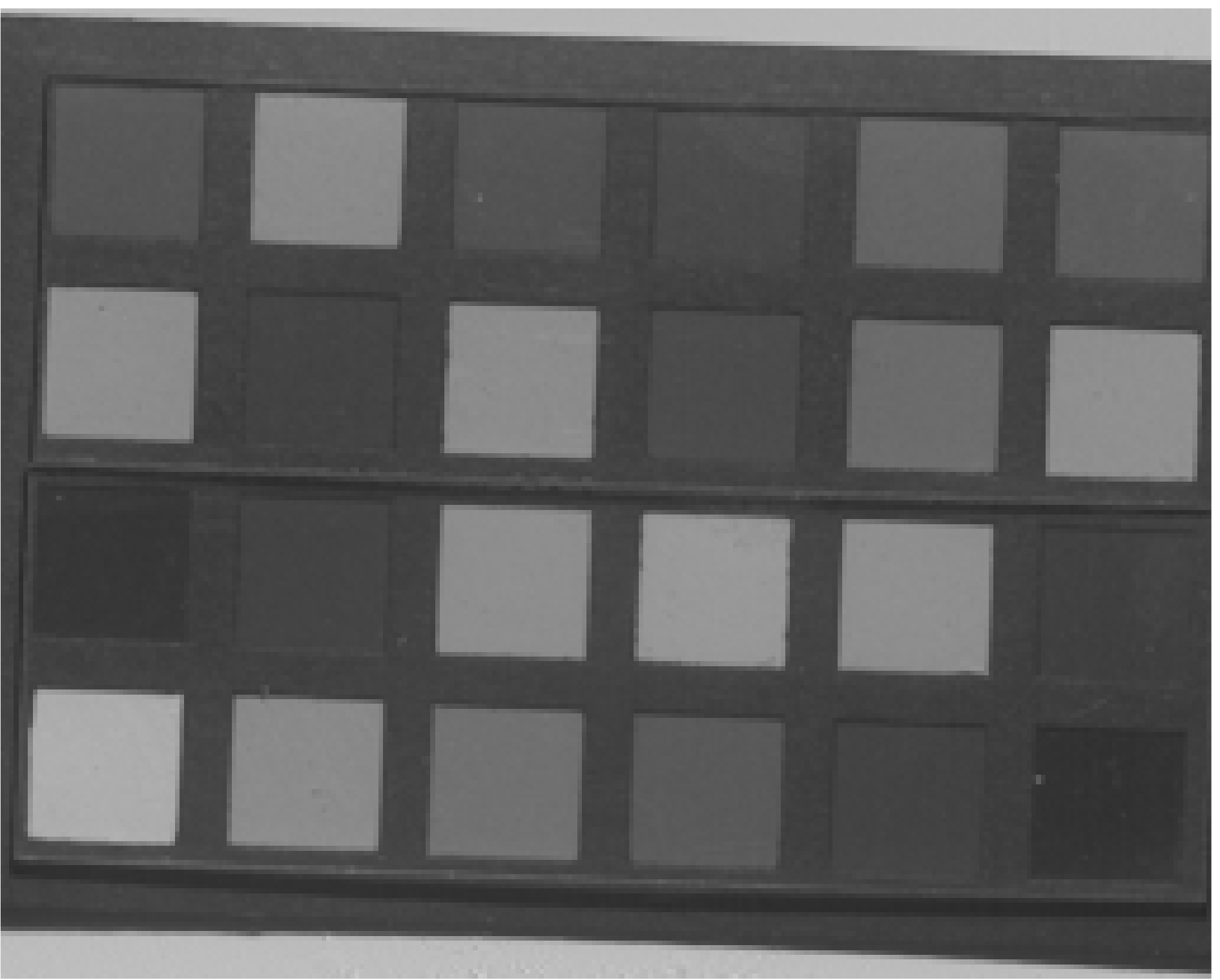}
\end{subfigure}%
\begin{subfigure}{0.06\textwidth}
\centering
\includegraphics[width=\textwidth]{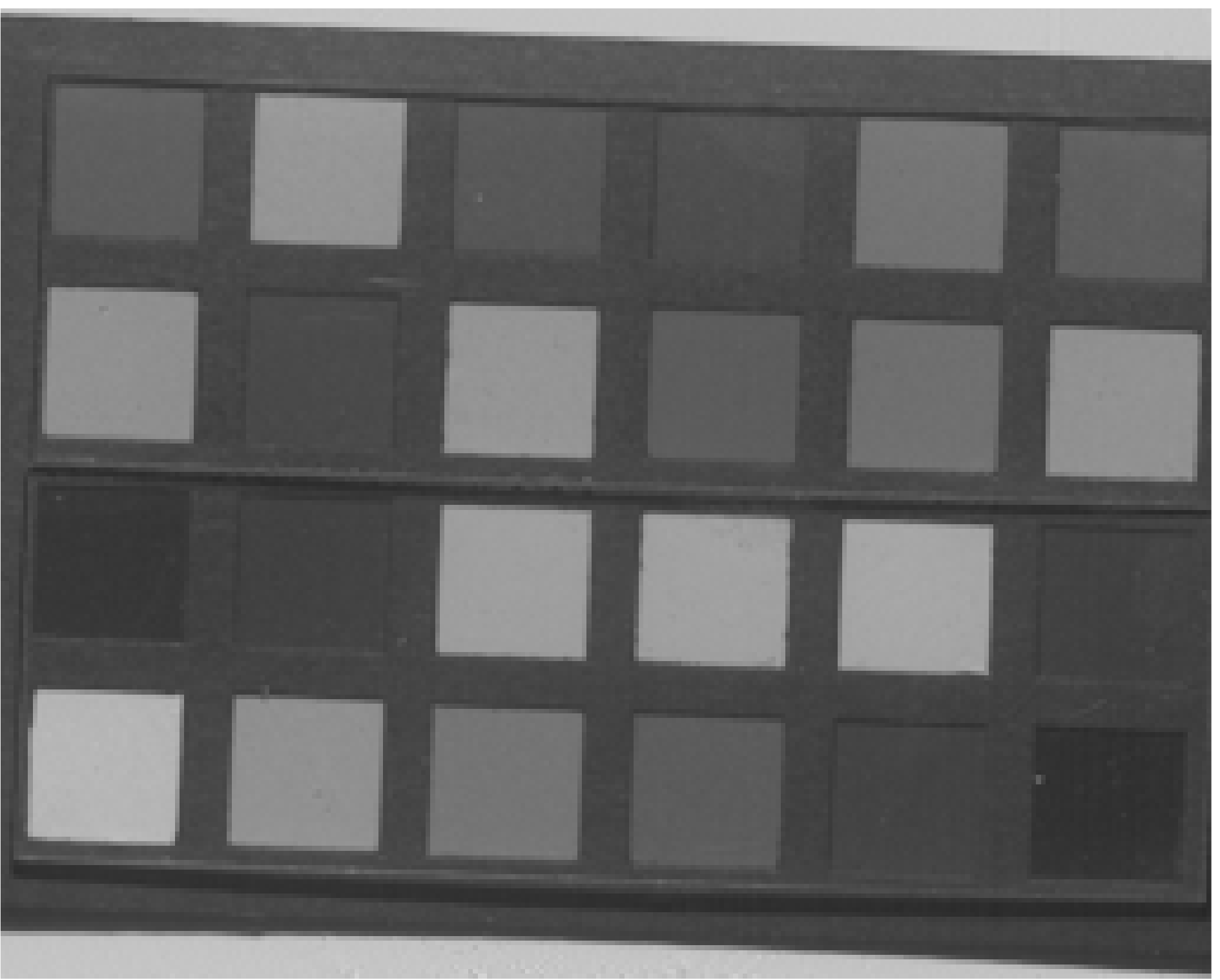}
\end{subfigure}%
\begin{subfigure}{0.06\textwidth}
\centering
\includegraphics[width=\textwidth]{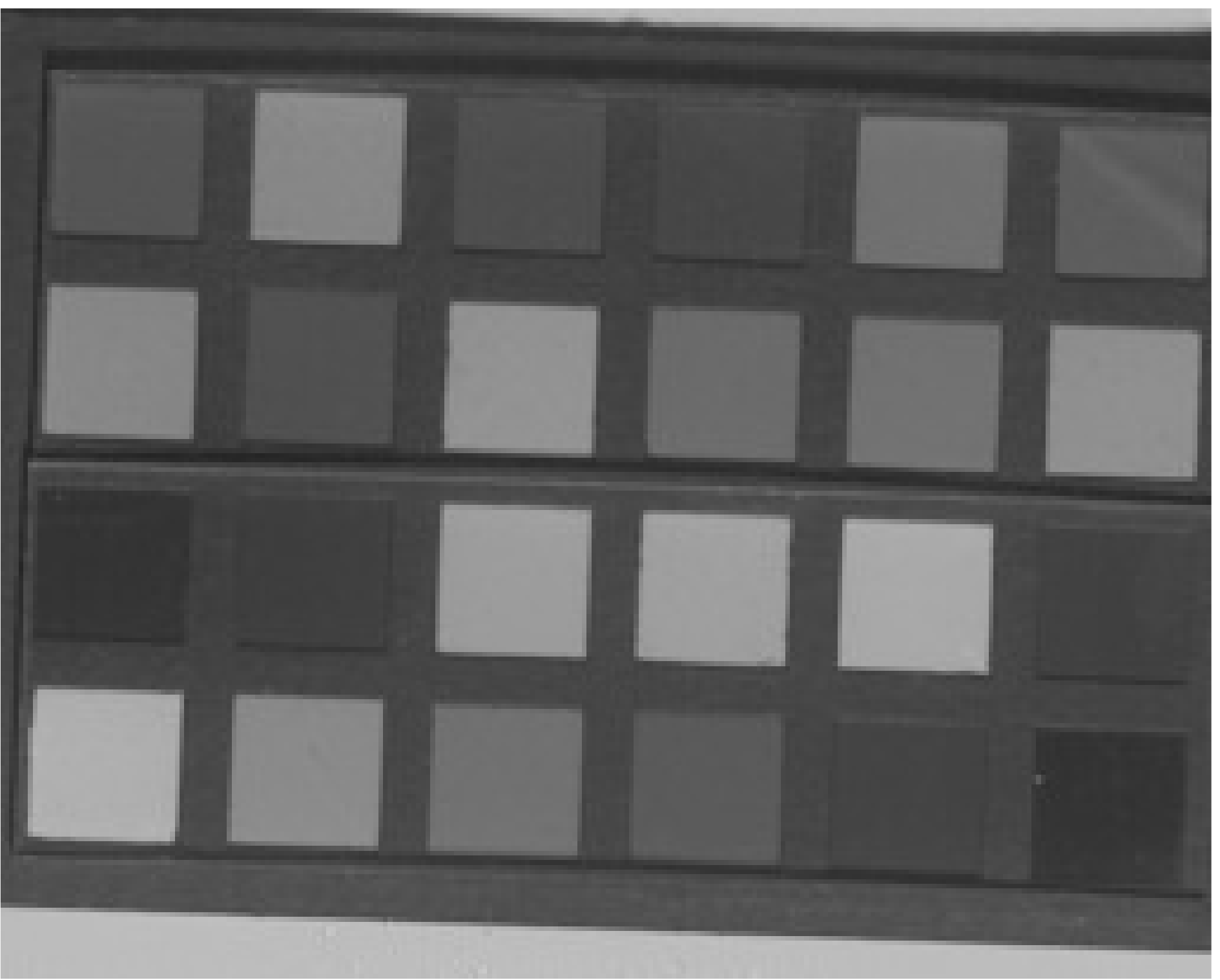}
\end{subfigure}%
\begin{subfigure}{0.06\textwidth}
\centering
\includegraphics[width=\textwidth]{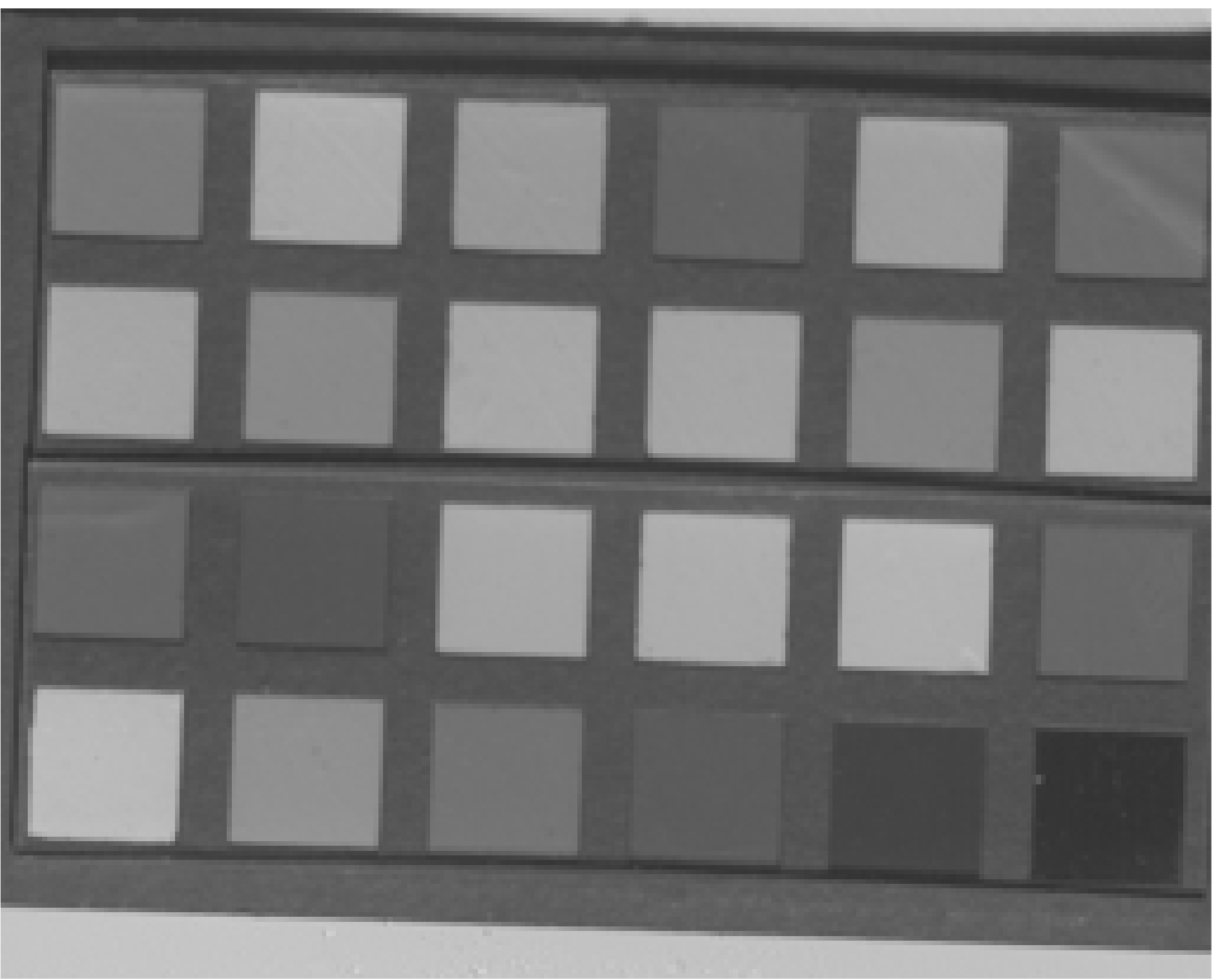}
\end{subfigure}%
\begin{subfigure}{0.06\textwidth}
\centering
\includegraphics[width=\textwidth]{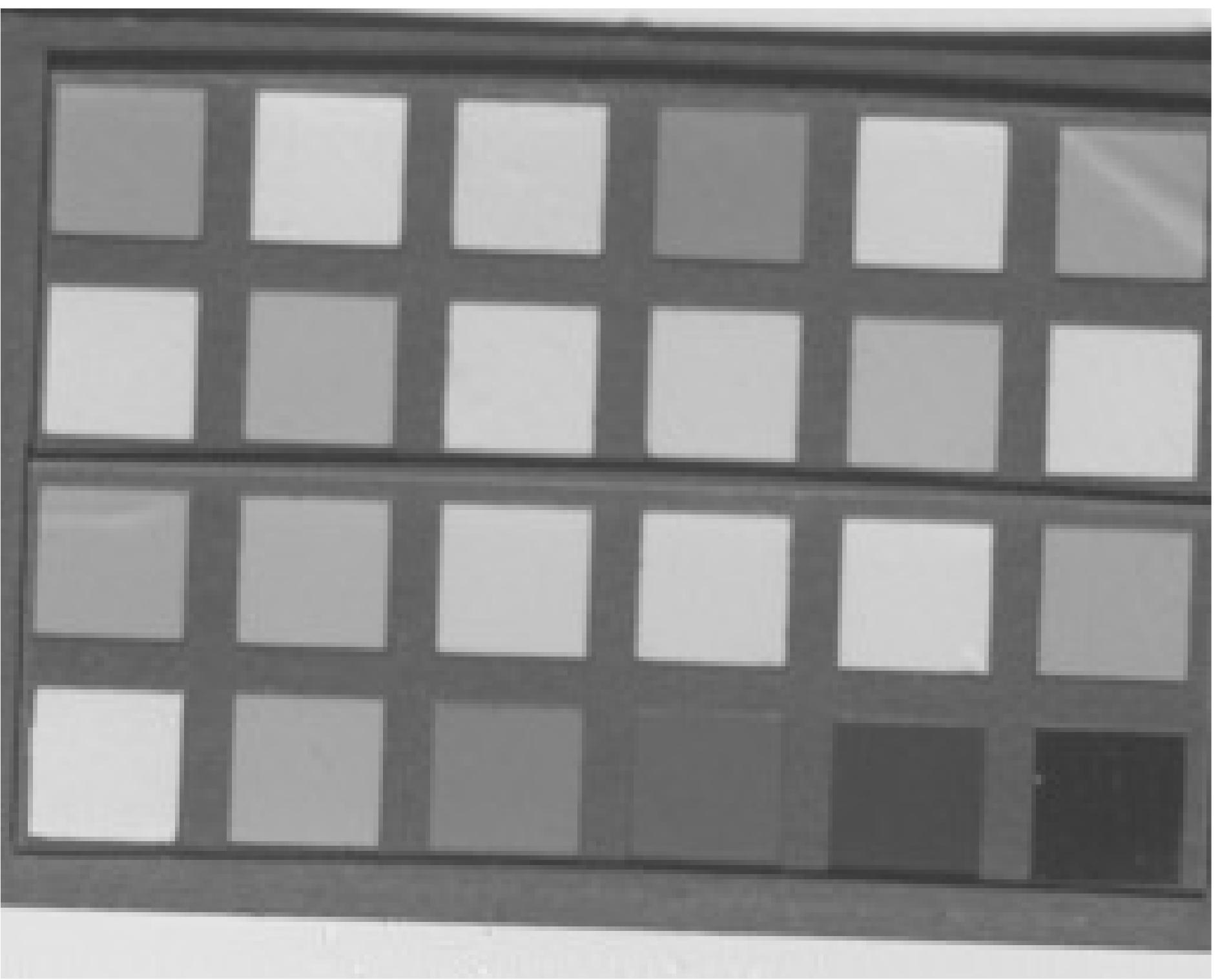}
\end{subfigure}%
\begin{subfigure}{0.06\textwidth}
\centering
\includegraphics[width=\textwidth]{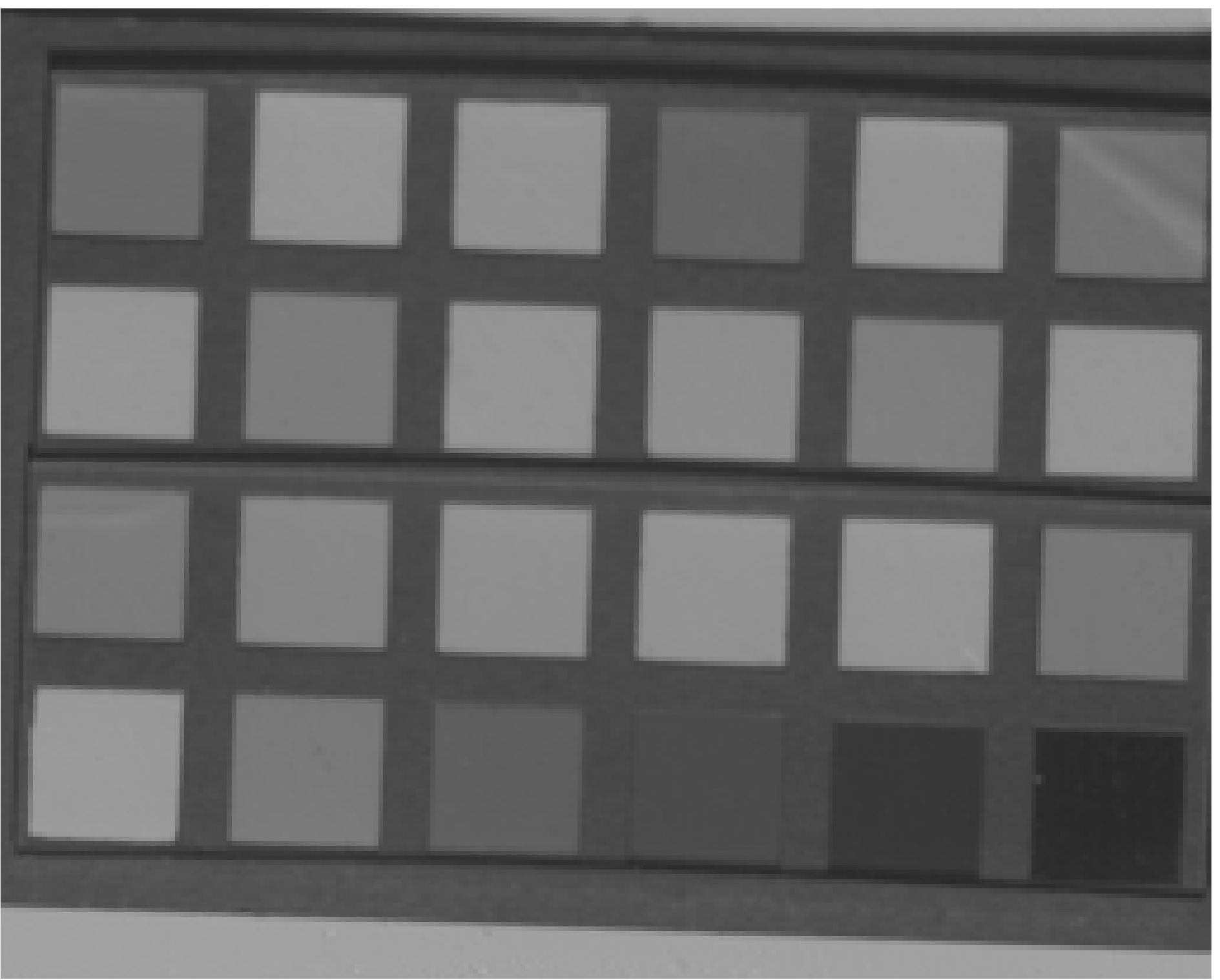}
\end{subfigure}%
\begin{subfigure}{0.06\textwidth}
\centering
\includegraphics[width=\textwidth]{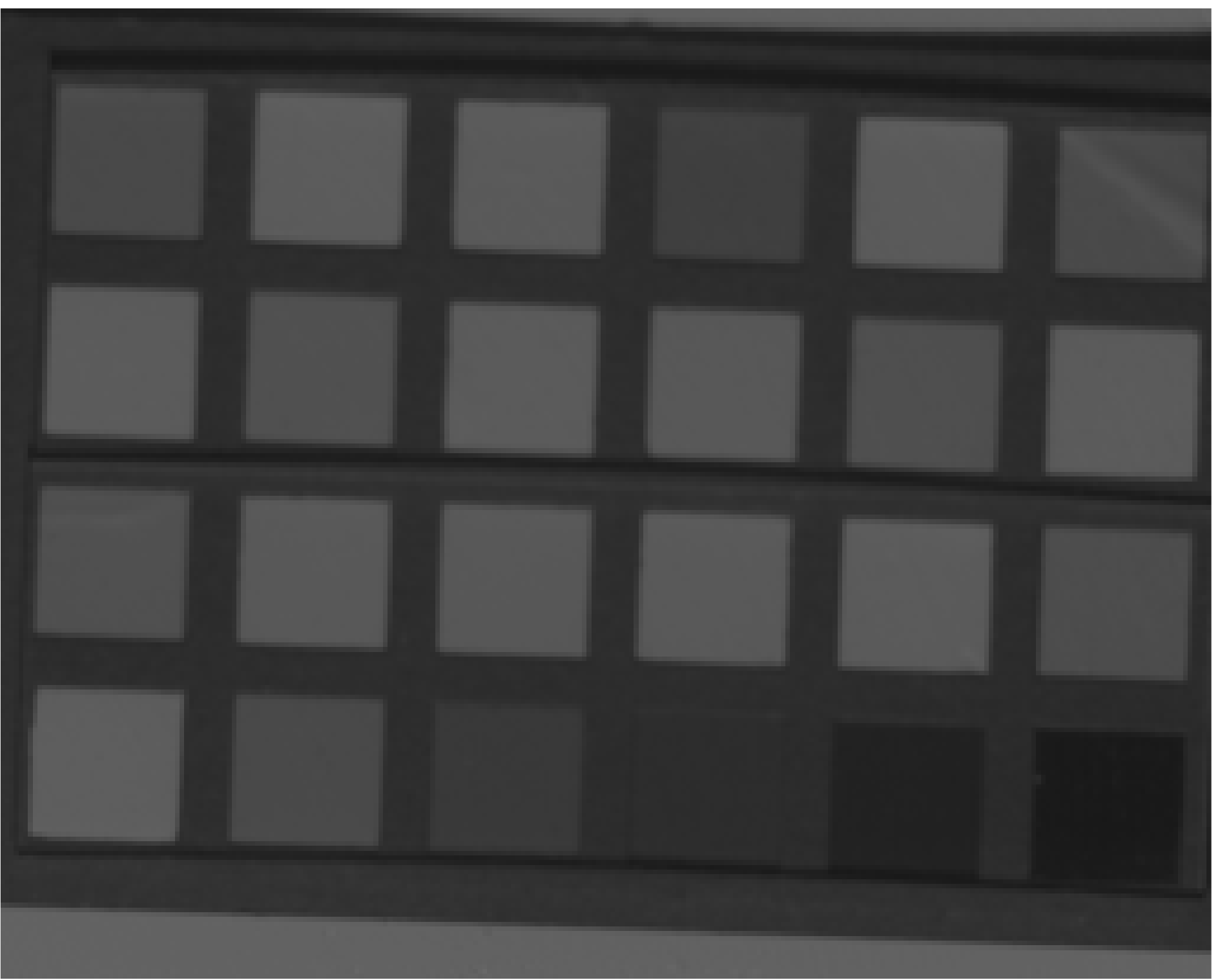}
\end{subfigure}\\%
\begin{subfigure}{0.06\textwidth}
\captionsetup{justification=raggedright,font=scriptsize}
\caption*{437 to 463nm}
\end{subfigure}%
\begin{subfigure}{0.06\textwidth}
\centering
\includegraphics[width=\textwidth]{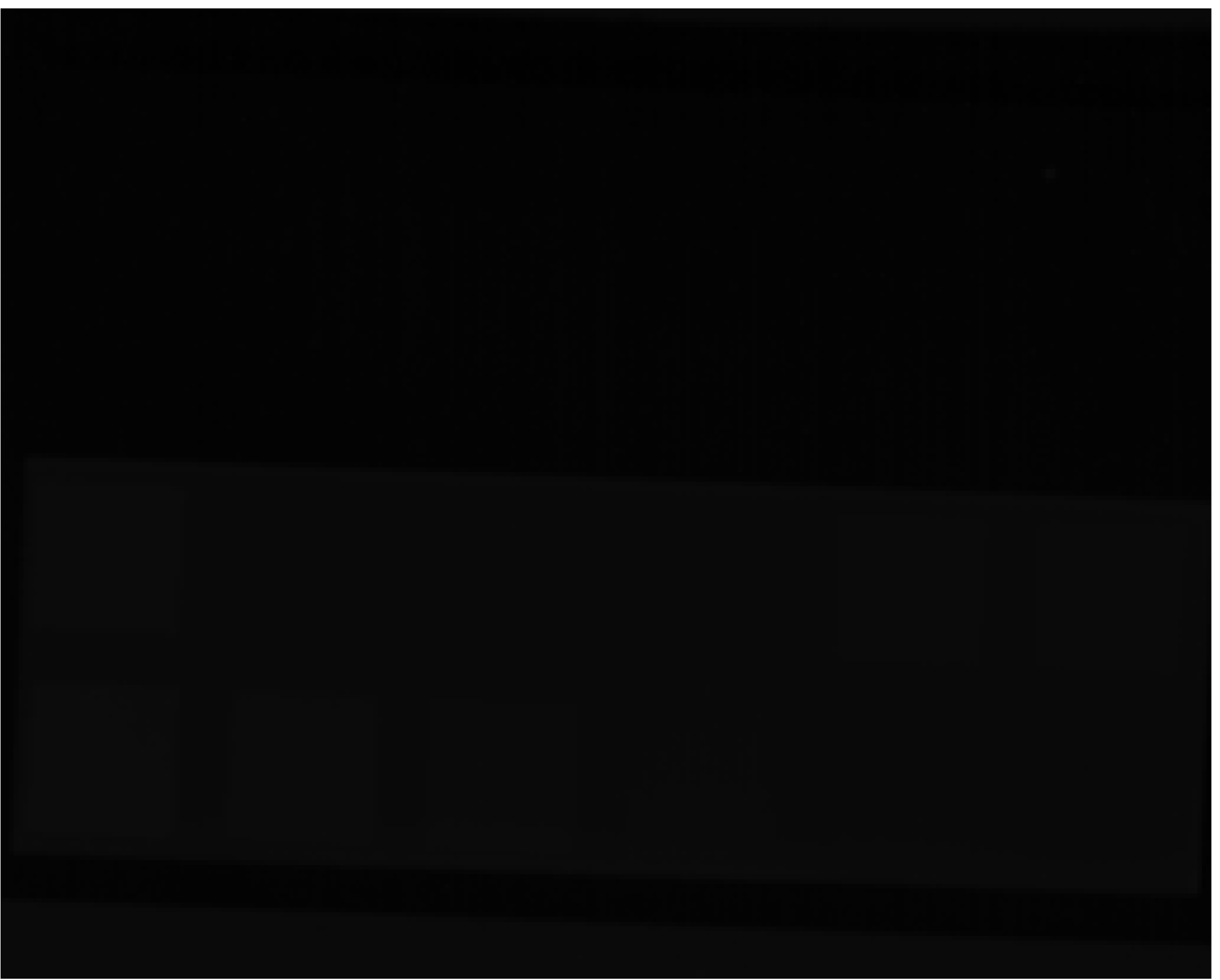}
\end{subfigure}%
\begin{subfigure}{0.06\textwidth}
\centering
\includegraphics[width=\textwidth]{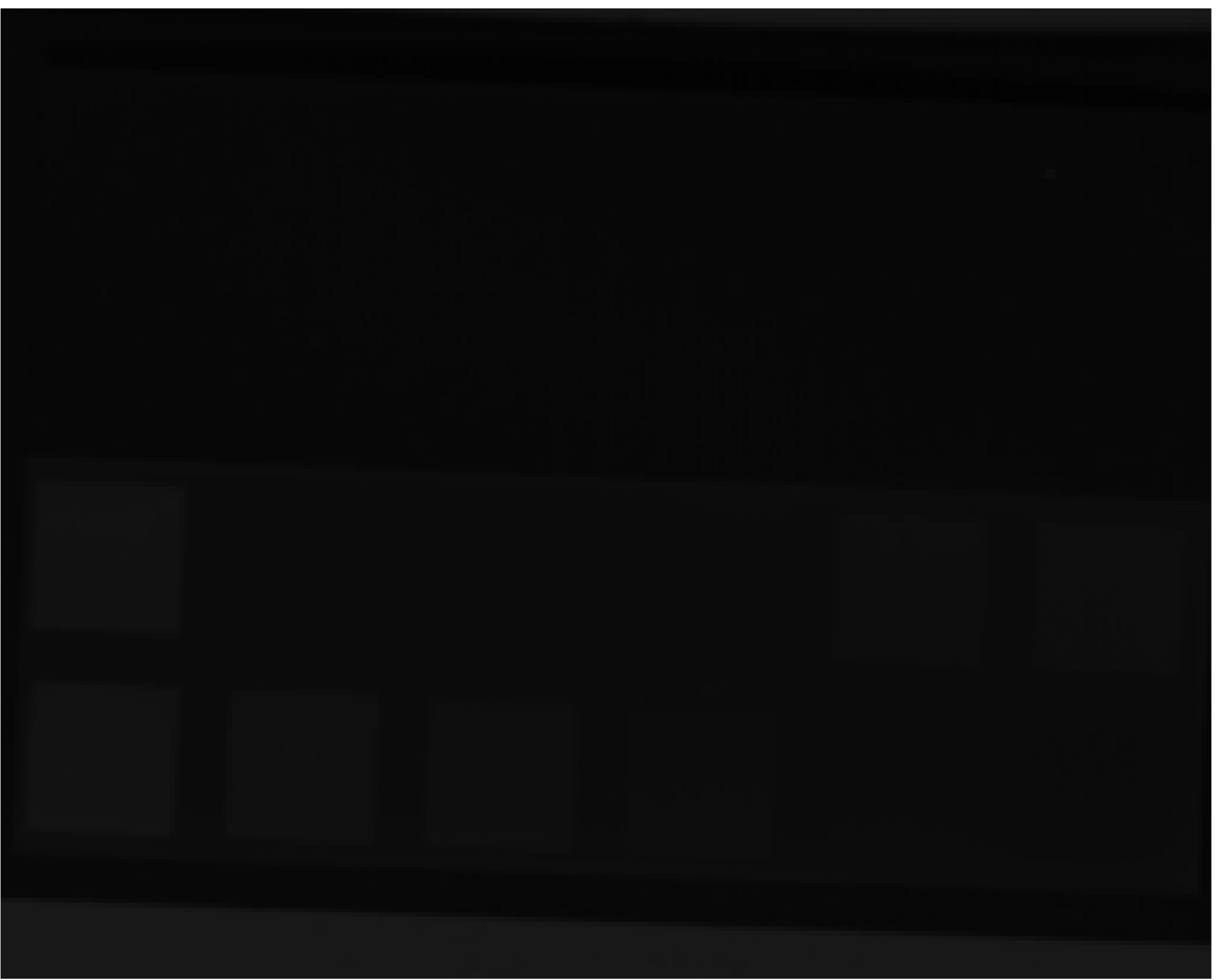}
\end{subfigure}%
\begin{subfigure}{0.06\textwidth}
\centering
\includegraphics[width=\textwidth]{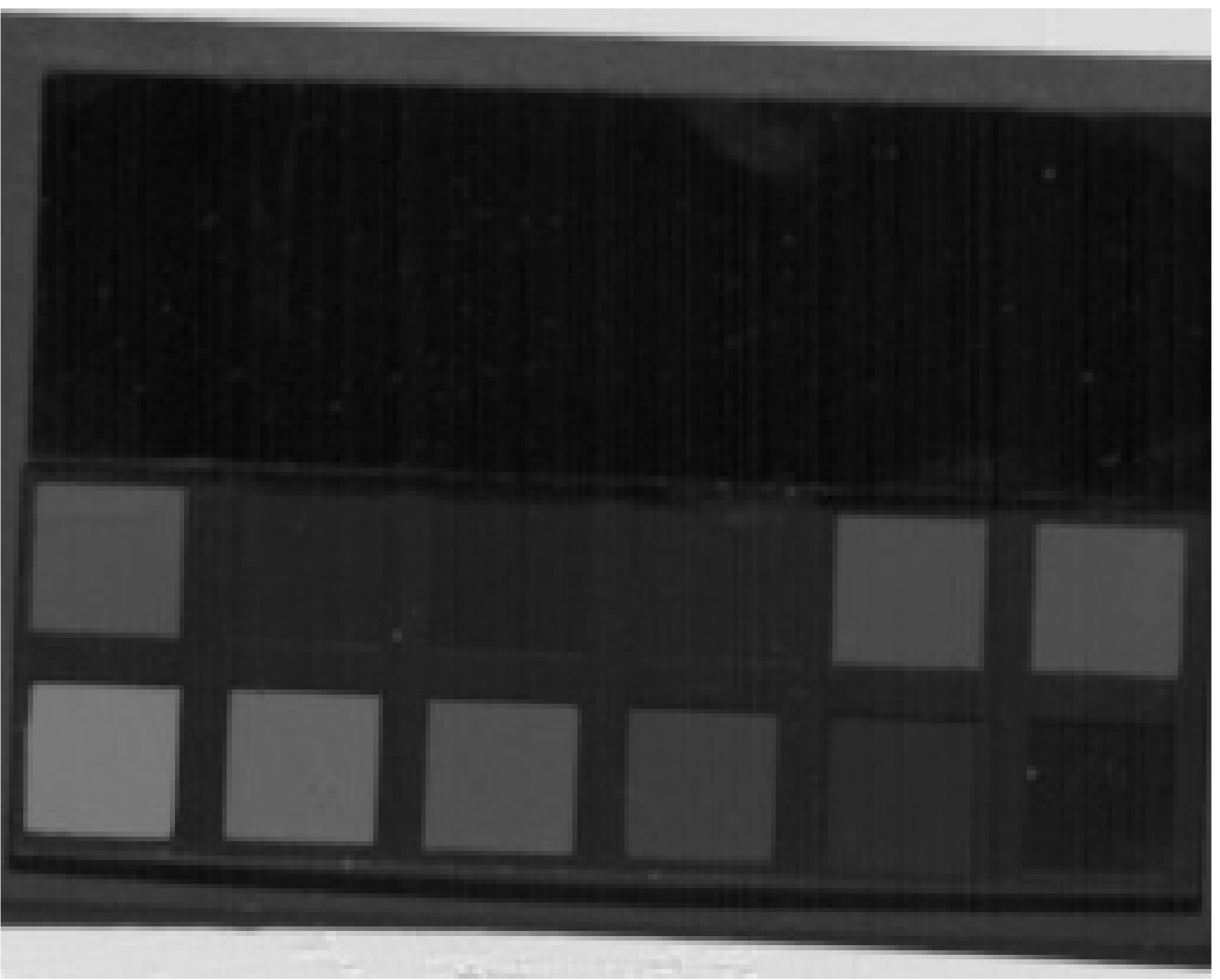}
\end{subfigure}%
\begin{subfigure}{0.06\textwidth}
\centering
\includegraphics[width=\textwidth]{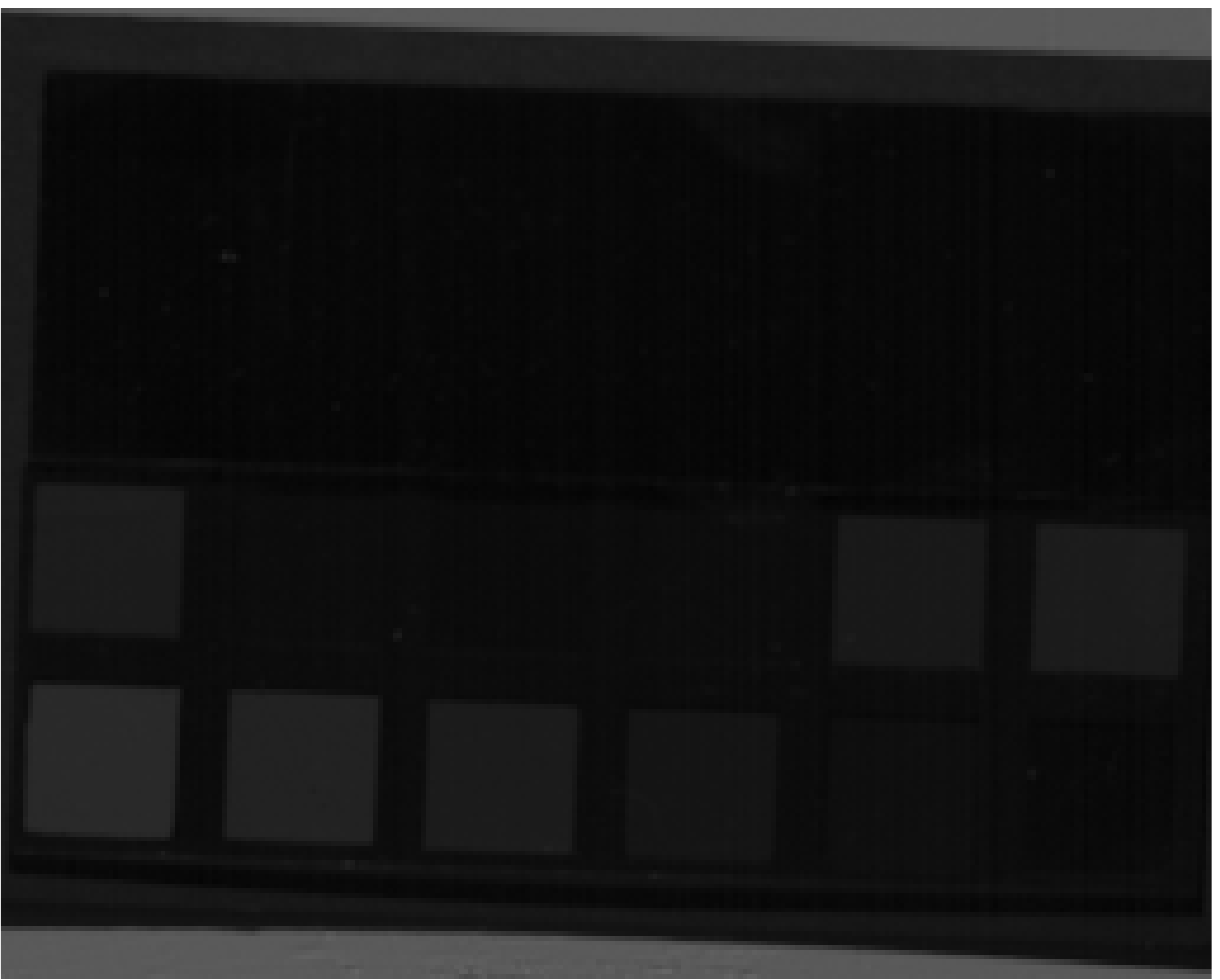}
\end{subfigure}%
\begin{subfigure}{0.06\textwidth}
\centering
\includegraphics[width=\textwidth]{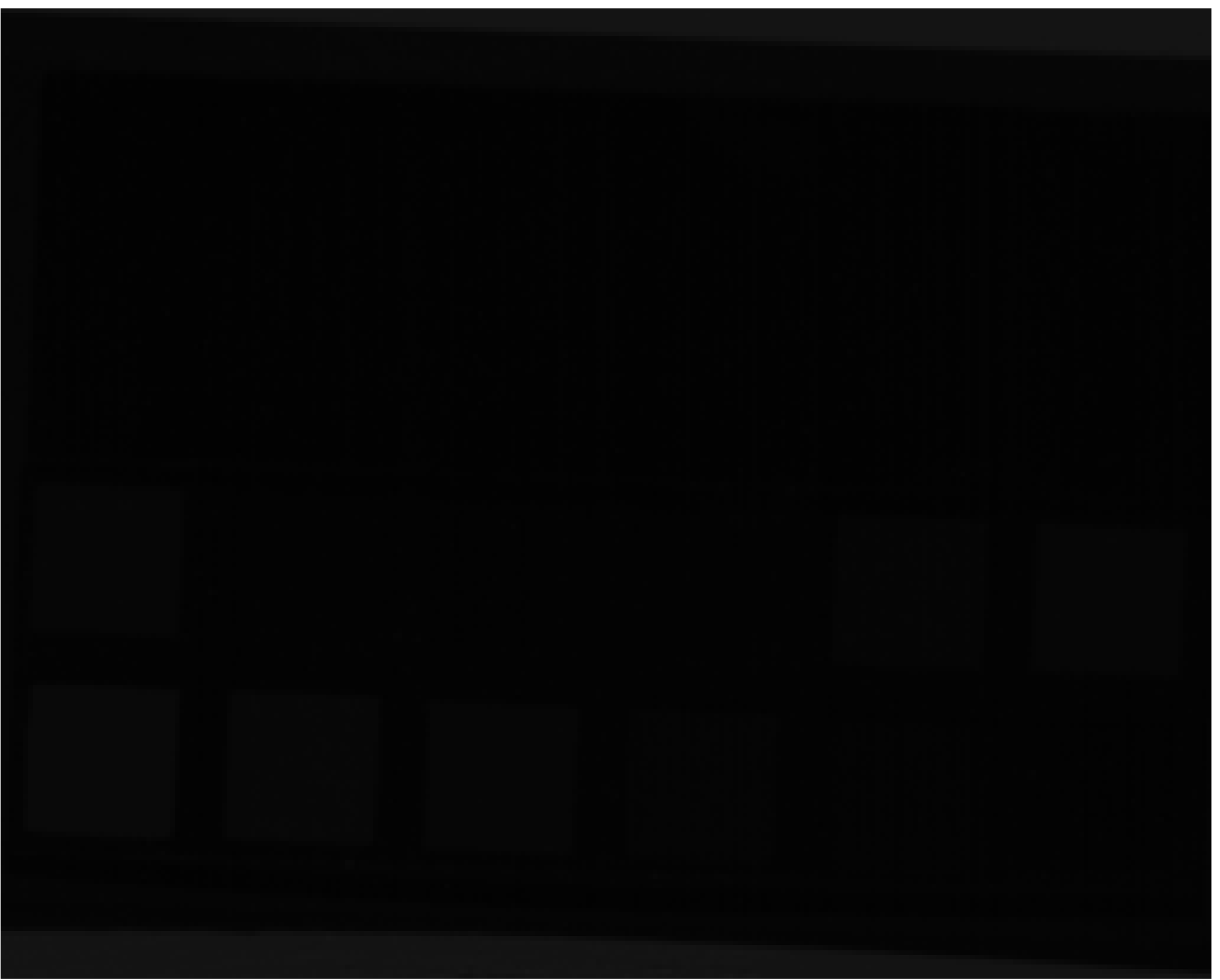}
\end{subfigure}%
\begin{subfigure}{0.06\textwidth}
\centering
\includegraphics[width=\textwidth]{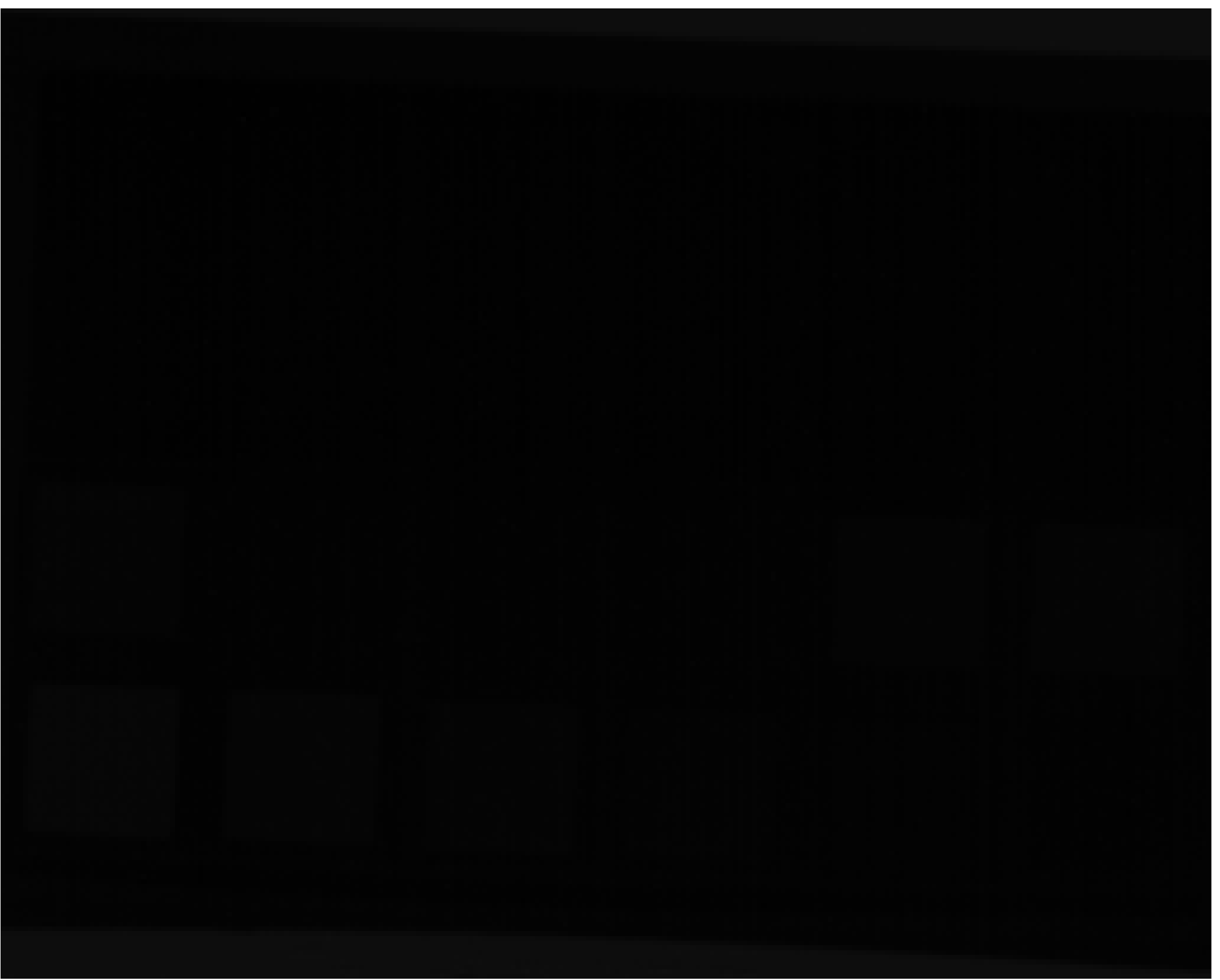}
\end{subfigure}%
\begin{subfigure}{0.06\textwidth}
\centering
\includegraphics[width=\textwidth]{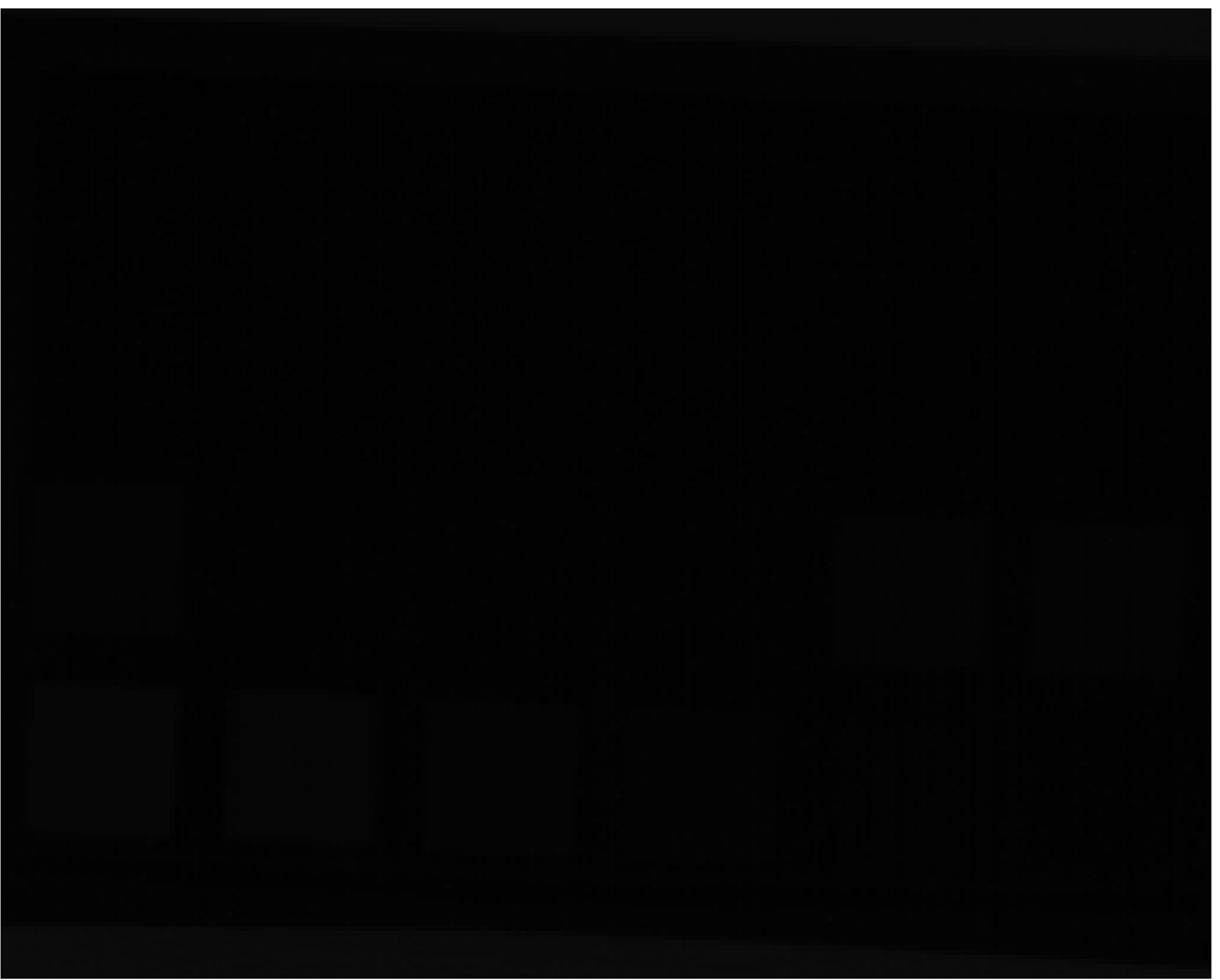}
\end{subfigure}%
\begin{subfigure}{0.06\textwidth}
\centering
\includegraphics[width=\textwidth]{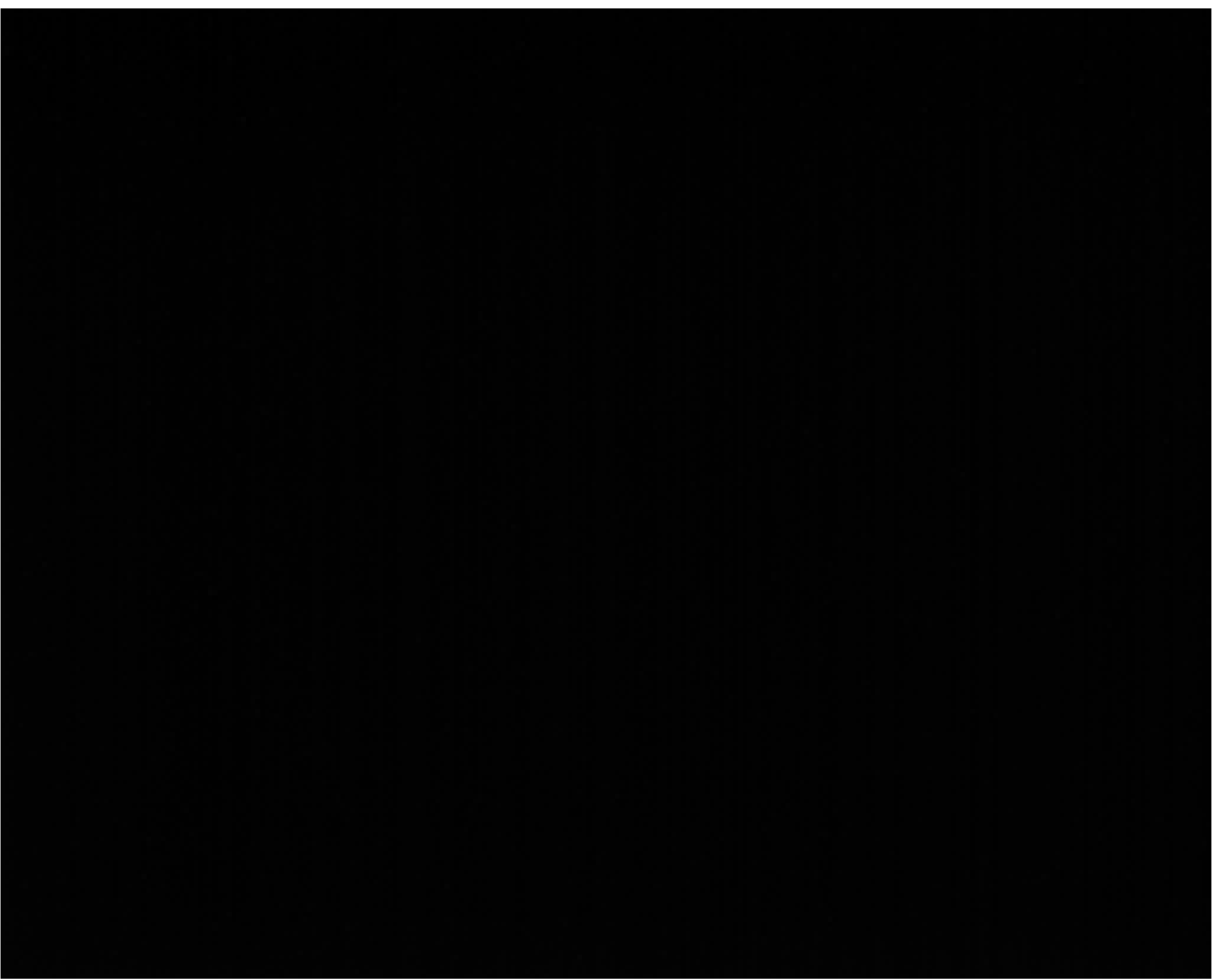}
\end{subfigure}%
\begin{subfigure}{0.06\textwidth}
\centering
\includegraphics[width=\textwidth]{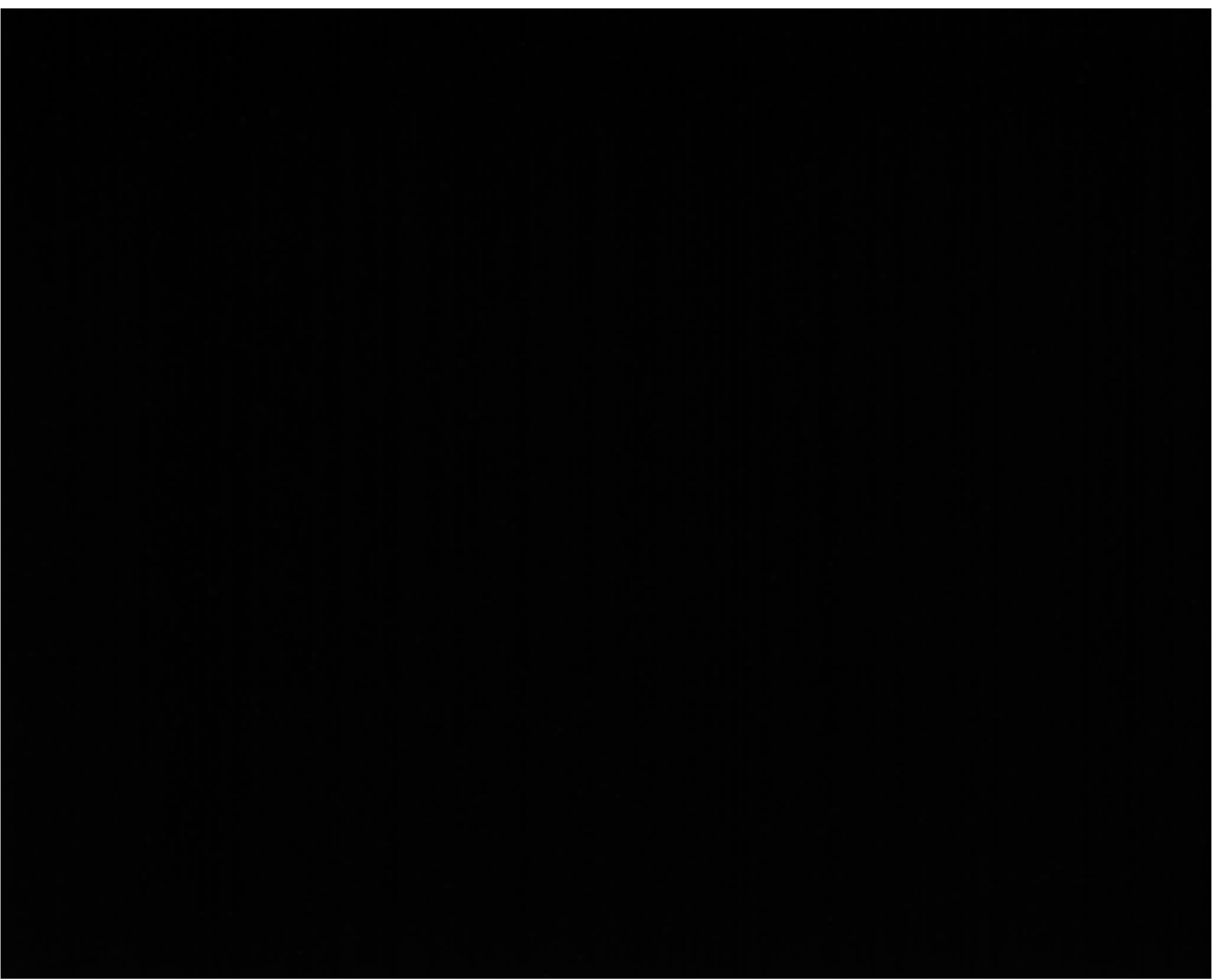}
\end{subfigure}%
\begin{subfigure}{0.06\textwidth}
\centering
\includegraphics[width=\textwidth]{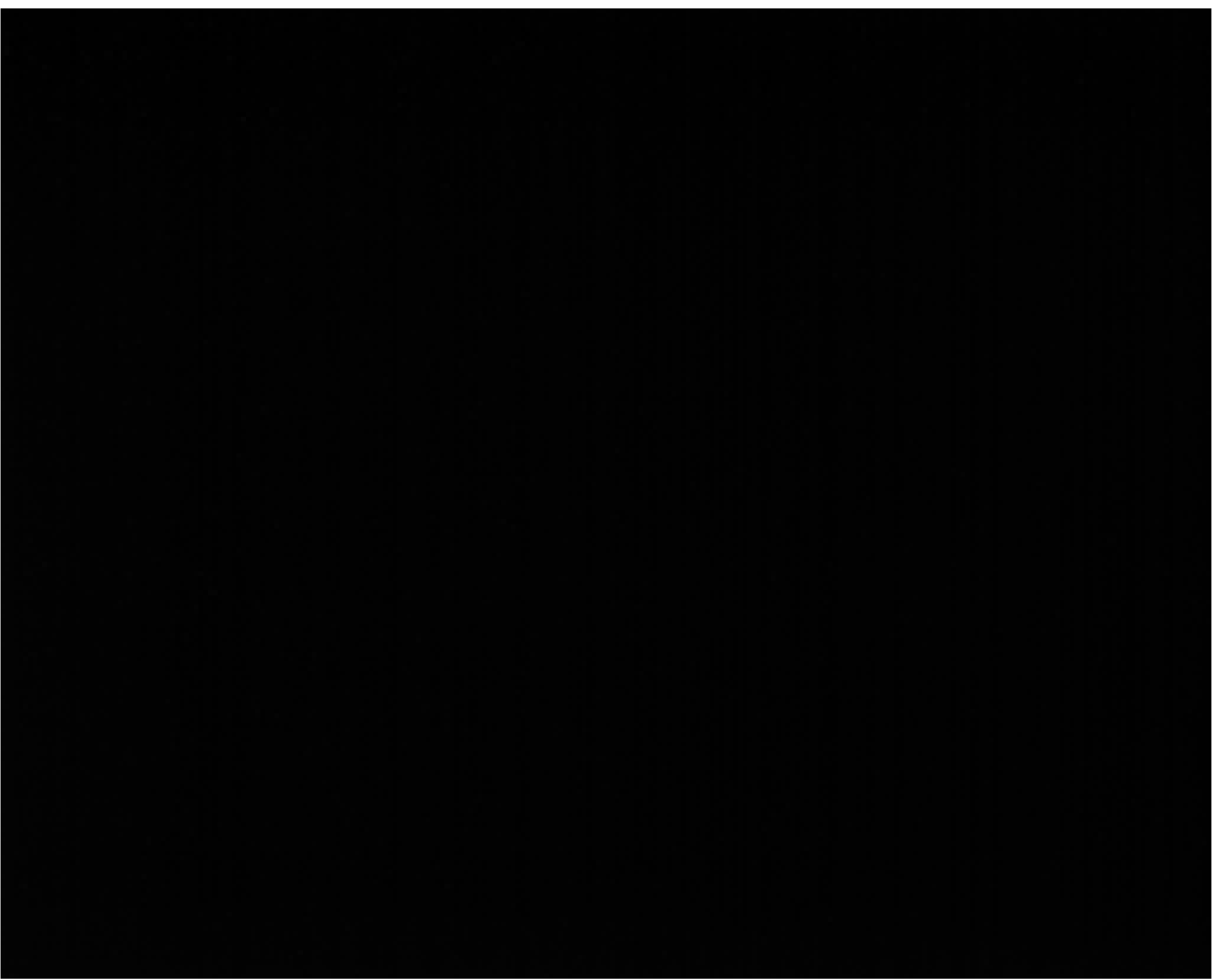}
\end{subfigure}%
\begin{subfigure}{0.06\textwidth}
\centering
\includegraphics[width=\textwidth]{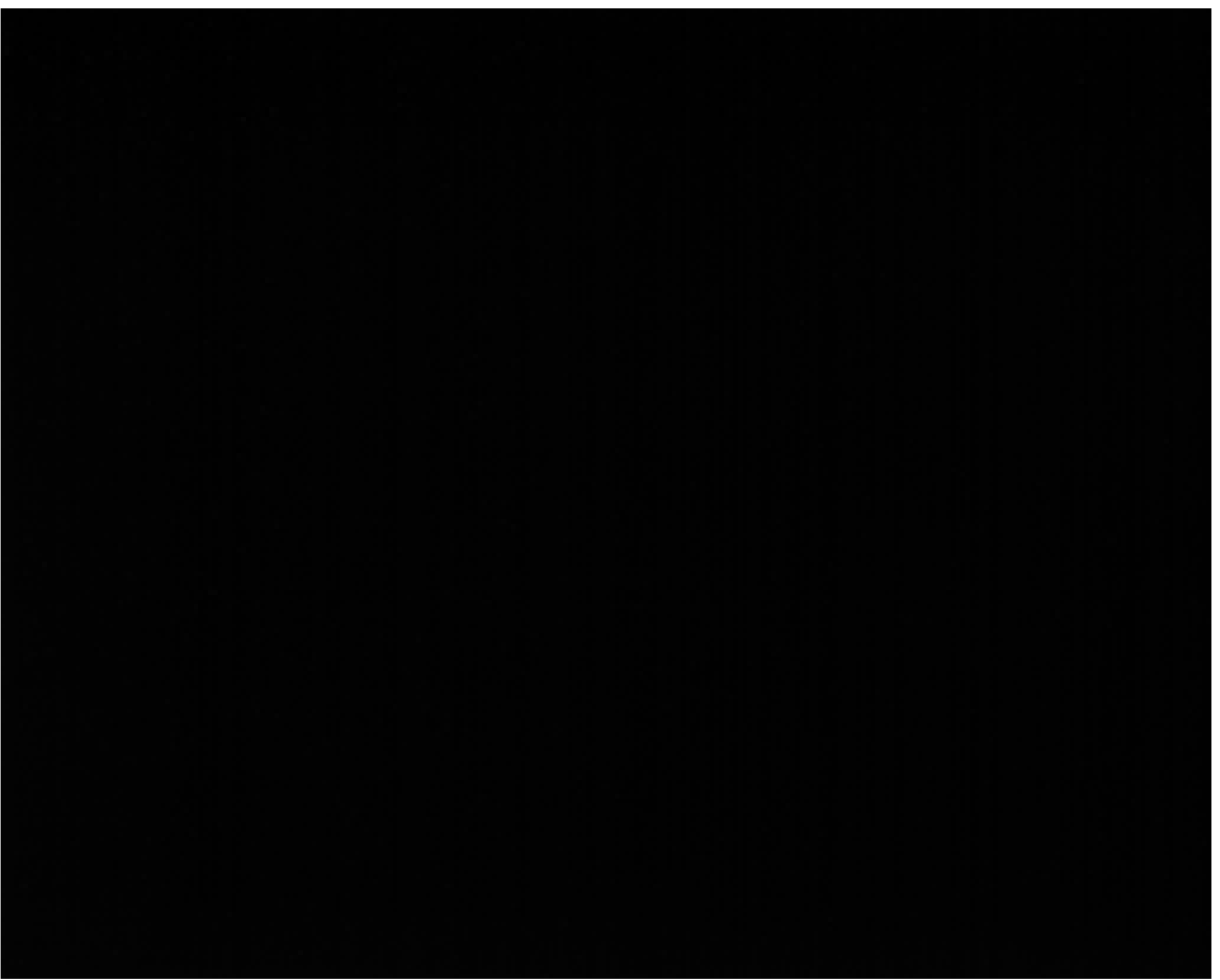}
\end{subfigure}%
\begin{subfigure}{0.06\textwidth}
\centering
\includegraphics[width=\textwidth]{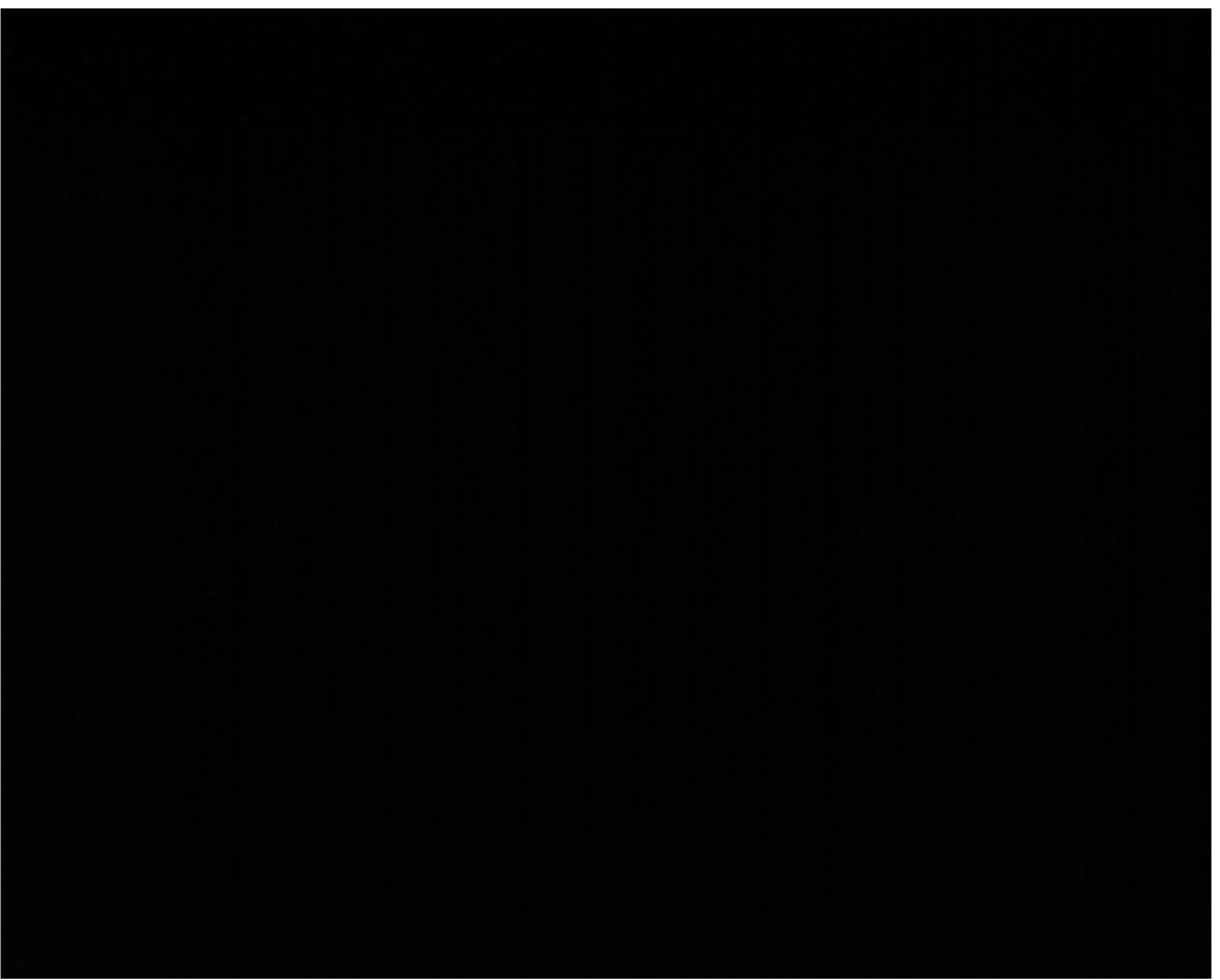}
\end{subfigure}%
\begin{subfigure}{0.06\textwidth}
\centering
\includegraphics[width=\textwidth]{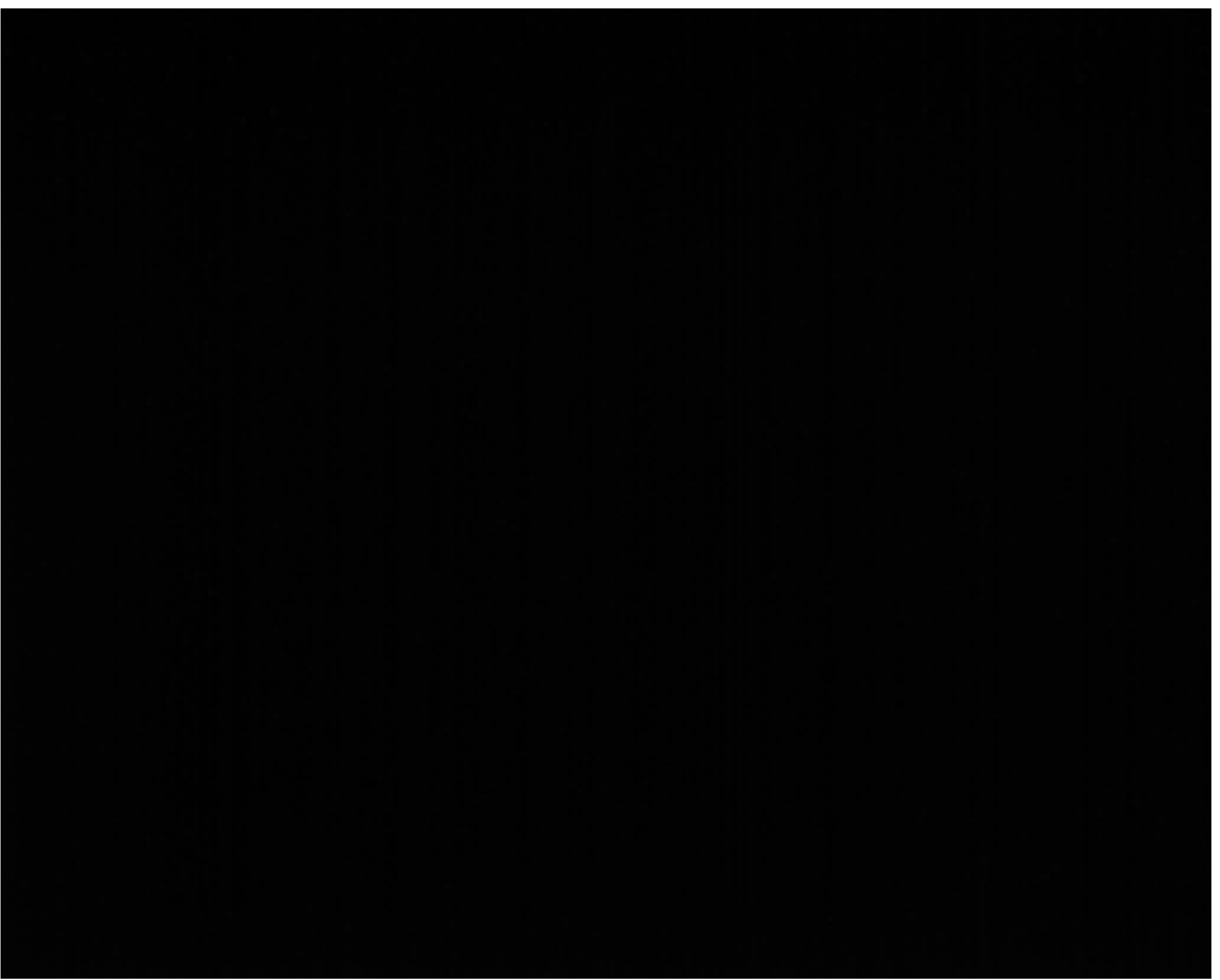}
\end{subfigure}%
\begin{subfigure}{0.06\textwidth}
\centering
\includegraphics[width=\textwidth]{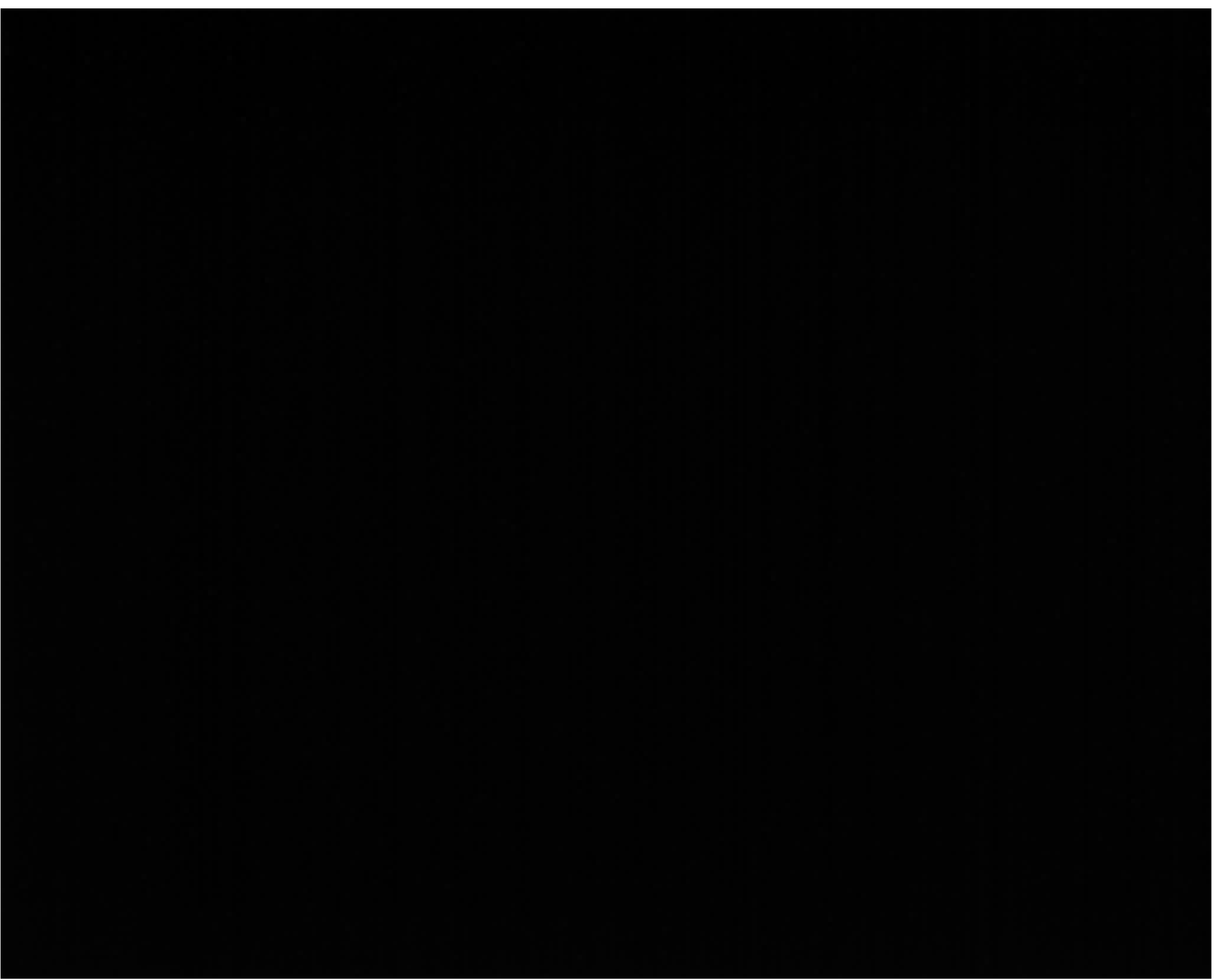}
\end{subfigure}\\%
\begin{subfigure}{0.06\textwidth}
\captionsetup{justification=raggedright,font=scriptsize}
\caption*{487 to 513nm}
\end{subfigure}%
\begin{subfigure}{0.06\textwidth}
\centering
\includegraphics[width=\textwidth]{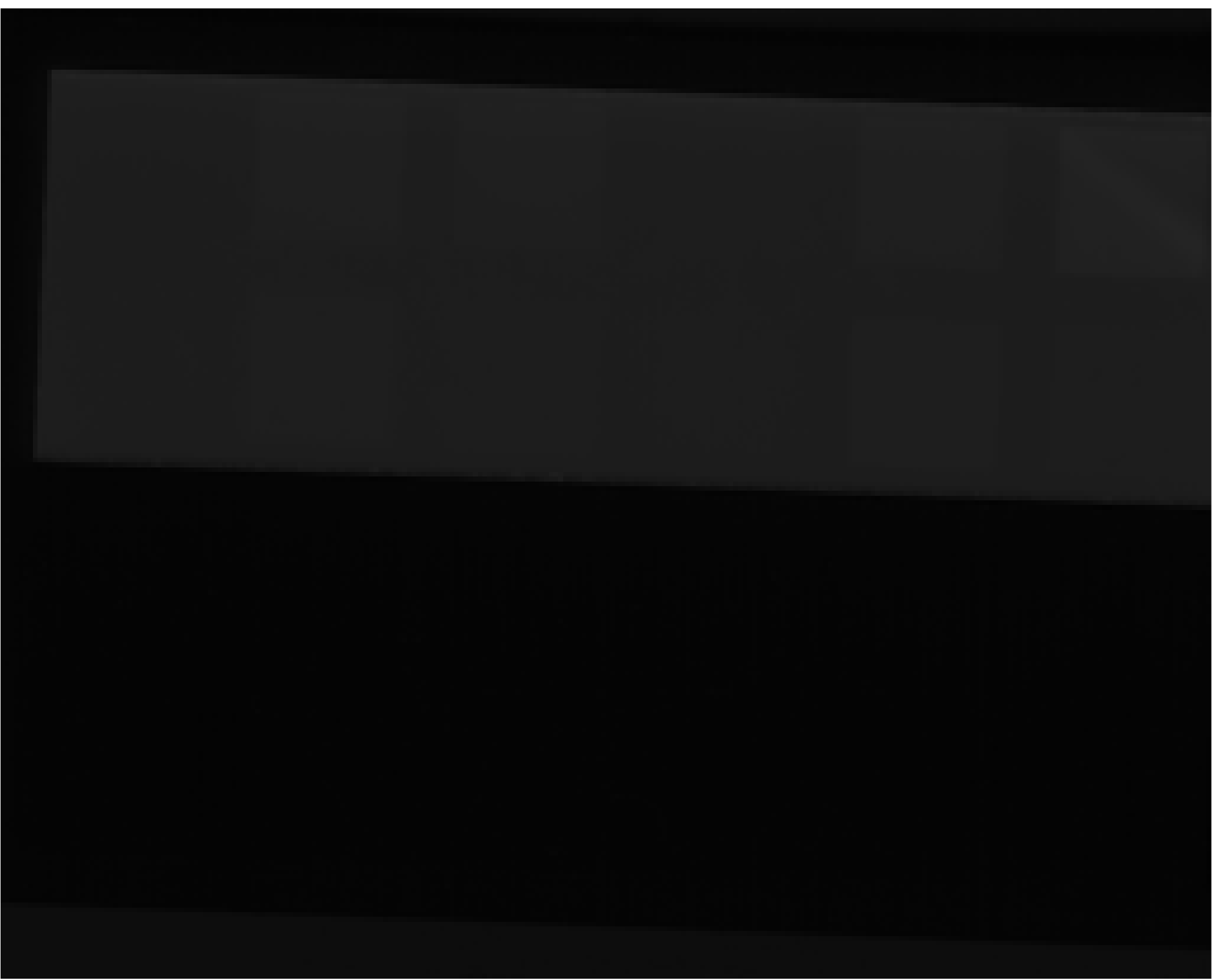}
\end{subfigure}%
\begin{subfigure}{0.06\textwidth}
\centering
\includegraphics[width=\textwidth]{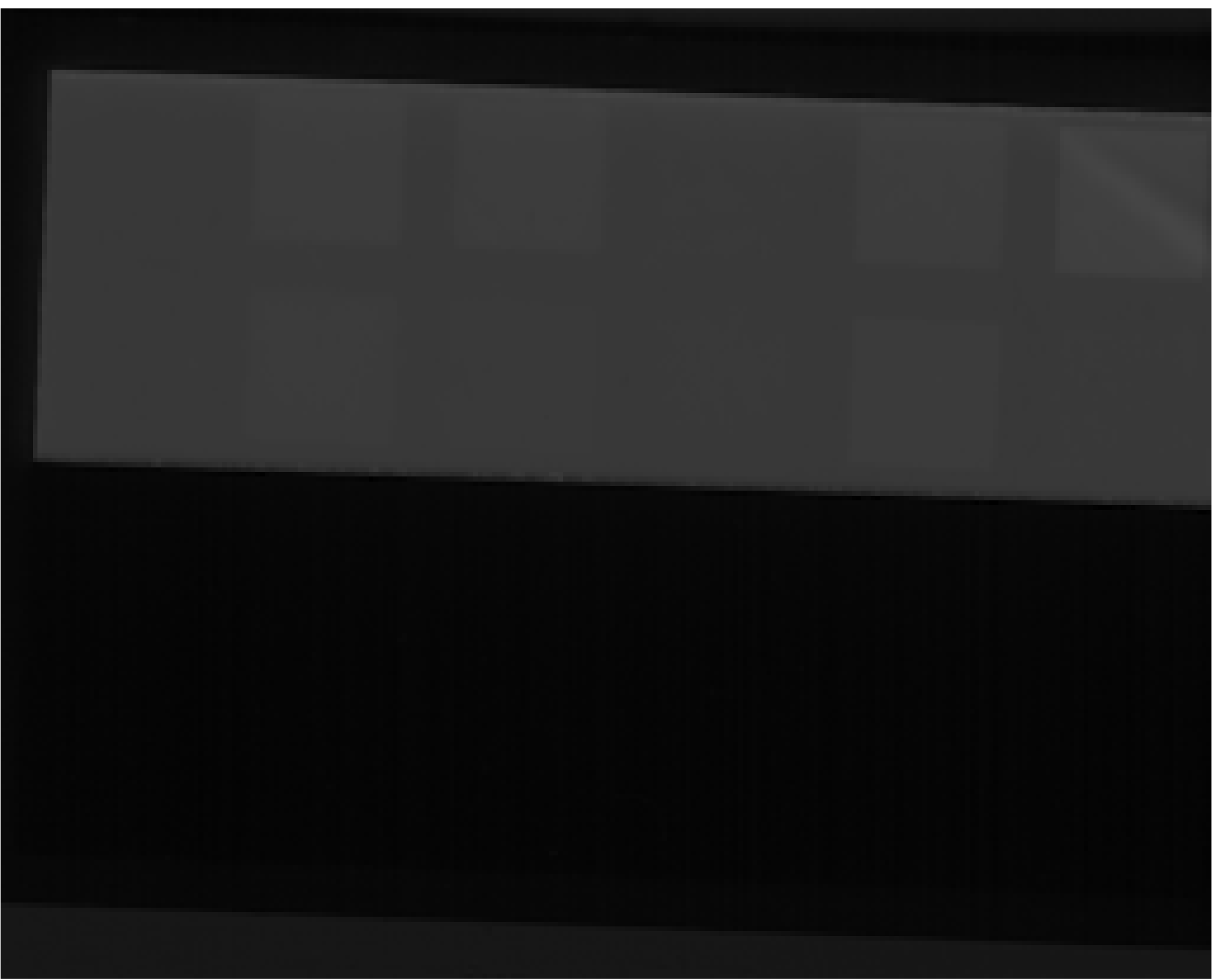}
\end{subfigure}%
\begin{subfigure}{0.06\textwidth}
\centering
\includegraphics[width=\textwidth]{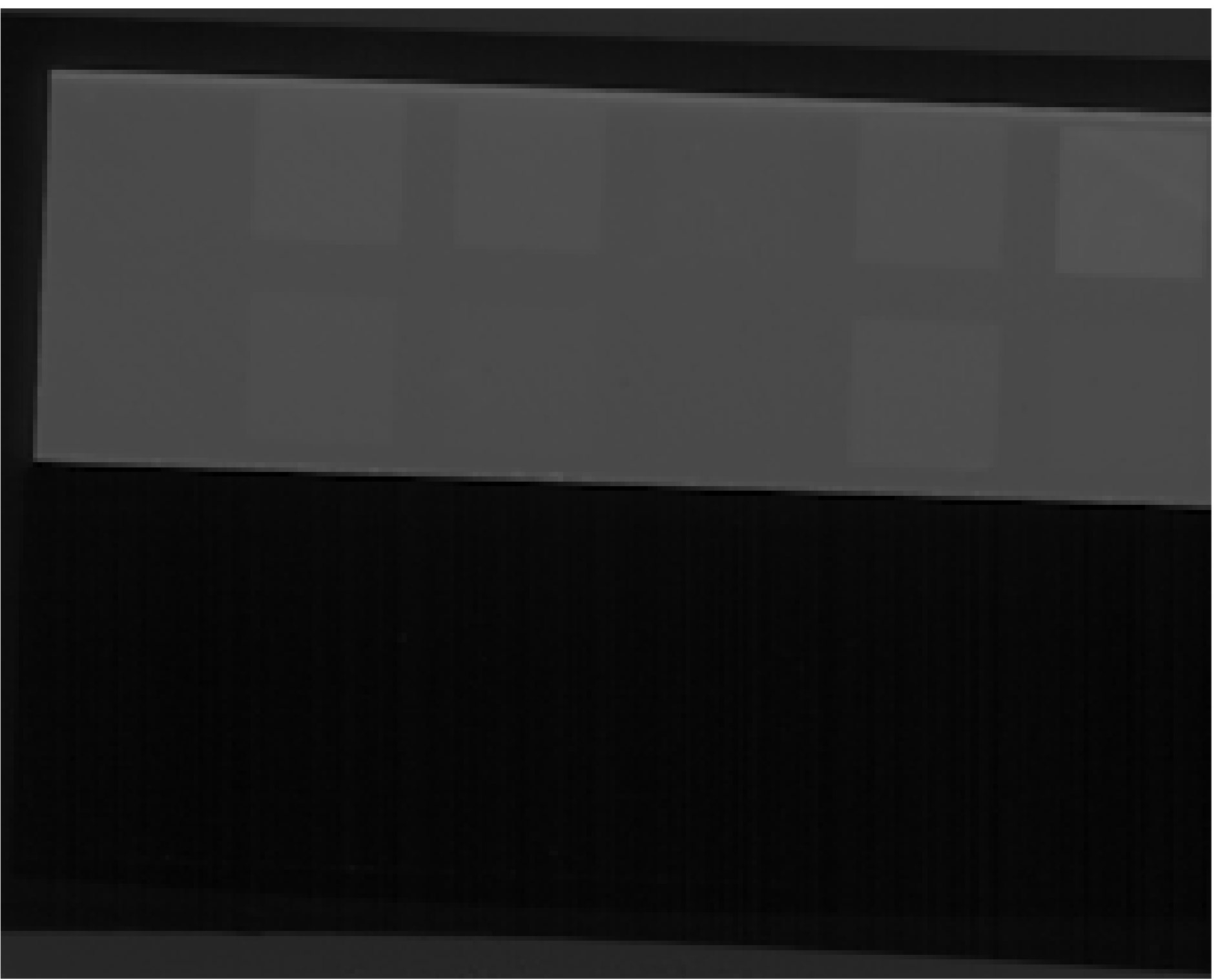}
\end{subfigure}%
\begin{subfigure}{0.06\textwidth}
\centering
\includegraphics[width=\textwidth]{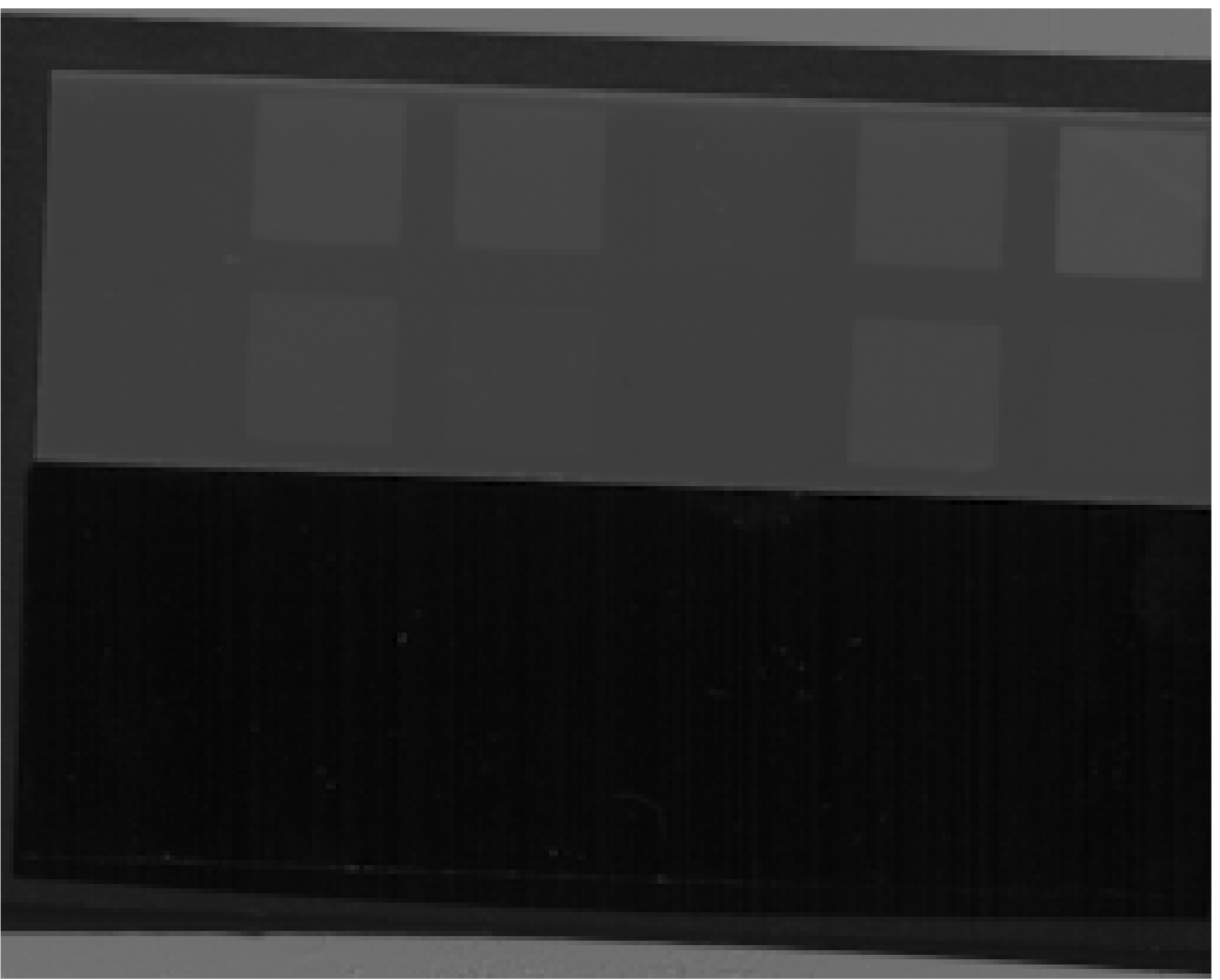}
\end{subfigure}%
\begin{subfigure}{0.06\textwidth}
\centering
\includegraphics[width=\textwidth]{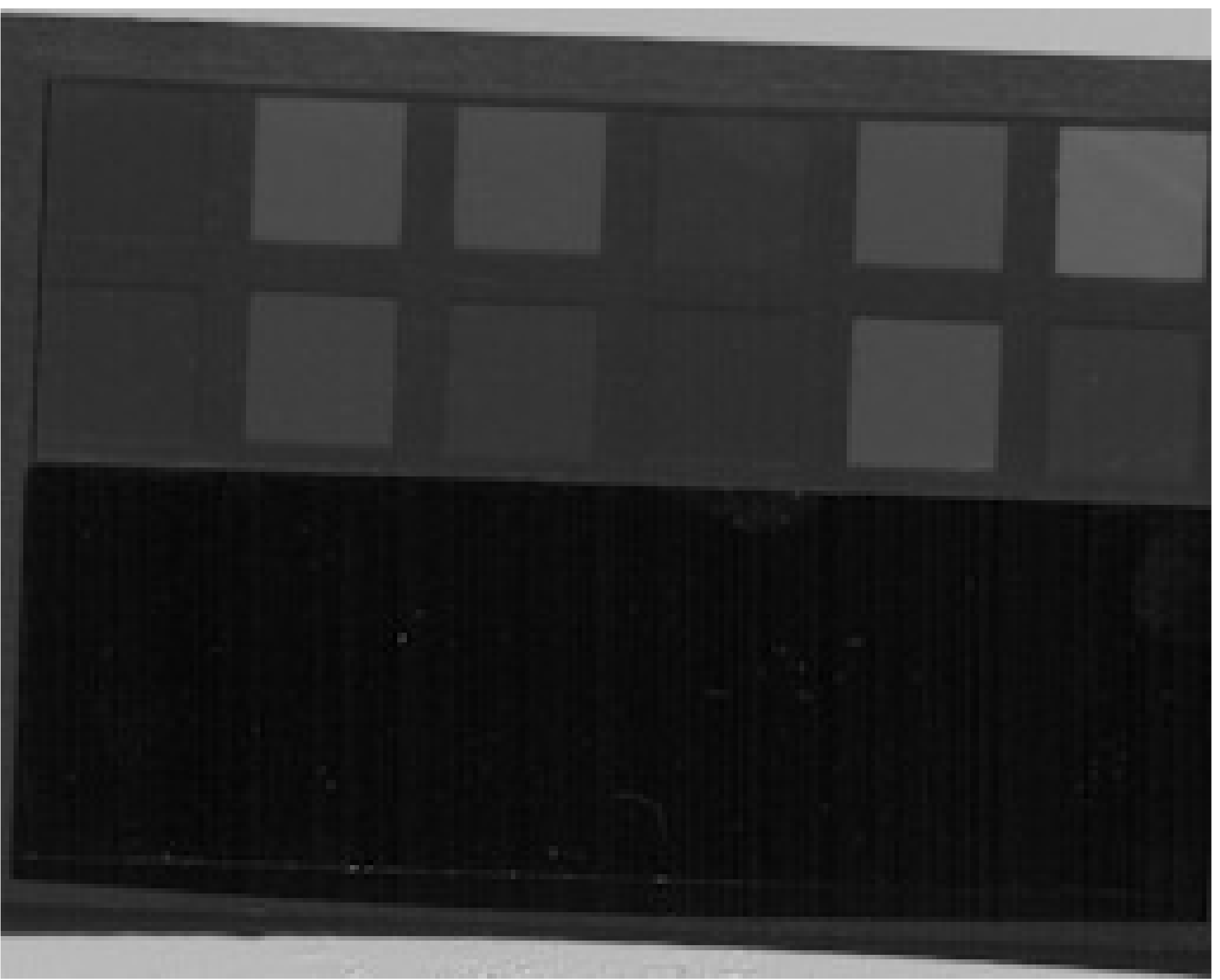}
\end{subfigure}%
\begin{subfigure}{0.06\textwidth}
\centering
\includegraphics[width=\textwidth]{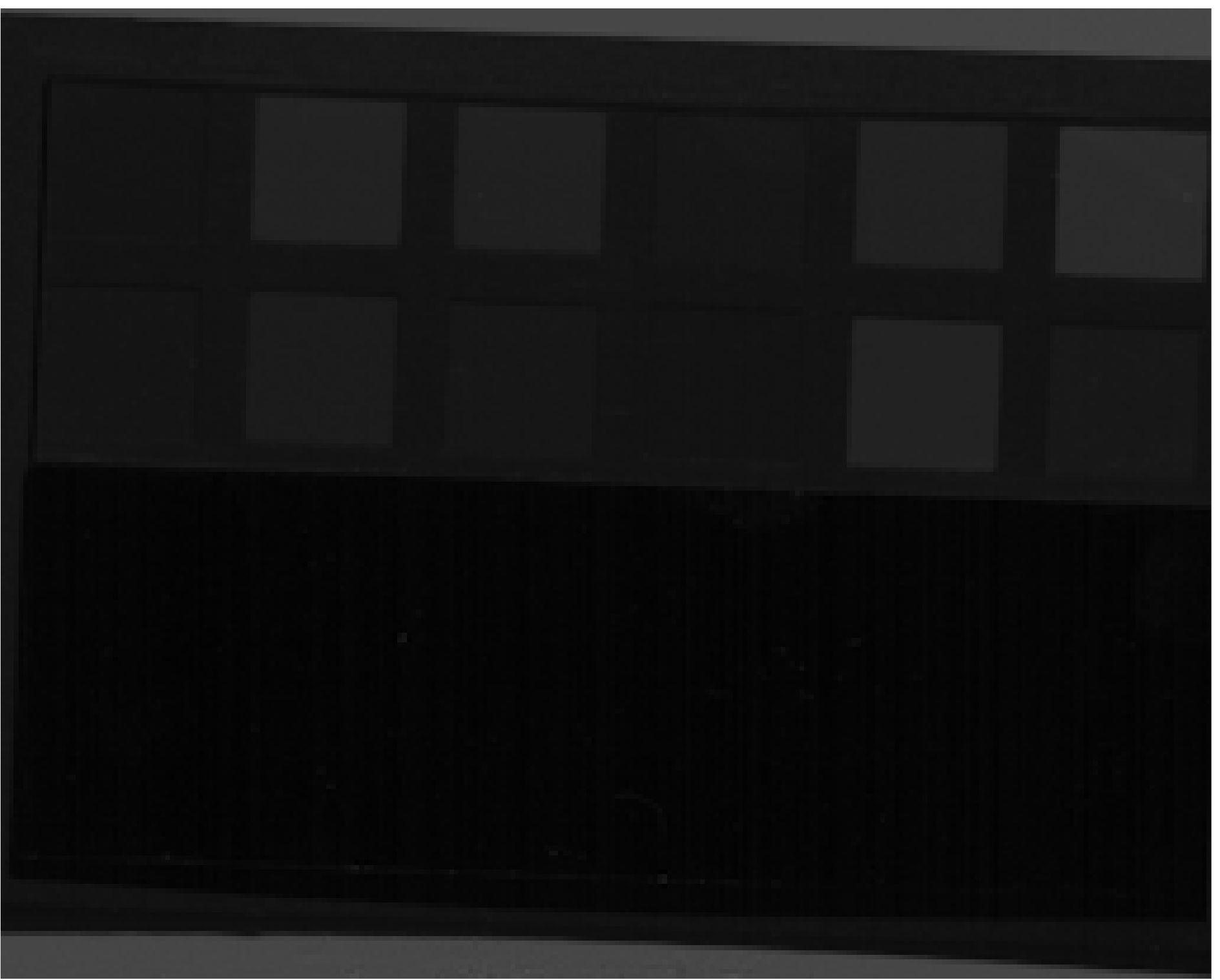}
\end{subfigure}%
\begin{subfigure}{0.06\textwidth}
\centering
\includegraphics[width=\textwidth]{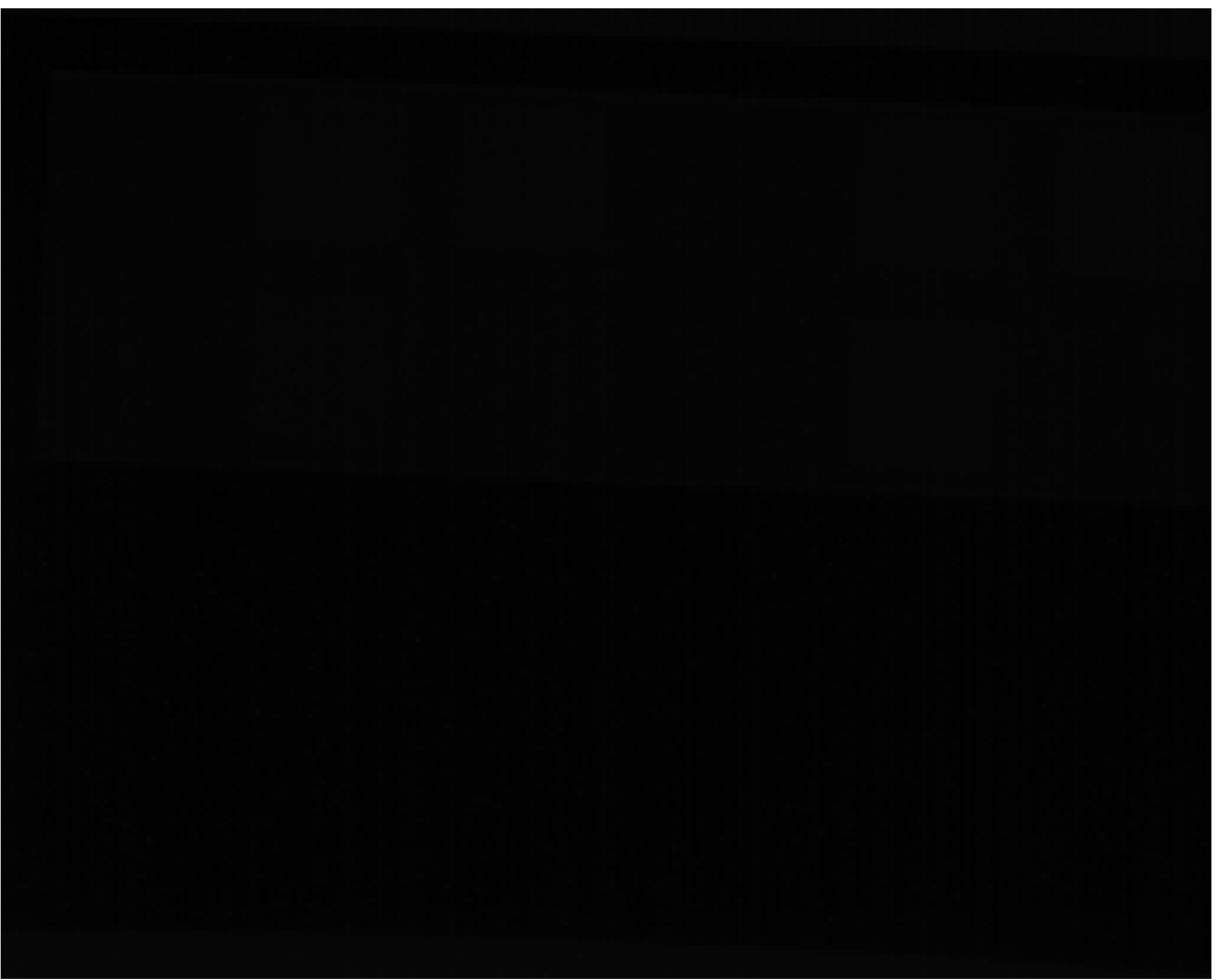}
\end{subfigure}%
\begin{subfigure}{0.06\textwidth}
\centering
\includegraphics[width=\textwidth]{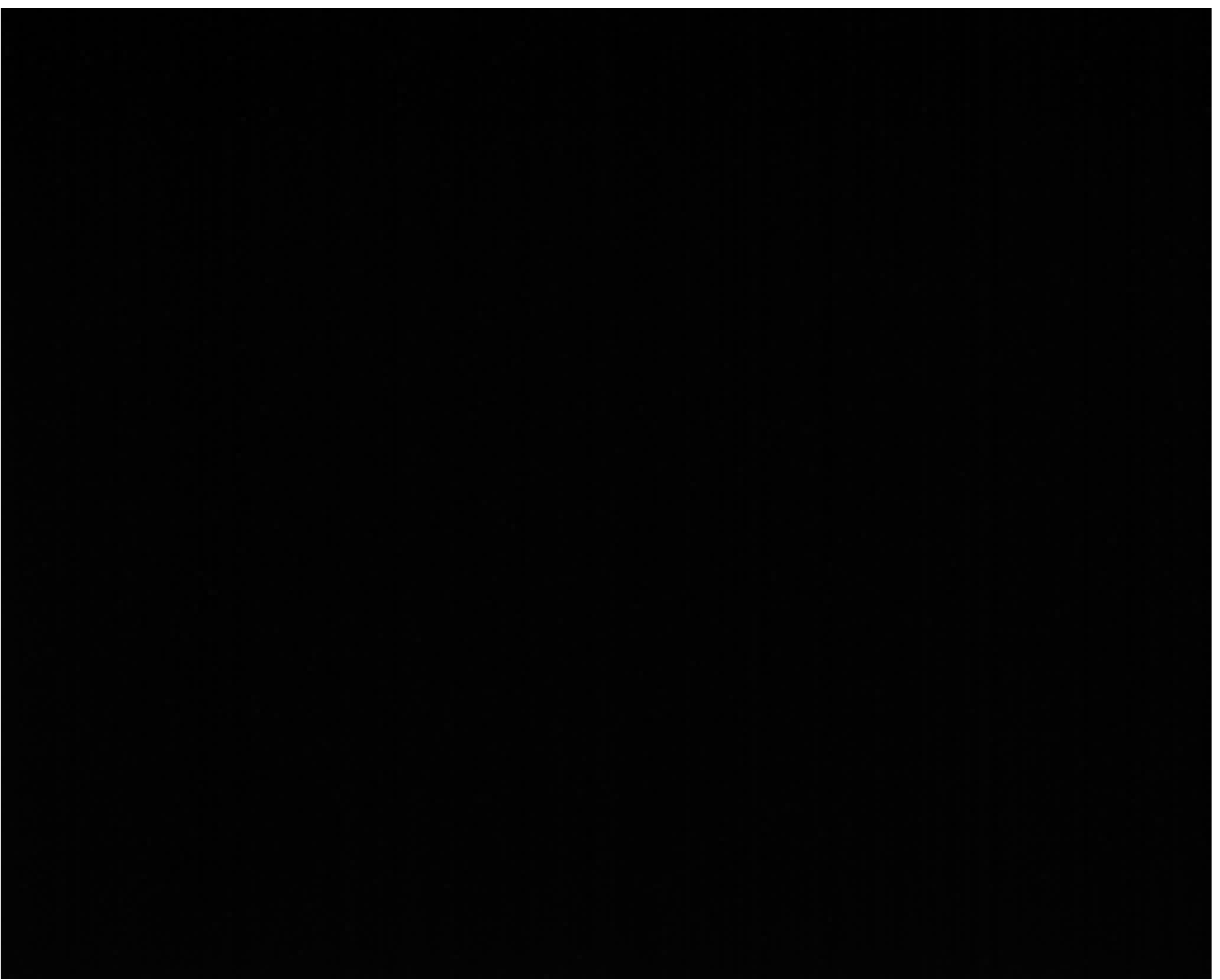}
\end{subfigure}%
\begin{subfigure}{0.06\textwidth}
\centering
\includegraphics[width=\textwidth]{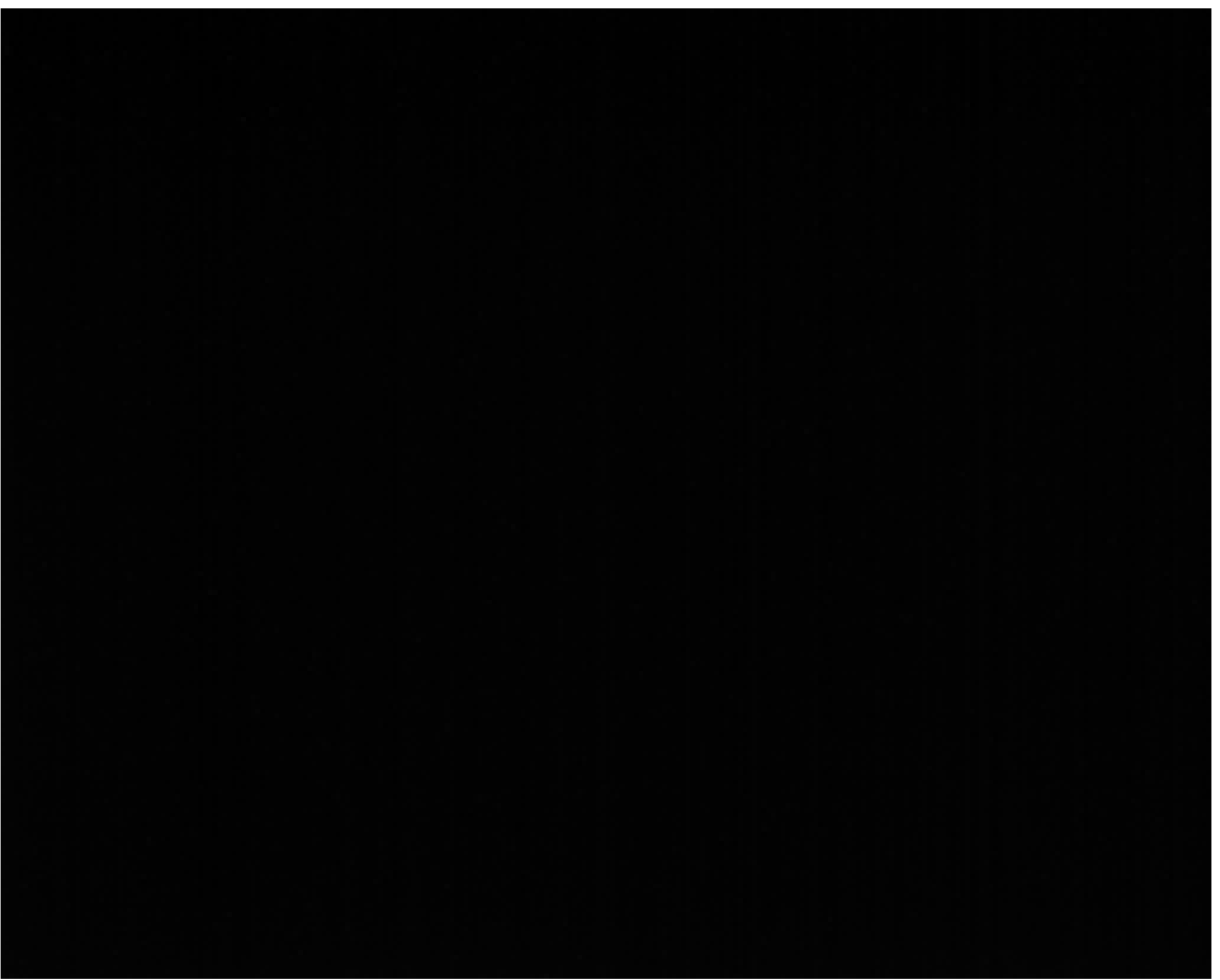}
\end{subfigure}%
\begin{subfigure}{0.06\textwidth}
\centering
\includegraphics[width=\textwidth]{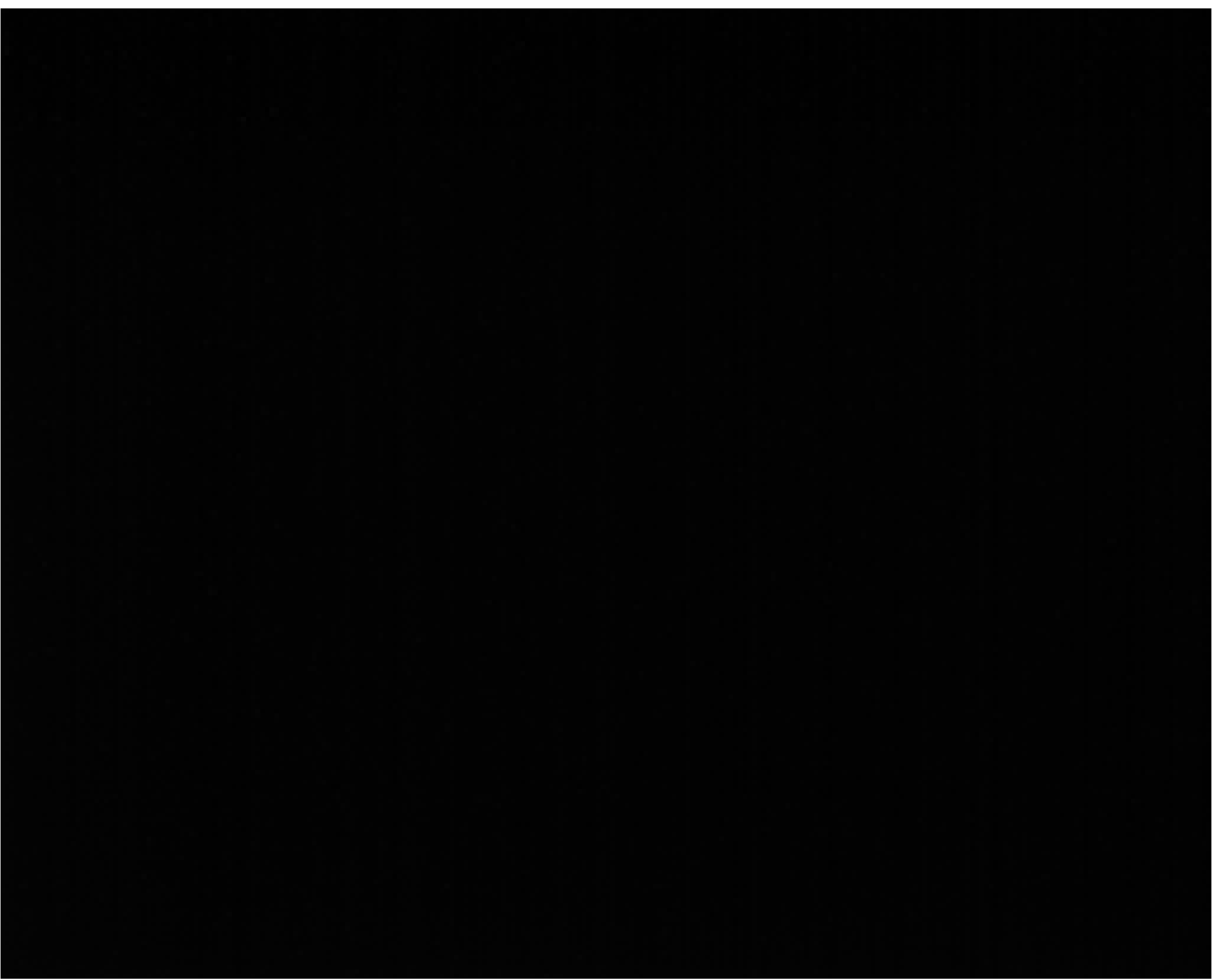}
\end{subfigure}%
\begin{subfigure}{0.06\textwidth}
\centering
\includegraphics[width=\textwidth]{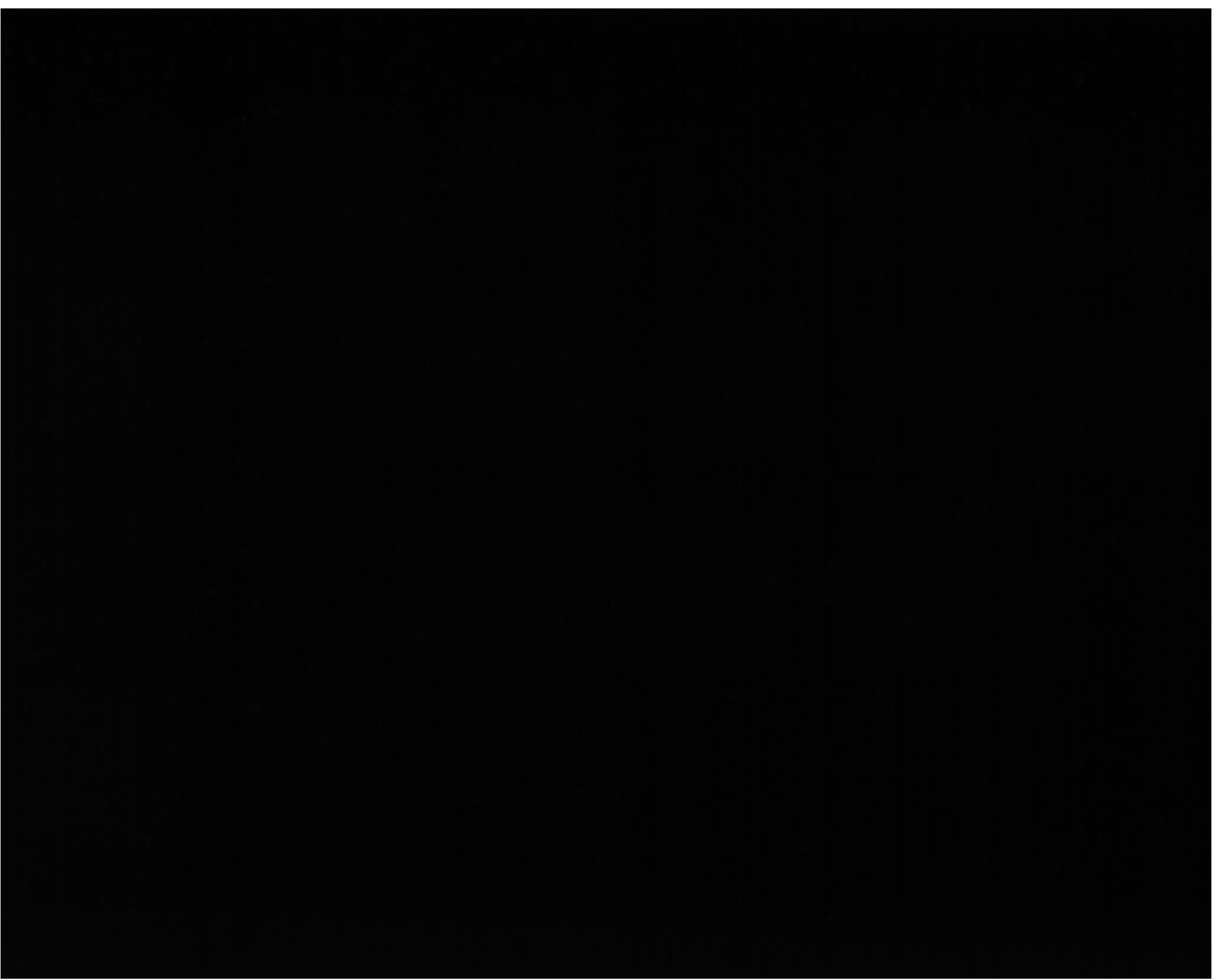}
\end{subfigure}%
\begin{subfigure}{0.06\textwidth}
\centering
\includegraphics[width=\textwidth]{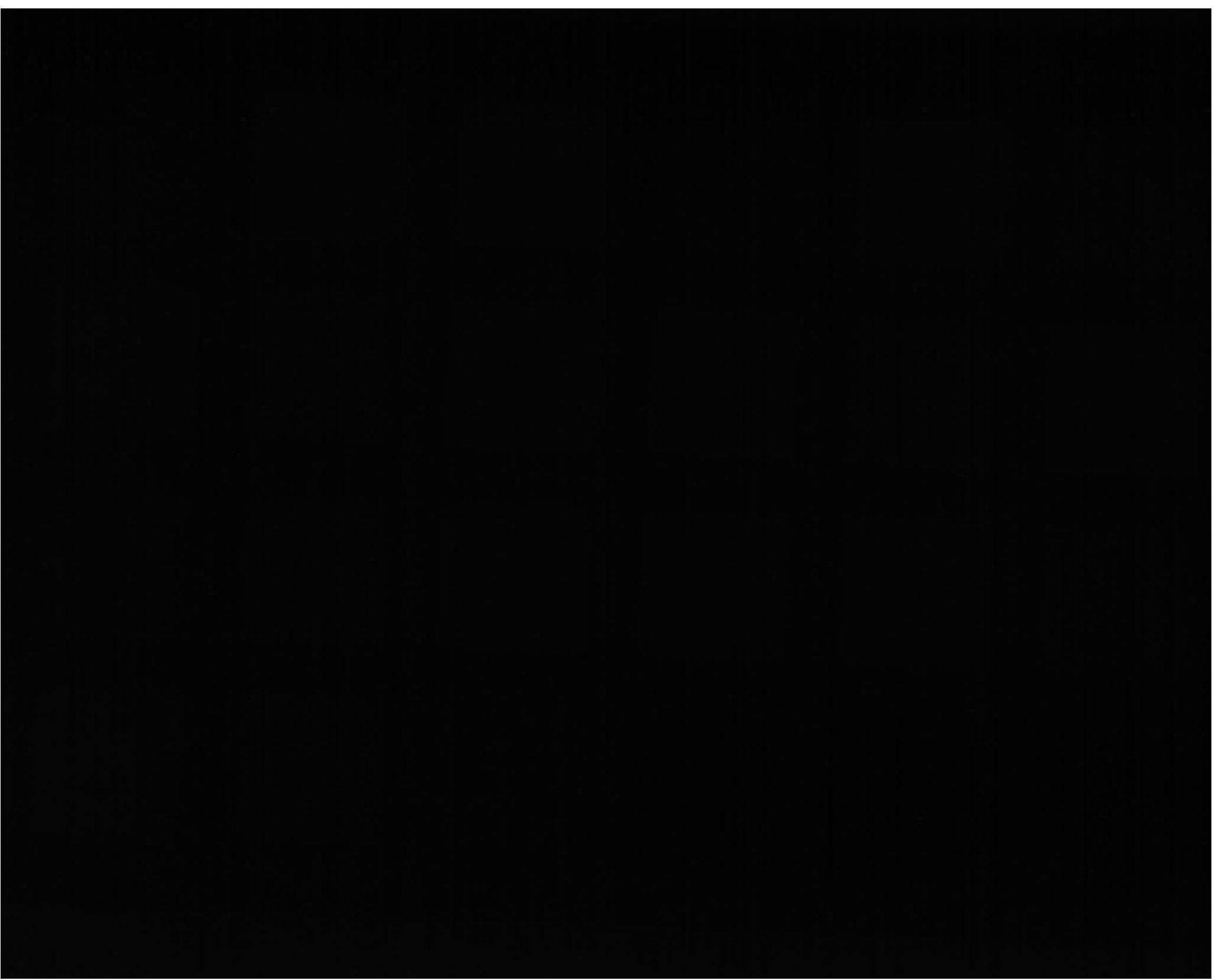}
\end{subfigure}%
\begin{subfigure}{0.06\textwidth}
\centering
\includegraphics[width=\textwidth]{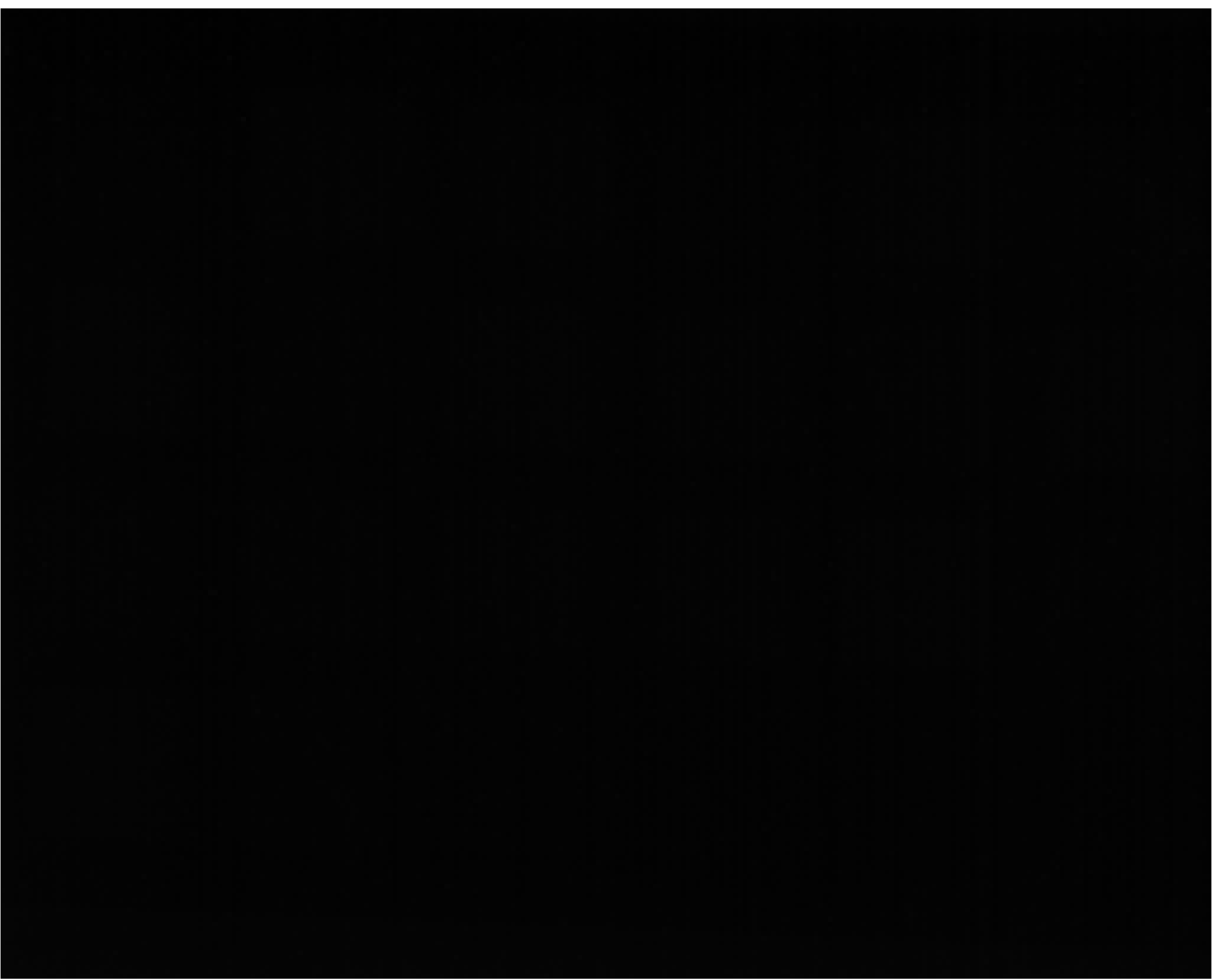}
\end{subfigure}%
\begin{subfigure}{0.06\textwidth}
\centering
\includegraphics[width=\textwidth]{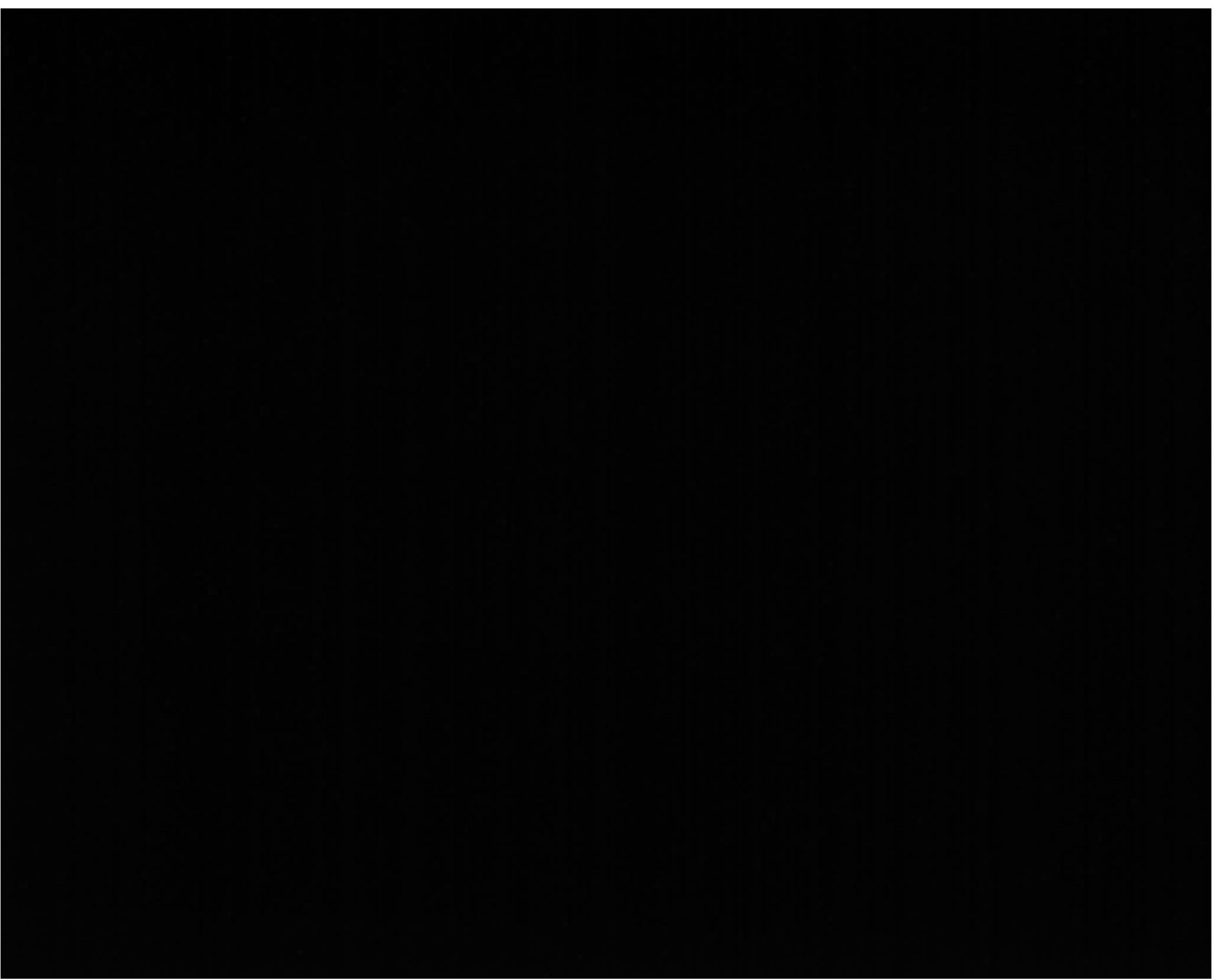}
\end{subfigure}\\%
\begin{subfigure}{0.06\textwidth}
\captionsetup{justification=raggedright,font=scriptsize}
\caption*{537 to 563nm}
\end{subfigure}%
\begin{subfigure}{0.06\textwidth}
\centering
\includegraphics[width=\textwidth]{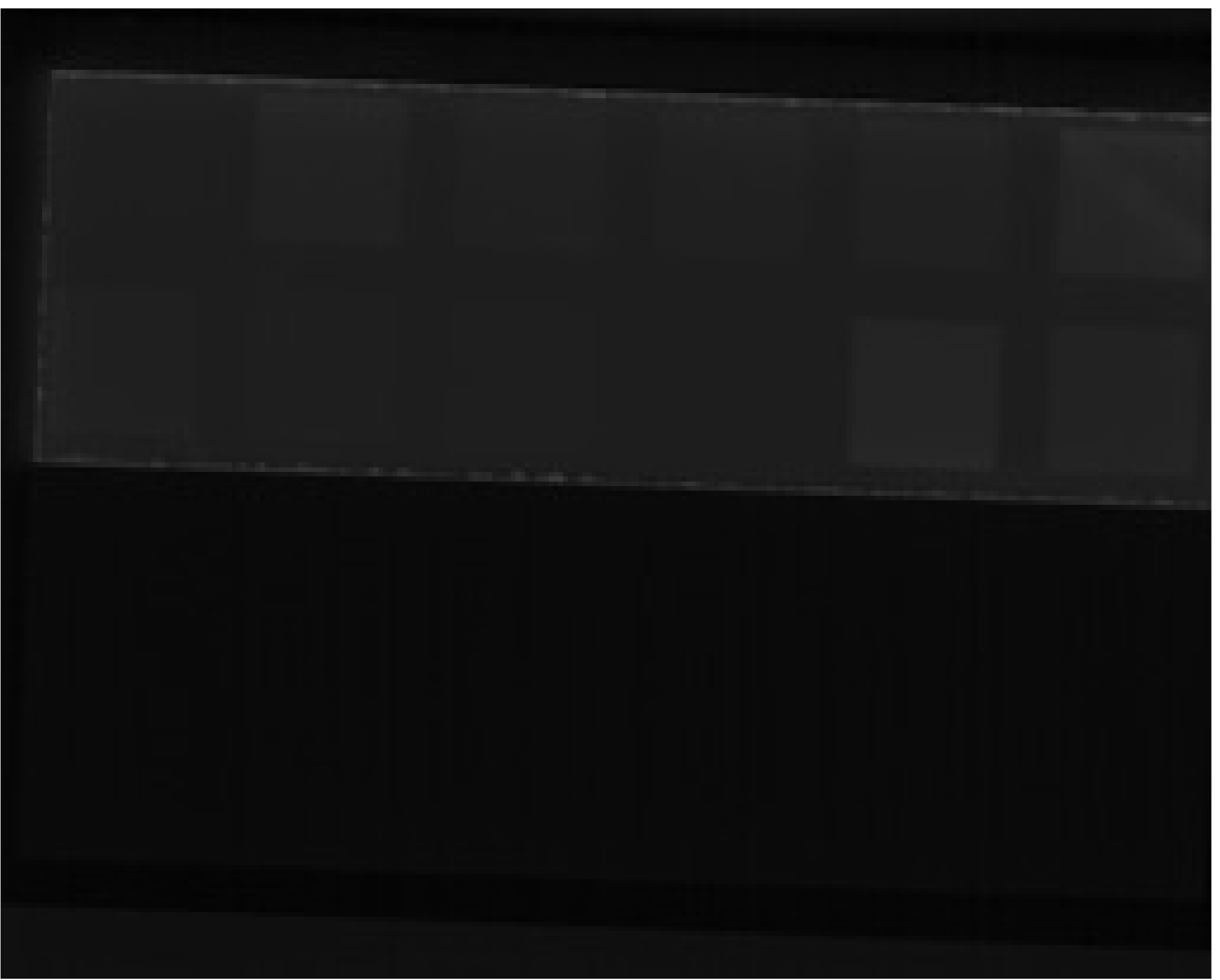}
\end{subfigure}%
\begin{subfigure}{0.06\textwidth}
\centering
\includegraphics[width=\textwidth]{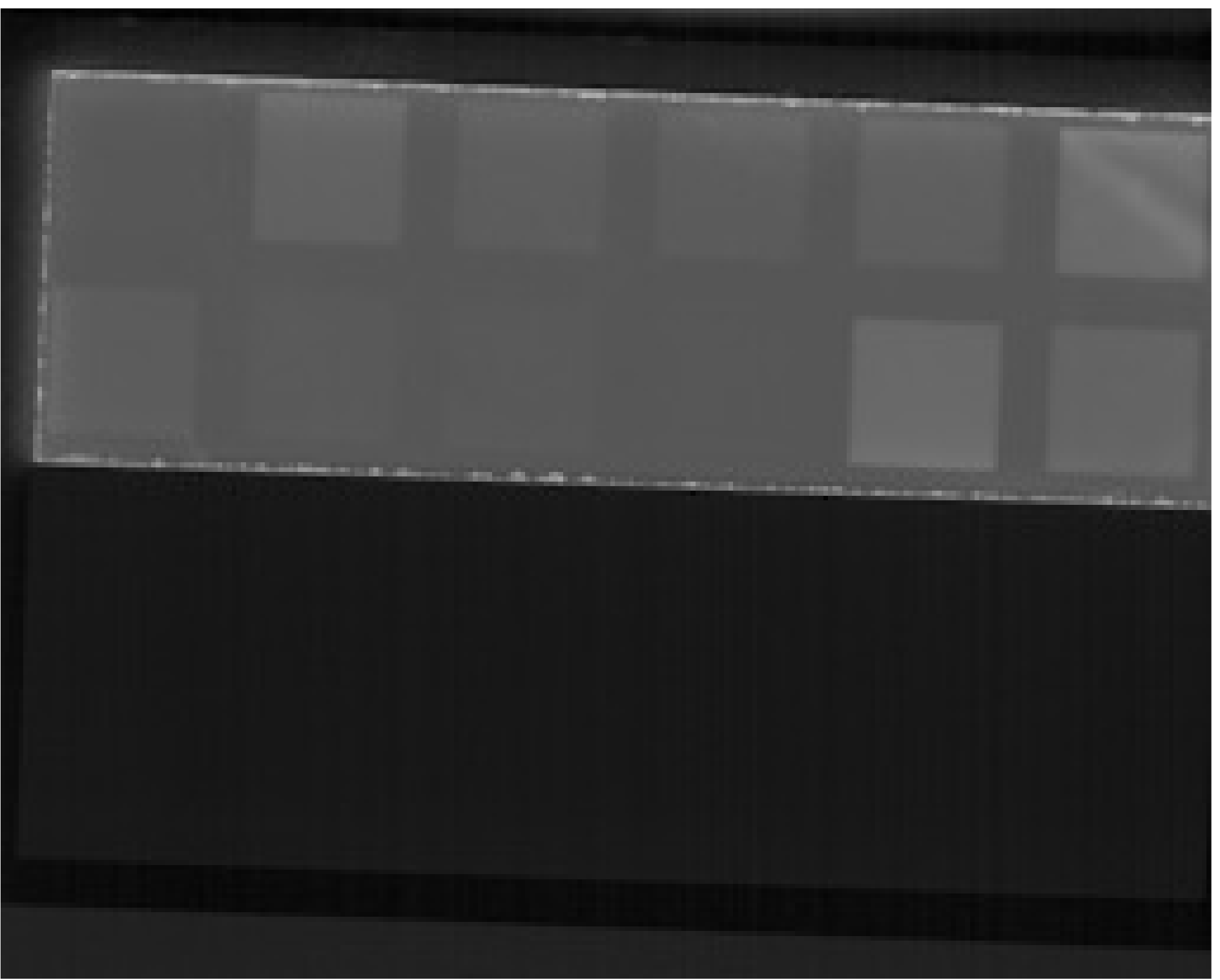}
\end{subfigure}%
\begin{subfigure}{0.06\textwidth}
\centering
\includegraphics[width=\textwidth]{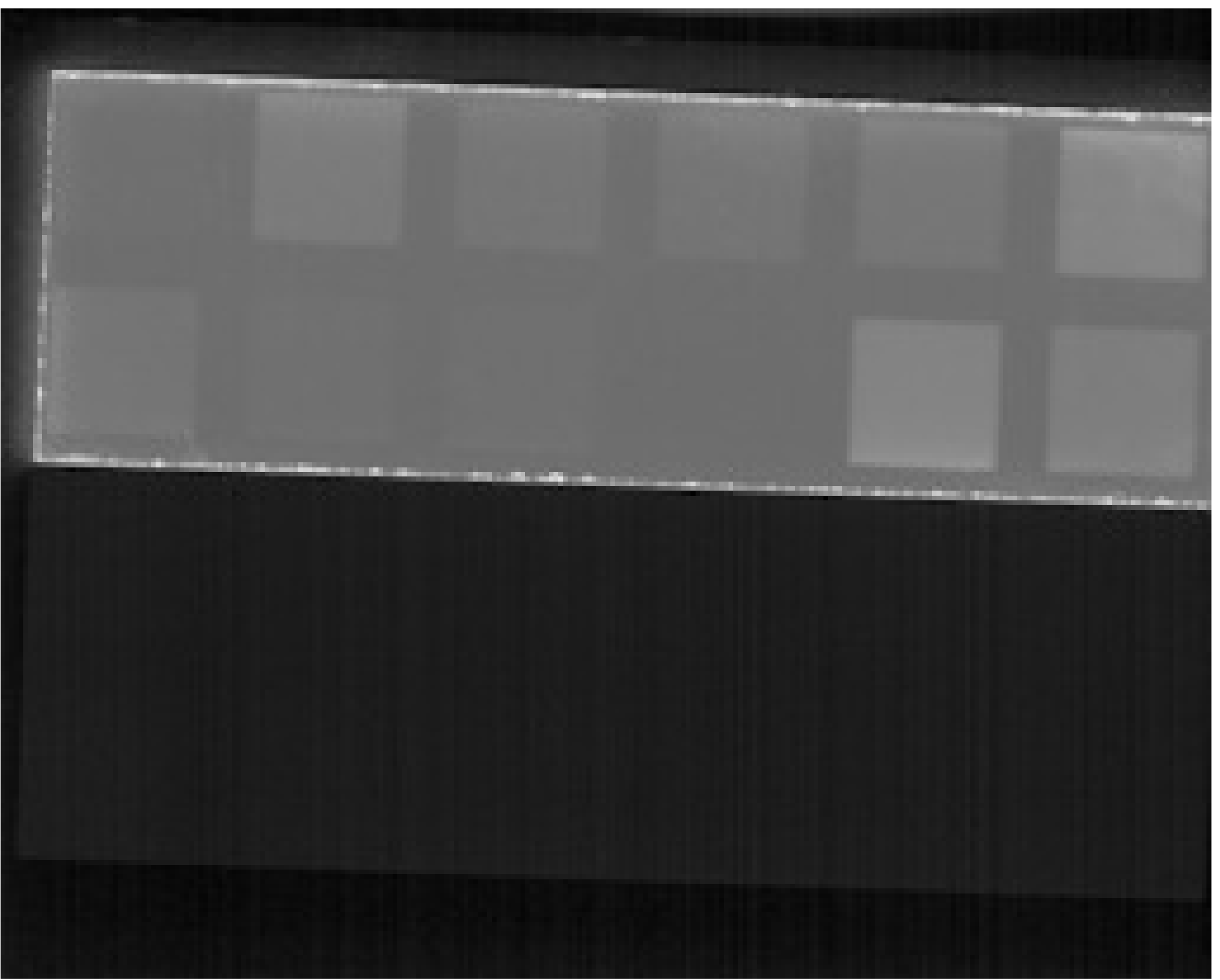}
\end{subfigure}%
\begin{subfigure}{0.06\textwidth}
\centering
\includegraphics[width=\textwidth]{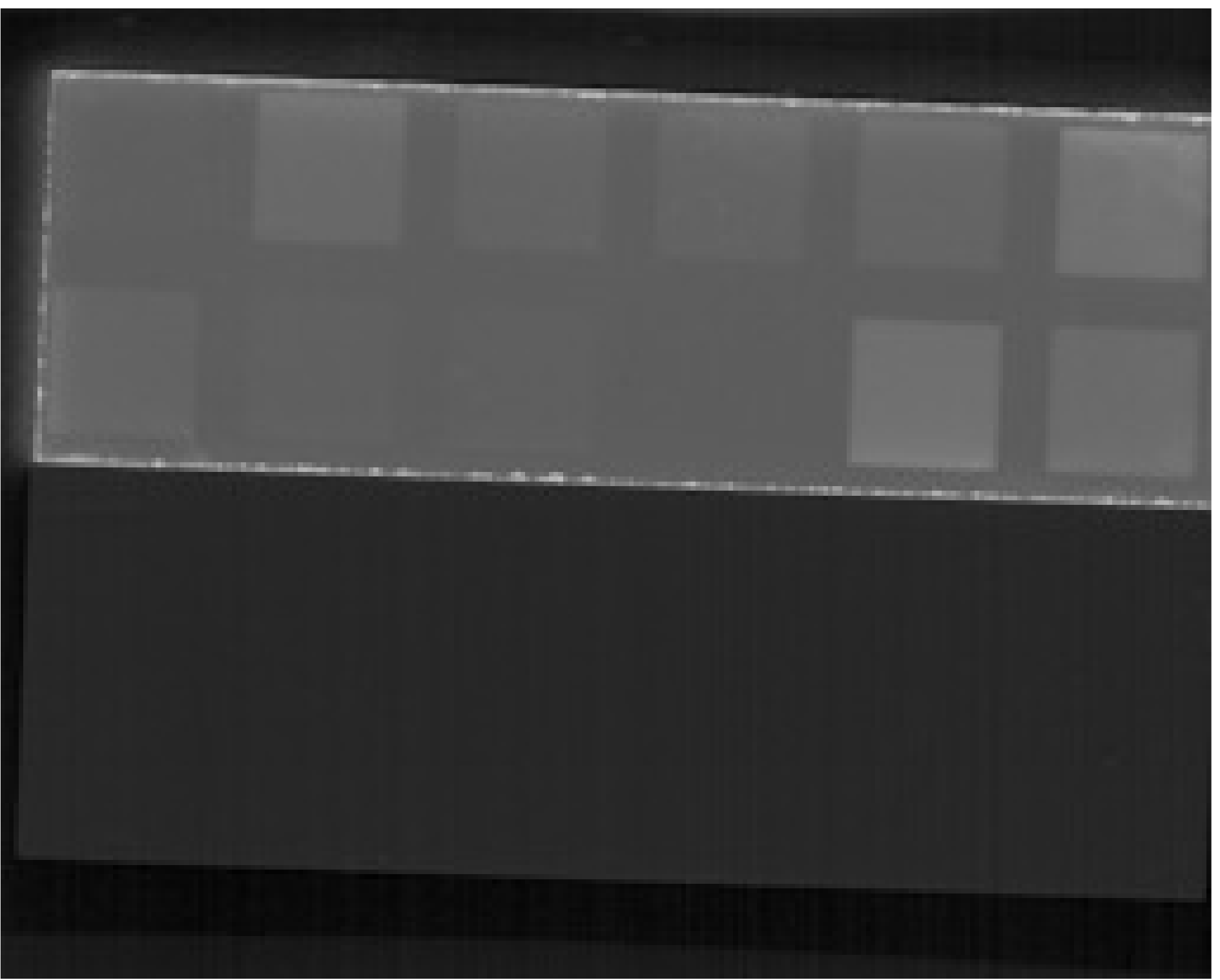}
\end{subfigure}%
\begin{subfigure}{0.06\textwidth}
\centering
\includegraphics[width=\textwidth]{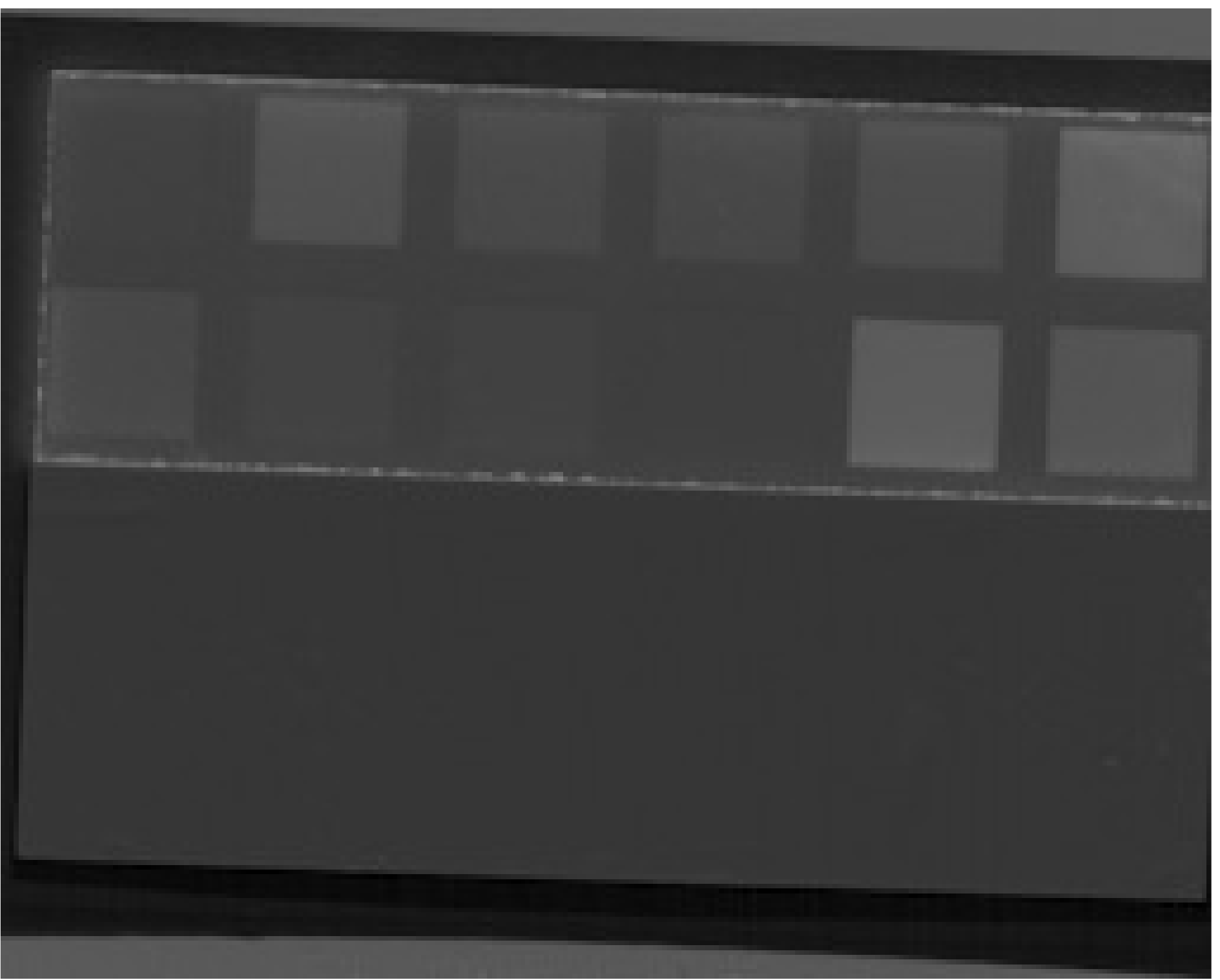}
\end{subfigure}%
\begin{subfigure}{0.06\textwidth}
\centering
\includegraphics[width=\textwidth]{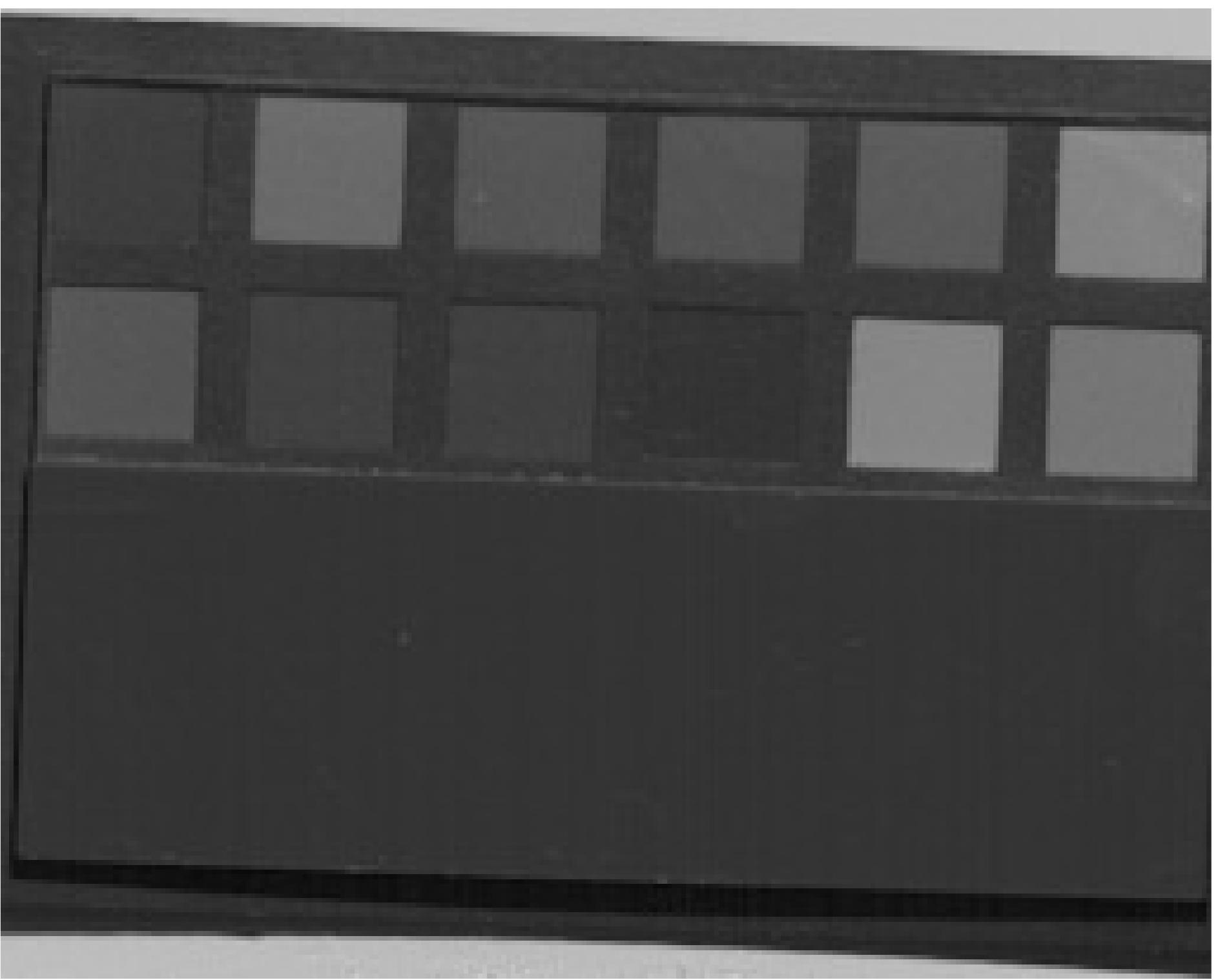}
\end{subfigure}%
\begin{subfigure}{0.06\textwidth}
\centering
\includegraphics[width=\textwidth]{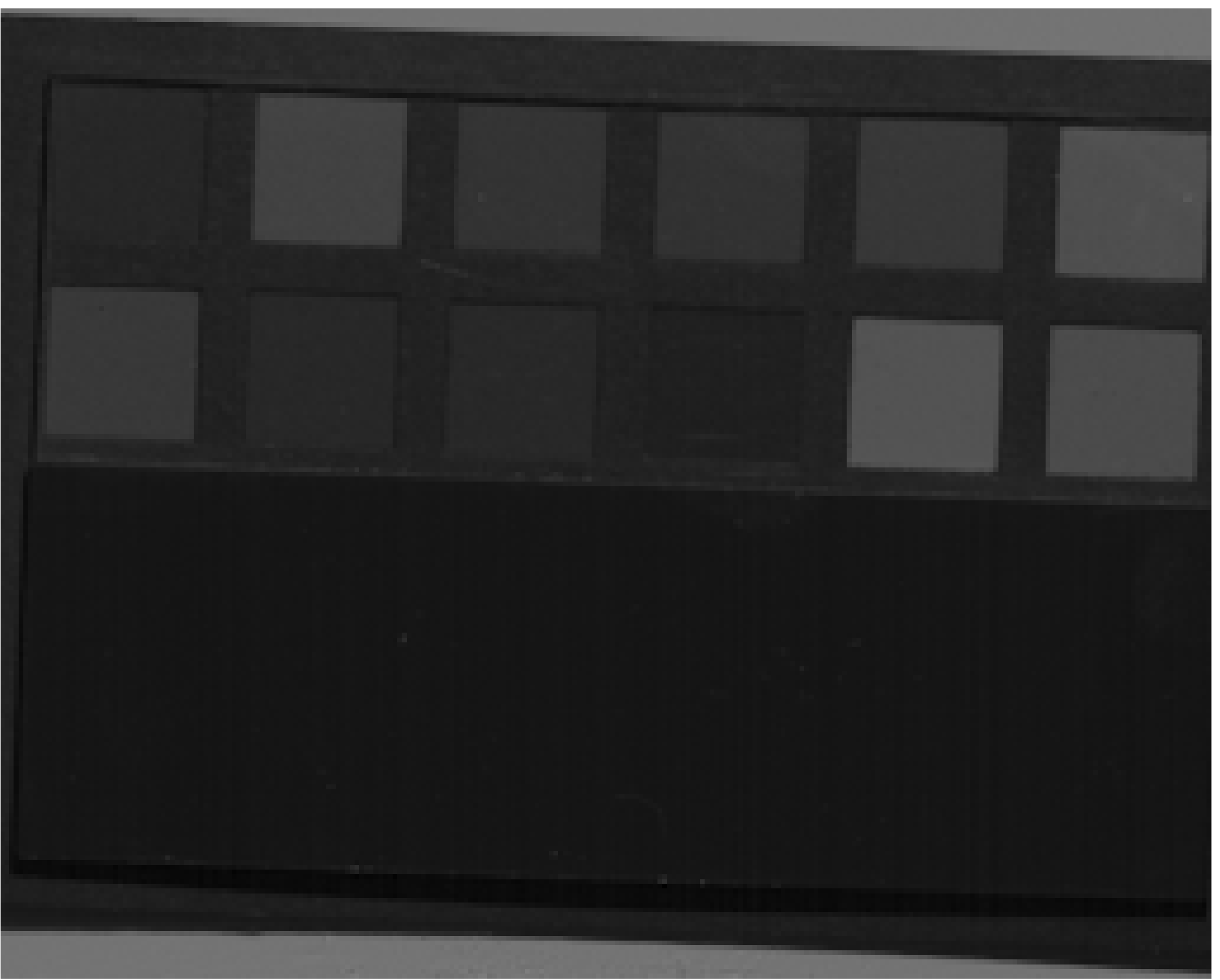}
\end{subfigure}%
\begin{subfigure}{0.06\textwidth}
\centering
\includegraphics[width=\textwidth]{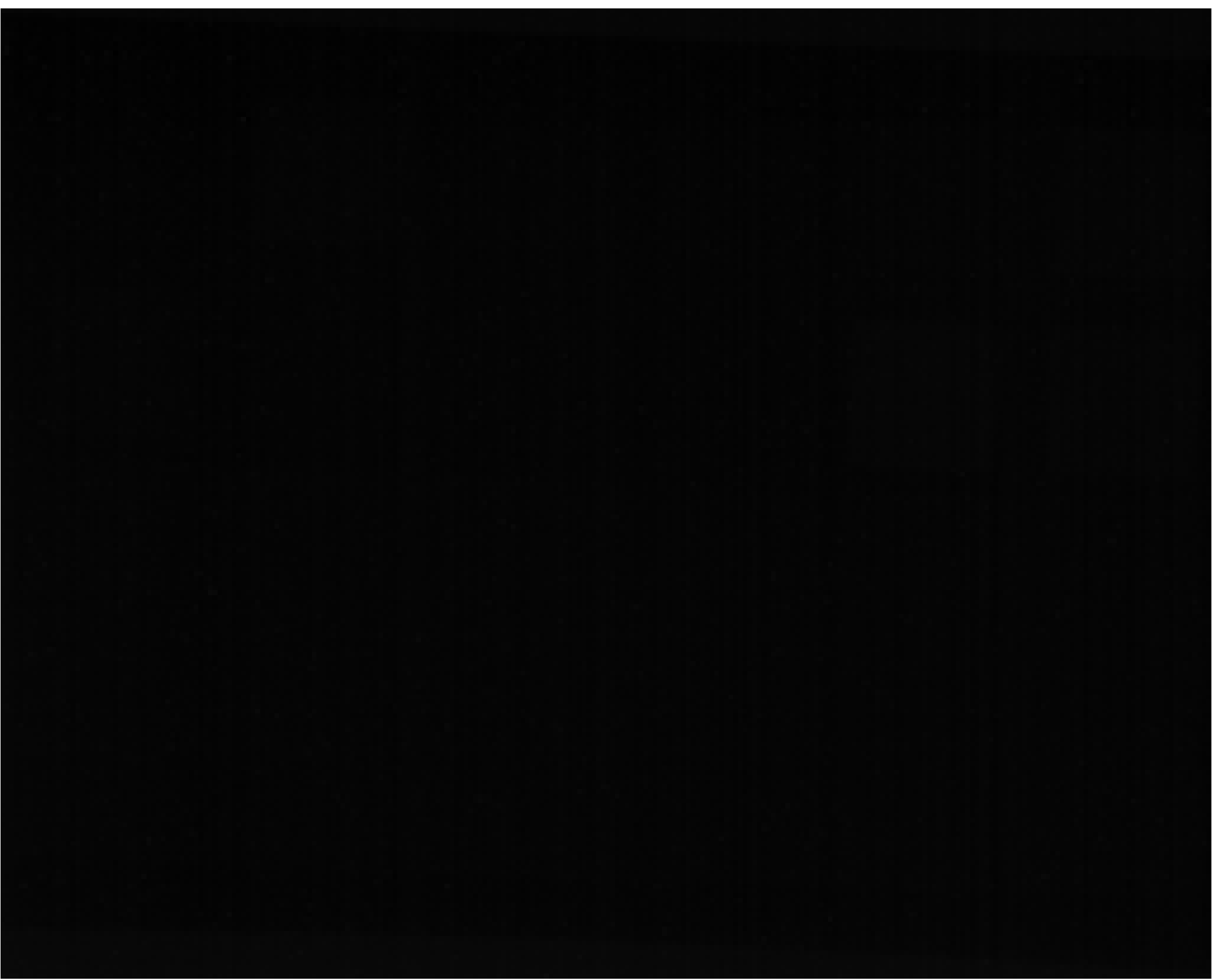}
\end{subfigure}%
\begin{subfigure}{0.06\textwidth}
\centering
\includegraphics[width=\textwidth]{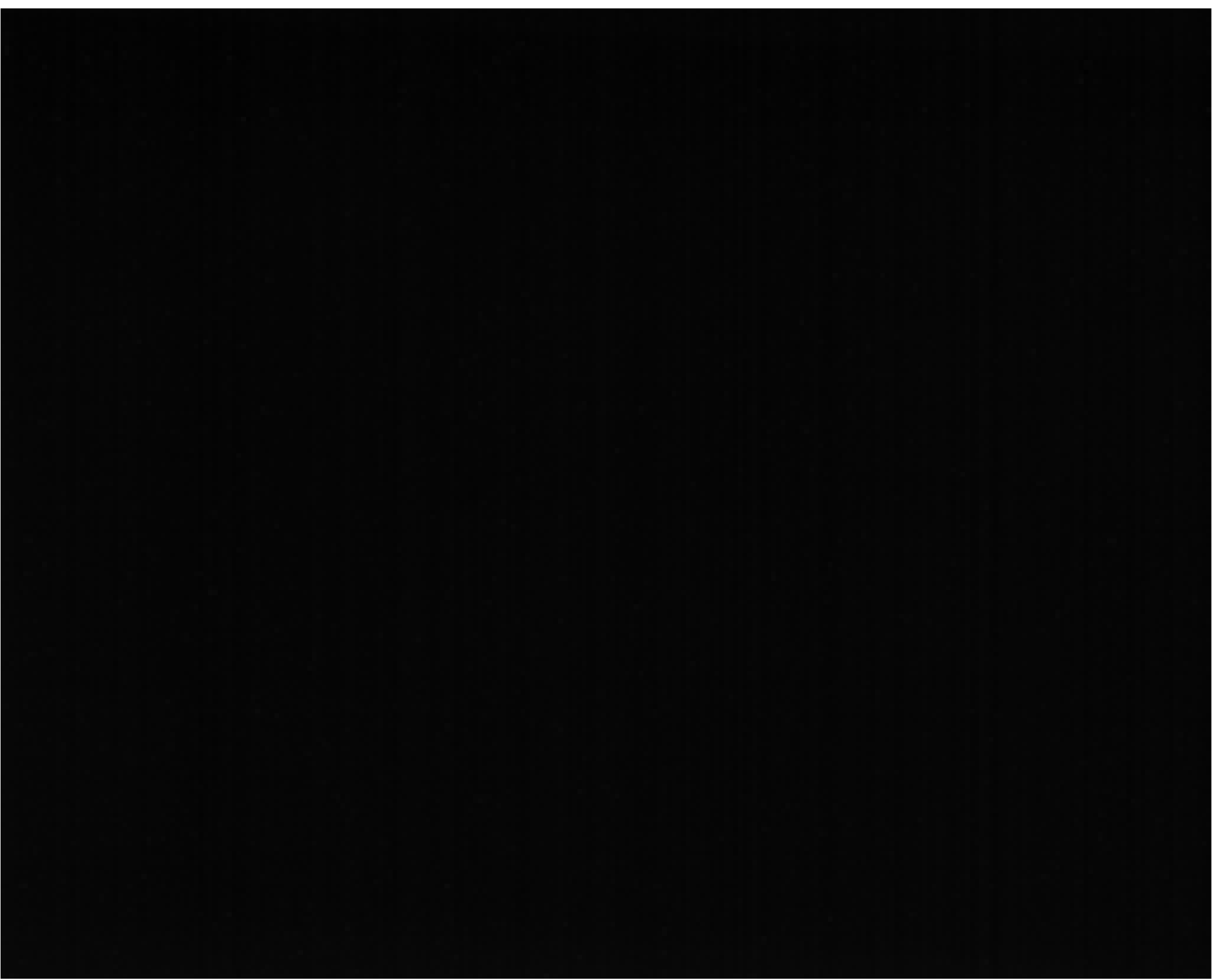}
\end{subfigure}%
\begin{subfigure}{0.06\textwidth}
\centering
\includegraphics[width=\textwidth]{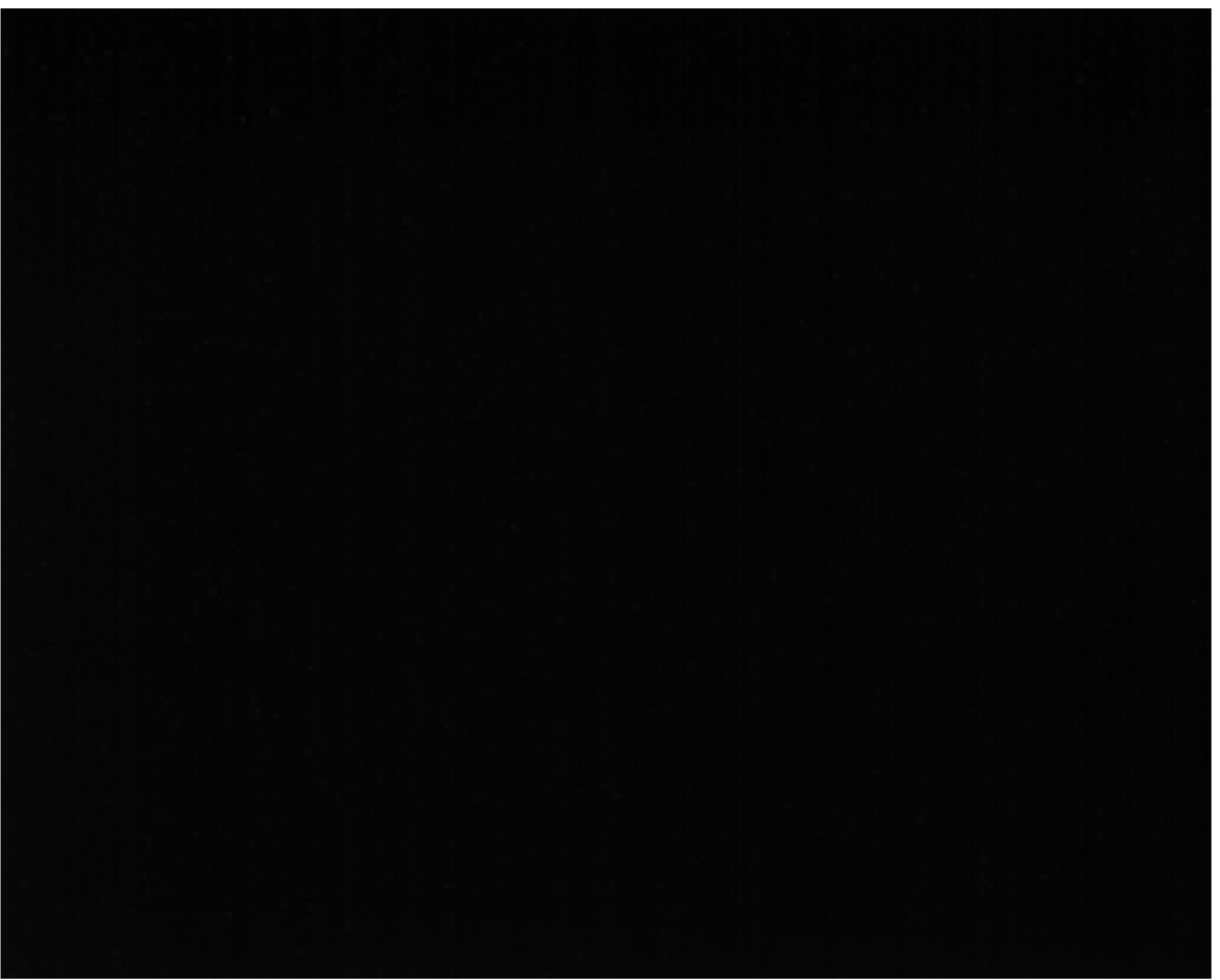}
\end{subfigure}%
\begin{subfigure}{0.06\textwidth}
\centering
\includegraphics[width=\textwidth]{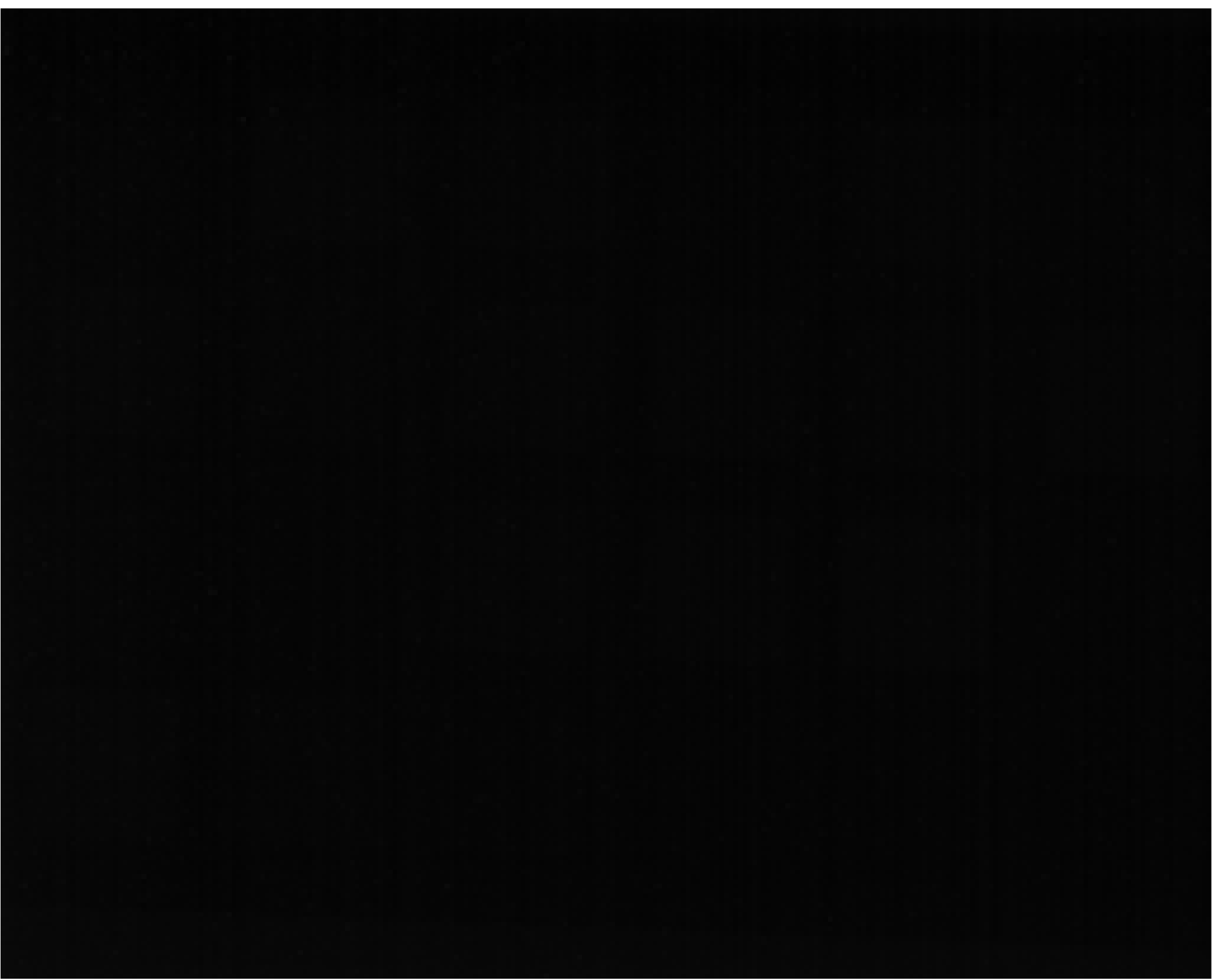}
\end{subfigure}%
\begin{subfigure}{0.06\textwidth}
\centering
\includegraphics[width=\textwidth]{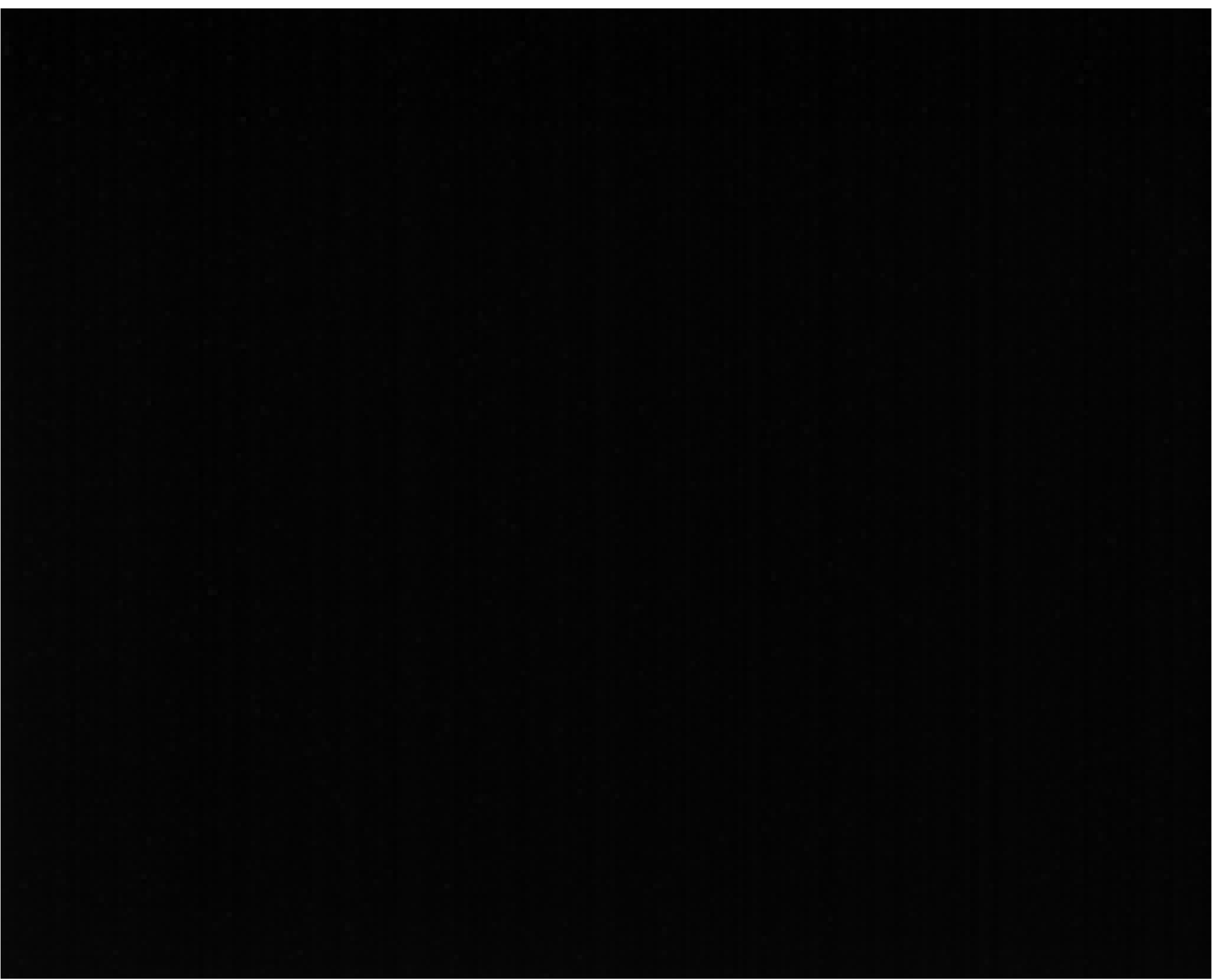}
\end{subfigure}%
\begin{subfigure}{0.06\textwidth}
\centering
\includegraphics[width=\textwidth]{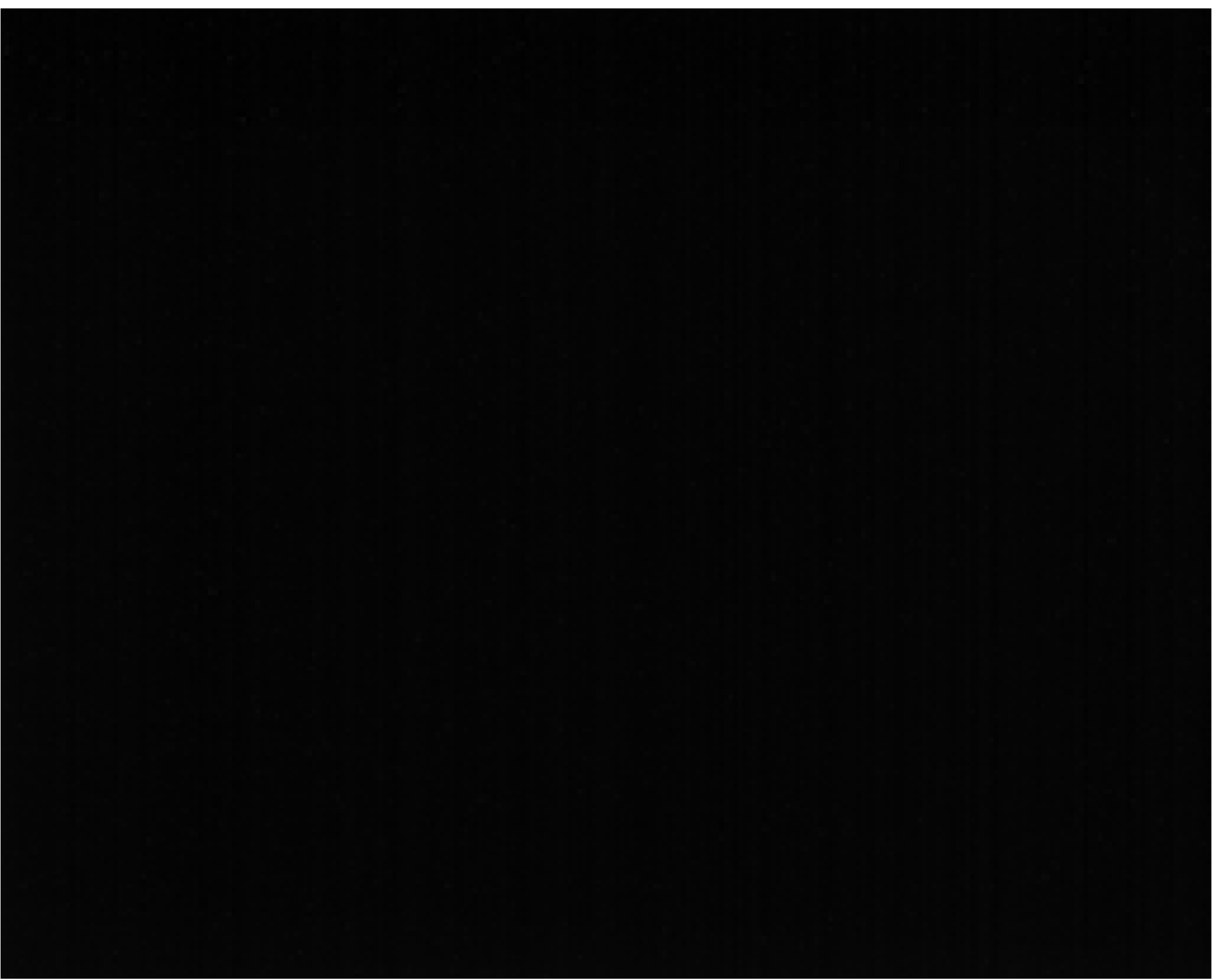}
\end{subfigure}%
\begin{subfigure}{0.06\textwidth}
\centering
\includegraphics[width=\textwidth]{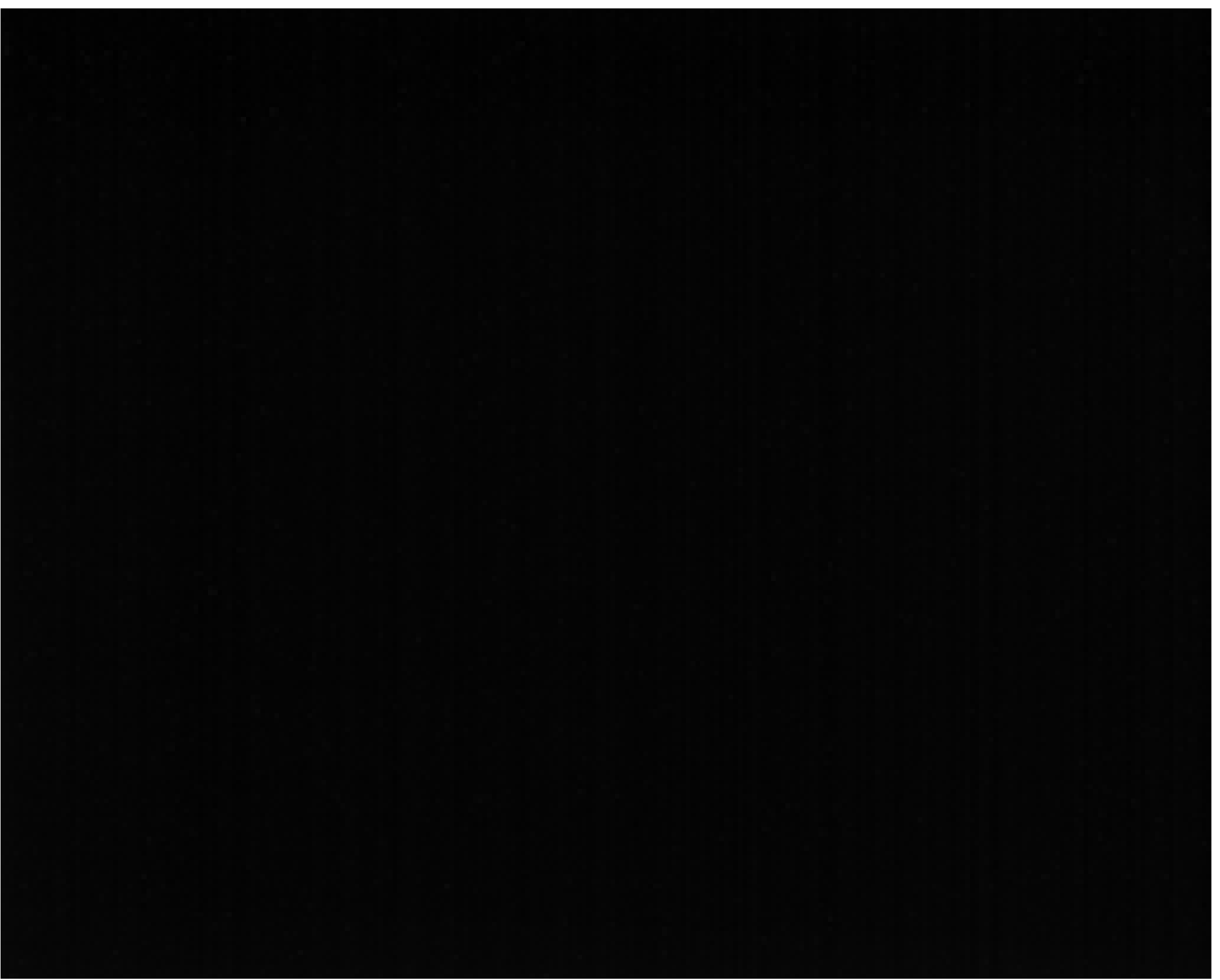}
\end{subfigure}\\%
\begin{subfigure}{0.06\textwidth}
\captionsetup{justification=raggedright,font=scriptsize}
\caption*{587 to 613nm}
\end{subfigure}%
\begin{subfigure}{0.06\textwidth}
\centering
\includegraphics[width=\textwidth]{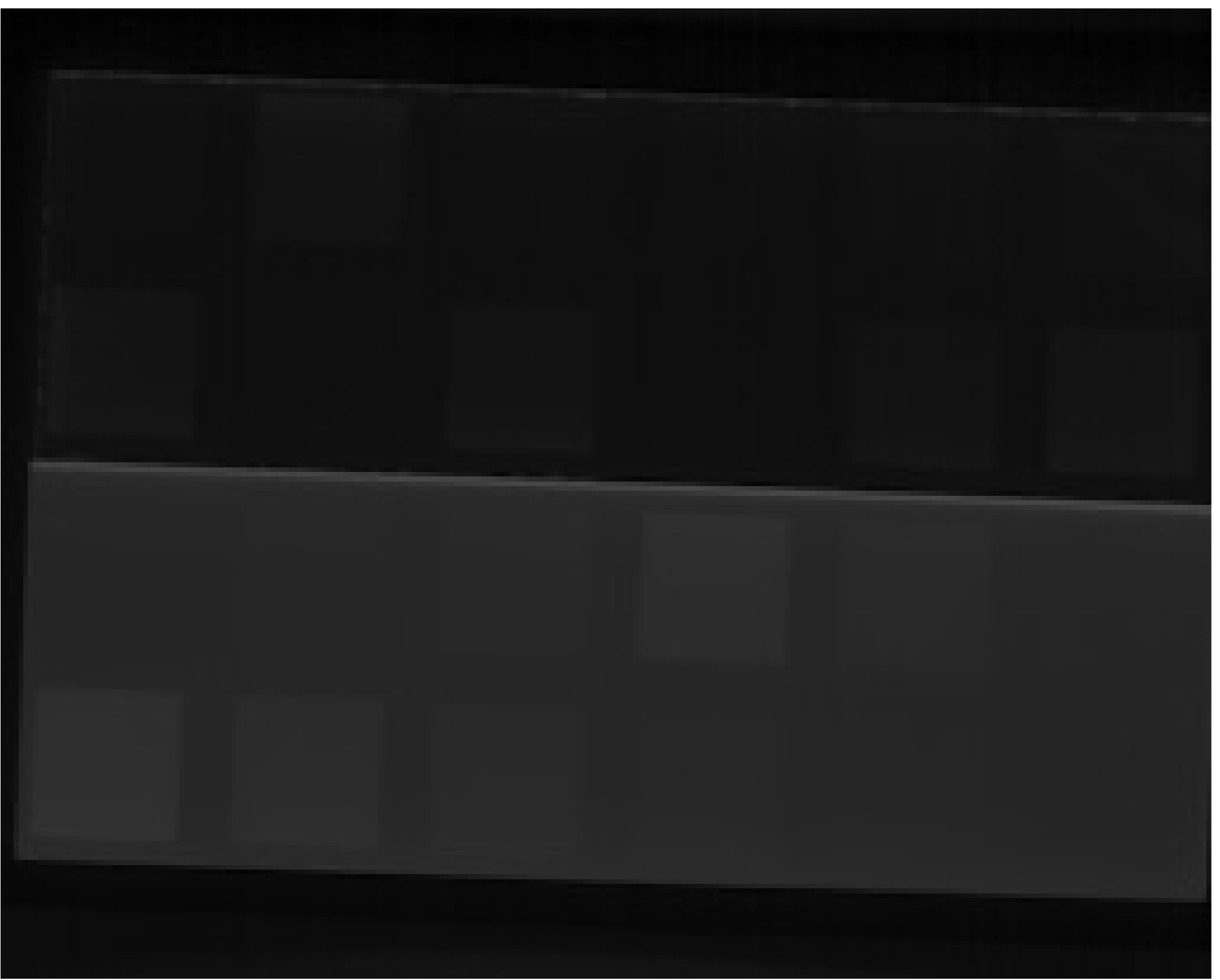}
\end{subfigure}%
\begin{subfigure}{0.06\textwidth}
\centering
\includegraphics[width=\textwidth]{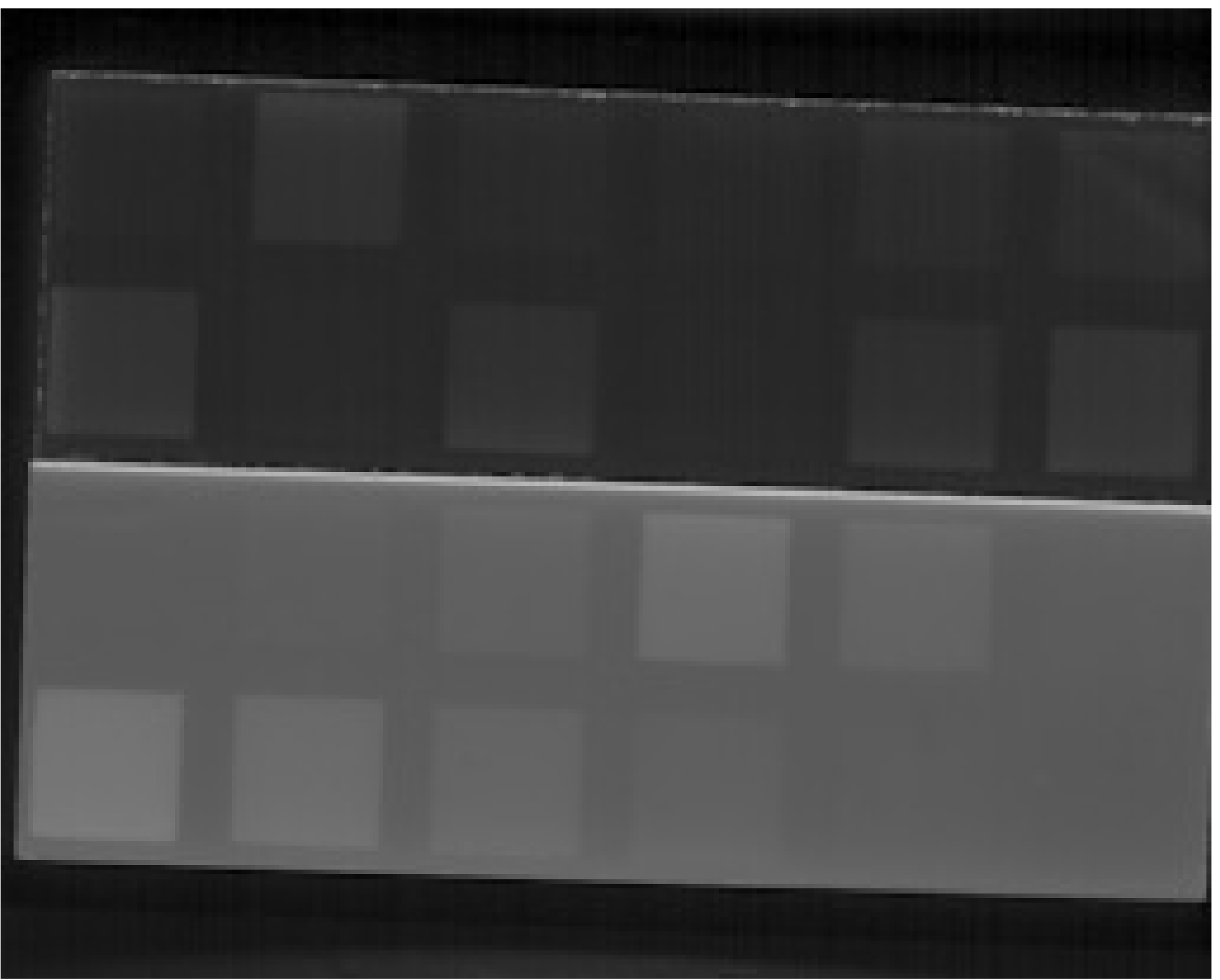}
\end{subfigure}%
\begin{subfigure}{0.06\textwidth}
\centering
\includegraphics[width=\textwidth]{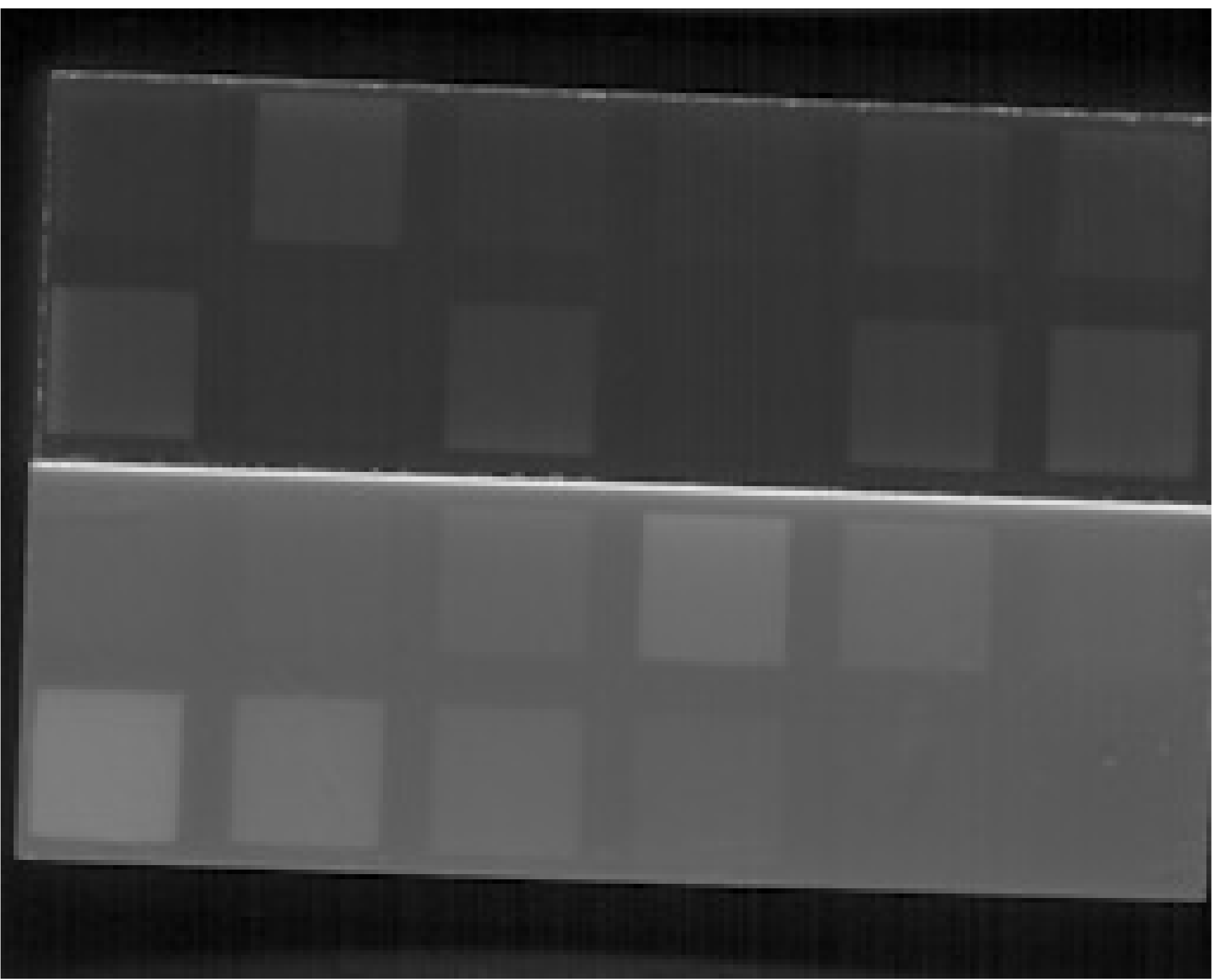}
\end{subfigure}%
\begin{subfigure}{0.06\textwidth}
\centering
\includegraphics[width=\textwidth]{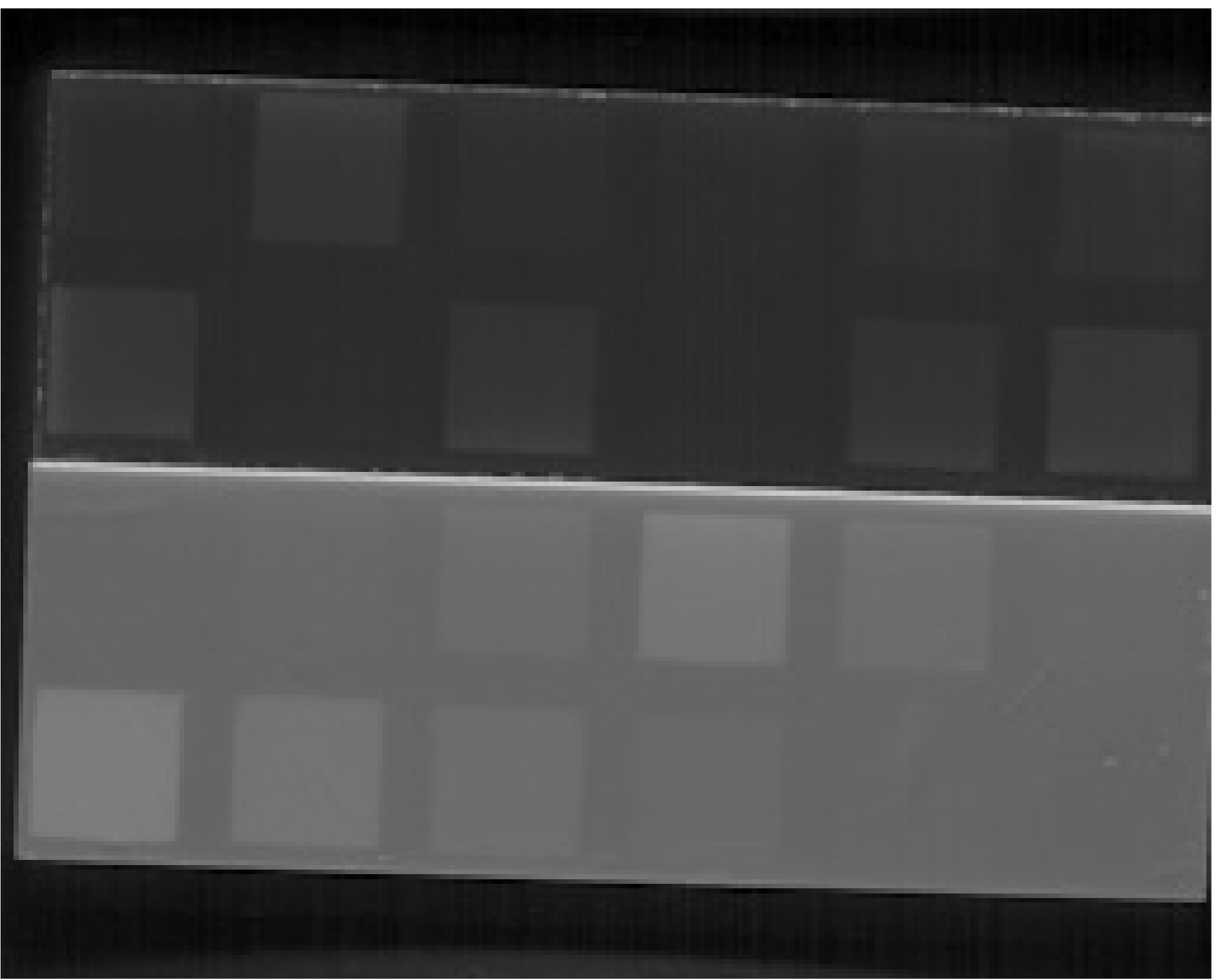}
\end{subfigure}%
\begin{subfigure}{0.06\textwidth}
\centering
\includegraphics[width=\textwidth]{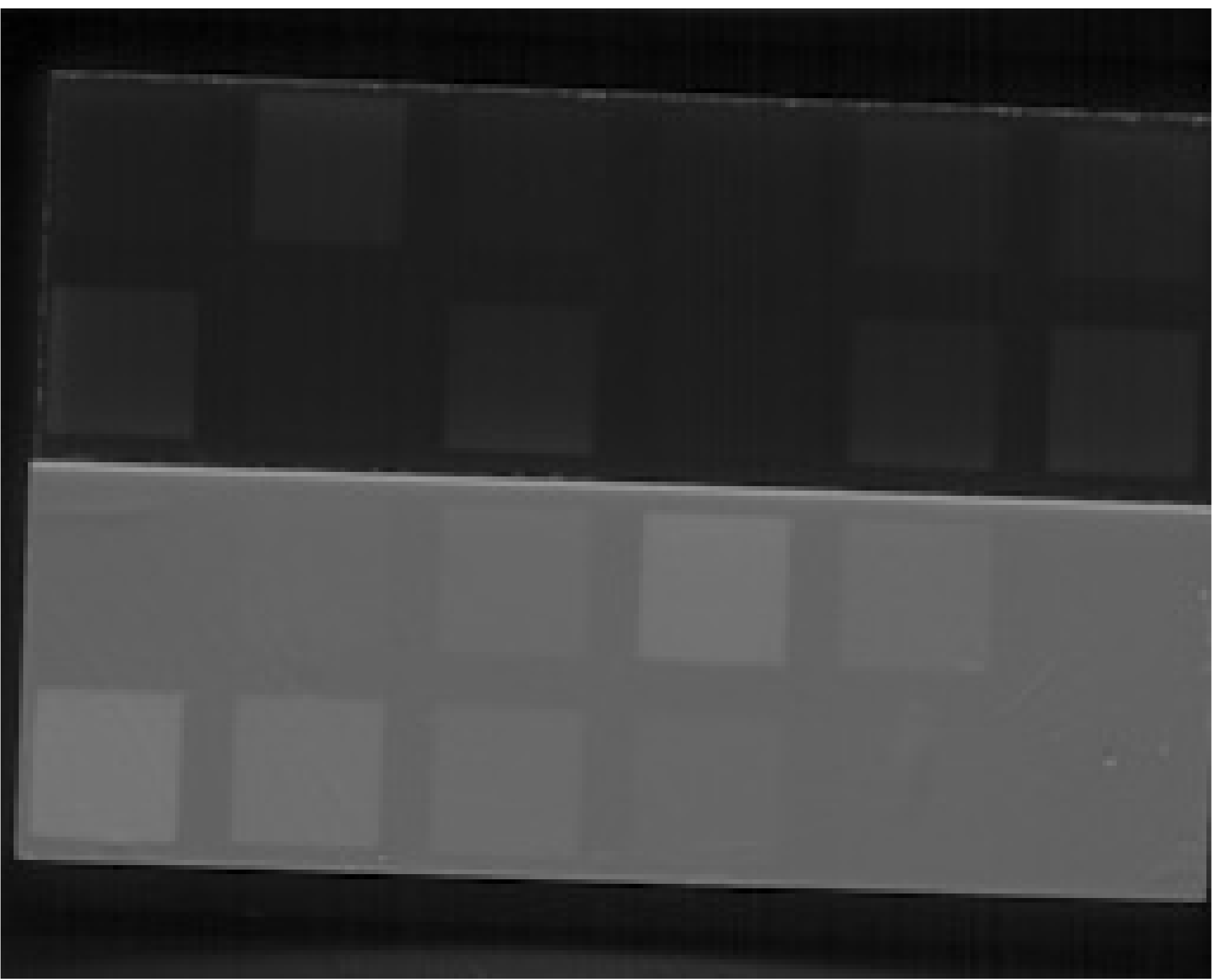}
\end{subfigure}%
\begin{subfigure}{0.06\textwidth}
\centering
\includegraphics[width=\textwidth]{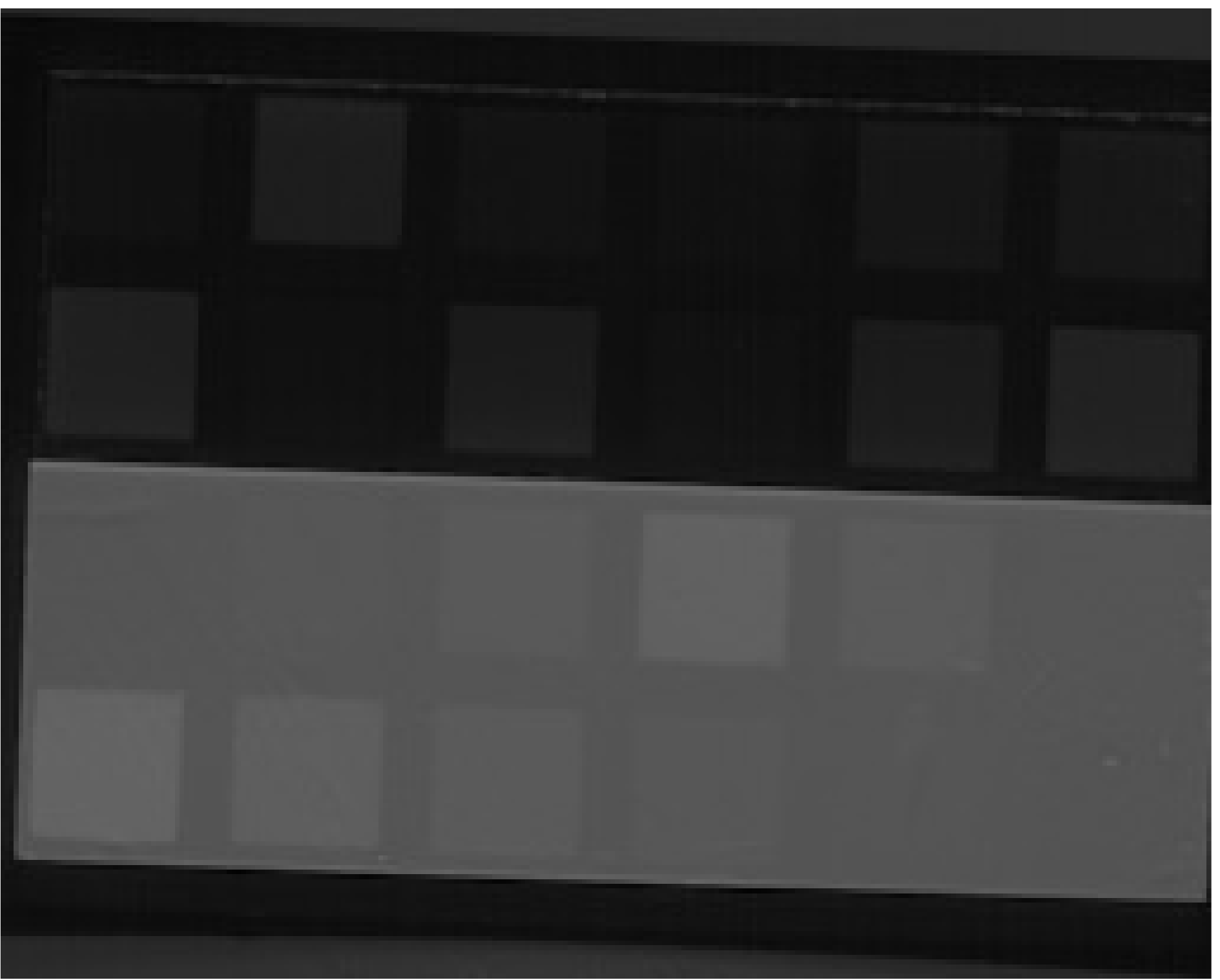}
\end{subfigure}%
\begin{subfigure}{0.06\textwidth}
\centering
\includegraphics[width=\textwidth]{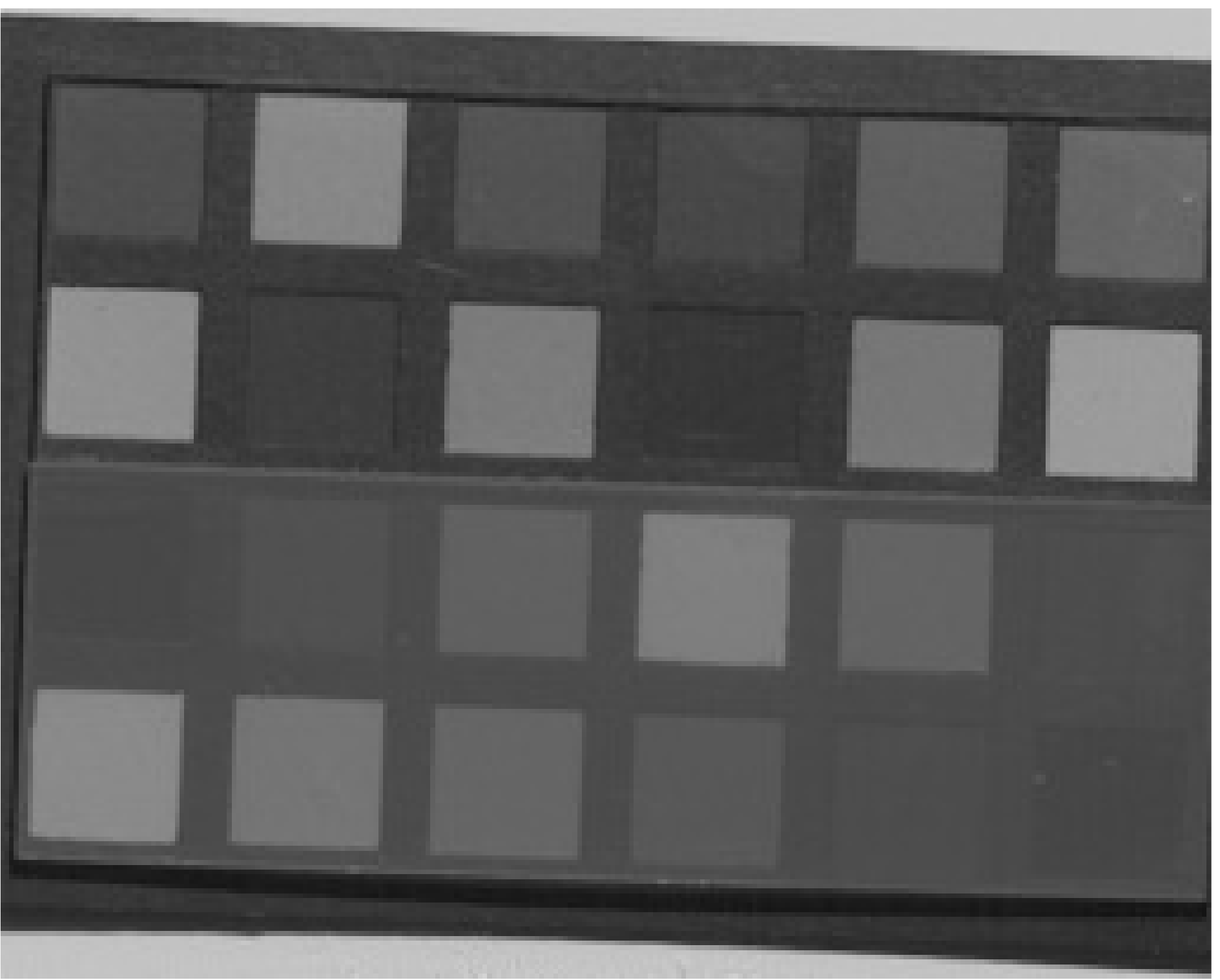}
\end{subfigure}%
\begin{subfigure}{0.06\textwidth}
\centering
\includegraphics[width=\textwidth]{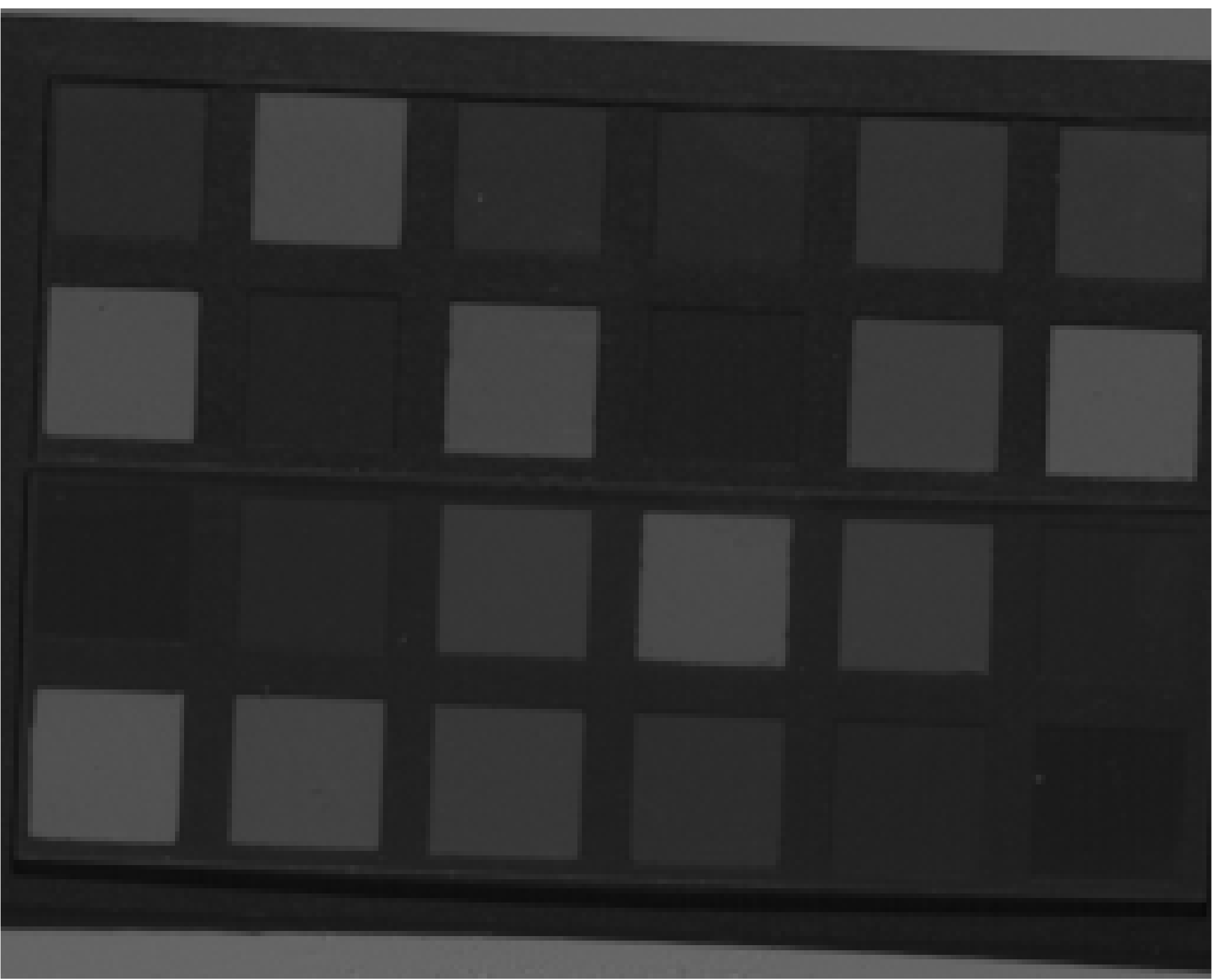}
\end{subfigure}%
\begin{subfigure}{0.06\textwidth}
\centering
\includegraphics[width=\textwidth]{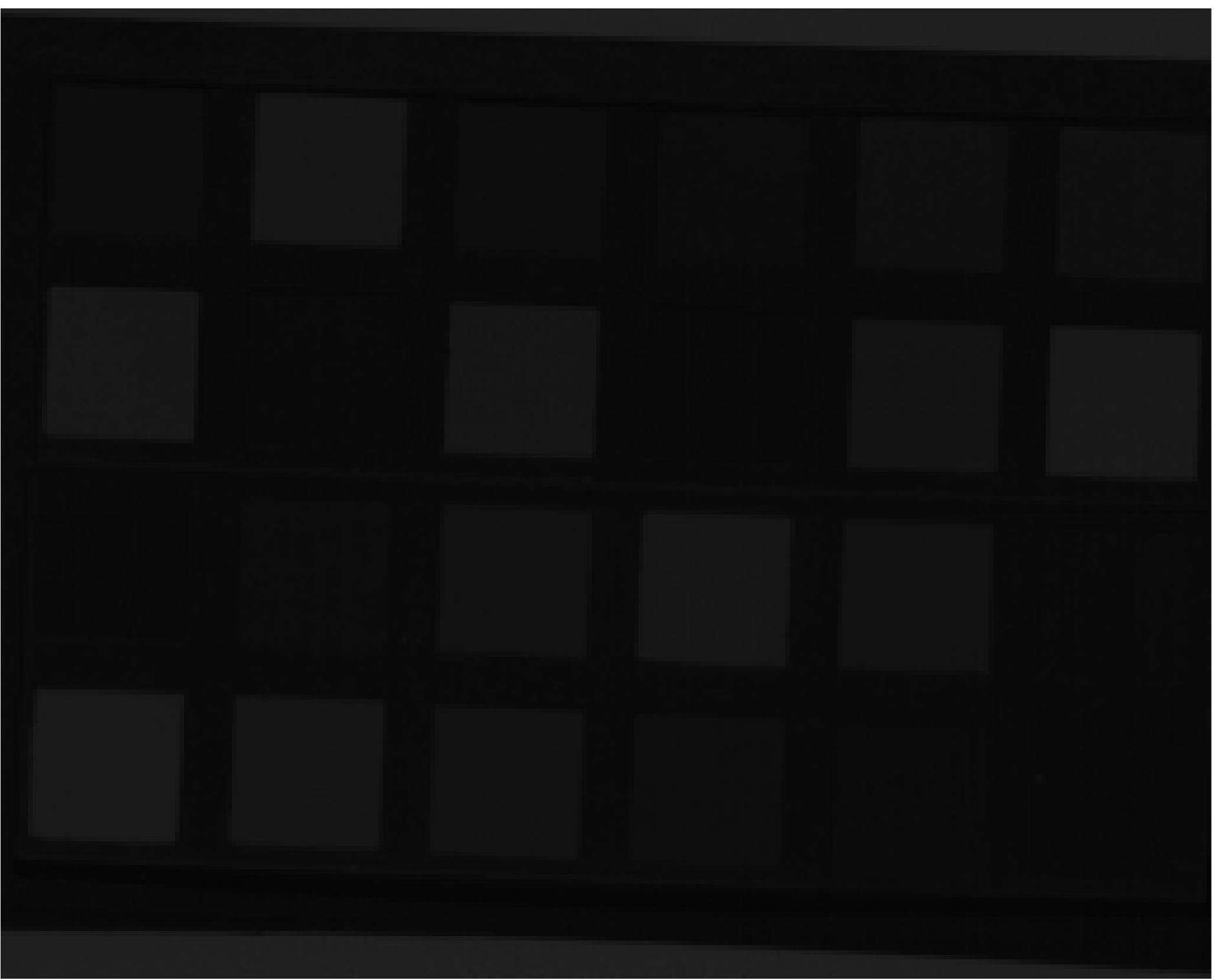}
\end{subfigure}%
\begin{subfigure}{0.06\textwidth}
\centering
\includegraphics[width=\textwidth]{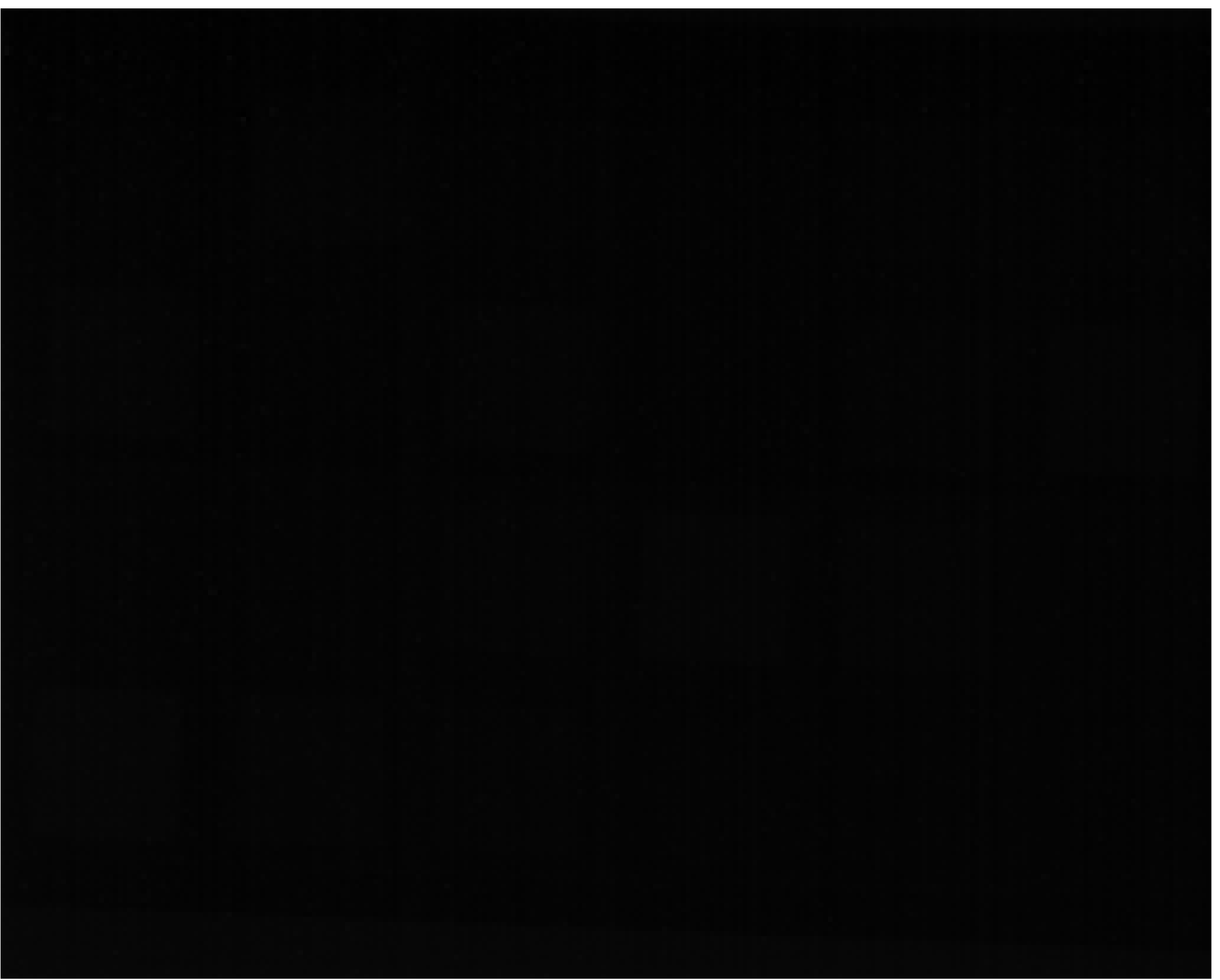}
\end{subfigure}%
\begin{subfigure}{0.06\textwidth}
\centering
\includegraphics[width=\textwidth]{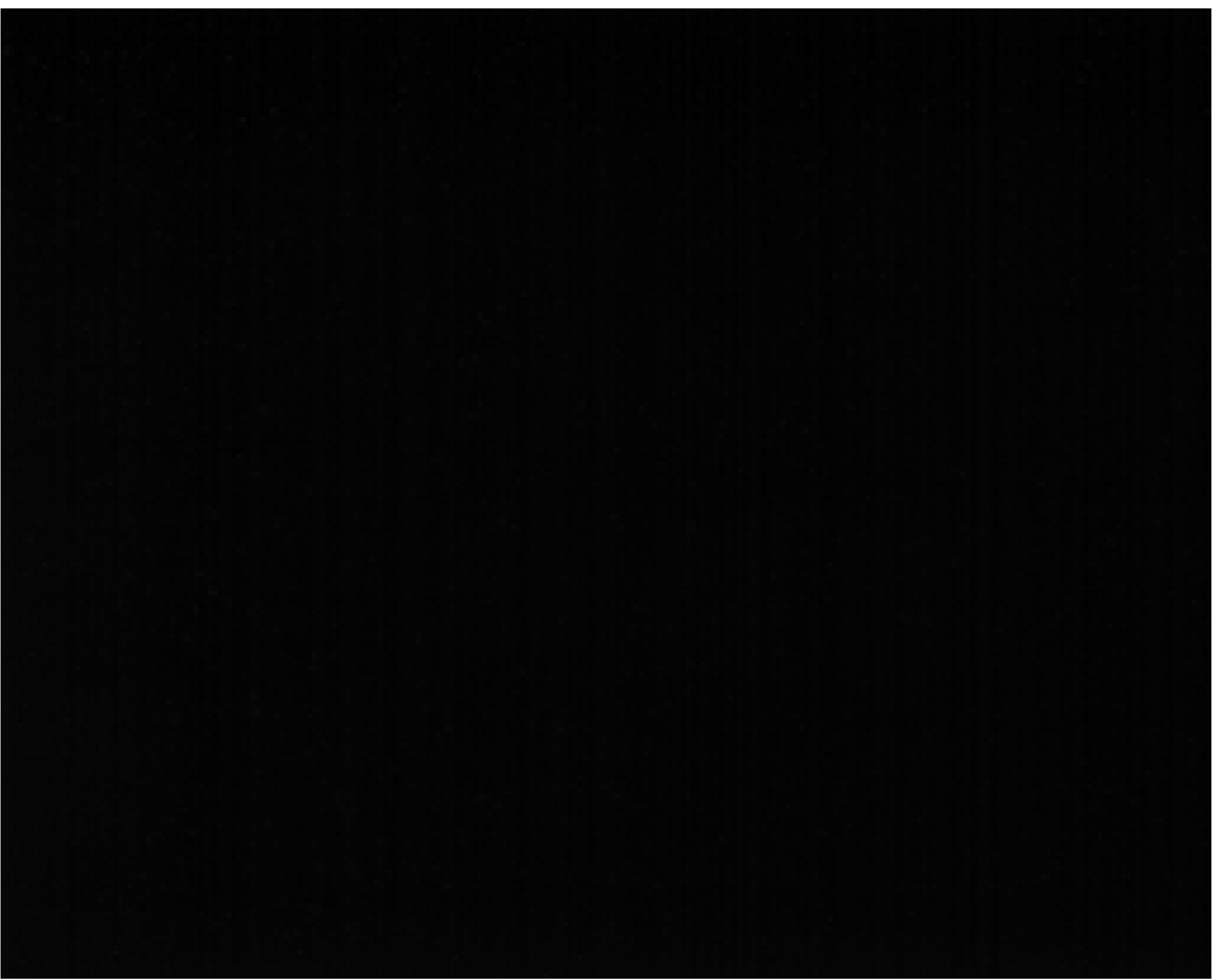}
\end{subfigure}%
\begin{subfigure}{0.06\textwidth}
\centering
\includegraphics[width=\textwidth]{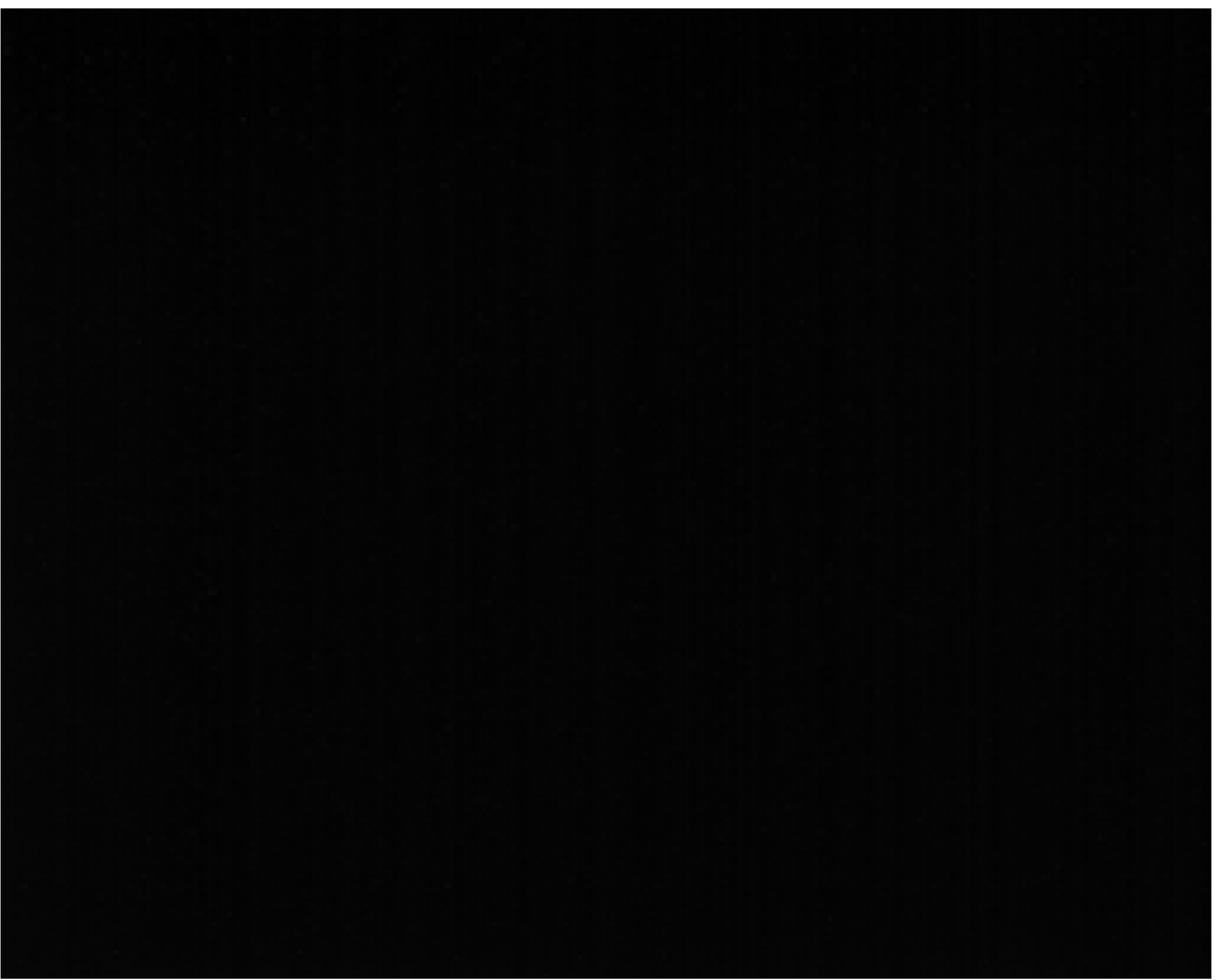}
\end{subfigure}%
\begin{subfigure}{0.06\textwidth}
\centering
\includegraphics[width=\textwidth]{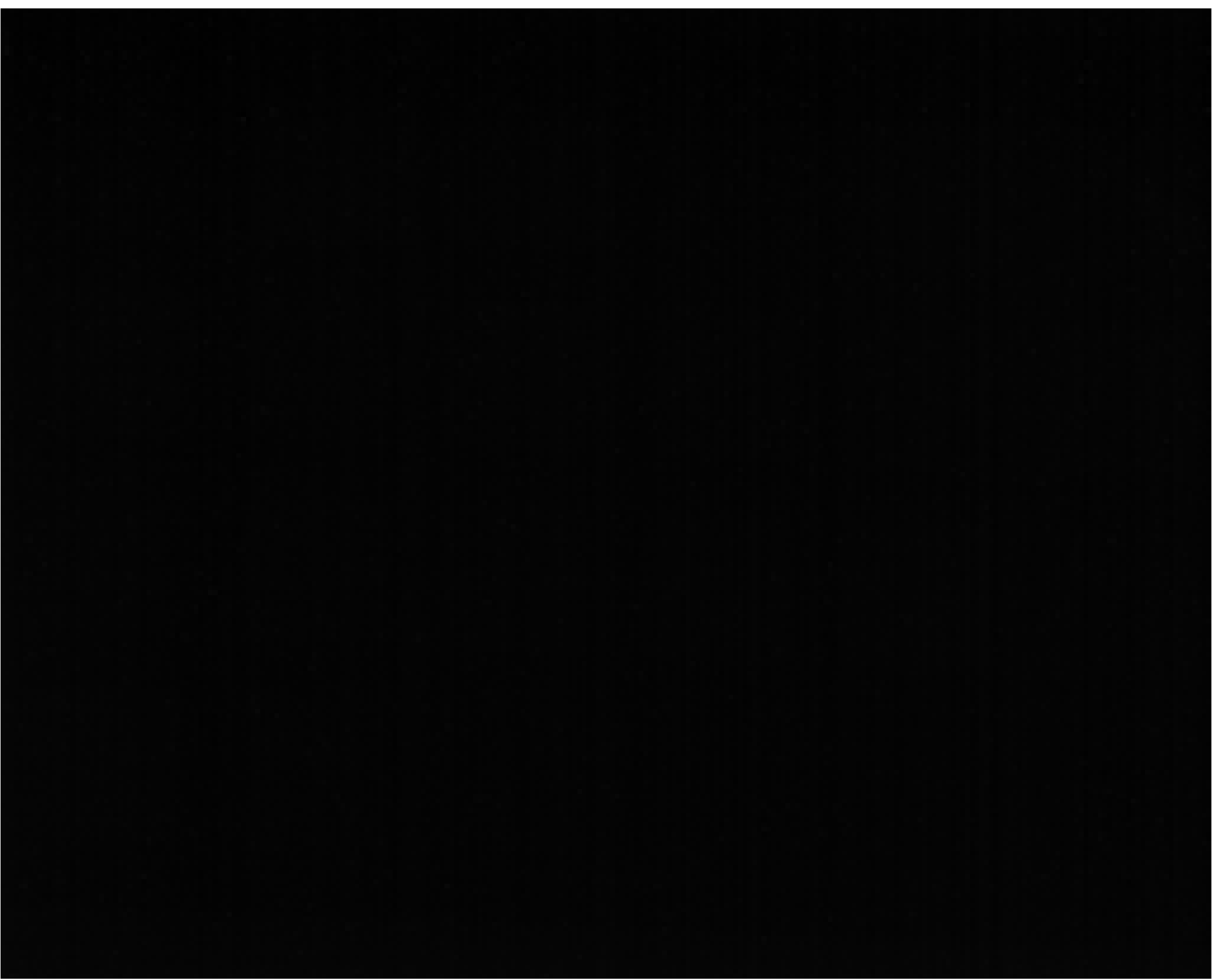}
\end{subfigure}%
\begin{subfigure}{0.06\textwidth}
\centering
\includegraphics[width=\textwidth]{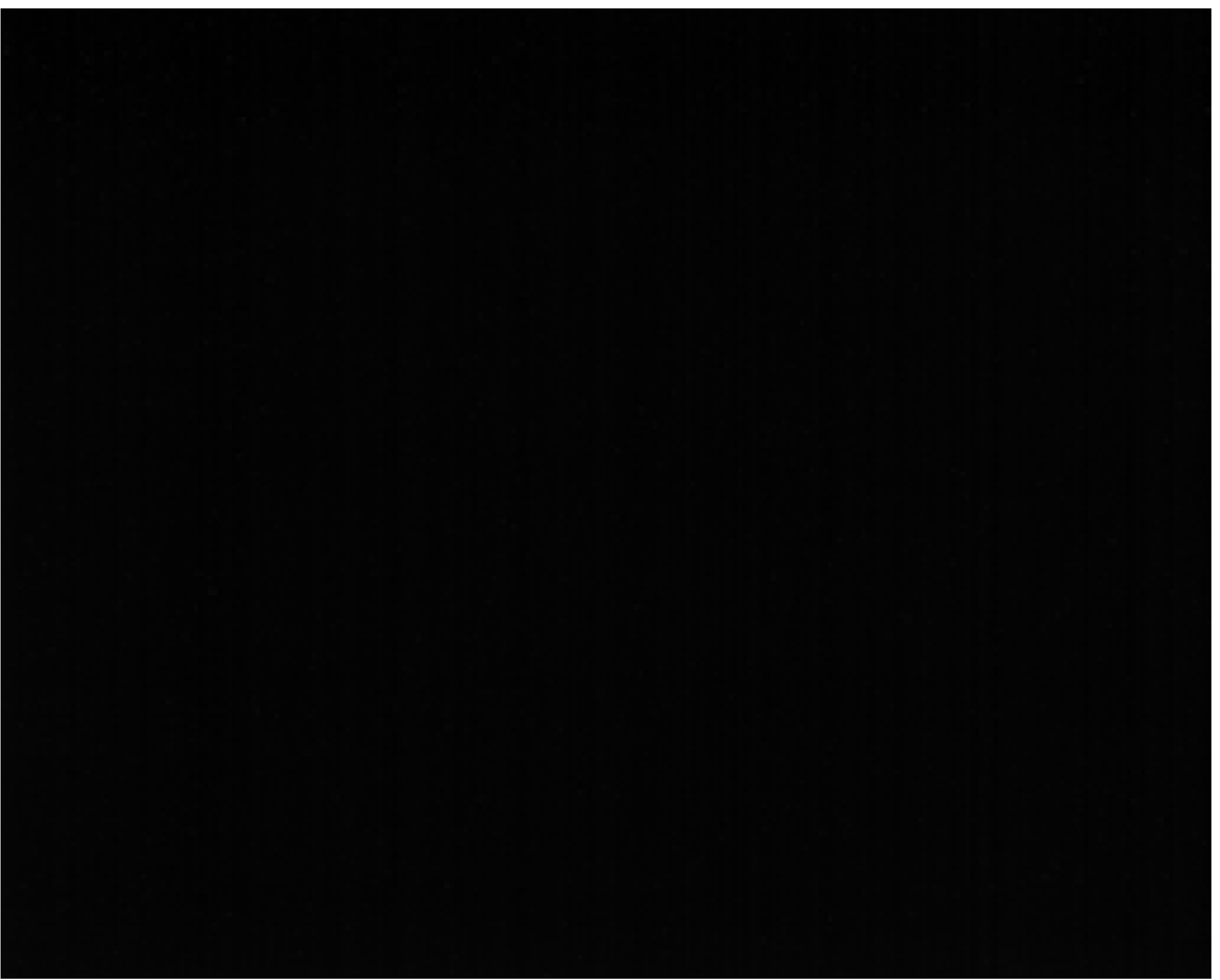}
\end{subfigure}\\%
\begin{subfigure}{0.06\textwidth}
\captionsetup{justification=raggedright,font=scriptsize}
\caption*{637 to 663nm}
\end{subfigure}%
\begin{subfigure}{0.06\textwidth}
\centering
\includegraphics[width=\textwidth]{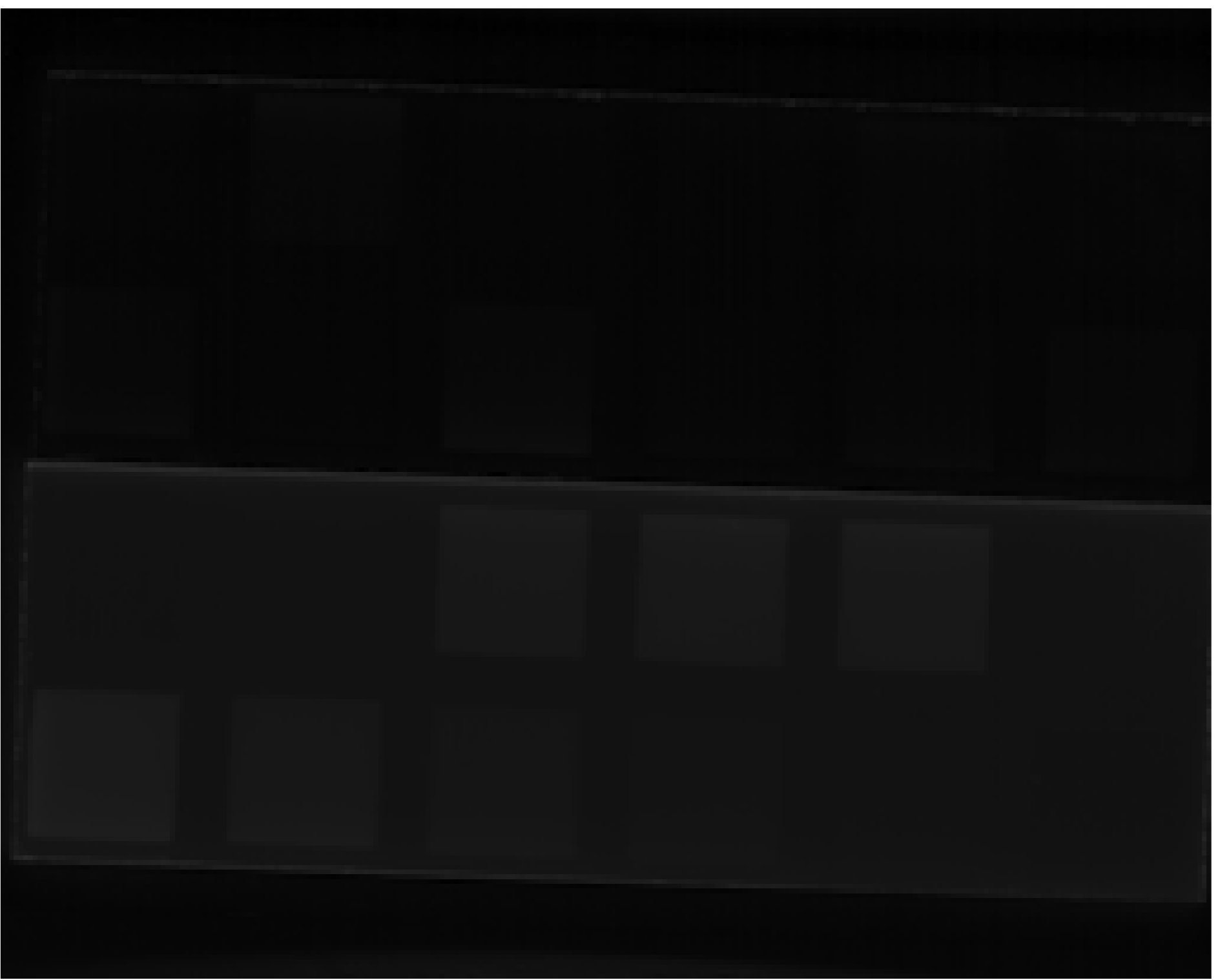}
\end{subfigure}%
\begin{subfigure}{0.06\textwidth}
\centering
\includegraphics[width=\textwidth]{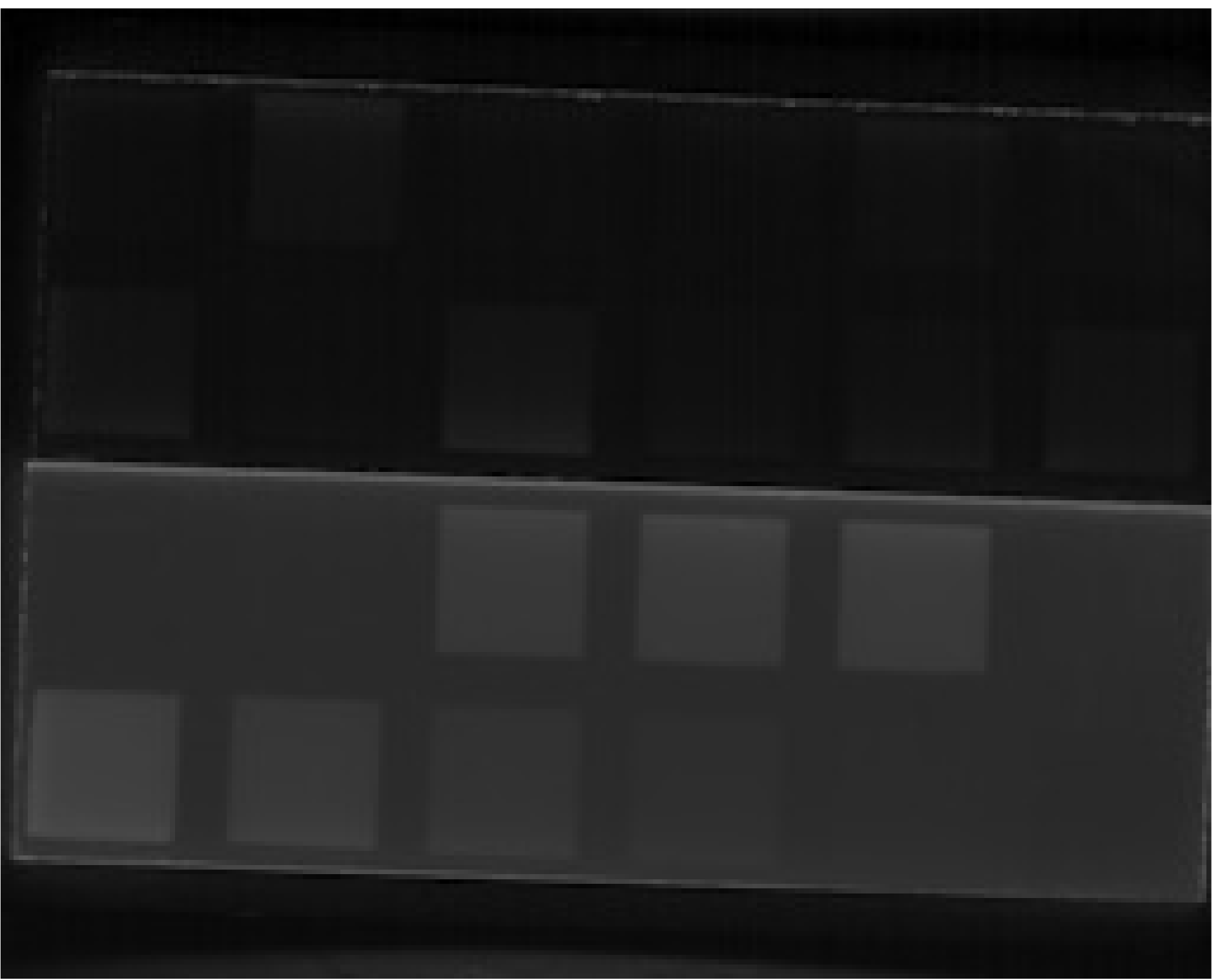}
\end{subfigure}%
\begin{subfigure}{0.06\textwidth}
\centering
\includegraphics[width=\textwidth]{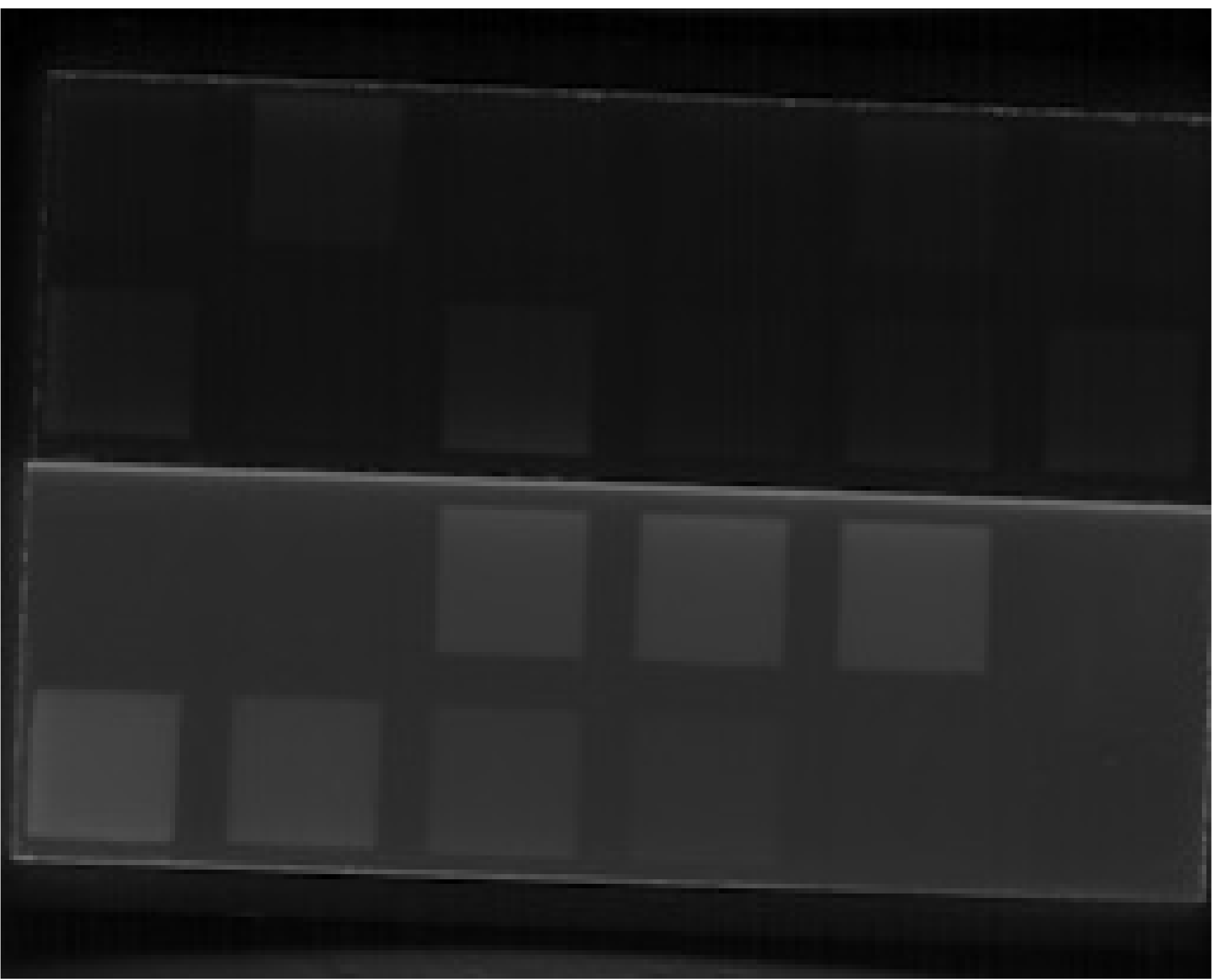}
\end{subfigure}%
\begin{subfigure}{0.06\textwidth}
\centering
\includegraphics[width=\textwidth]{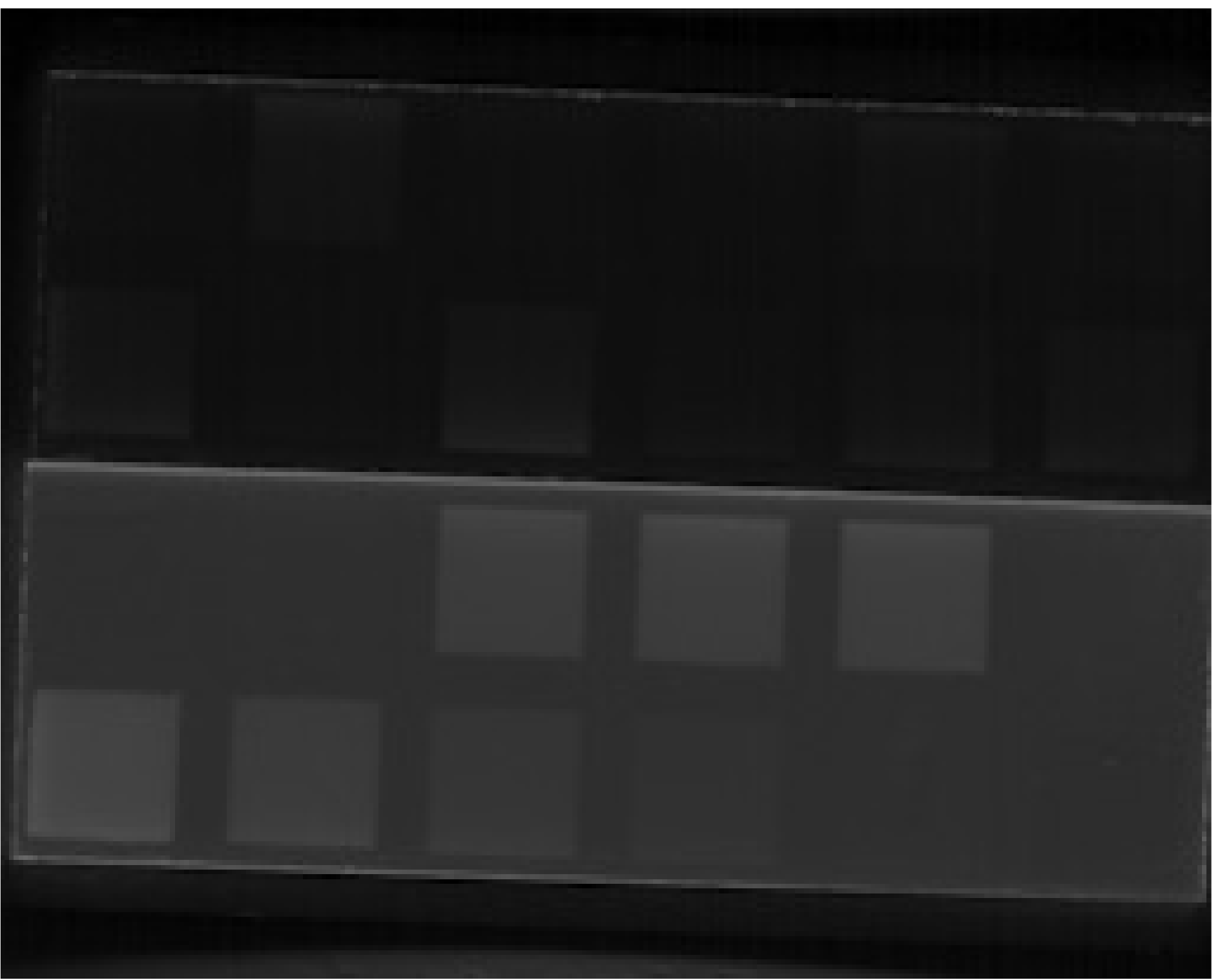}
\end{subfigure}%
\begin{subfigure}{0.06\textwidth}
\centering
\includegraphics[width=\textwidth]{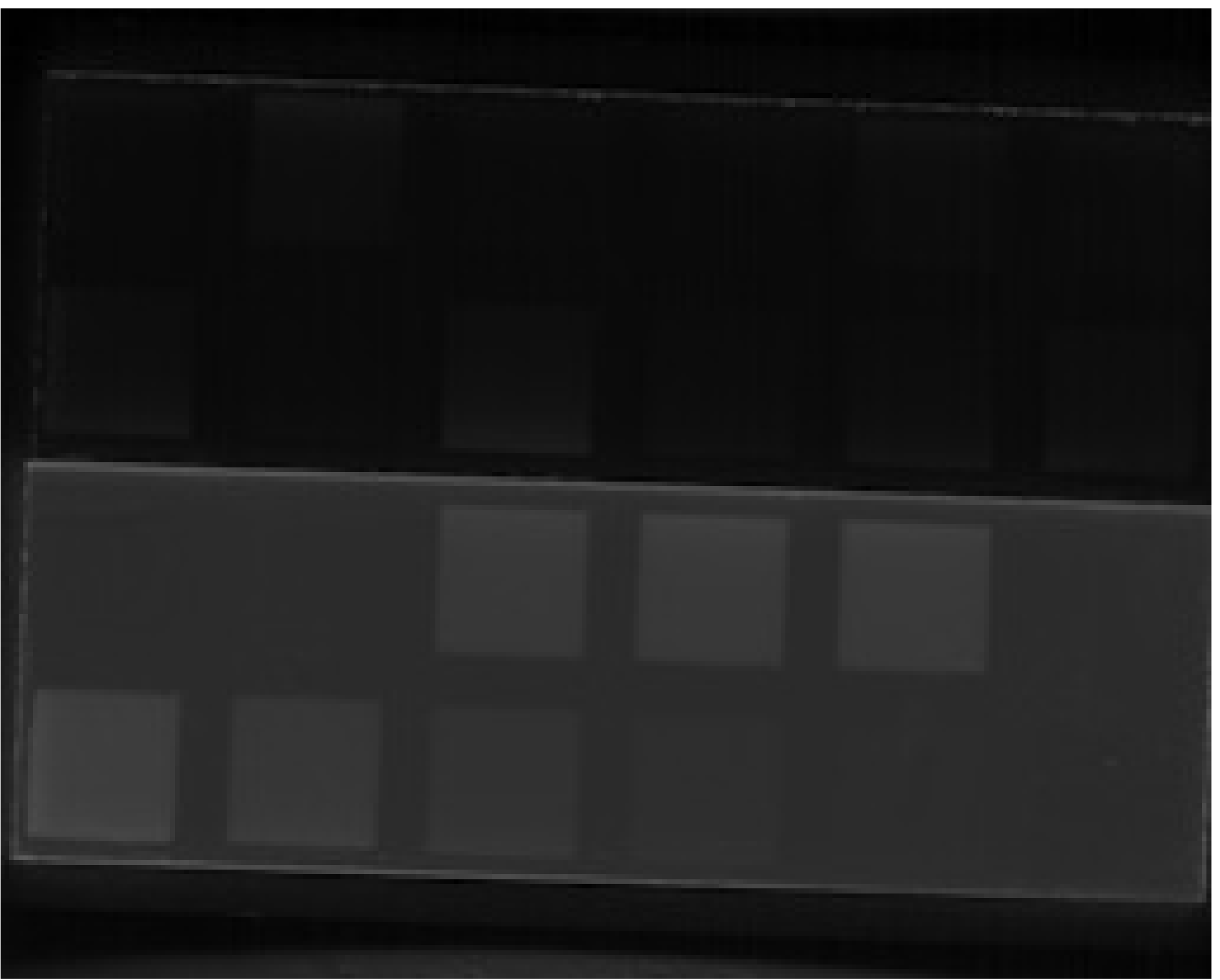}
\end{subfigure}%
\begin{subfigure}{0.06\textwidth}
\centering
\includegraphics[width=\textwidth]{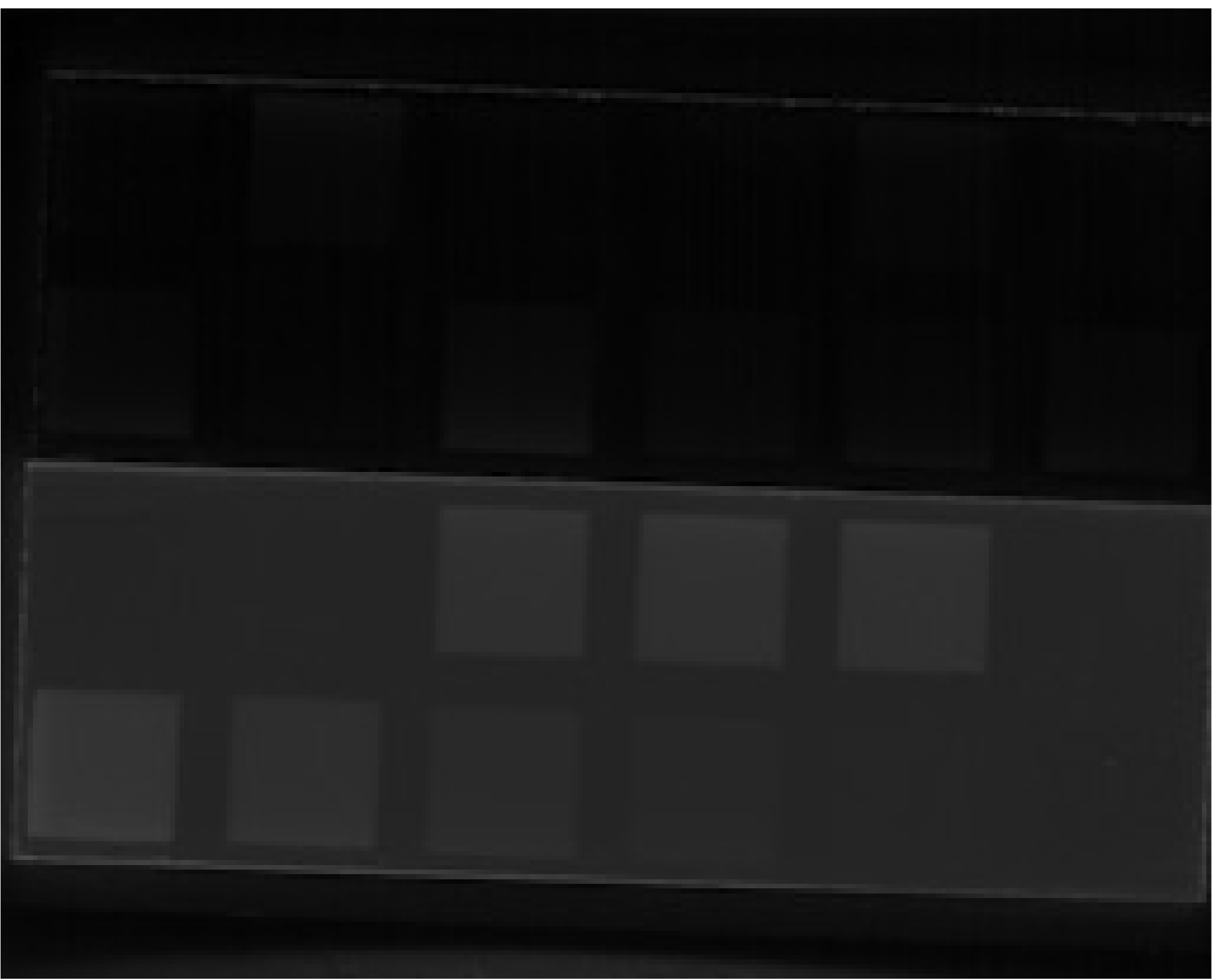}
\end{subfigure}%
\begin{subfigure}{0.06\textwidth}
\centering
\includegraphics[width=\textwidth]{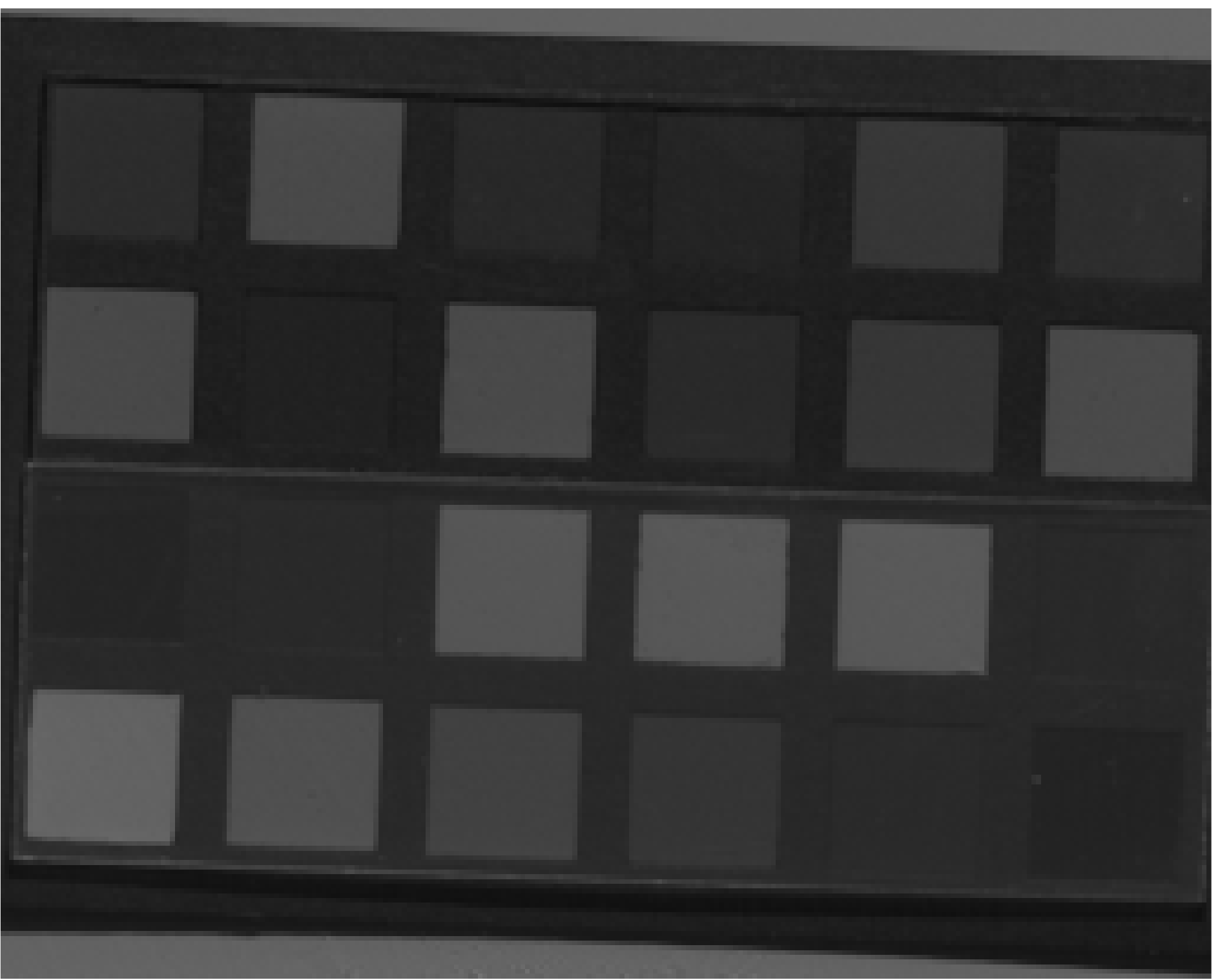}
\end{subfigure}%
\begin{subfigure}{0.06\textwidth}
\centering
\includegraphics[width=\textwidth]{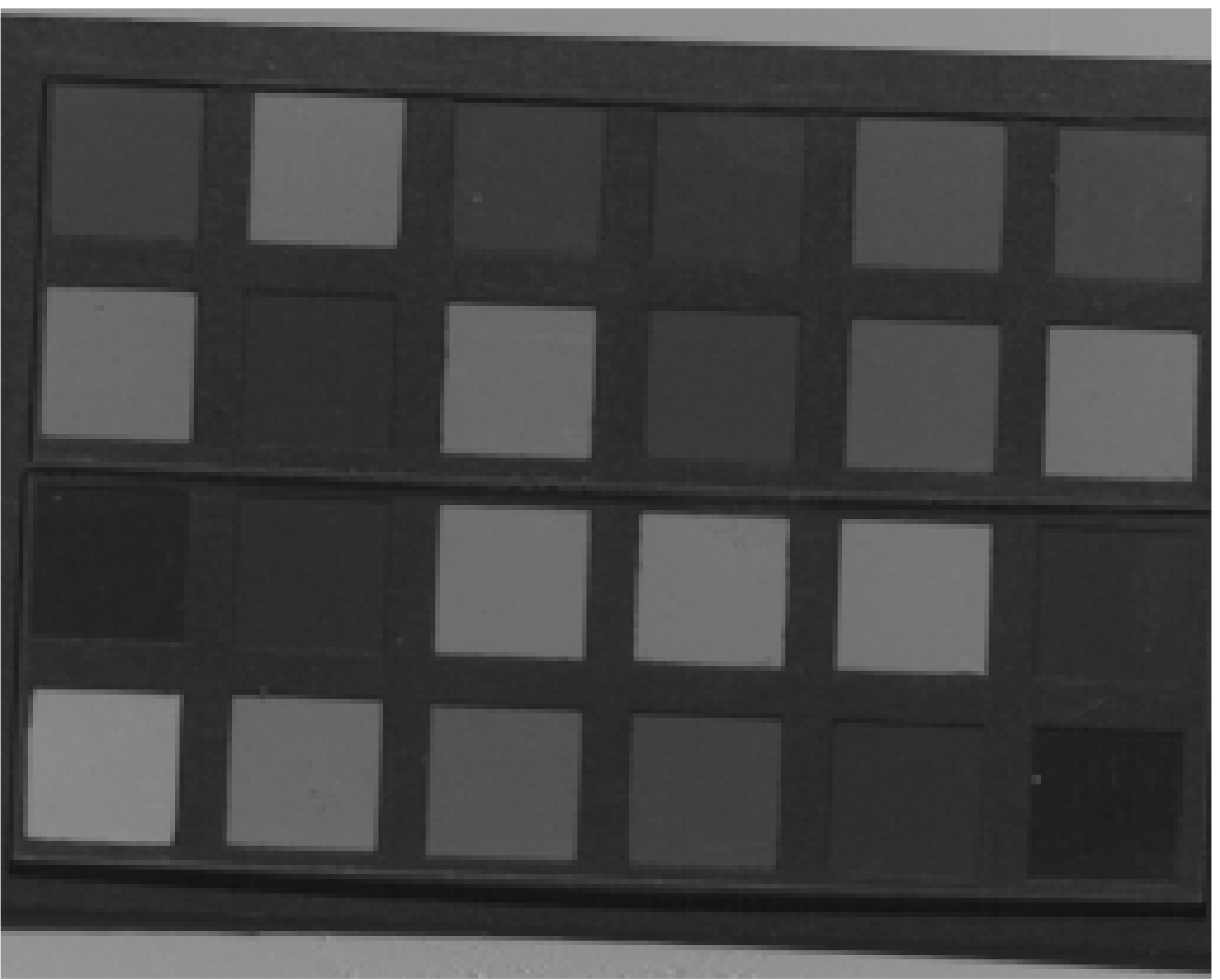}
\end{subfigure}%
\begin{subfigure}{0.06\textwidth}
\centering
\includegraphics[width=\textwidth]{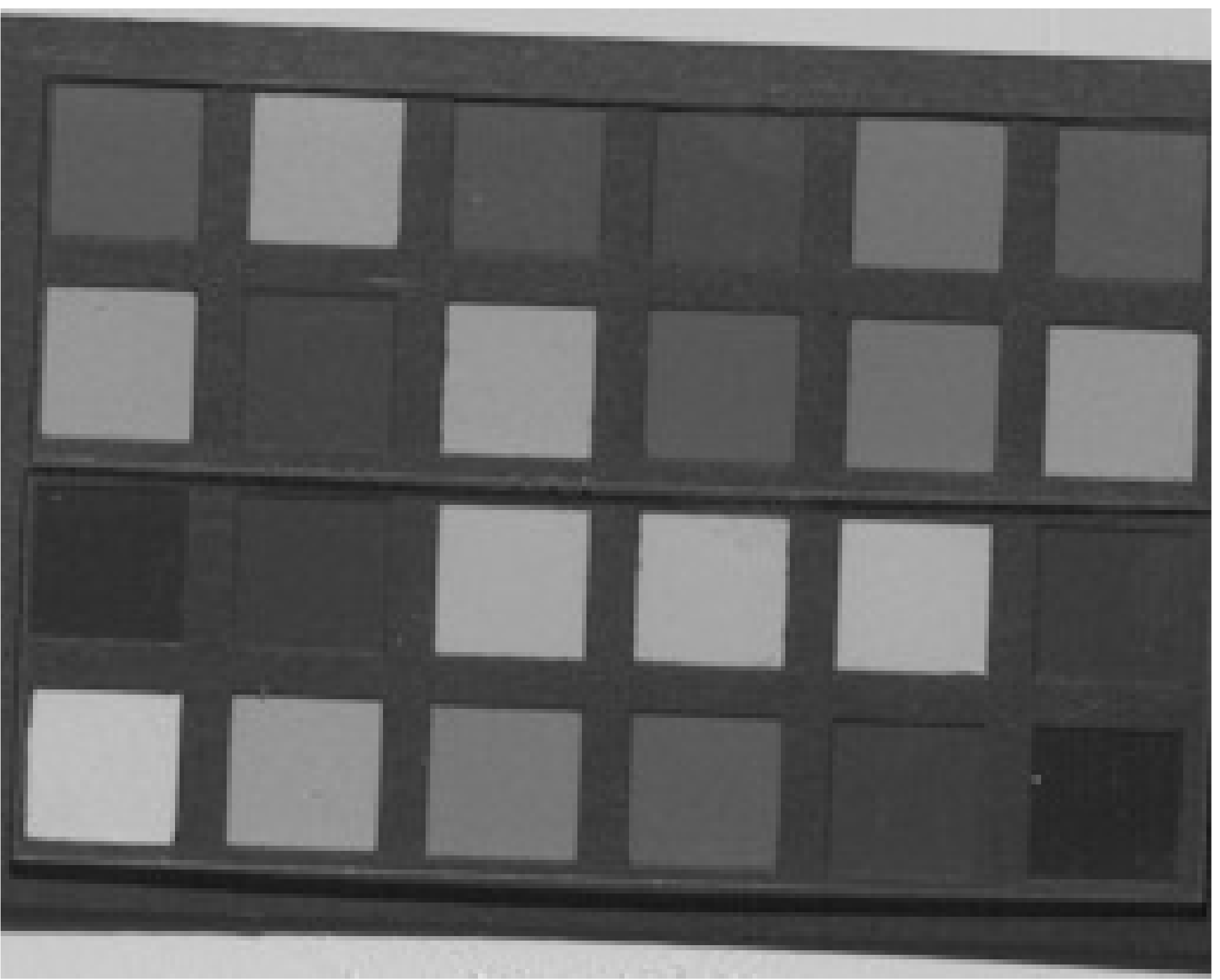}
\end{subfigure}%
\begin{subfigure}{0.06\textwidth}
\centering
\includegraphics[width=\textwidth]{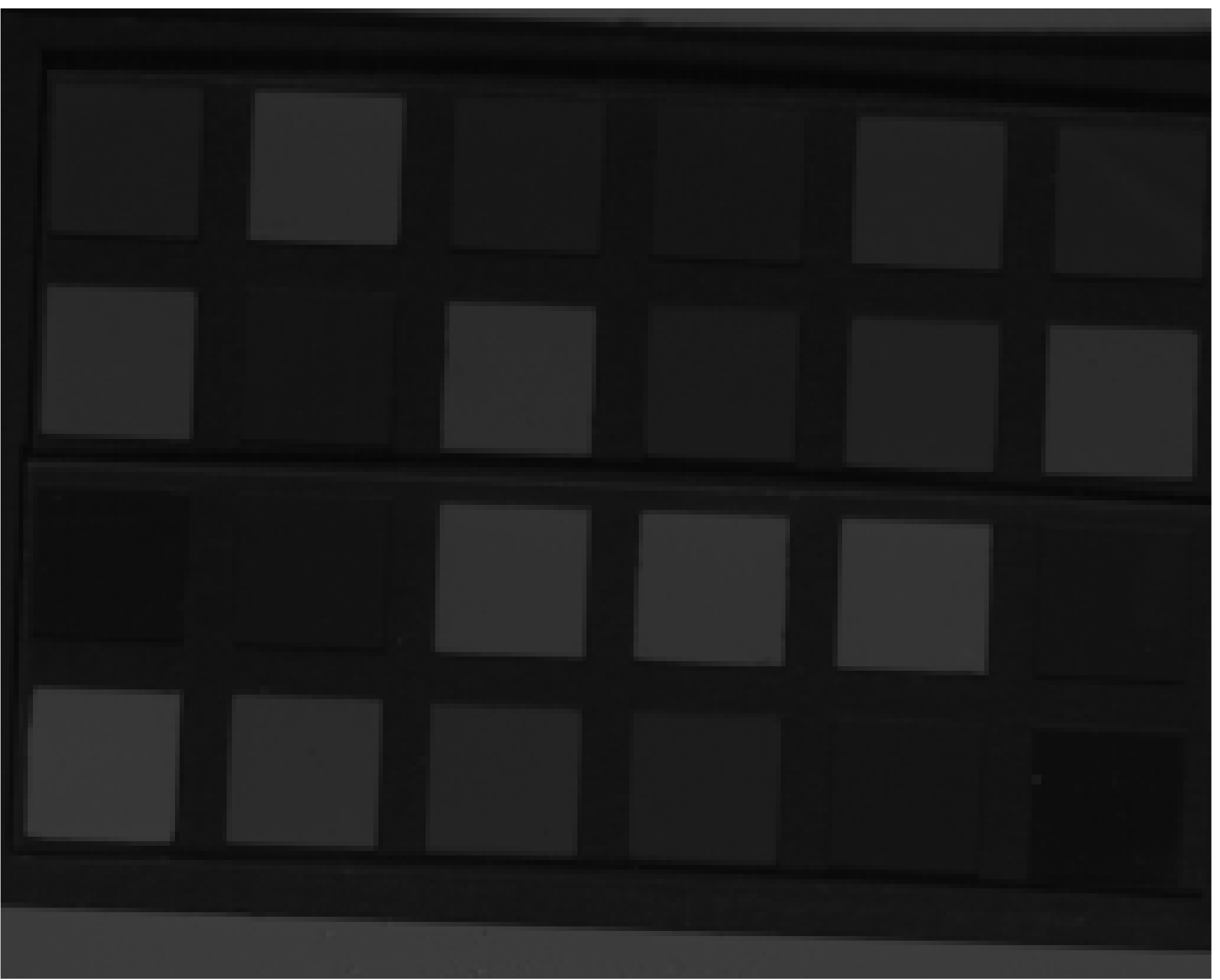}
\end{subfigure}%
\begin{subfigure}{0.06\textwidth}
\centering
\includegraphics[width=\textwidth]{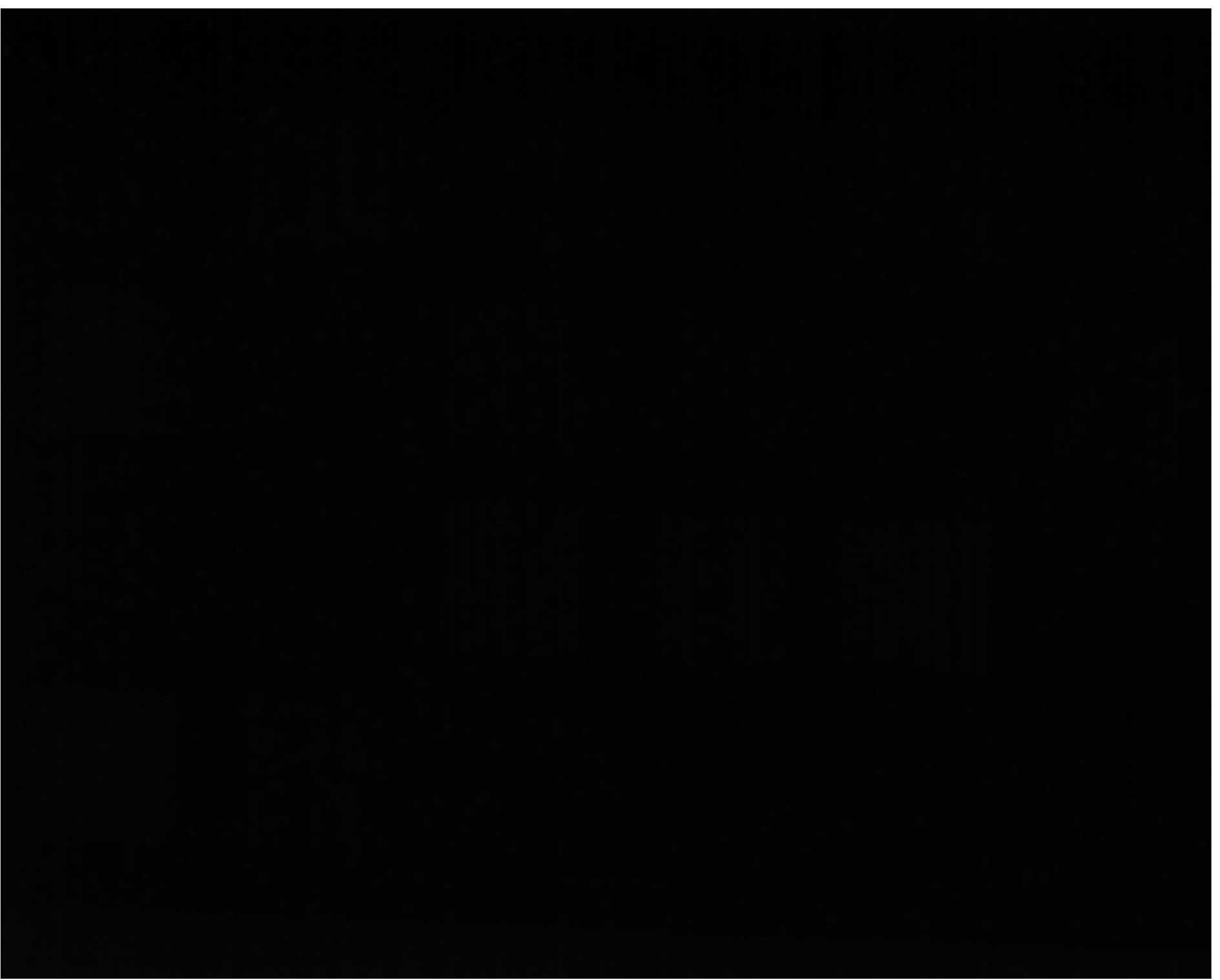}
\end{subfigure}%
\begin{subfigure}{0.06\textwidth}
\centering
\includegraphics[width=\textwidth]{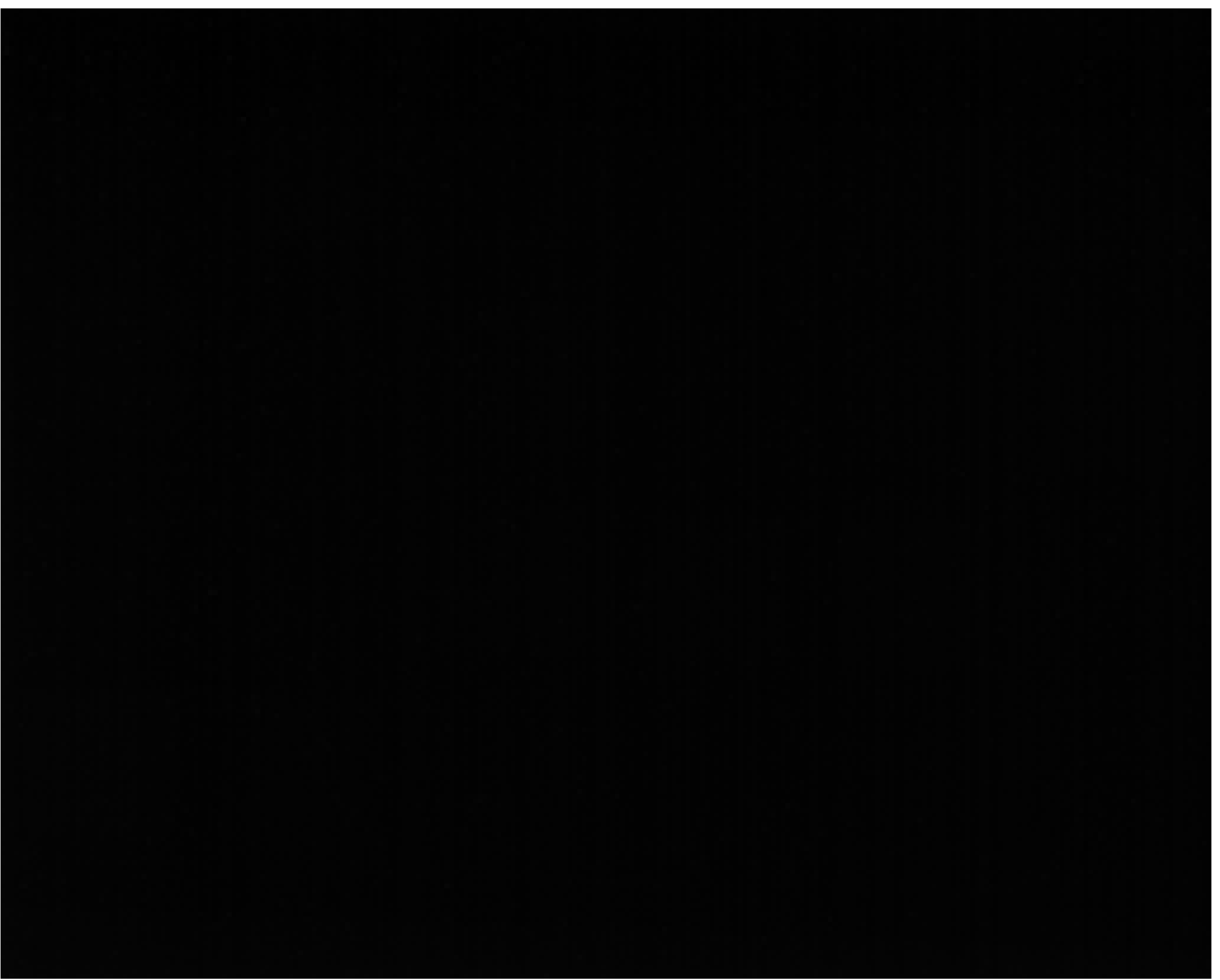}
\end{subfigure}%
\begin{subfigure}{0.06\textwidth}
\centering
\includegraphics[width=\textwidth]{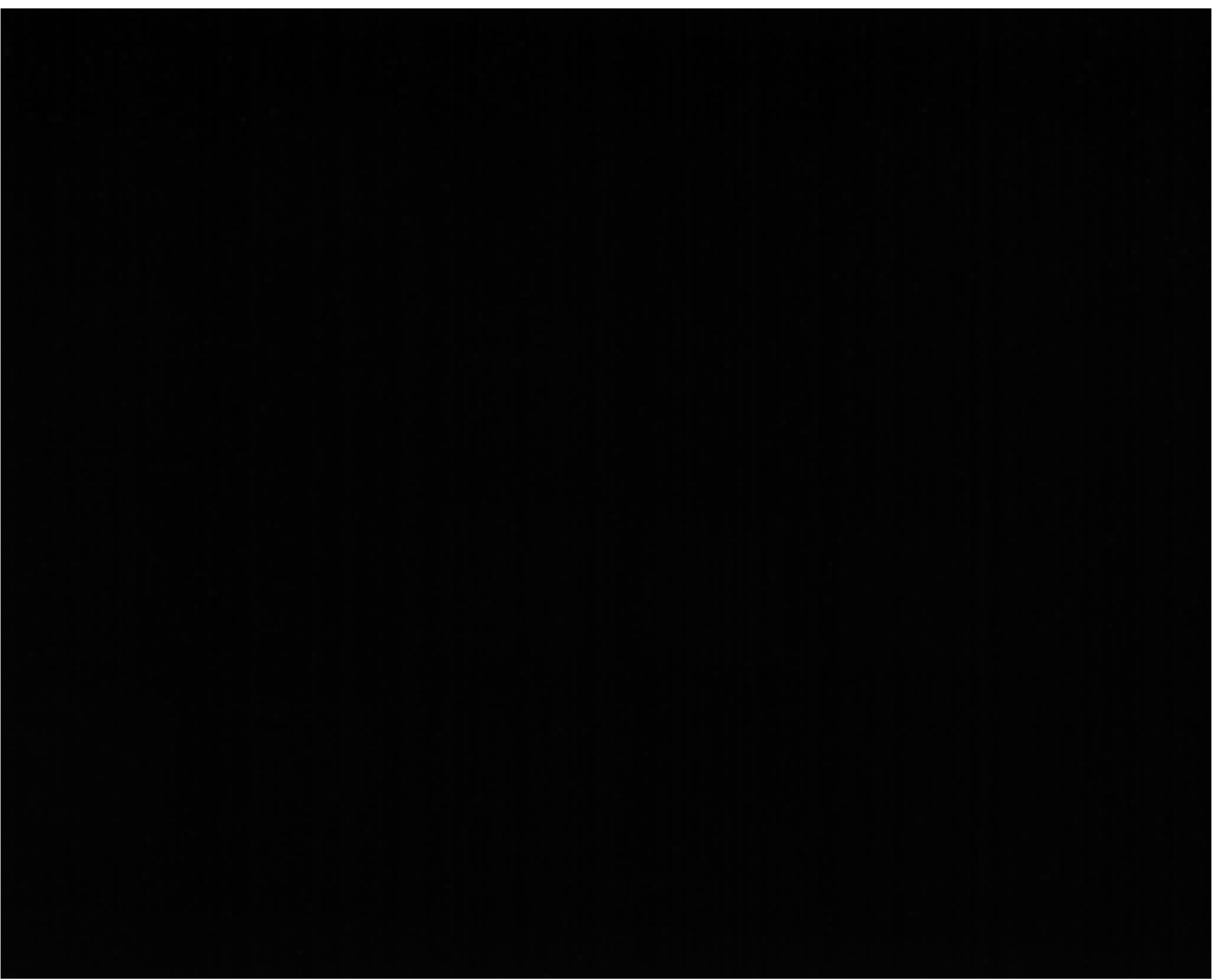}
\end{subfigure}%
\begin{subfigure}{0.06\textwidth}
\centering
\includegraphics[width=\textwidth]{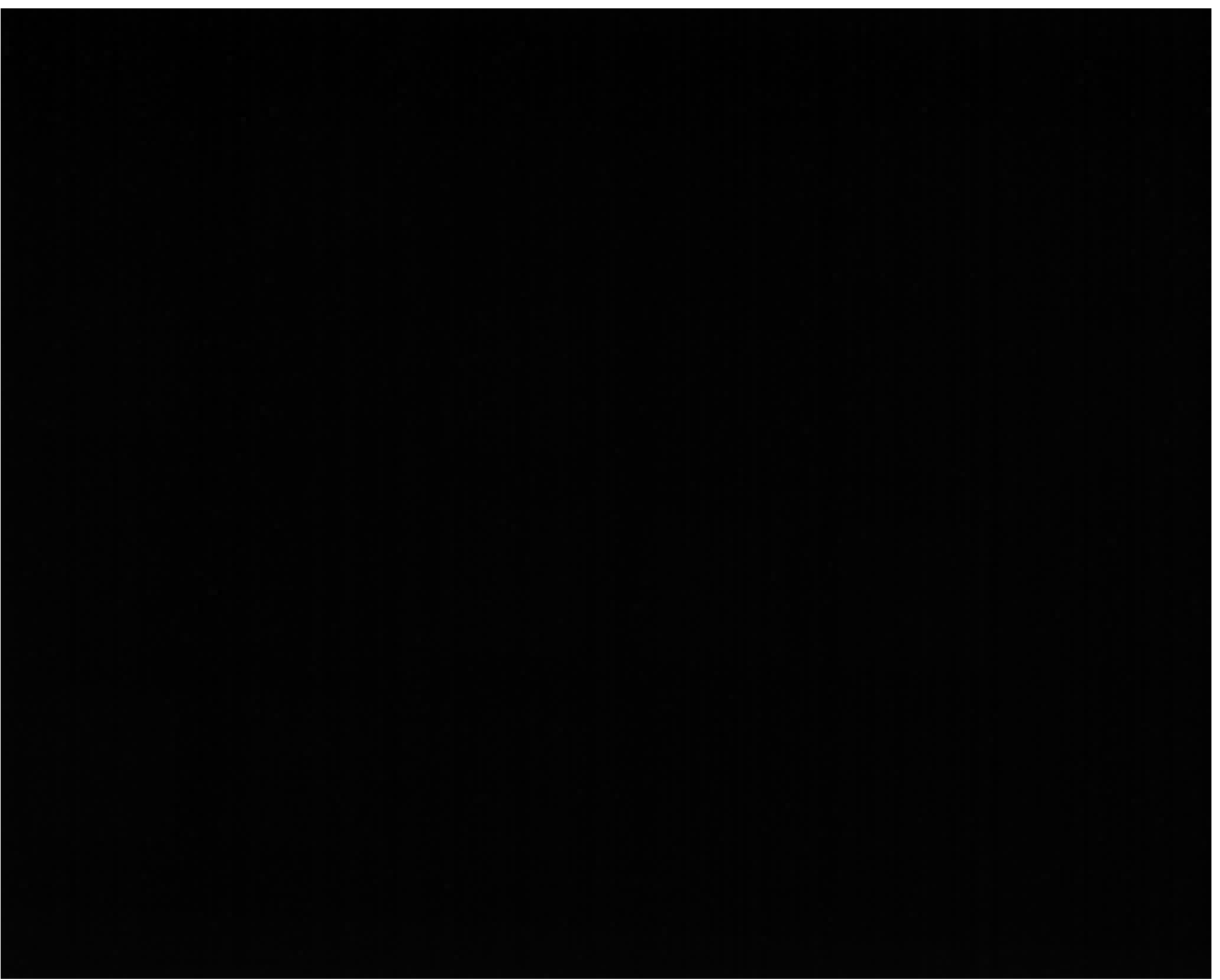}
\end{subfigure}\\%
\begin{subfigure}{0.06\textwidth}
\captionsetup{justification=raggedright,font=scriptsize}
\caption*{687 to 713nm}
\end{subfigure}%
\begin{subfigure}{0.06\textwidth}
\centering
\includegraphics[width=\textwidth]{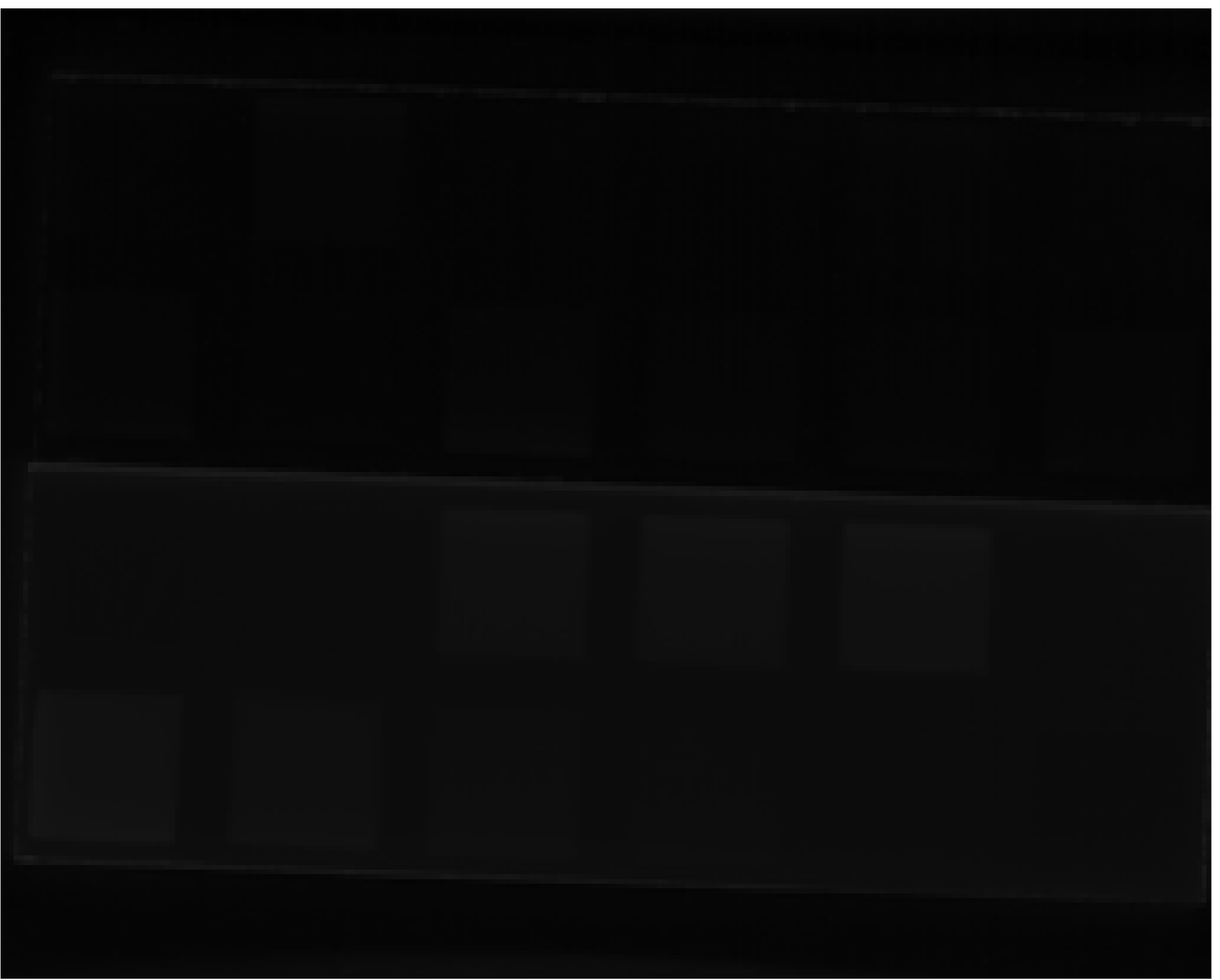}
\end{subfigure}%
\begin{subfigure}{0.06\textwidth}
\centering
\includegraphics[width=\textwidth]{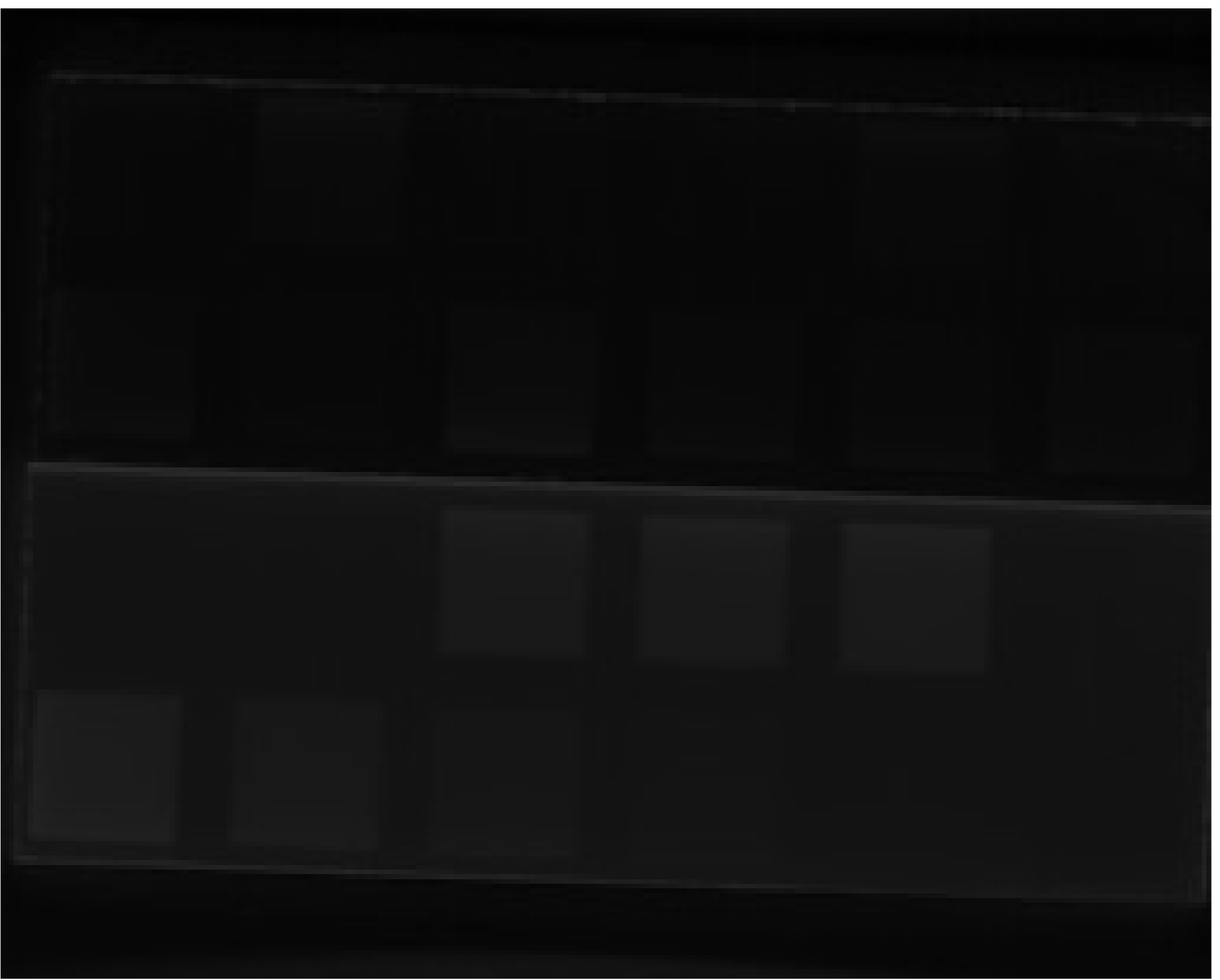}
\end{subfigure}%
\begin{subfigure}{0.06\textwidth}
\centering
\includegraphics[width=\textwidth]{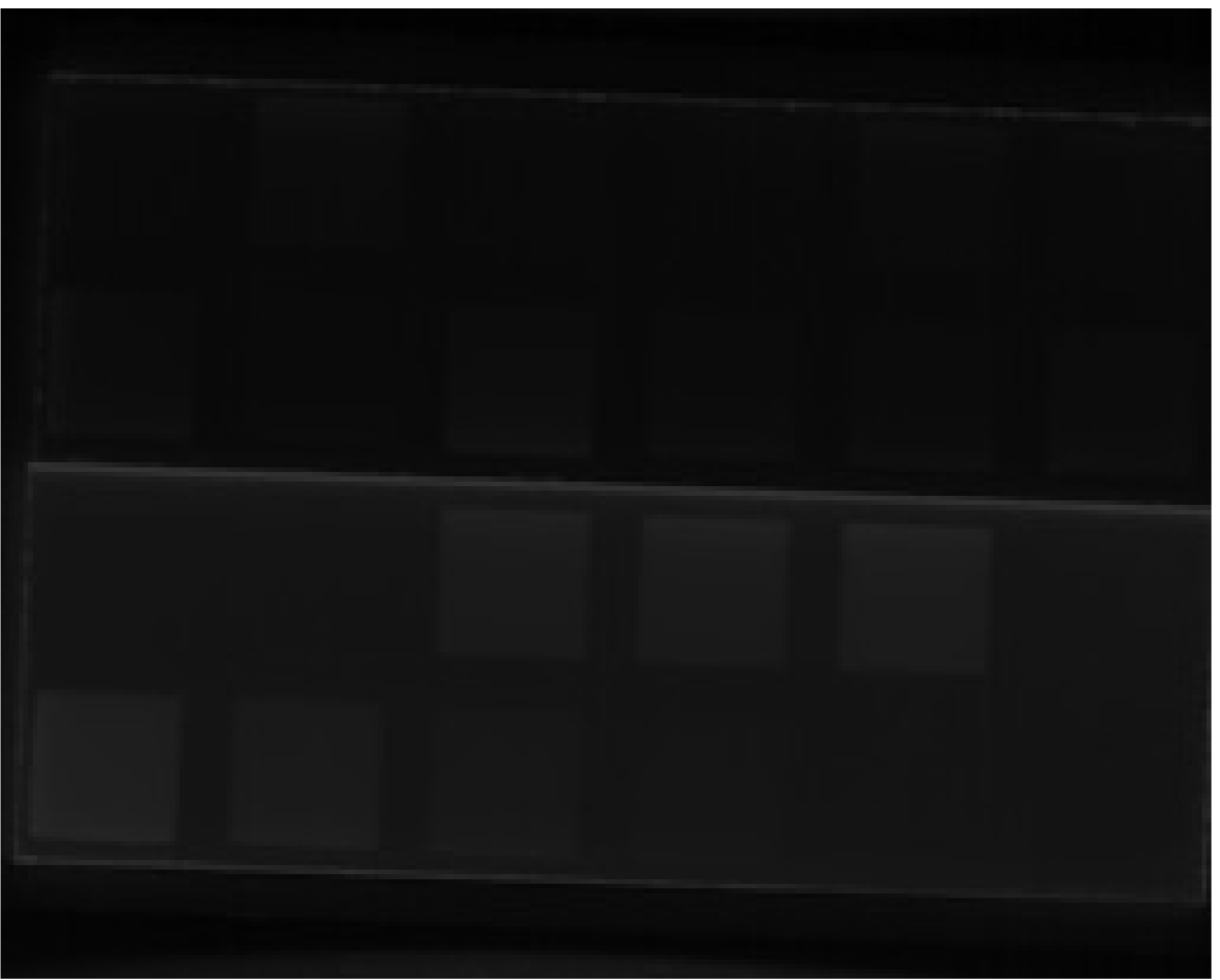}
\end{subfigure}%
\begin{subfigure}{0.06\textwidth}
\centering
\includegraphics[width=\textwidth]{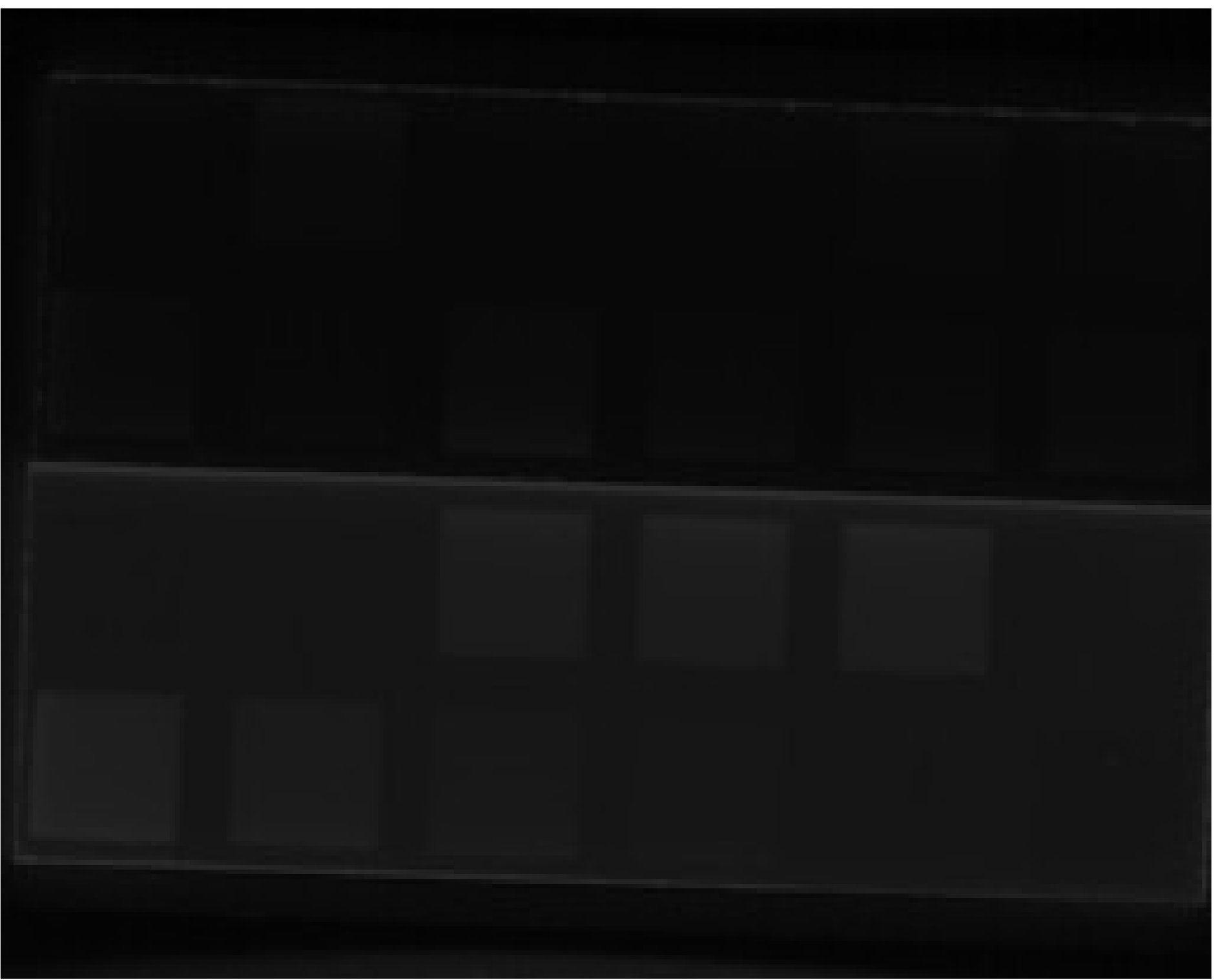}
\end{subfigure}%
\begin{subfigure}{0.06\textwidth}
\centering
\includegraphics[width=\textwidth]{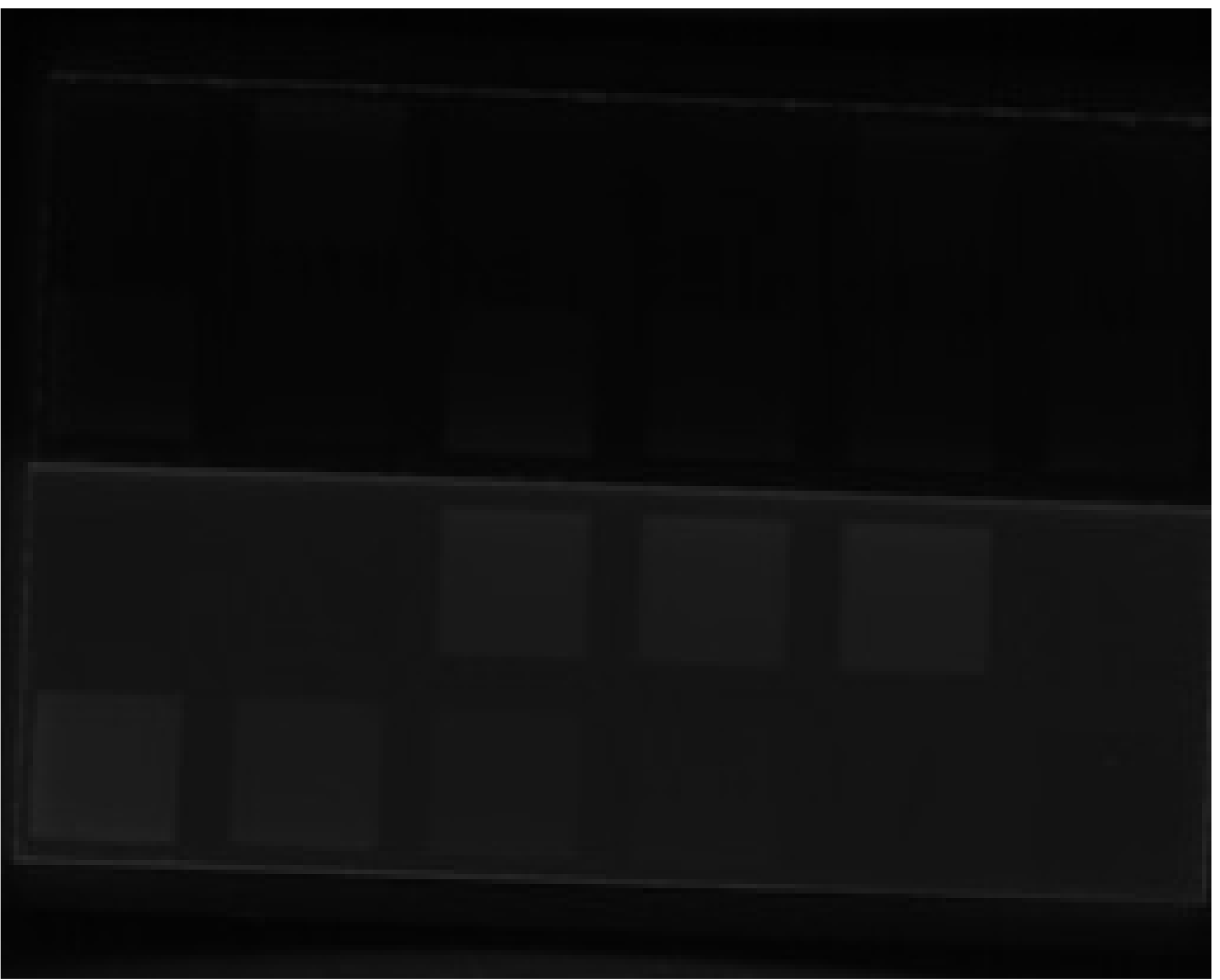}
\end{subfigure}%
\begin{subfigure}{0.06\textwidth}
\centering
\includegraphics[width=\textwidth]{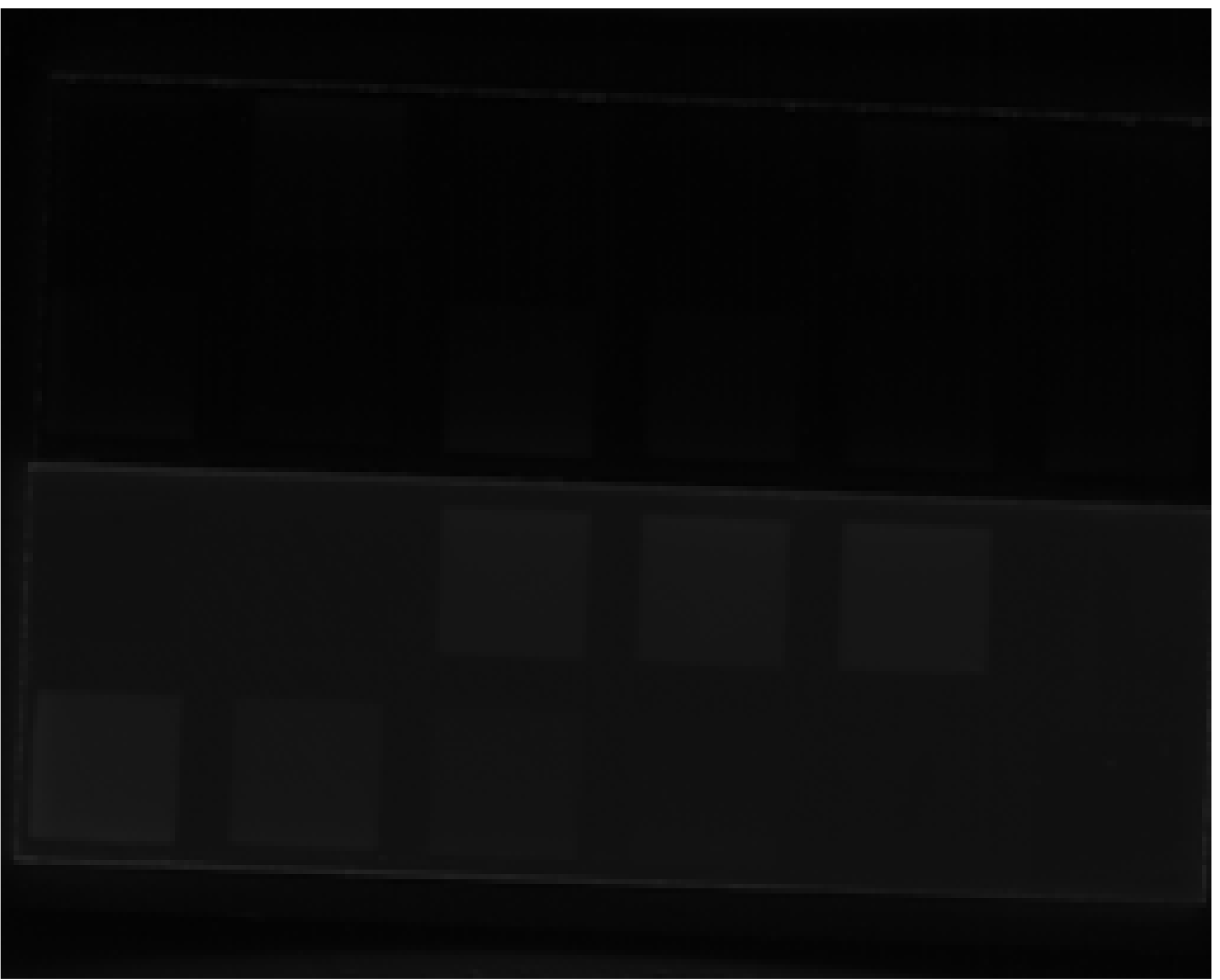}
\end{subfigure}%
\begin{subfigure}{0.06\textwidth}
\centering
\includegraphics[width=\textwidth]{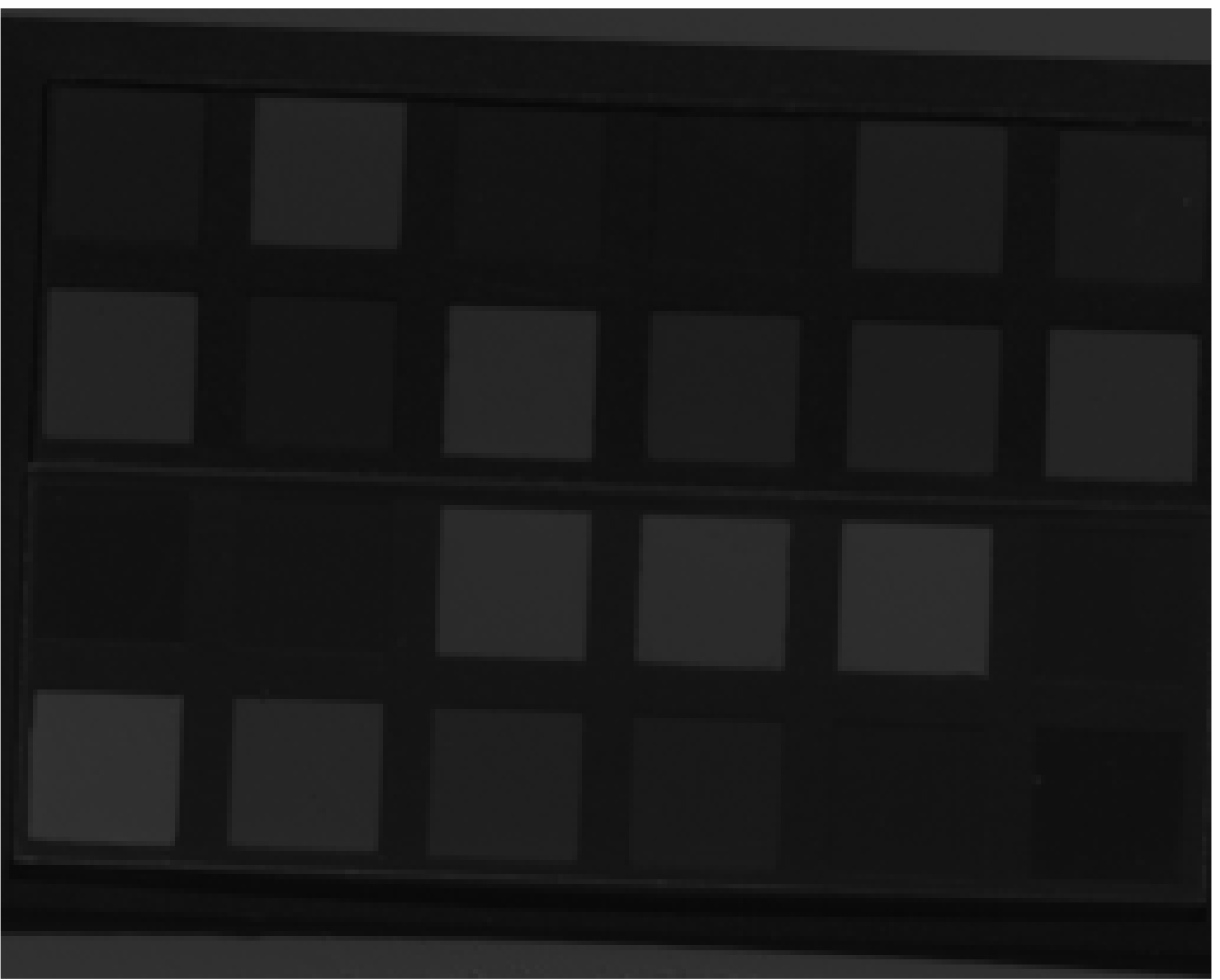}
\end{subfigure}%
\begin{subfigure}{0.06\textwidth}
\centering
\includegraphics[width=\textwidth]{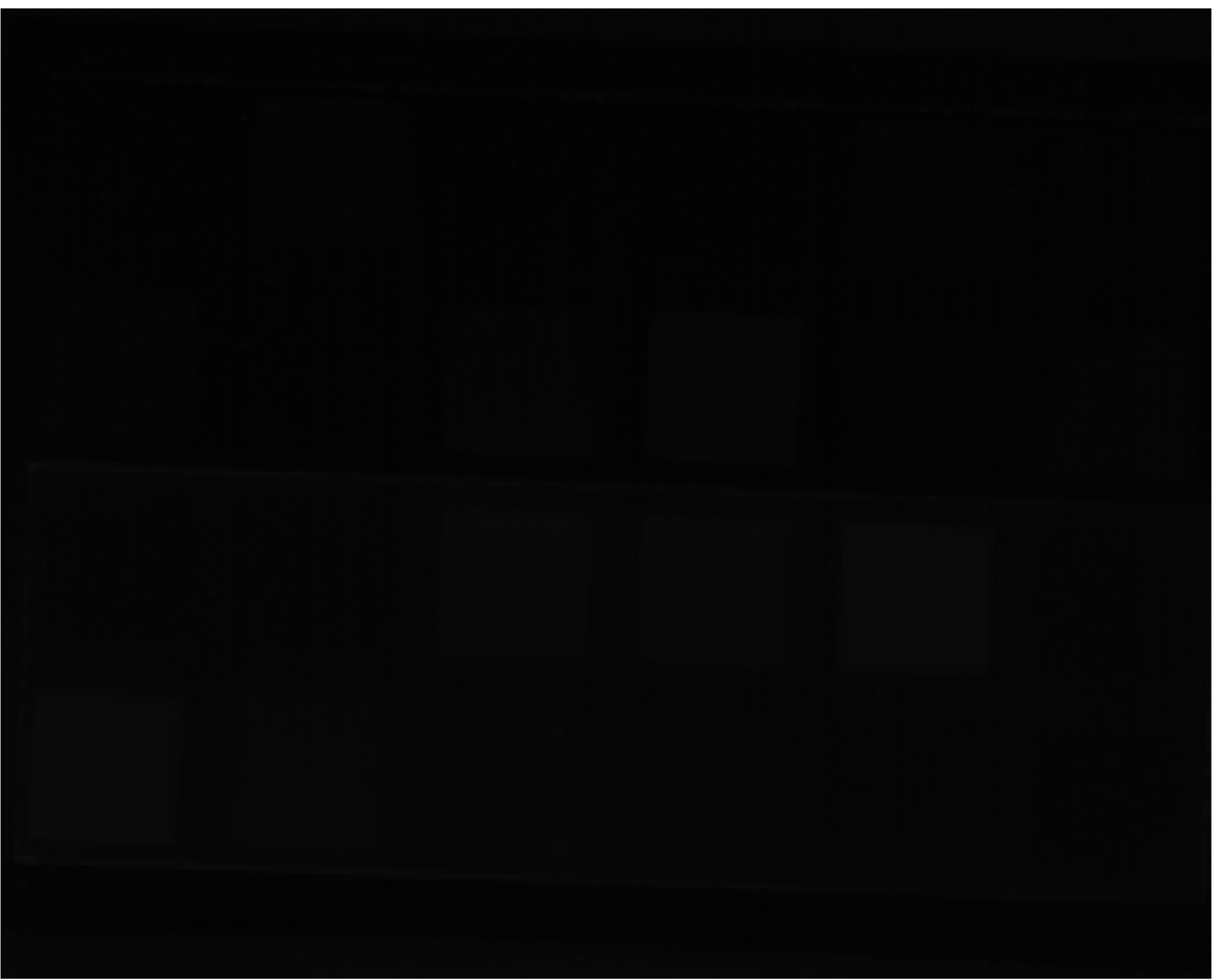}
\end{subfigure}%
\begin{subfigure}{0.06\textwidth}
\centering
\includegraphics[width=\textwidth]{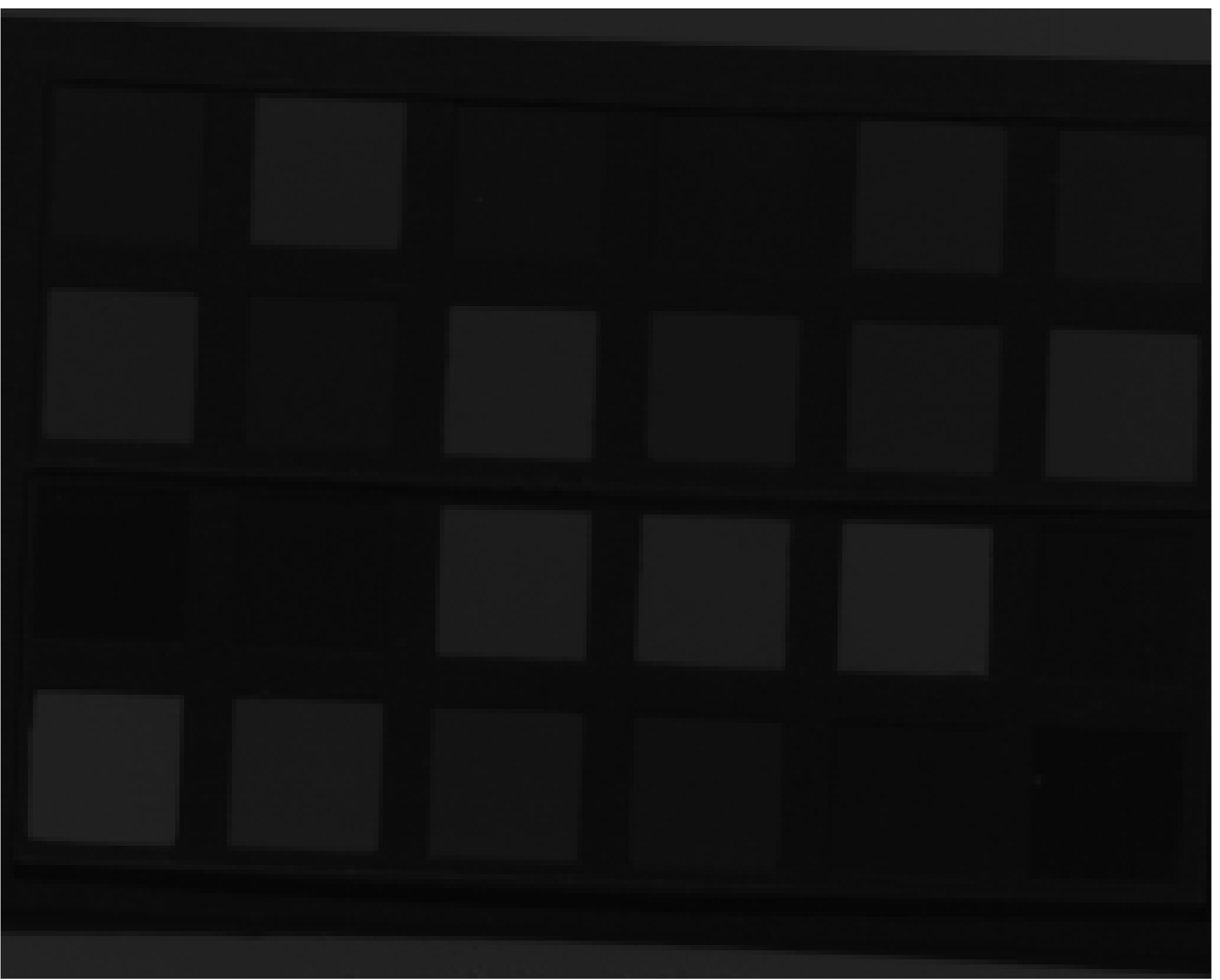}
\end{subfigure}%
\begin{subfigure}{0.06\textwidth}
\centering
\includegraphics[width=\textwidth]{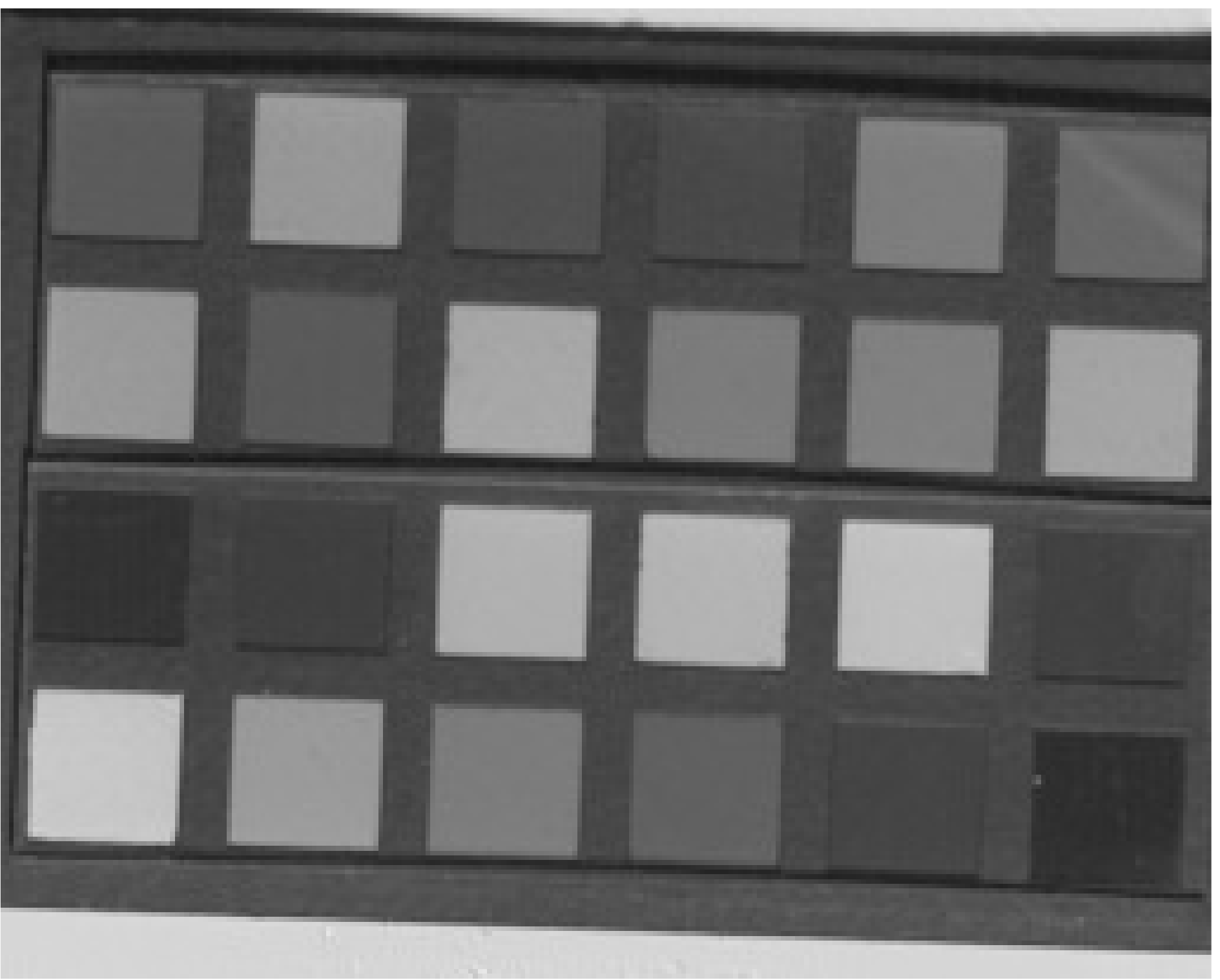}
\end{subfigure}%
\begin{subfigure}{0.06\textwidth}
\centering
\includegraphics[width=\textwidth]{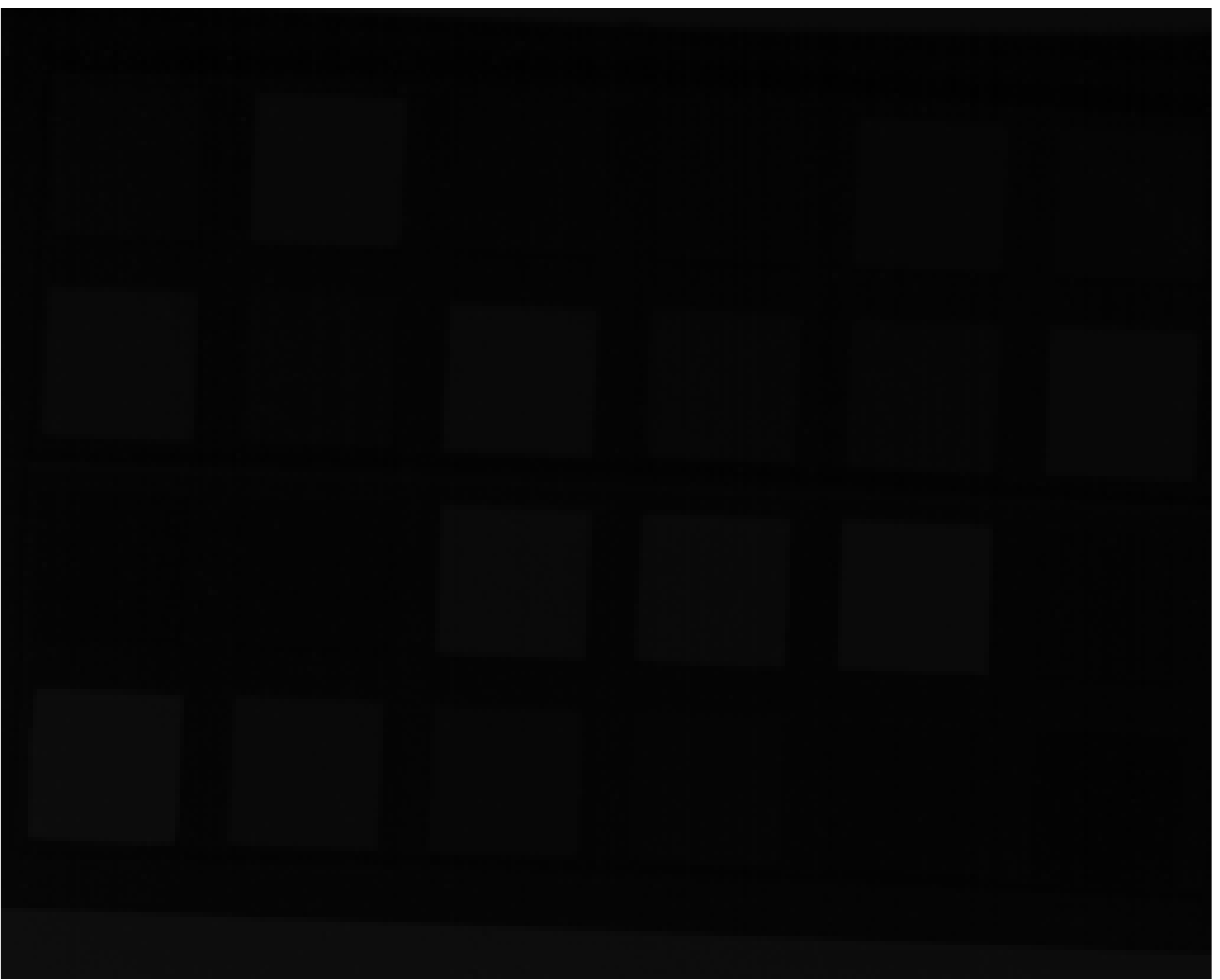}
\end{subfigure}%
\begin{subfigure}{0.06\textwidth}
\centering
\includegraphics[width=\textwidth]{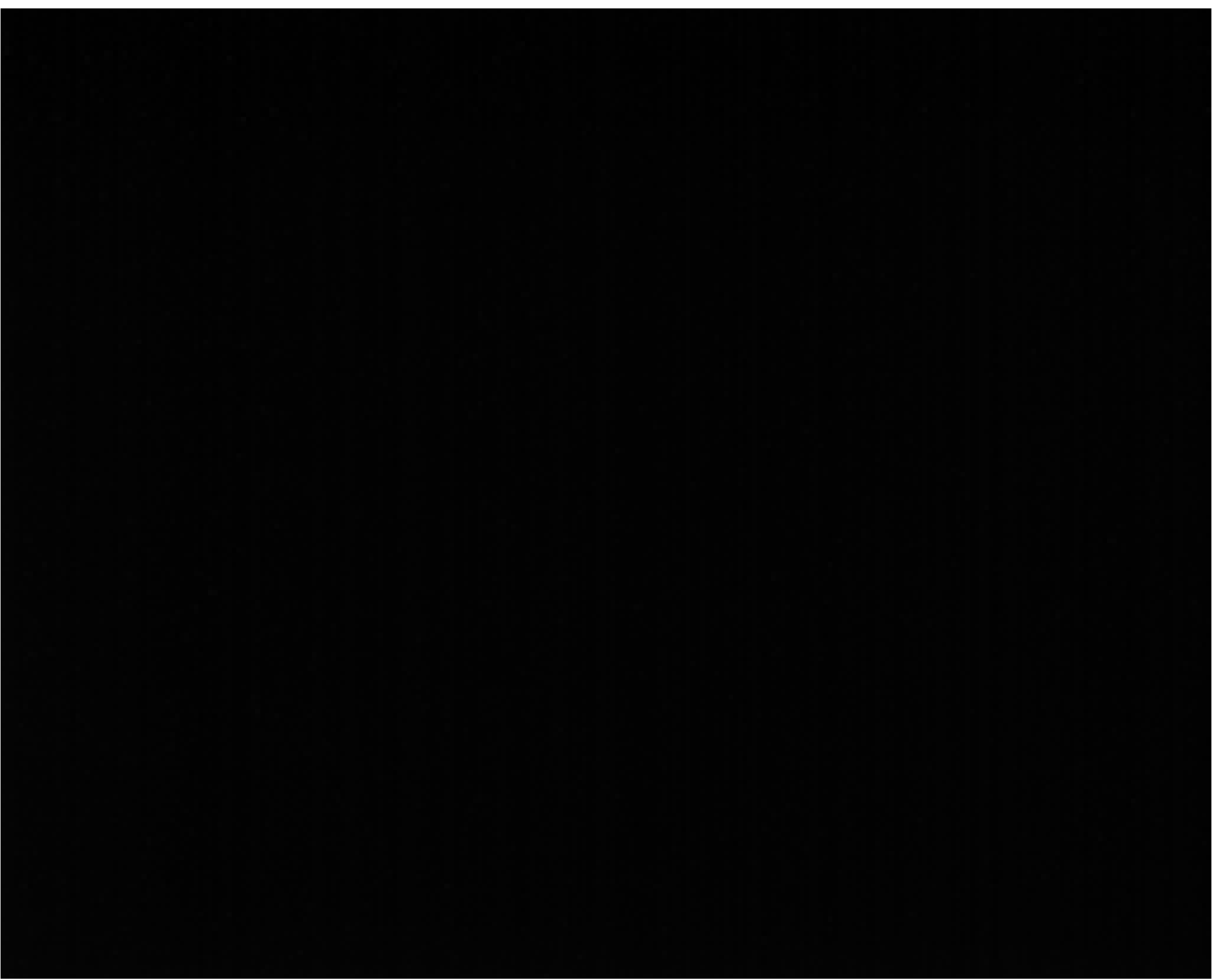}
\end{subfigure}%
\begin{subfigure}{0.06\textwidth}
\centering
\includegraphics[width=\textwidth]{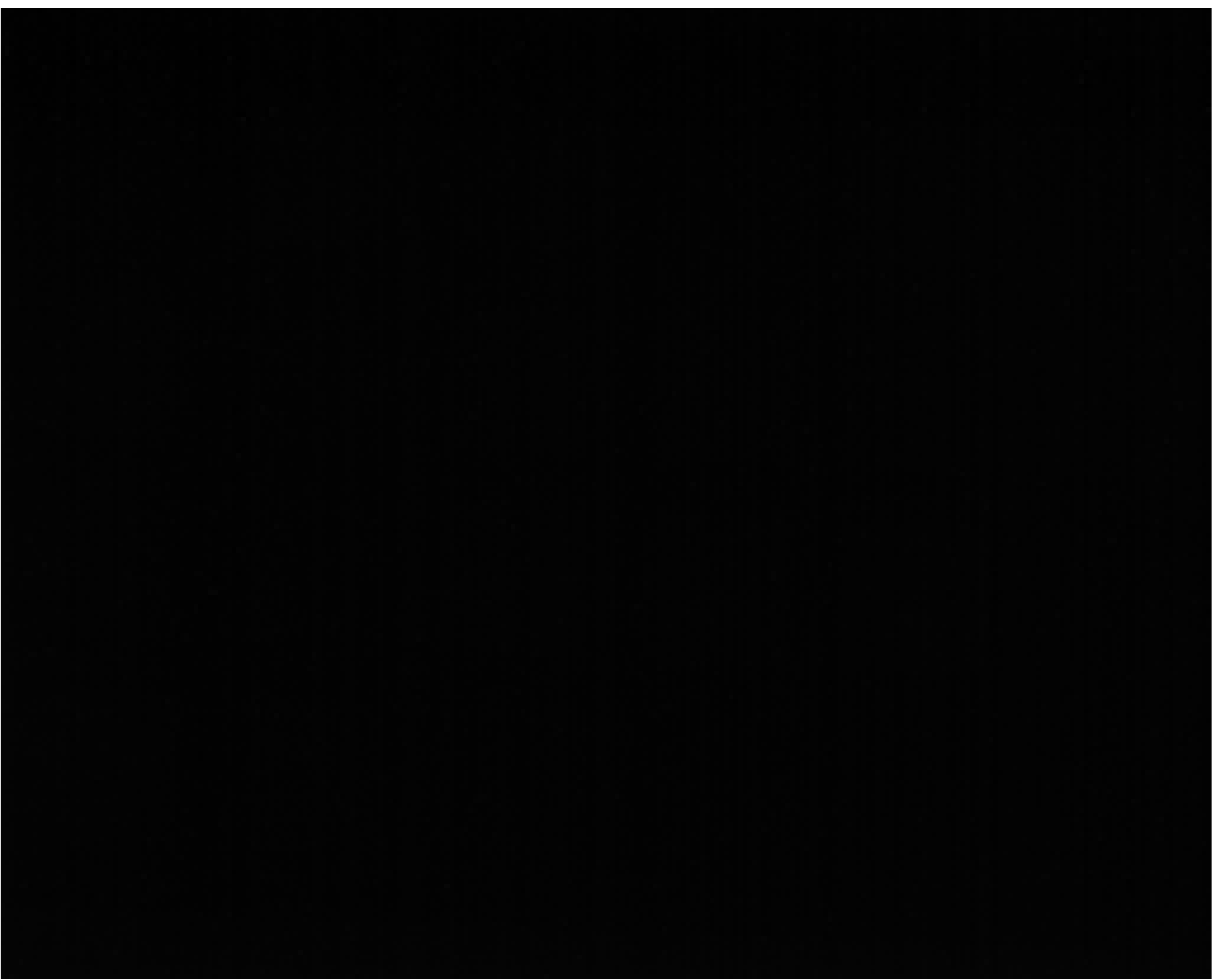}
\end{subfigure}%
\begin{subfigure}{0.06\textwidth}
\centering
\includegraphics[width=\textwidth]{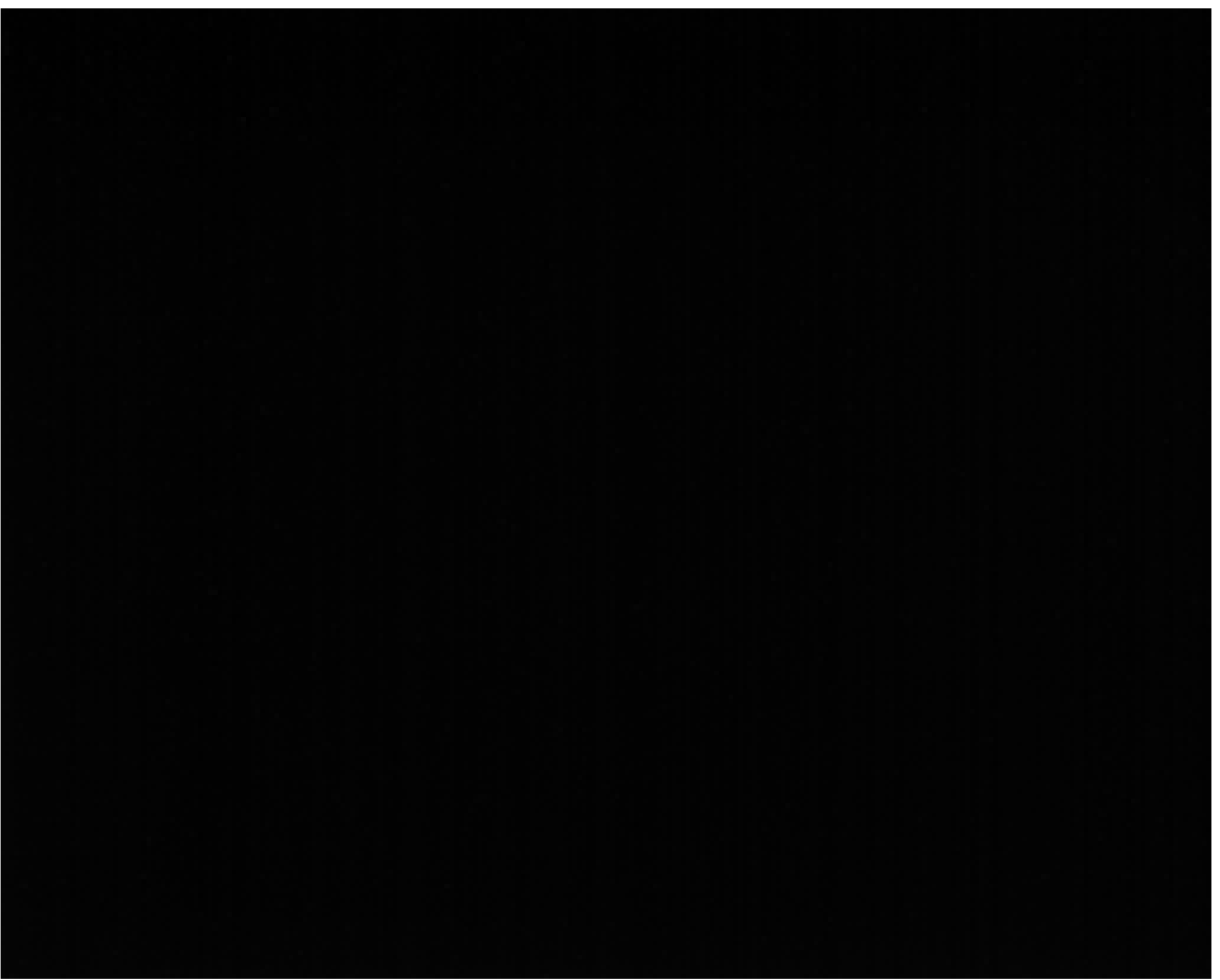}
\end{subfigure}\\%
\begin{subfigure}{0.06\textwidth}
\captionsetup{justification=raggedright,font=scriptsize}
\caption*{787 to 813nm}
\caption*{}
\end{subfigure}%
\begin{subfigure}{0.06\textwidth}
\centering
\includegraphics[width=\textwidth]{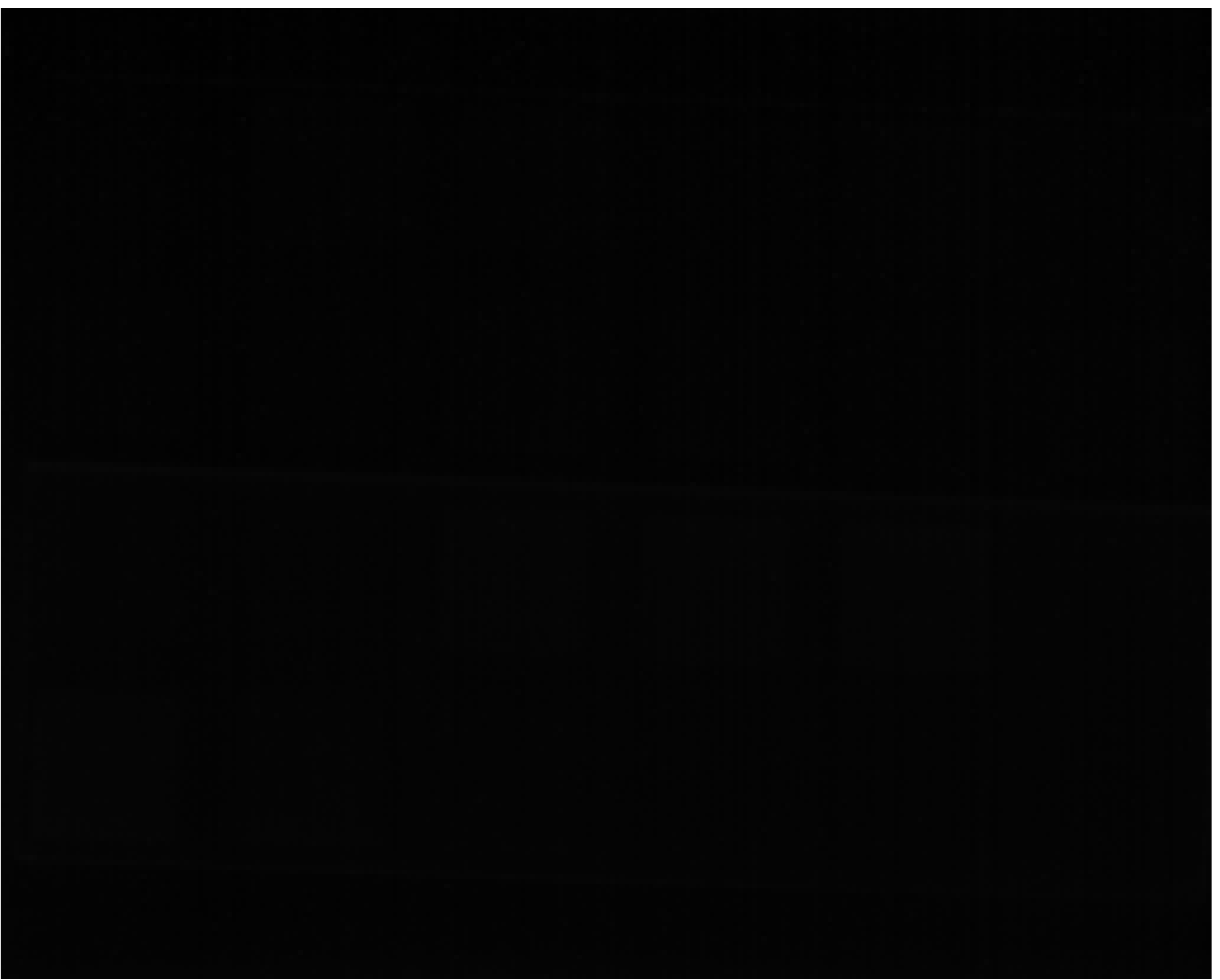}
\caption*{365nm}
\end{subfigure}%
\begin{subfigure}{0.06\textwidth}
\centering
\includegraphics[width=\textwidth]{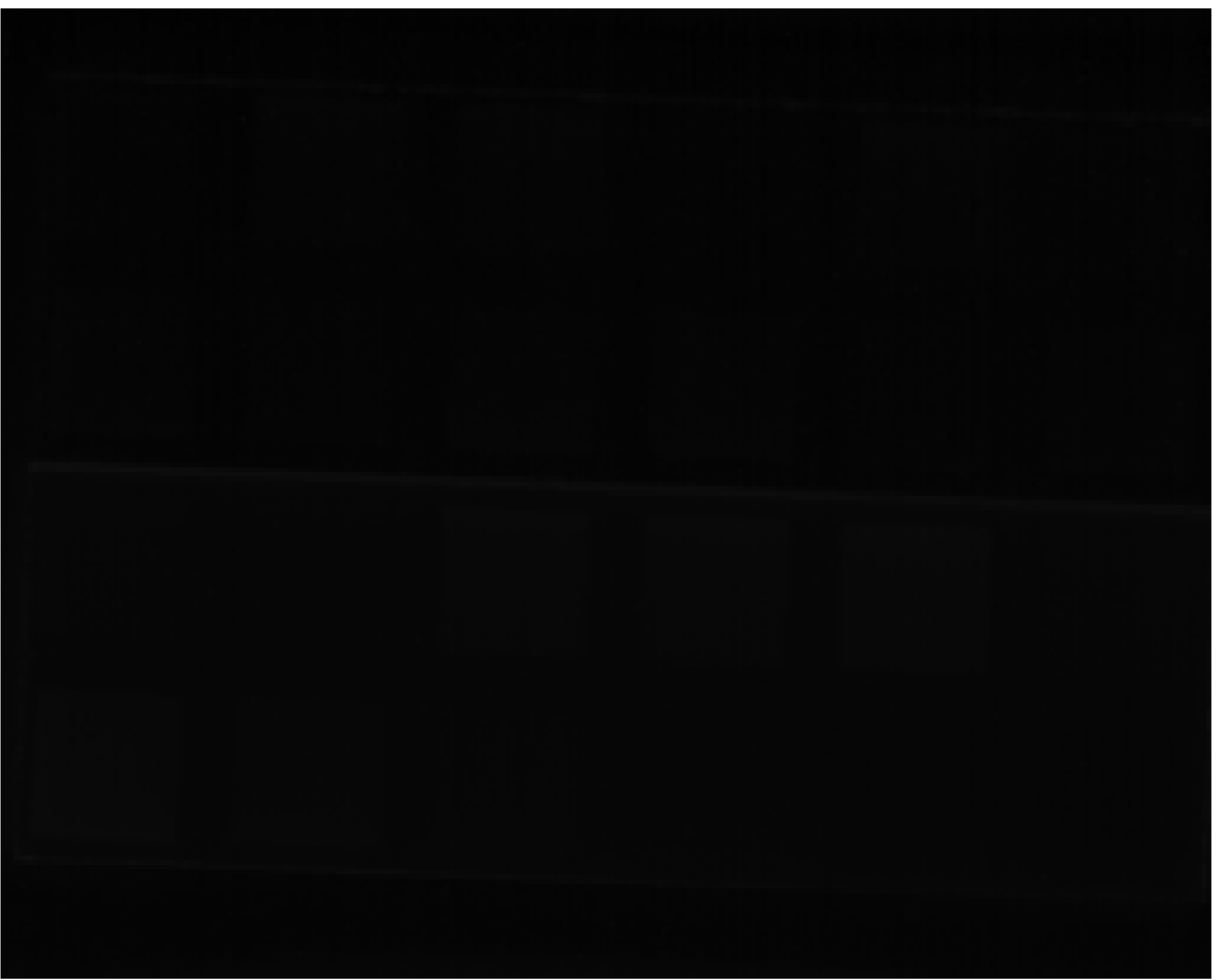}
\caption*{395nm}
\end{subfigure}%
\begin{subfigure}{0.06\textwidth}
\centering
\includegraphics[width=\textwidth]{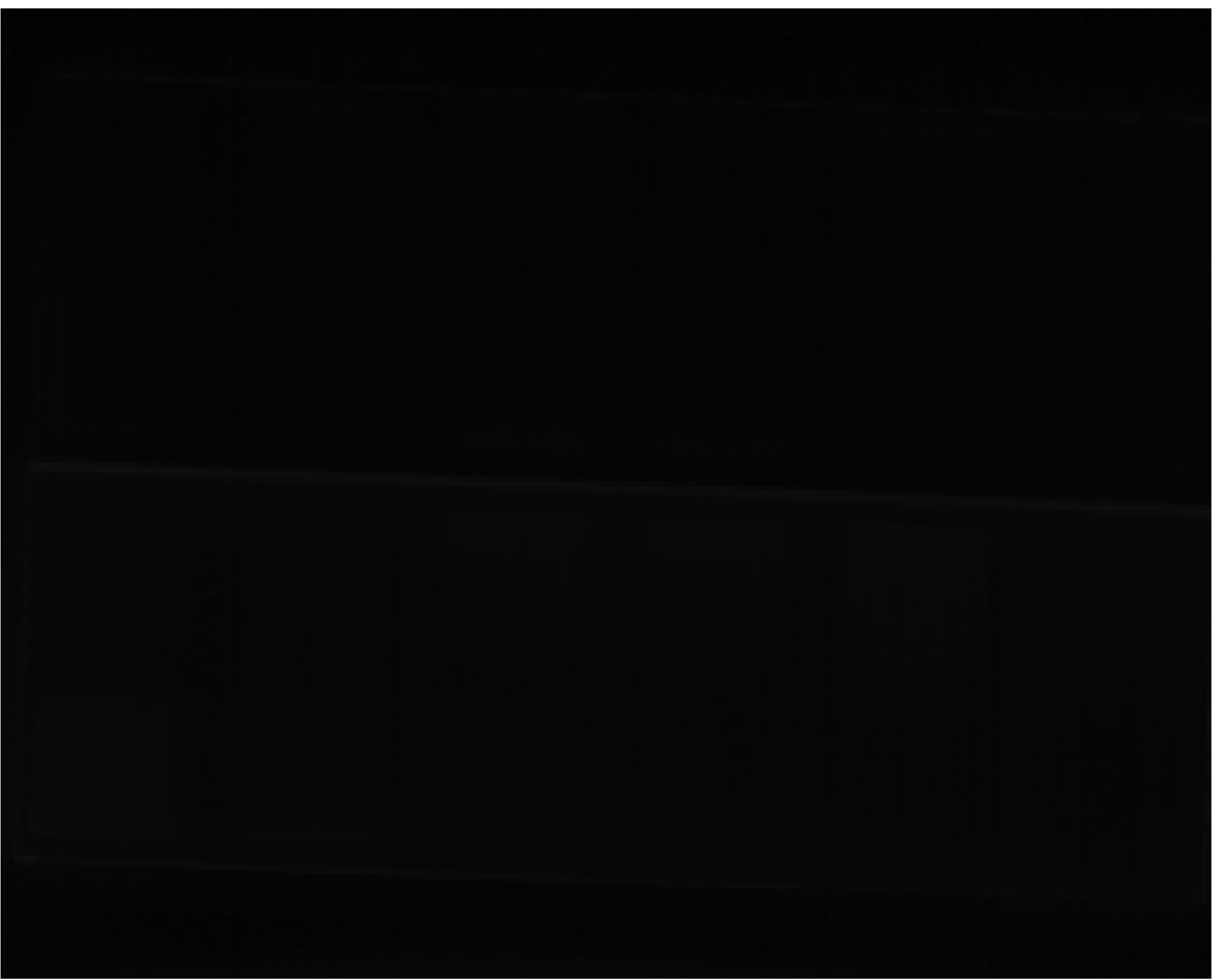}
\caption*{447nm}
\end{subfigure}%
\begin{subfigure}{0.06\textwidth}
\centering
\includegraphics[width=\textwidth]{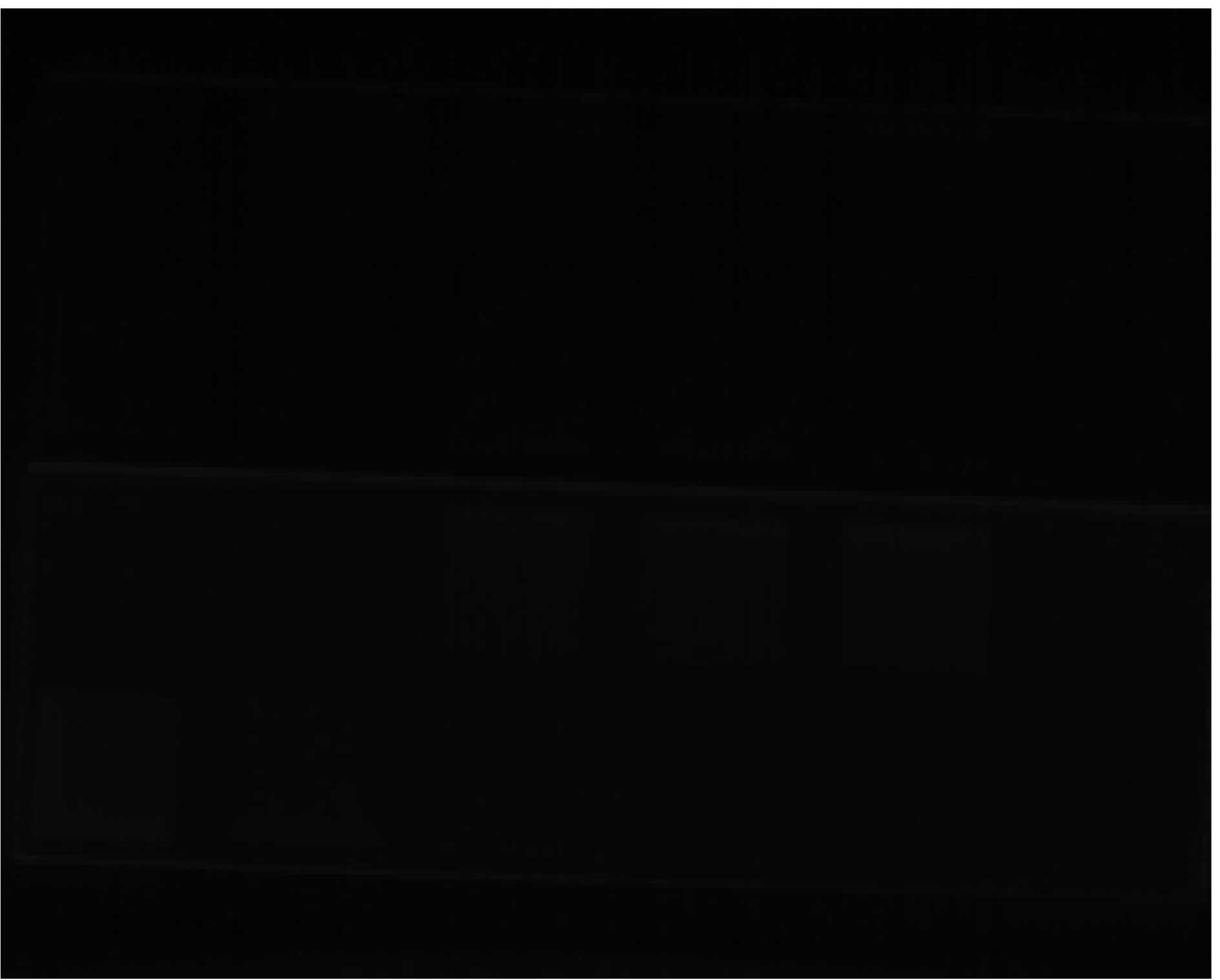}
\caption*{470nm}
\end{subfigure}%
\begin{subfigure}{0.06\textwidth}
\centering
\includegraphics[width=\textwidth]{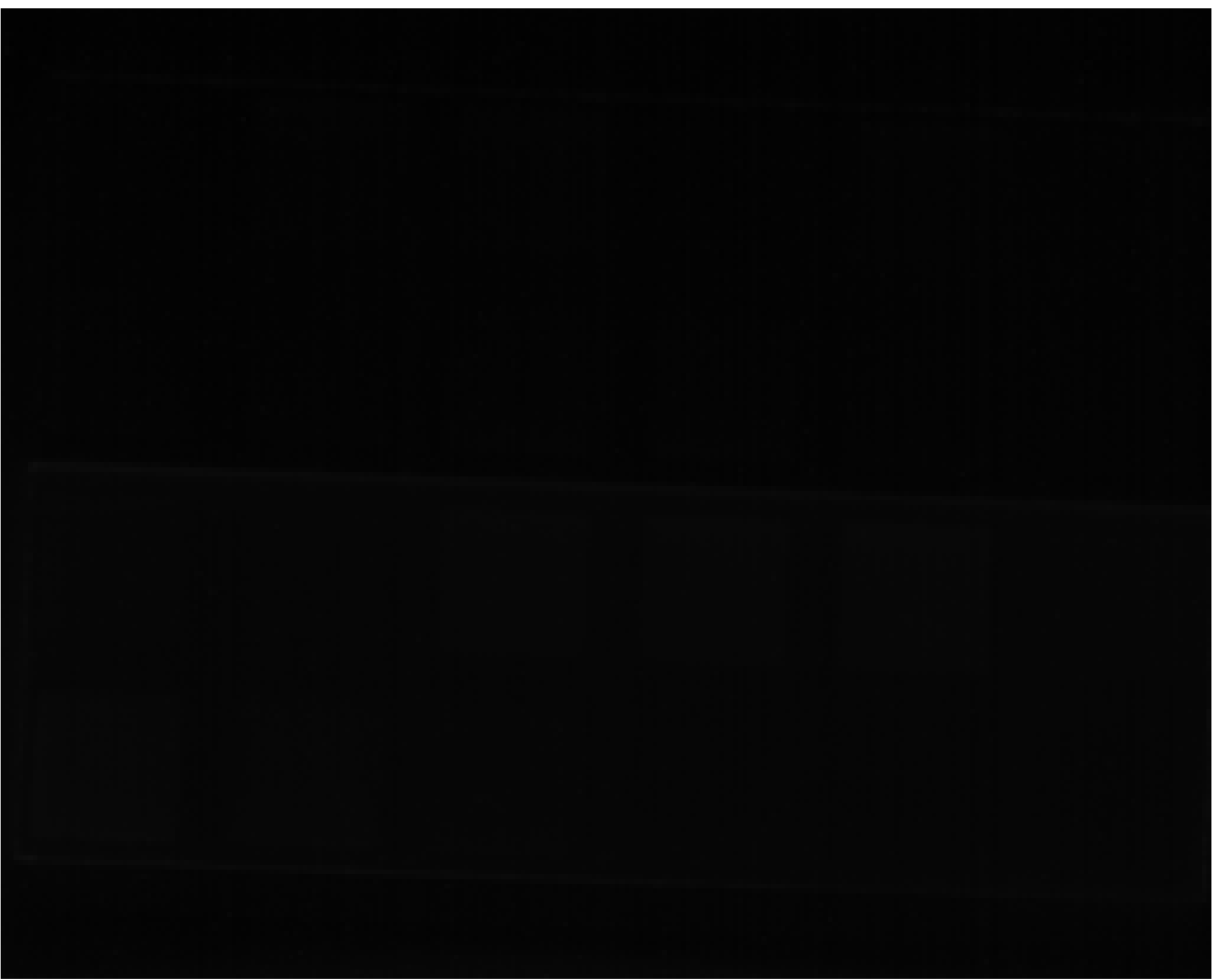}
\caption*{505nm}
\end{subfigure}%
\begin{subfigure}{0.06\textwidth}
\centering
\includegraphics[width=\textwidth]{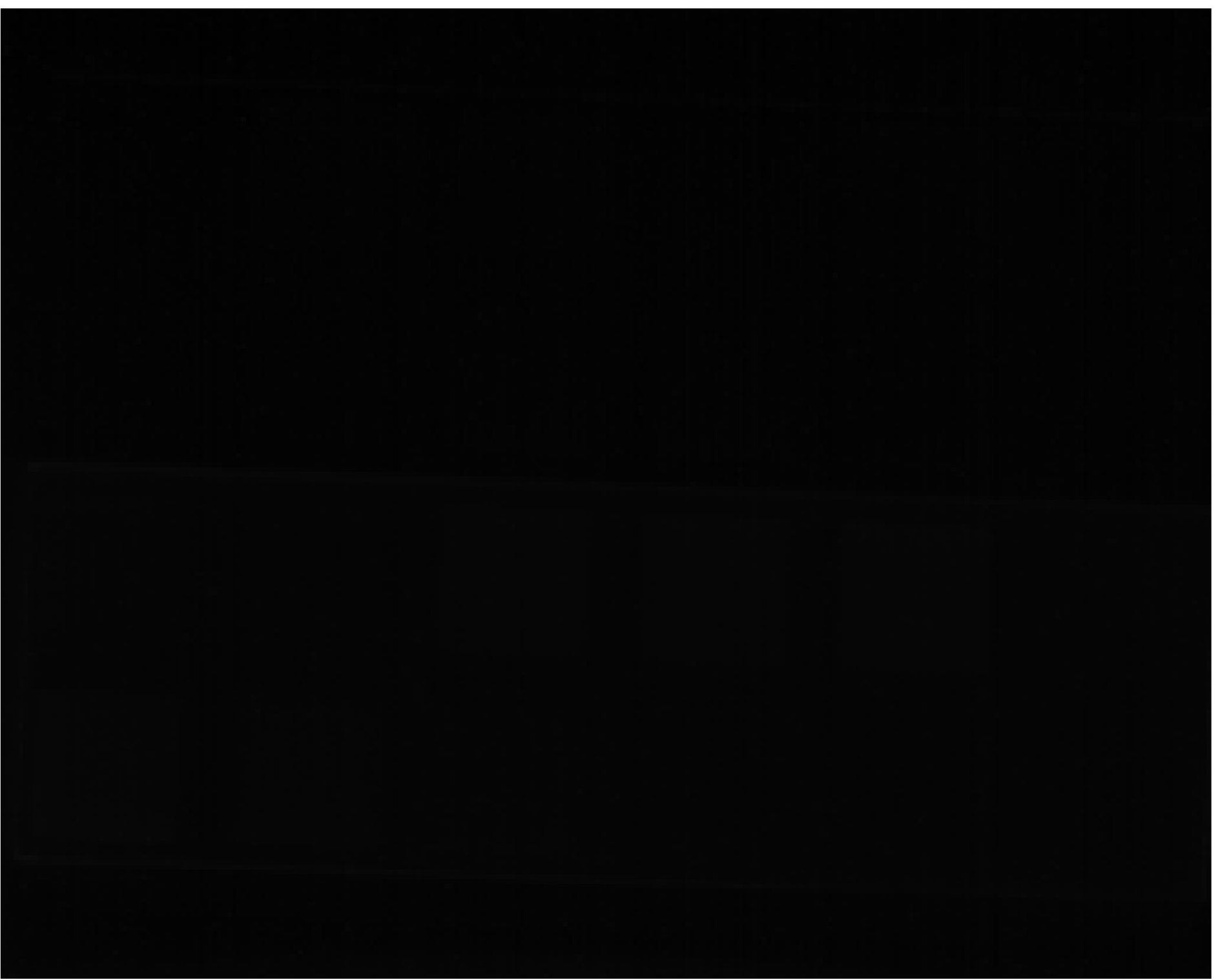}
\caption*{530nm}
\end{subfigure}%
\begin{subfigure}{0.06\textwidth}
\centering
\includegraphics[width=\textwidth]{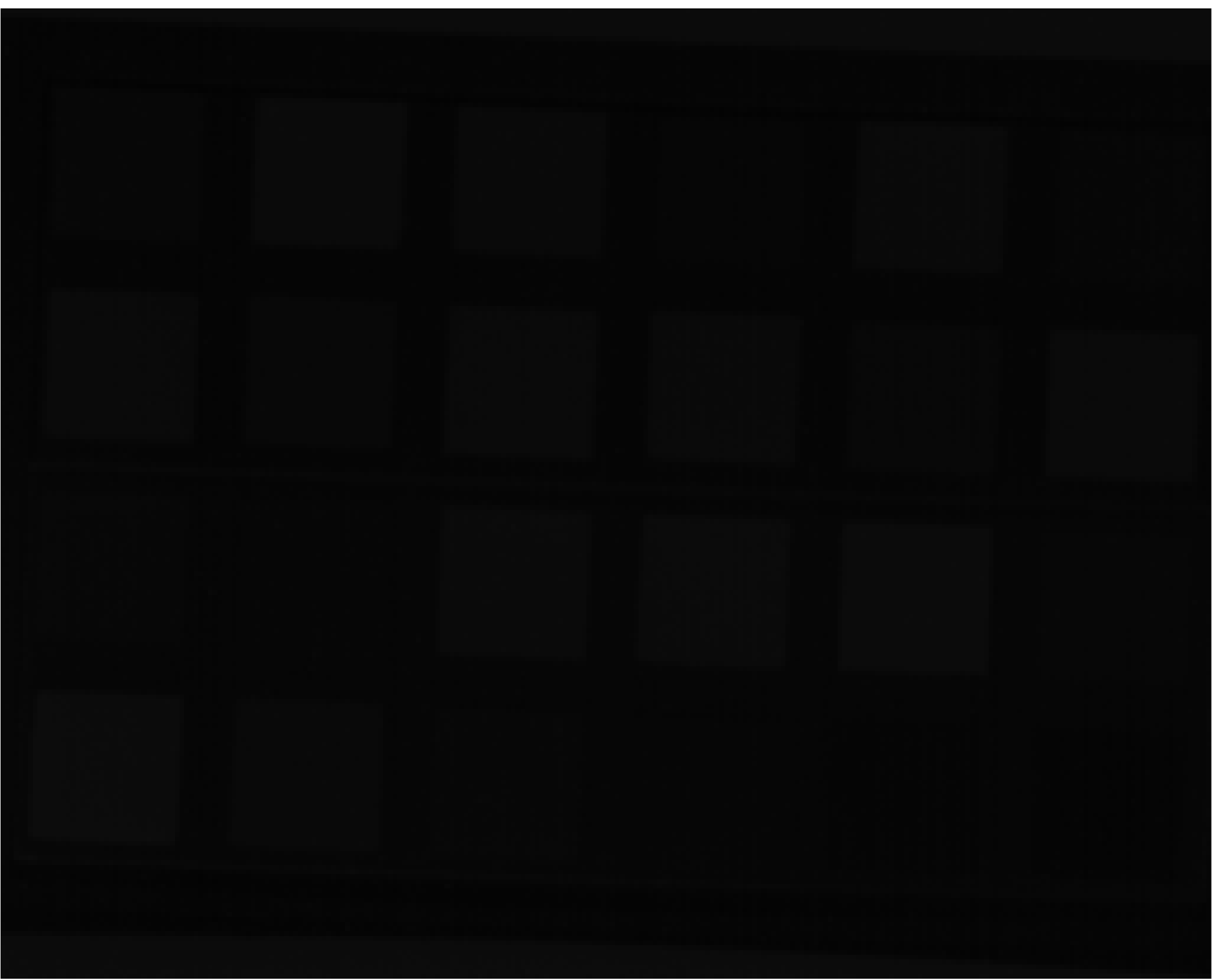}
\caption*{590nm}
\end{subfigure}%
\begin{subfigure}{0.06\textwidth}
\centering
\includegraphics[width=\textwidth]{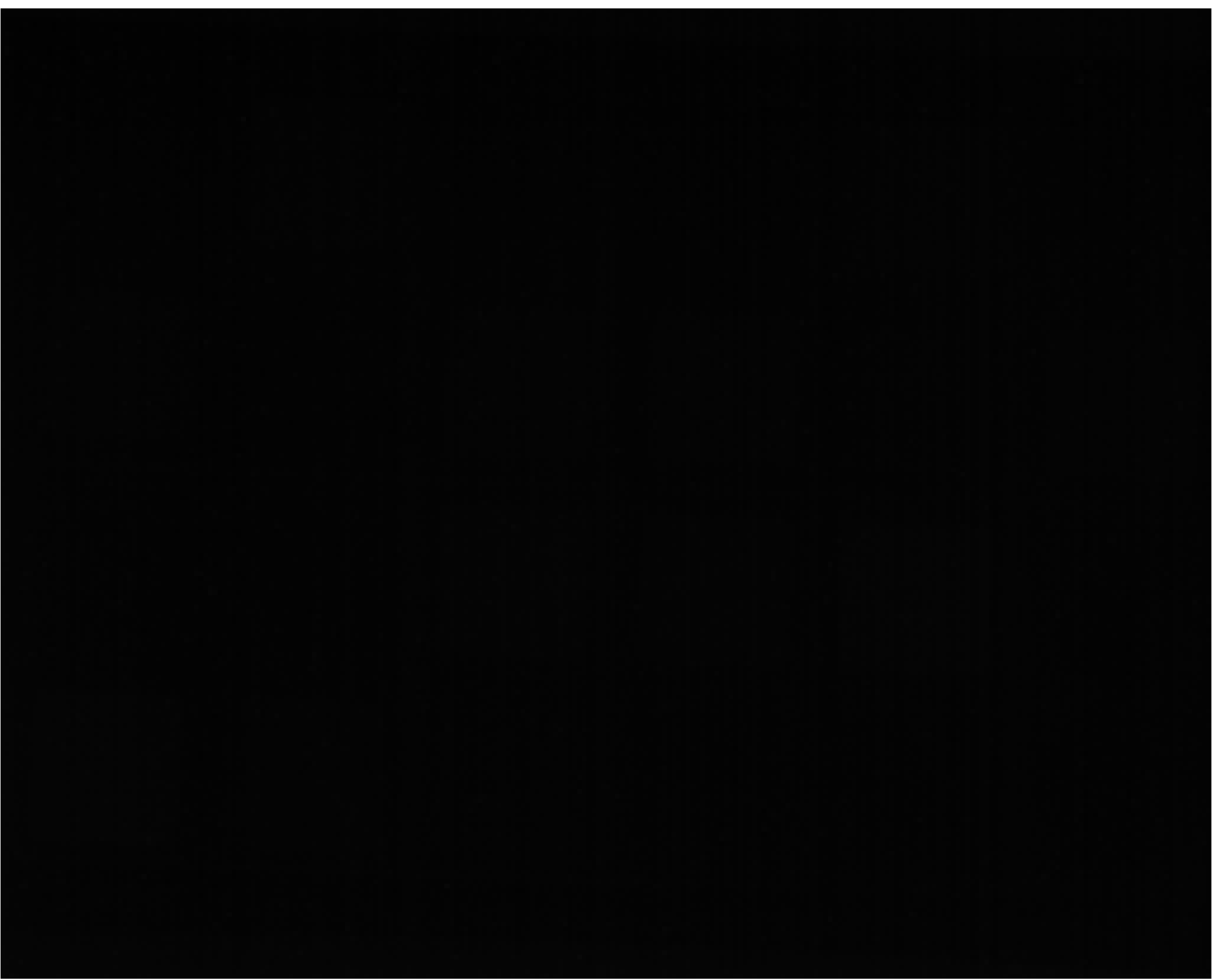}
\caption*{627nm}
\end{subfigure}%
\begin{subfigure}{0.06\textwidth}
\centering
\includegraphics[width=\textwidth]{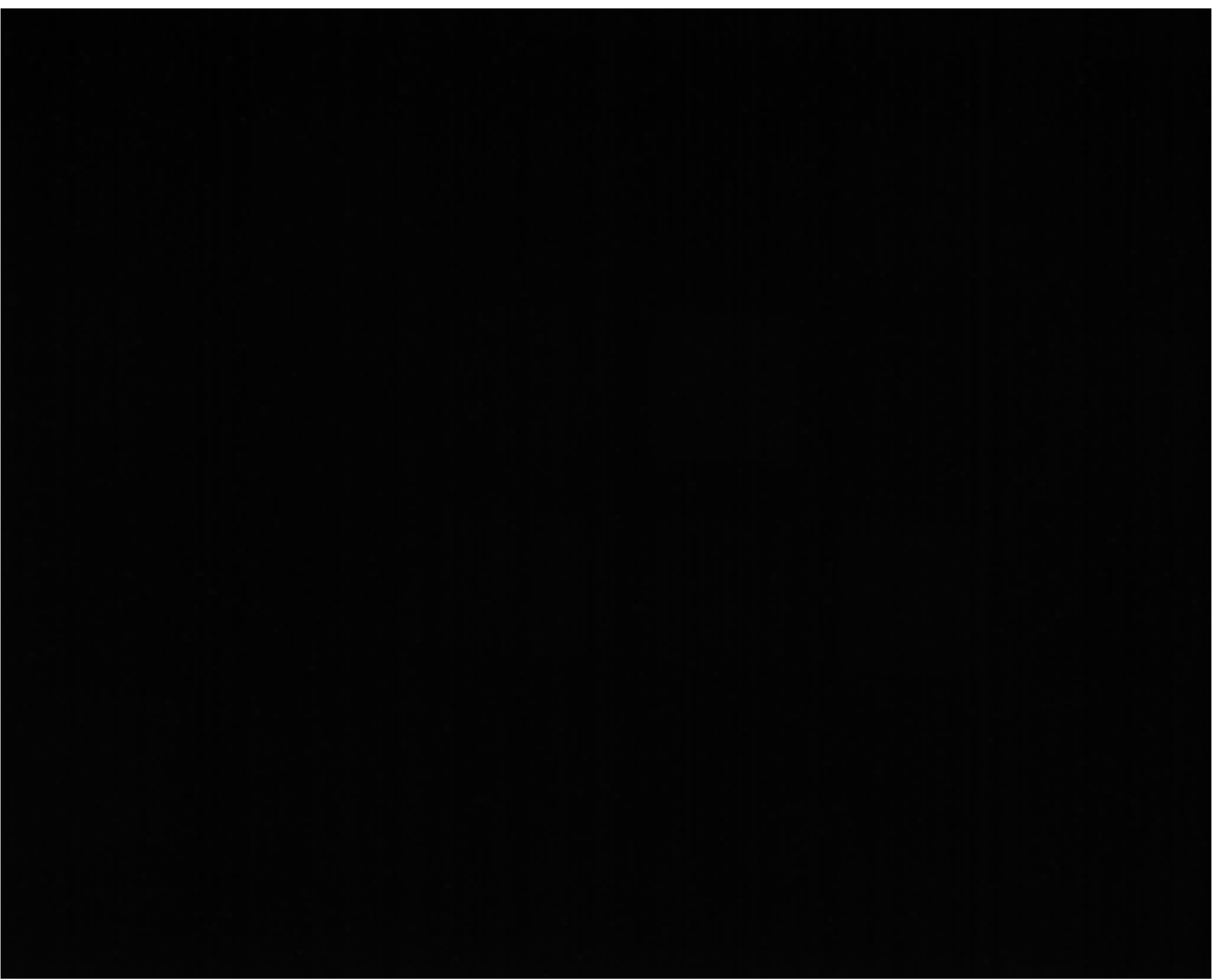}
\caption*{655nm}
\end{subfigure}%
\begin{subfigure}{0.06\textwidth}
\centering
\includegraphics[width=\textwidth]{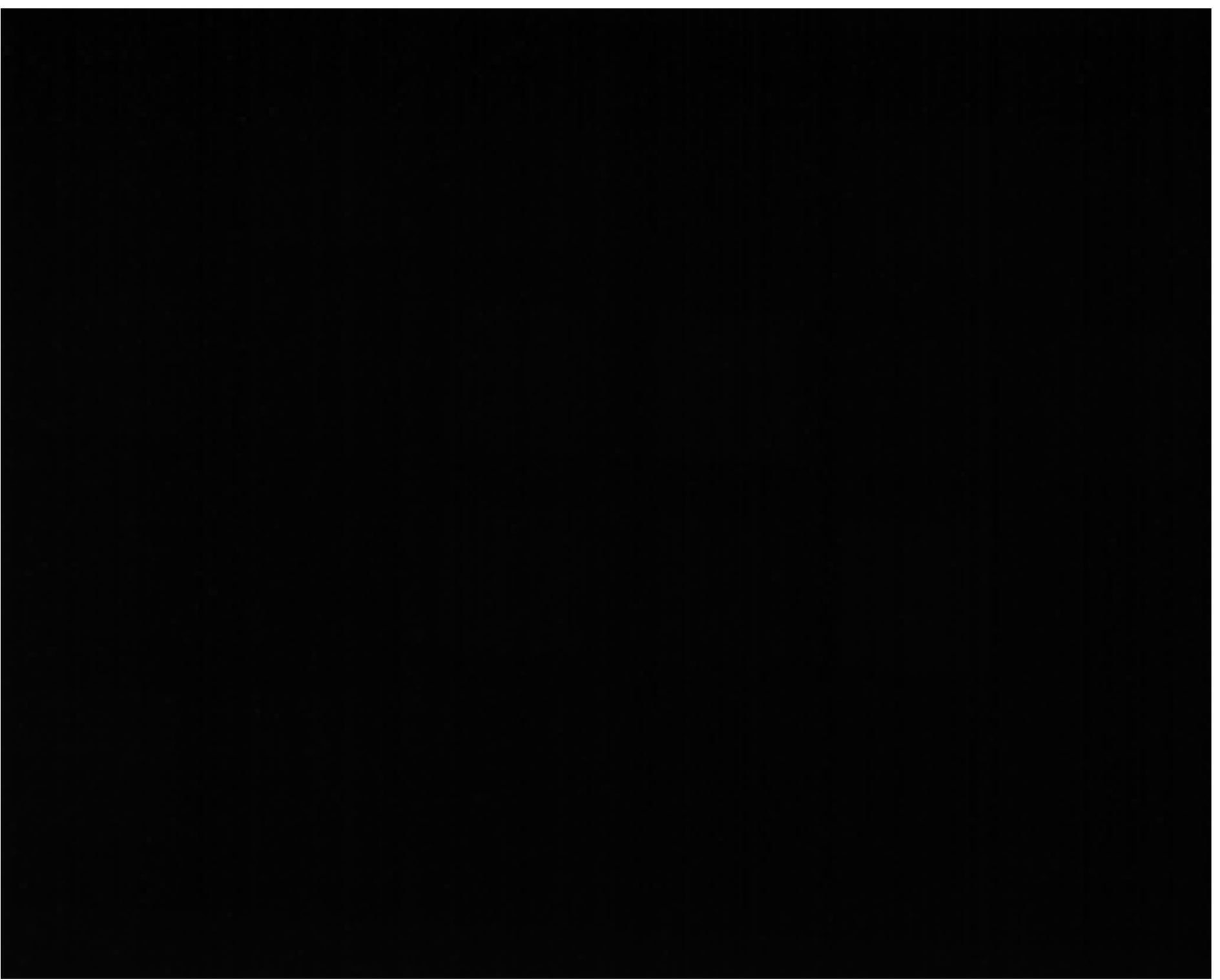}
\caption*{680nm}
\end{subfigure}%
\begin{subfigure}{0.06\textwidth}
\centering
\includegraphics[width=\textwidth]{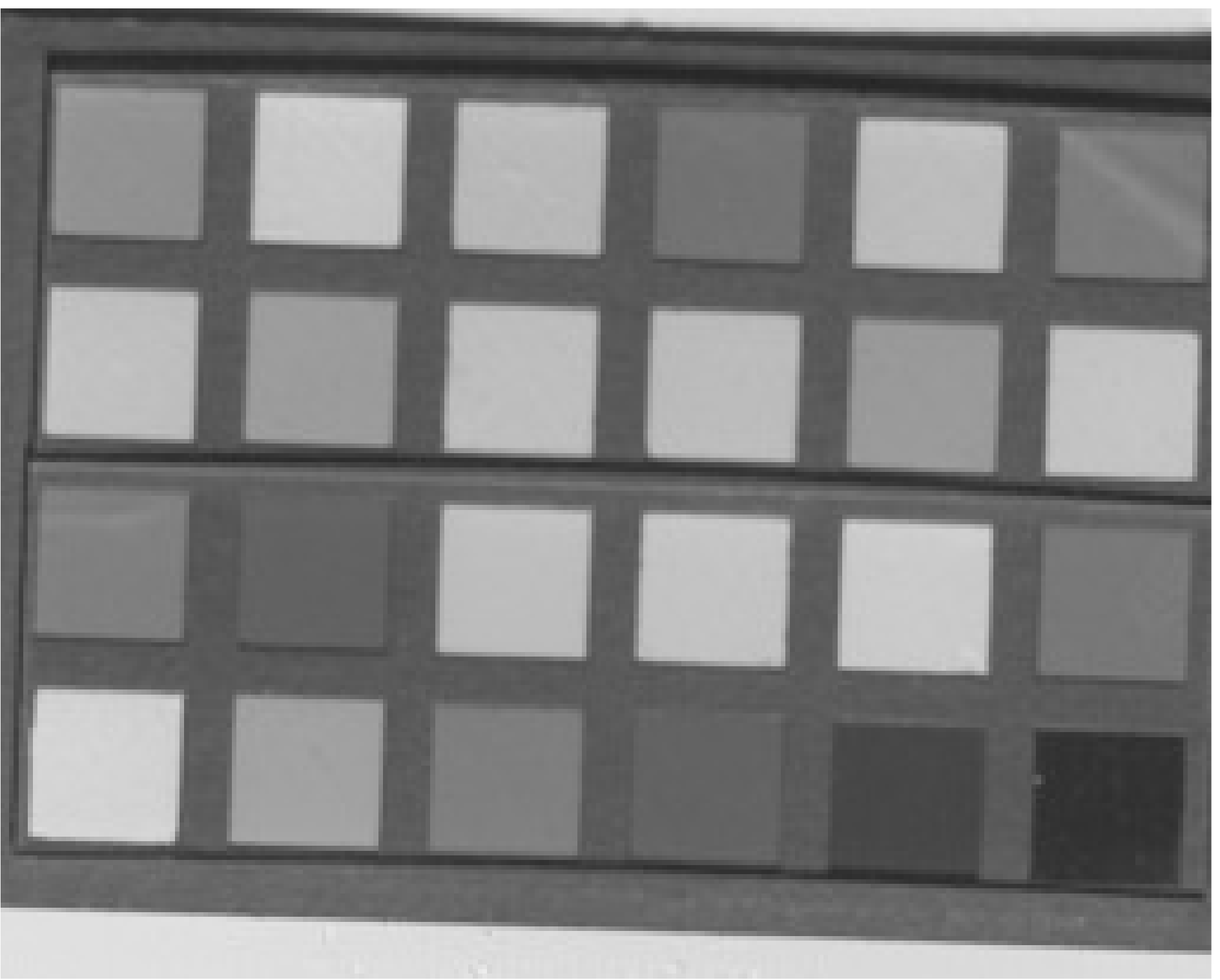}
\caption*{780nm}
\end{subfigure}%
\begin{subfigure}{0.06\textwidth}
\centering
\includegraphics[width=\textwidth]{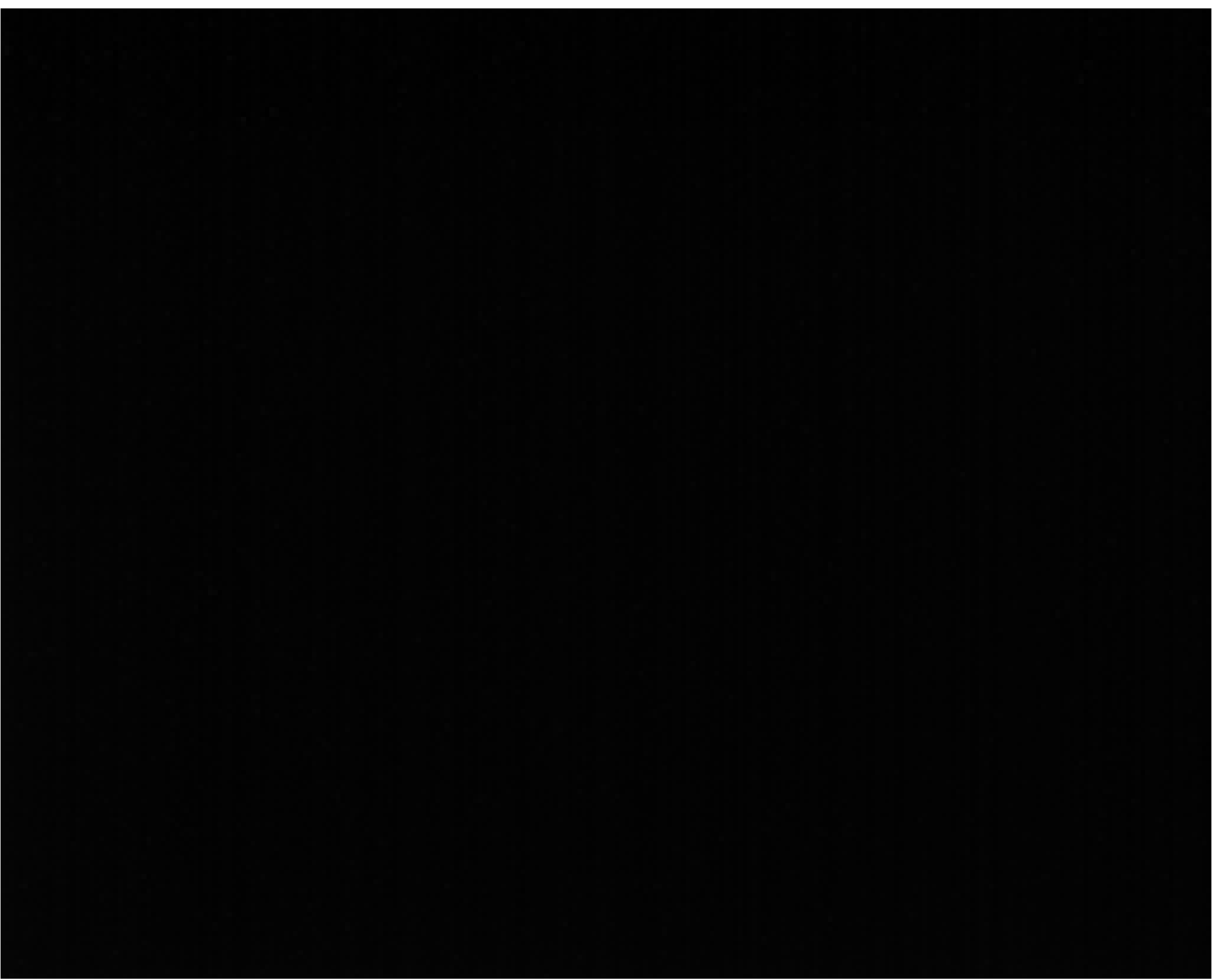}
\caption*{850nm}
\end{subfigure}%
\begin{subfigure}{0.06\textwidth}
\centering
\includegraphics[width=\textwidth]{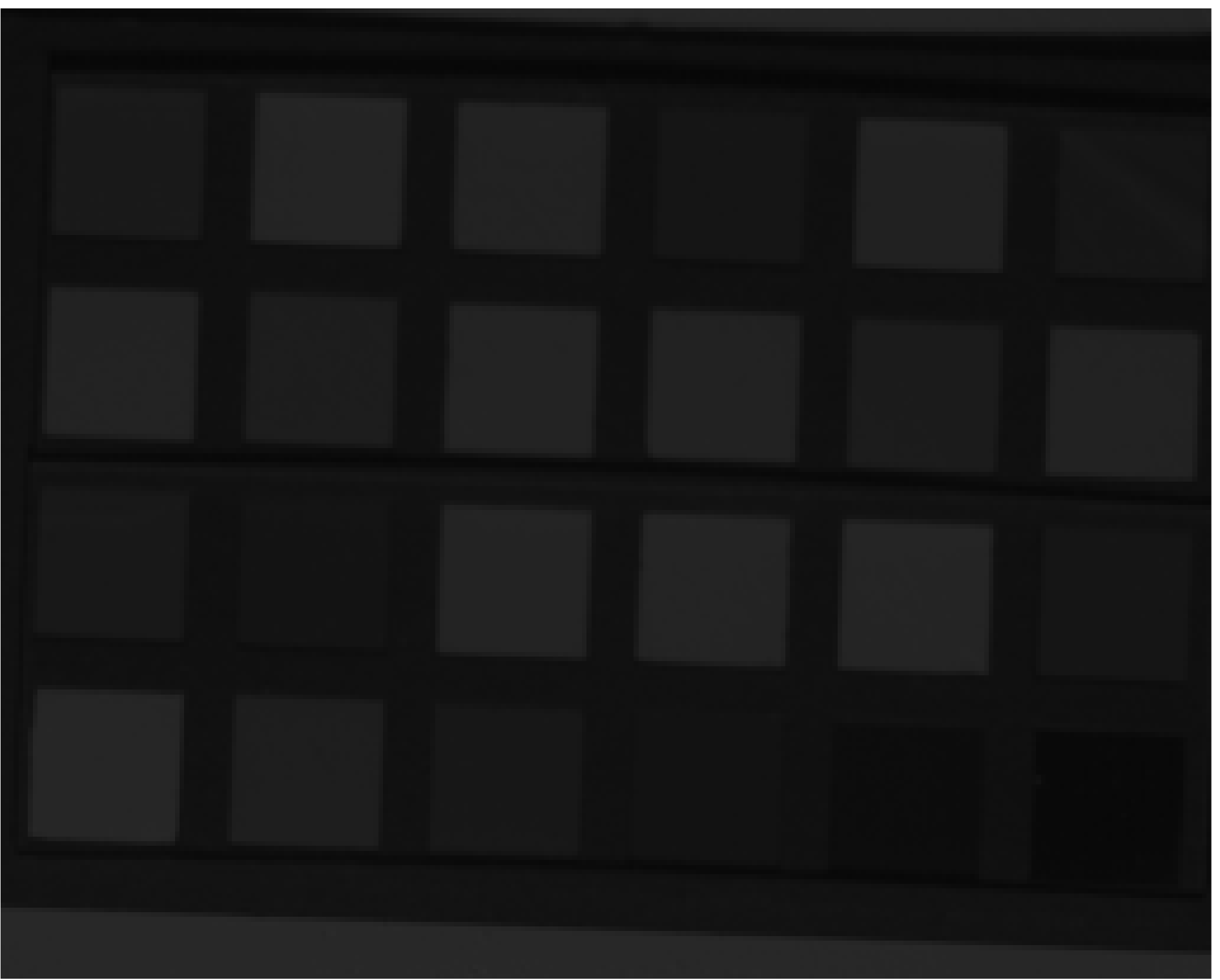}
\caption*{880nm}
\end{subfigure}%
\begin{subfigure}{0.06\textwidth}
\centering
\includegraphics[width=\textwidth]{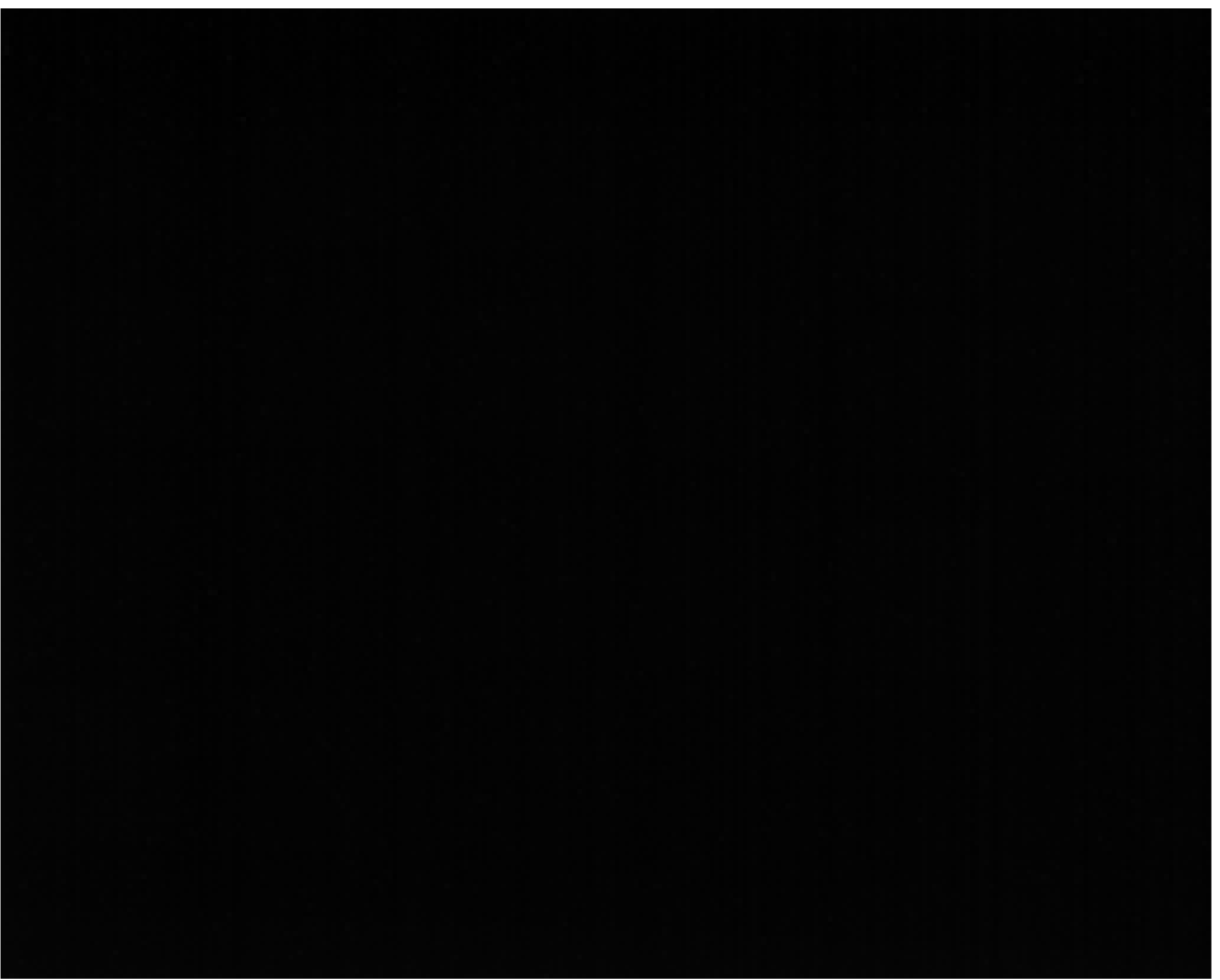}
\caption*{940nm}
\end{subfigure}
\caption{Experimental target image stack used as input to the estimation algorithm. Rows are data from the monochromatic channel (MC) and the seven camera filters, with filter pass bands specified next to each row. The columns show image illuminated by different narrowband lights, with peak emission wavelengths given below each column. Images near the 'diagonal' represent the reflected component, while 'off diagonal' images represent fluorescence signals. The measured image data are scaled and encoded with $\gamma = 2.2$ for visualization.}
\label{fig:sampleTarget}
\end{figure*}

%% file: ResultImage.tex
\begin{figure*}
\centering
\begin{subfigure}{0.030000\textwidth}
\centering
\rotcaption*{Patch A}
\end{subfigure}%
\begin{subfigure}{0.180000\textwidth}
\includegraphics[width = \textwidth]{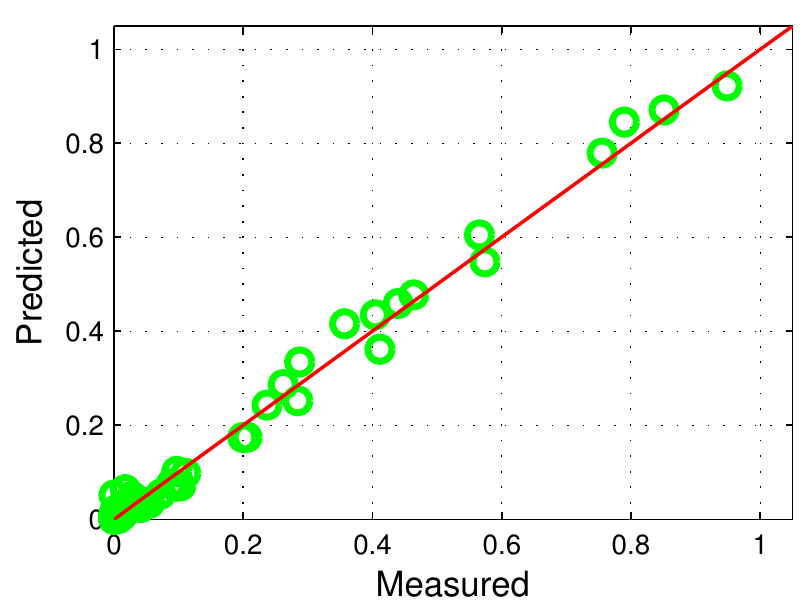}
\end{subfigure}%
\begin{subfigure}{0.180000\textwidth}
\centering
\includegraphics[width=\textwidth]{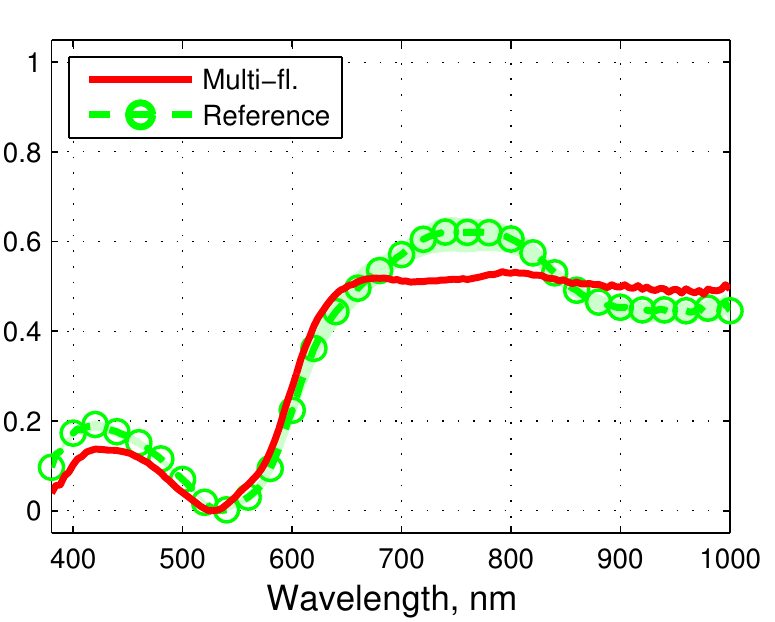}
\end{subfigure}%
\begin{subfigure}{0.180000\textwidth}
\centering
\includegraphics[width=\textwidth]{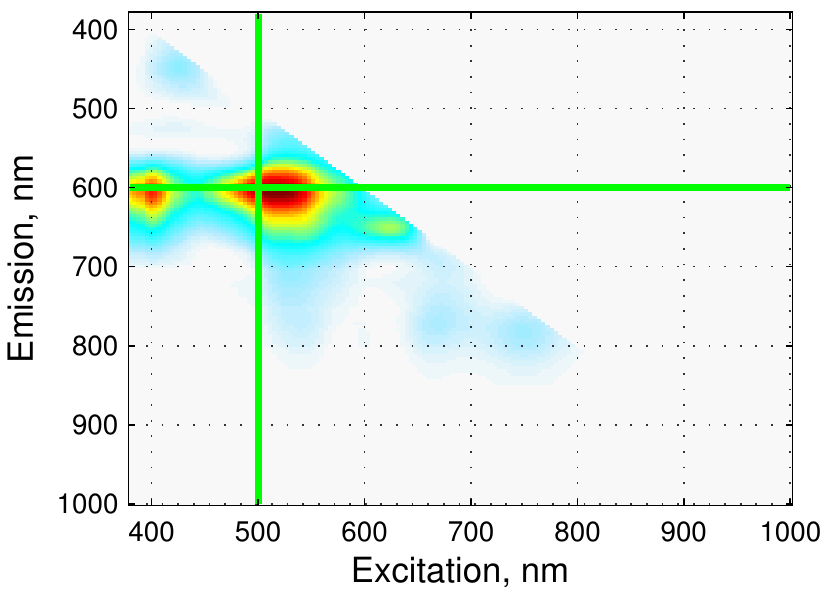}
\end{subfigure}%
\begin{subfigure}{0.180000\textwidth}
\centering
\includegraphics[width=\textwidth]{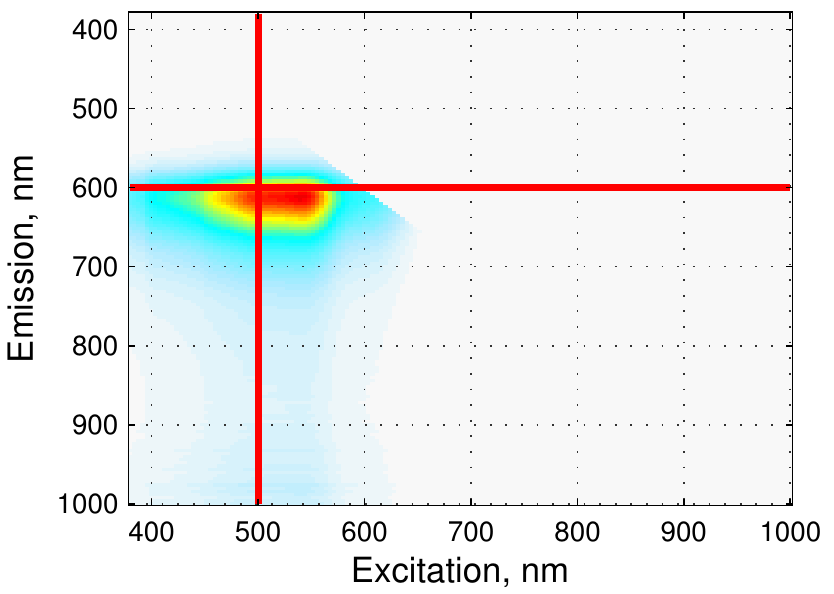}
\end{subfigure}%
\begin{subfigure}{0.050000\textwidth}
\centering
\raisebox{0.2in}{\includegraphics[width=\textwidth]{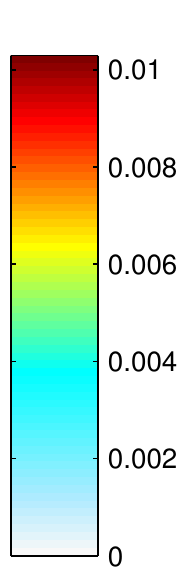}}
\caption*{}
\end{subfigure}%
\begin{subfigure}{0.180000\textwidth}
\centering
\includegraphics[width=\textwidth]{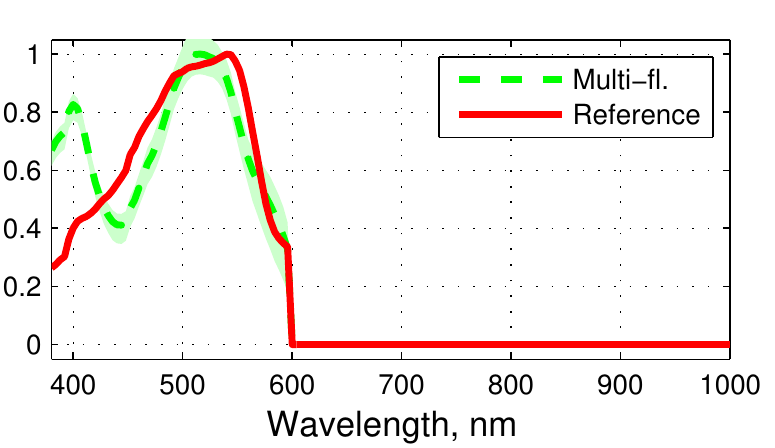}
\includegraphics[width=\textwidth]{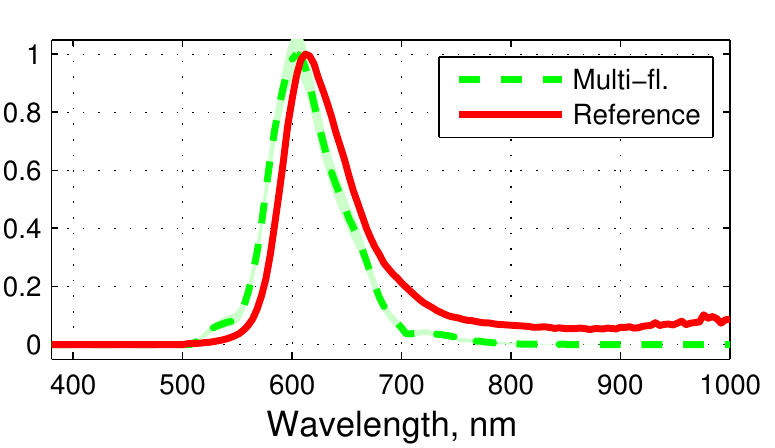}
\end{subfigure}\\
\rule{\textwidth}{1pt}
\begin{subfigure}{0.030000\textwidth}
\centering
\rotcaption*{Patch B}
\end{subfigure}%
\begin{subfigure}{0.180000\textwidth}
\includegraphics[width = \textwidth]{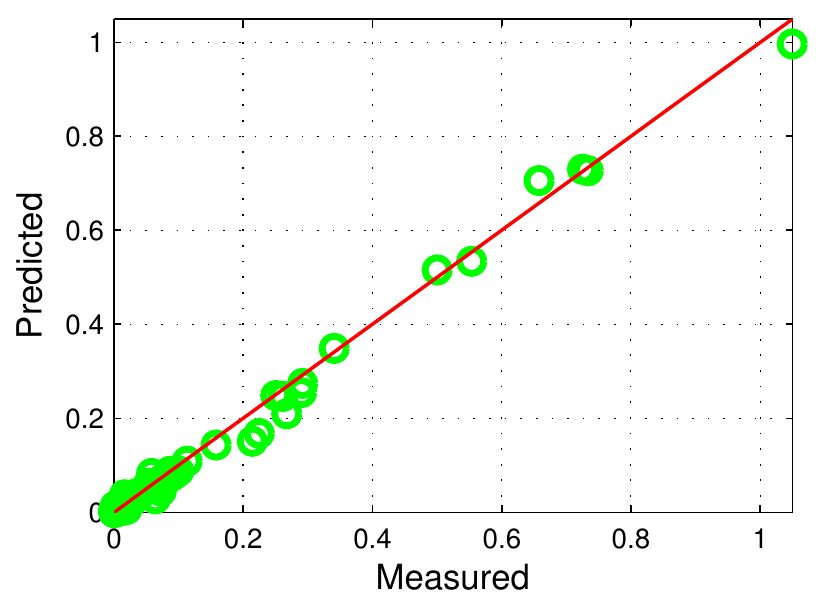}
\end{subfigure}%
\begin{subfigure}{0.180000\textwidth}
\centering
\includegraphics[width=\textwidth]{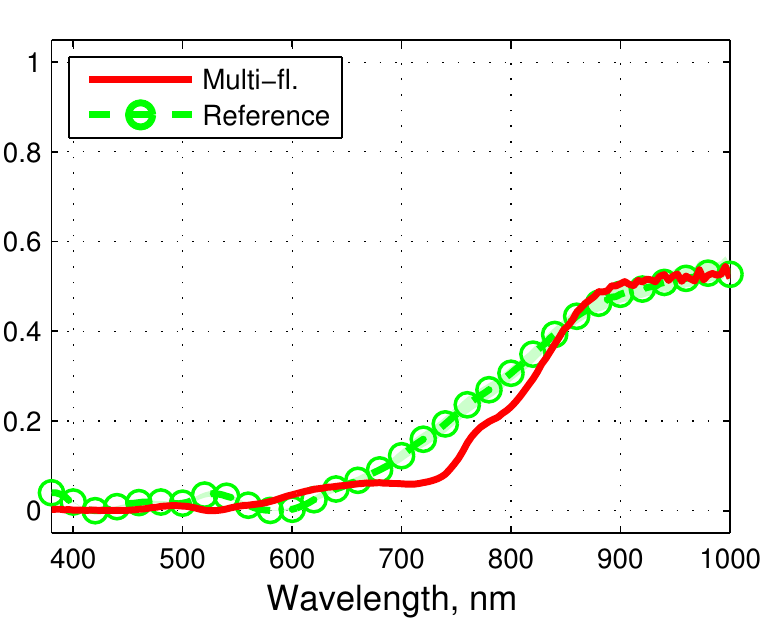}
\end{subfigure}%
\begin{subfigure}{0.180000\textwidth}
\centering
\includegraphics[width=\textwidth]{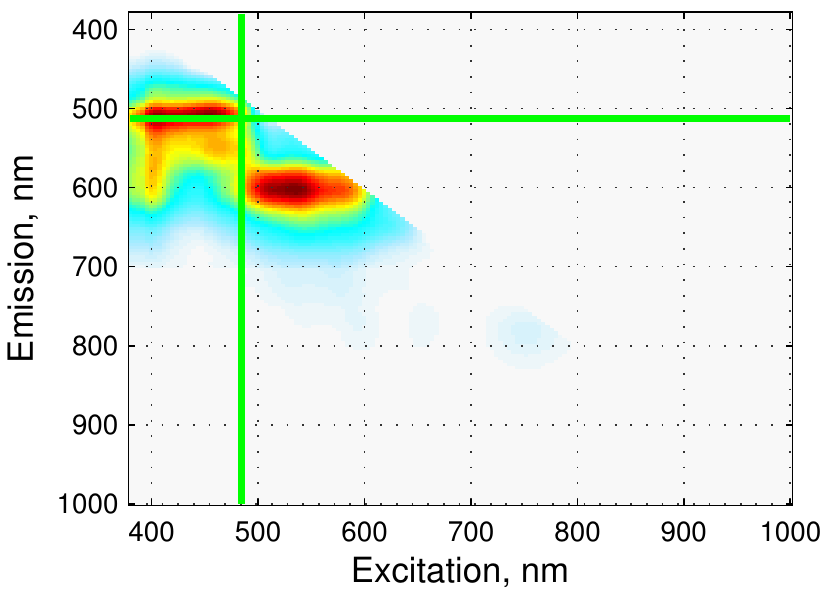}
\end{subfigure}%
\begin{subfigure}{0.180000\textwidth}
\centering
\includegraphics[width=\textwidth]{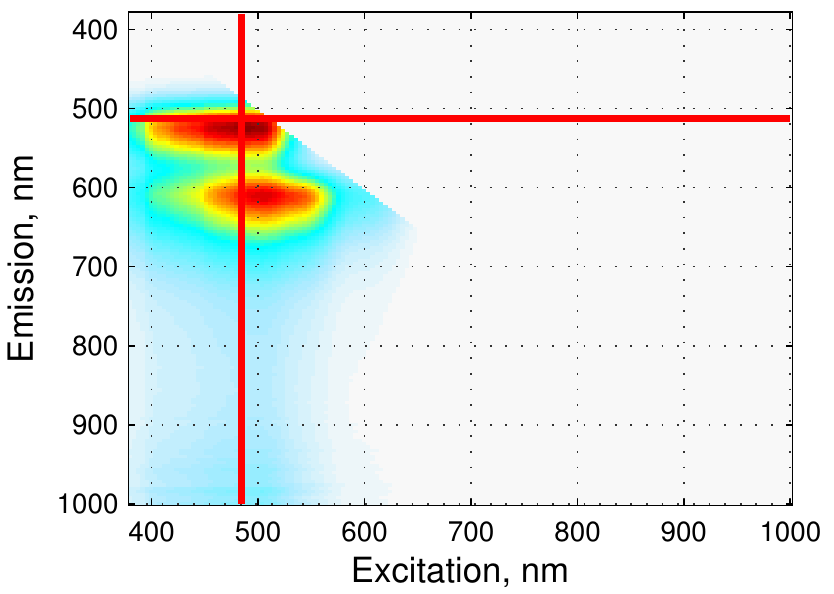}
\end{subfigure}%
\begin{subfigure}{0.050000\textwidth}
\centering
\raisebox{0.2in}{\includegraphics[width=\textwidth]{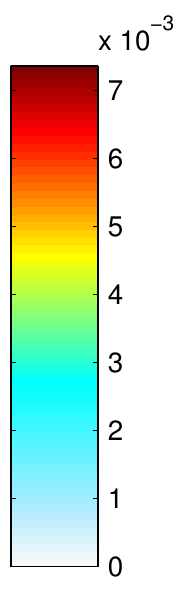}}
\end{subfigure}%
\begin{subfigure}{0.180000\textwidth}
\centering
\includegraphics[width=\textwidth]{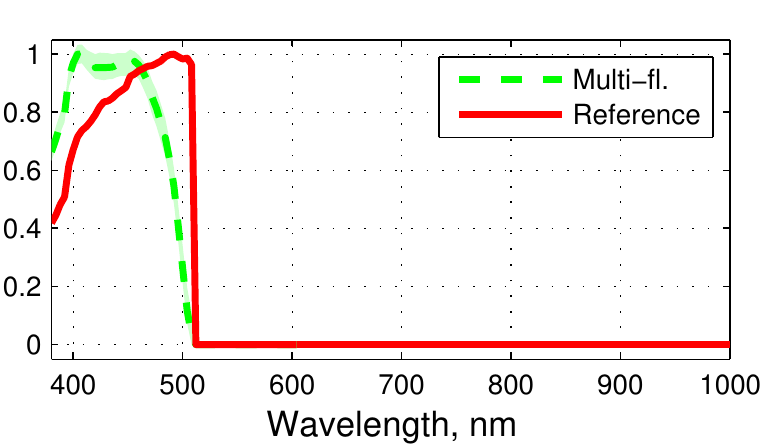}
\includegraphics[width=\textwidth]{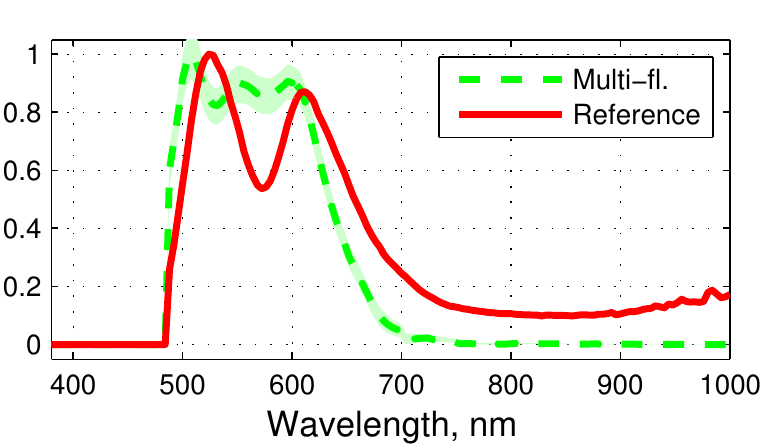}
\end{subfigure}\\
\rule{\textwidth}{1pt}
\begin{subfigure}{0.030000\textwidth}
\centering
\rotcaption*{Patch C}
\end{subfigure}%
\begin{subfigure}{0.180000\textwidth}
\includegraphics[width = \textwidth]{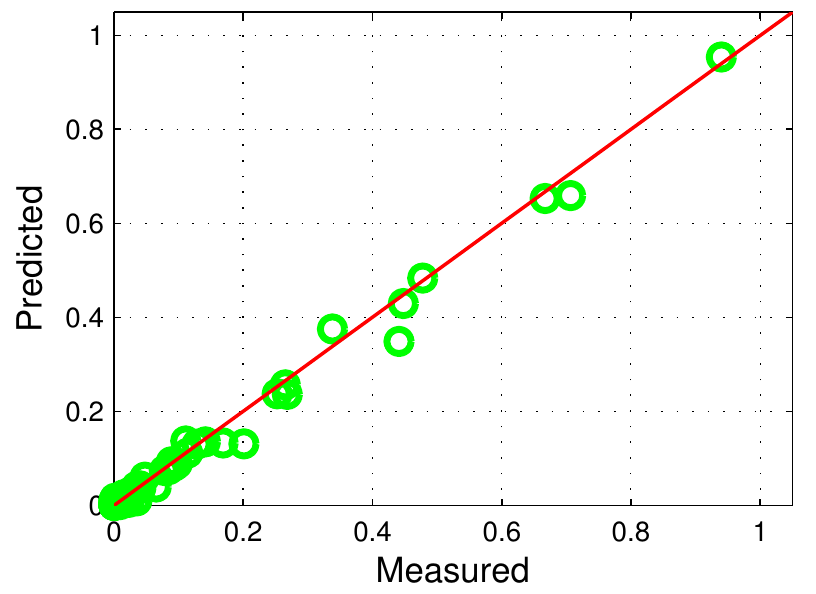}
\vfill
\end{subfigure}%
\begin{subfigure}{0.180000\textwidth}
\centering
\includegraphics[width=\textwidth]{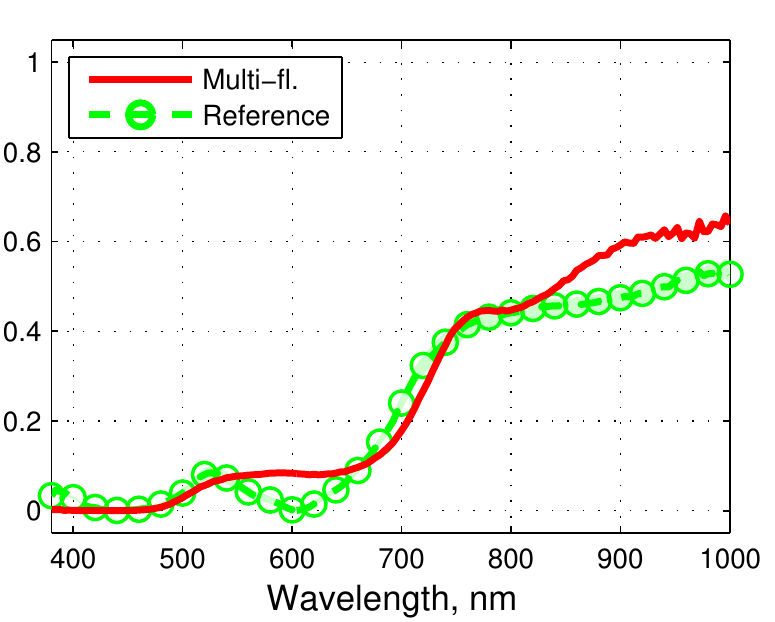}
\vfill
\end{subfigure}%
\begin{subfigure}{0.180000\textwidth}
\centering
\includegraphics[width=\textwidth]{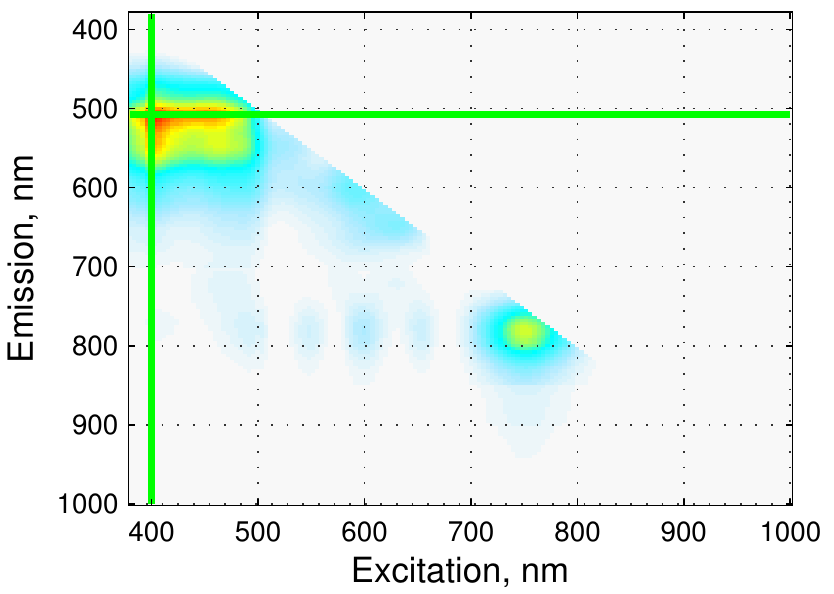}
\vfill
\end{subfigure}%
\begin{subfigure}{0.180000\textwidth}
\centering
\includegraphics[width=\textwidth]{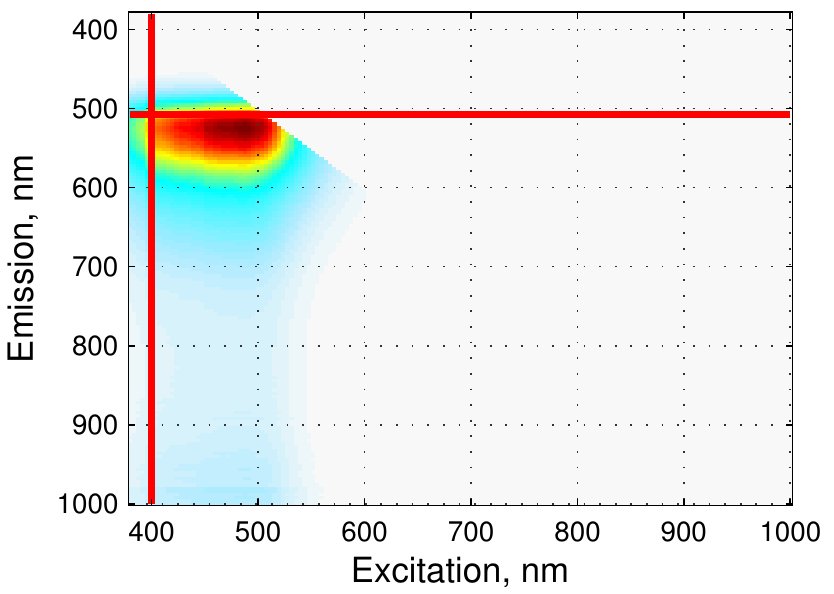}
\vfill
\end{subfigure}%
\begin{subfigure}{0.050000\textwidth}
\centering
\raisebox{0.2in}{\includegraphics[width=\textwidth]{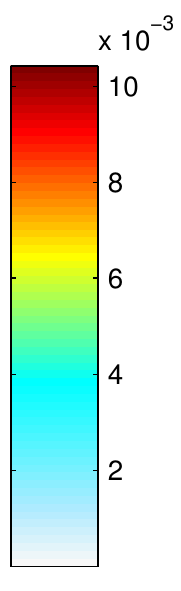}}
\end{subfigure}%
\begin{subfigure}{0.180000\textwidth}
\centering
\includegraphics[width=\textwidth]{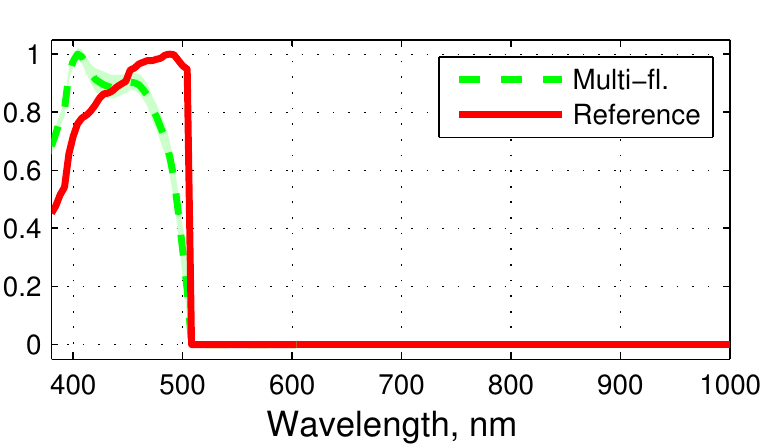}
\includegraphics[width=\textwidth]{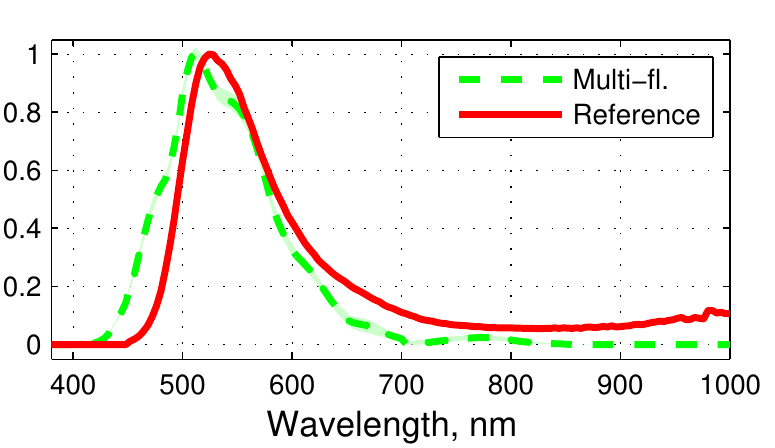}
\vfill
\end{subfigure}\\
\begin{subfigure}{0.030000\textwidth}
\centering
\caption*{}
\end{subfigure}%
\begin{subfigure}{0.180000\textwidth}
\caption{Pixels}
\end{subfigure}%
\begin{subfigure}{0.180000\textwidth}
\caption{Reflectance}
\end{subfigure}%
\begin{subfigure}{0.180000\textwidth}
\captionsetup{justification=centering}
\caption{Donaldson estimate (absolute)}
\end{subfigure}%
\begin{subfigure}{0.180000\textwidth}
\captionsetup{justification=centering}
\caption{Donaldson reference (absolute)}
\end{subfigure}%
\begin{subfigure}{0.050000\textwidth}
\caption*{}
\end{subfigure}%
\begin{subfigure}{0.180000\textwidth}
\captionsetup{justification=centering}
\caption{Spectra (normalized)}
\end{subfigure}
\caption{Multi-fluorophore estimation results for patches A, B and C outlined in Fig.~\ref{fig:sampleTargetRGB}a. Columns show the estimates and ground truth reference for pixel values (a) reflectance (b) the Donaldson matrix (c,d) and  (e) the Donaldson matrix cross-sections along rows and columns (c, green lines; d, red lines). Shaded areas denote the $95\%$ confidence intervals. Patch A contains one, while patches B and C two fluorophores. The separation between orange and green fluorophores in patch B is clearly visible, as the Donaldson matrix estimate is bi-modal. The green and amber fluorophores in patch C have similar emission spectra, and thus disguise as a single fluorescent compound.}
\label{fig:multiExExp}
\end{figure*}

%% file: multiFlAccuracy.tex
\begin{table*}
\renewcommand{\arraystretch}{1.3}
\centering
\caption{Comparison between multi-fluorophore algorithms. The RMSE $\pm$1 sd are shown. Suo et al. \cite{Suo:14} represent reflectance as a square matrix.  Although their estimate sometimes includes non-zero off-diagonal terms, we estimate the RMSE reflectance error using only the diagonal terms.}
\label{tab:multiAccExp}
\begin{tabular}{| l | c | c | c | c |}
\hline
\multirow{2}{*}{\diagbox{Algorithm}{Quantity}} & \multirow{2}{*}{Pixel values} & \multirow{2}{*}{Reflectance} & \multicolumn{2}{ c |}{Donaldson matrix} \\
\cline{4-5}
& &  & {Absolute $\times 10^{-2}$} & {Normalized} \\
\hline
\hline
Ours -- Multi-fluorophore & $\mathbf{0.02 \pm 0.00}$ & $\mathbf{0.07 \pm 0.03}$ & $\mathbf{0.08 \pm 0.01}$ & $\mathbf{0.09 \pm 0.02}$ \\
Suo et al. \cite{Suo:14} & $\mathbf{0.02 \pm 0.00}$ & $ 0.40 \pm 0.18$ & $0.13 \pm 0.01$ & $ 0.21\pm 0.04$ \\
\hline
\end{tabular}
\end{table*}

%% file: RelightImagesV2.tex
\begin{figure*}
\centering
\setlength\tabcolsep{1pt}
\begin{tabular}{cc cc cc c}
\begin{subfigure}{0.020000\textwidth}
\centering
\rotcaption*{$412$\si{\nm}}
\end{subfigure}&
\begin{subfigure}{0.180000\textwidth}
\centering
\includegraphics[width=\textwidth]{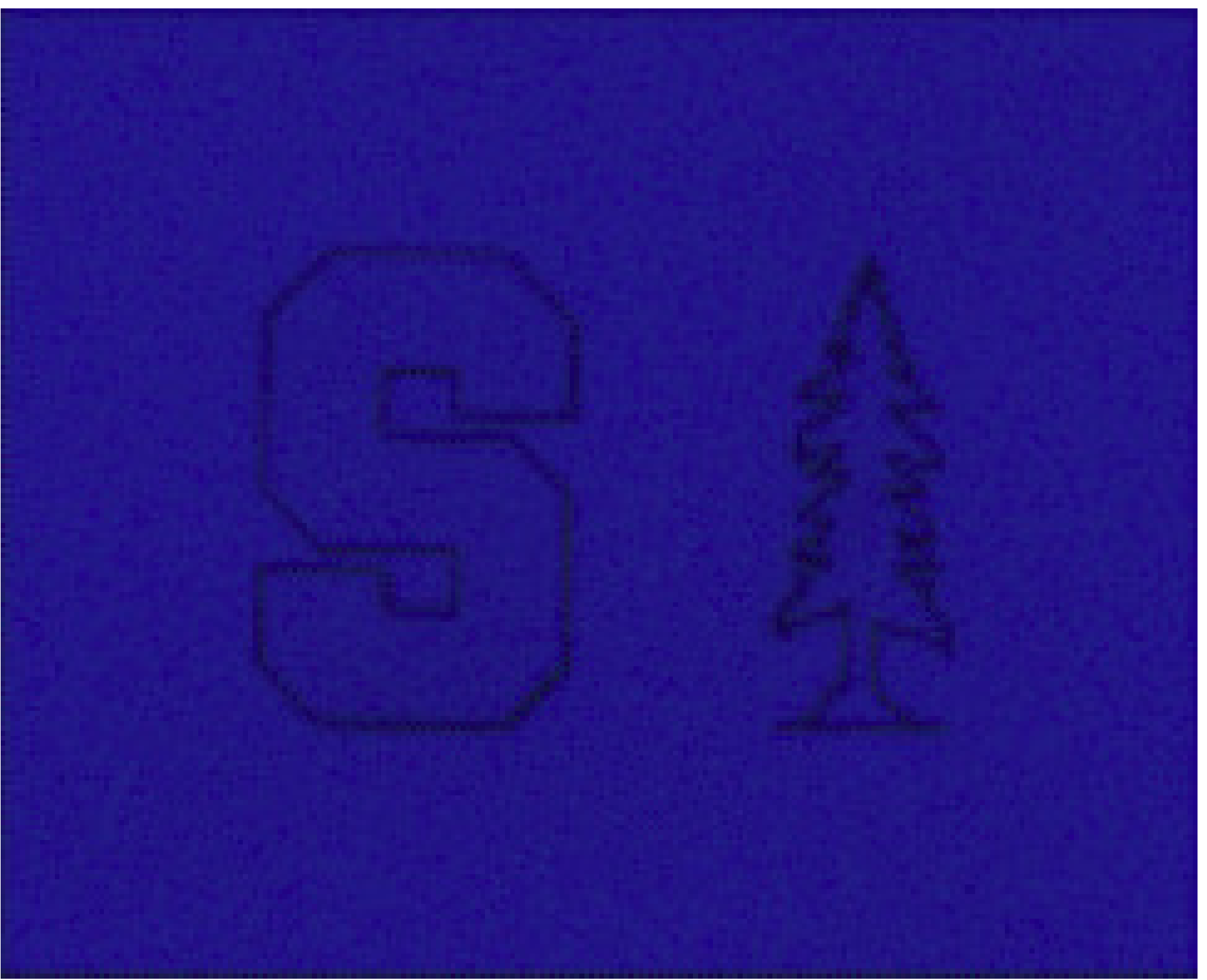}
\end{subfigure}&
\begin{subfigure}{0.180000\textwidth}
\centering
\includegraphics[width=\textwidth]{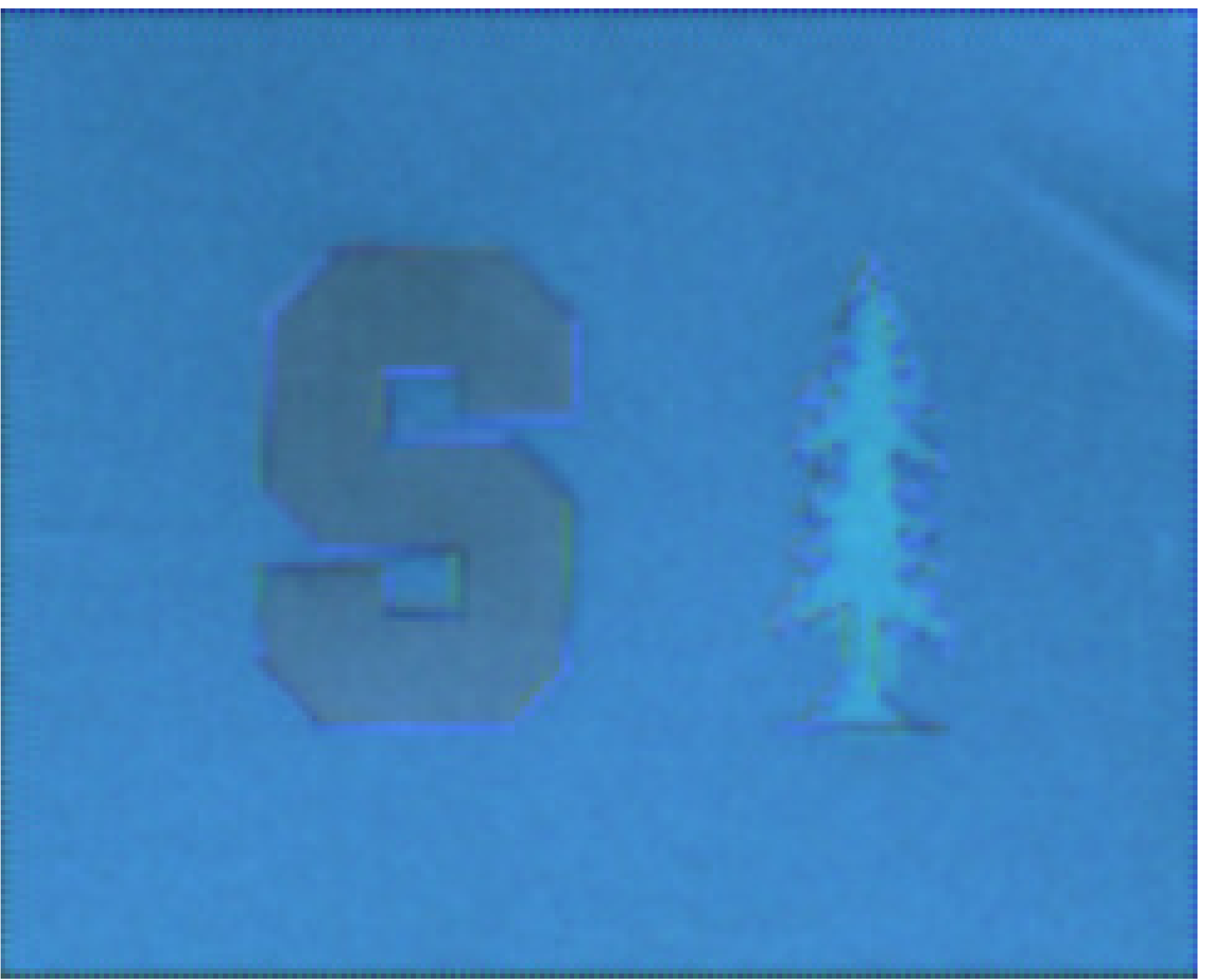}
\end{subfigure}&
\begin{subfigure}{0.180000\textwidth}
\centering
\includegraphics[width=\textwidth]{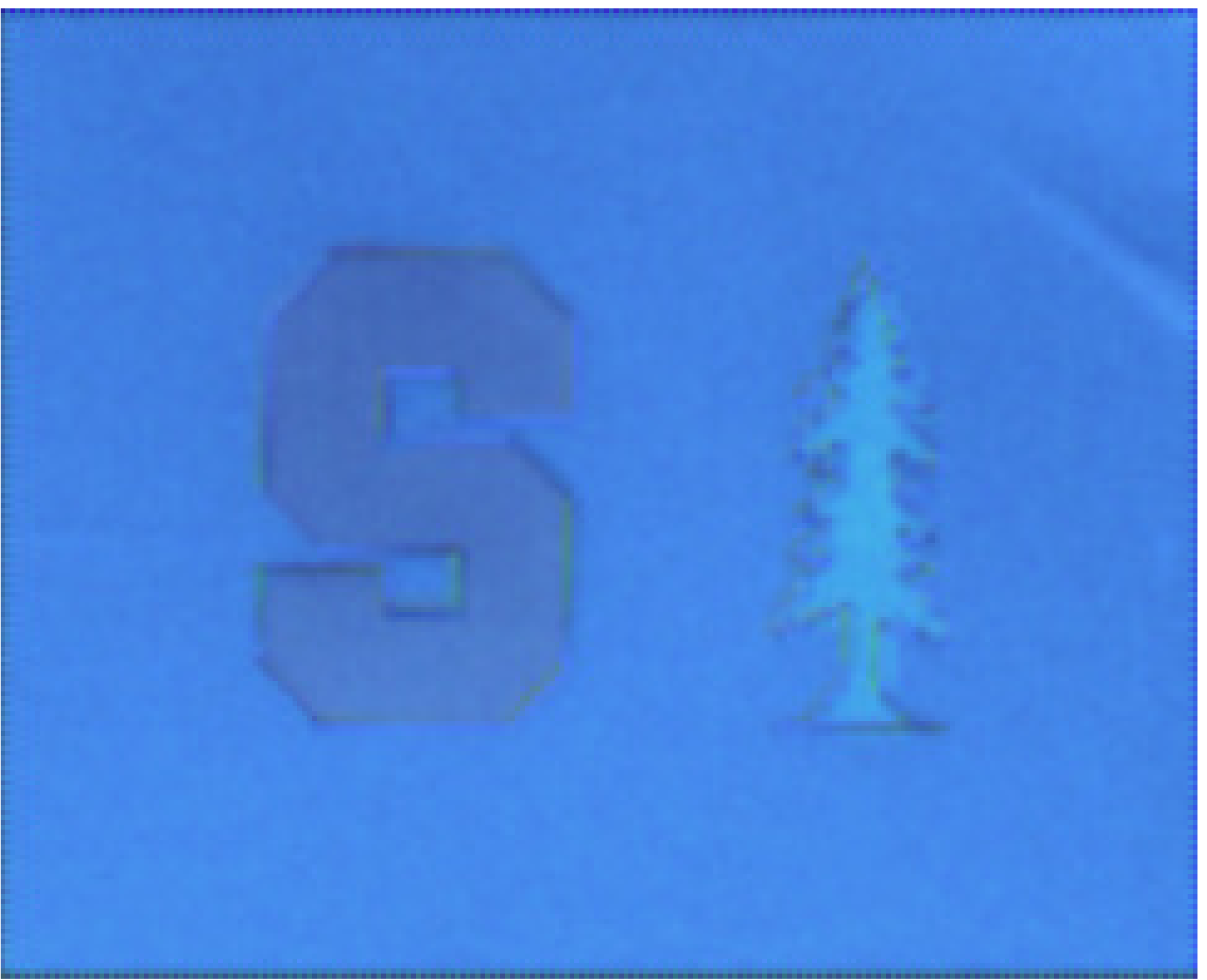}
\end{subfigure}&
\begin{subfigure}{0.180000\textwidth}
\centering
\includegraphics[width=\textwidth]{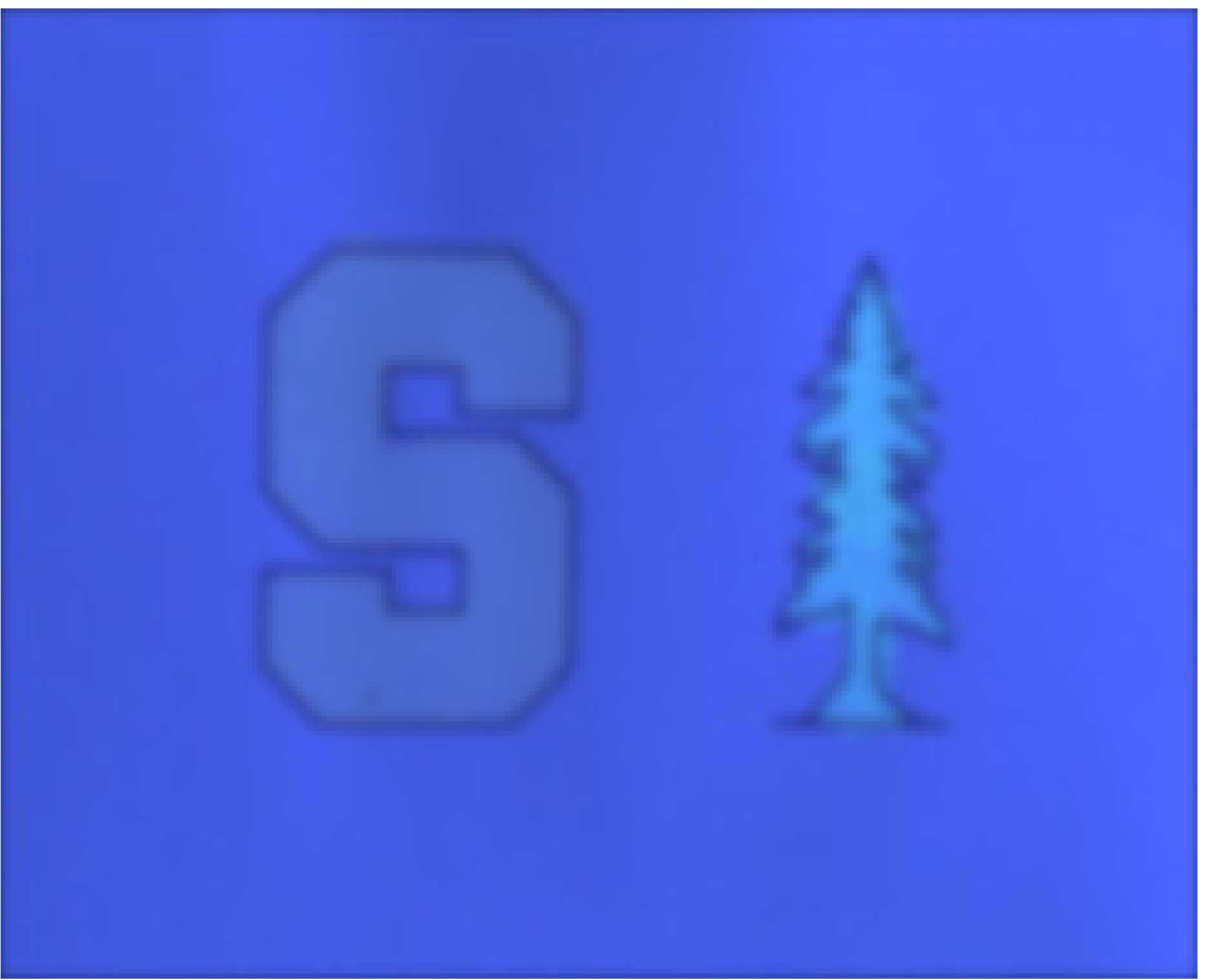}
\end{subfigure}&
\begin{subfigure}{0.180000\textwidth}
\centering
\includegraphics[width=\textwidth]{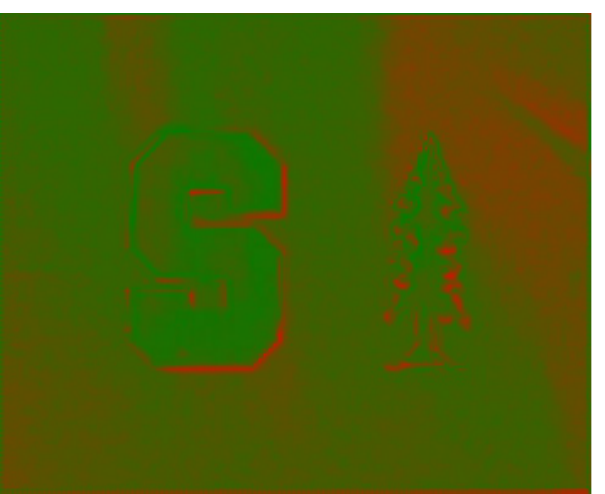}
\end{subfigure}&
\begin{subfigure}{0.030000\textwidth}
\centering
\raisebox{-0.5in}{\includegraphics[width=2.15\textwidth]{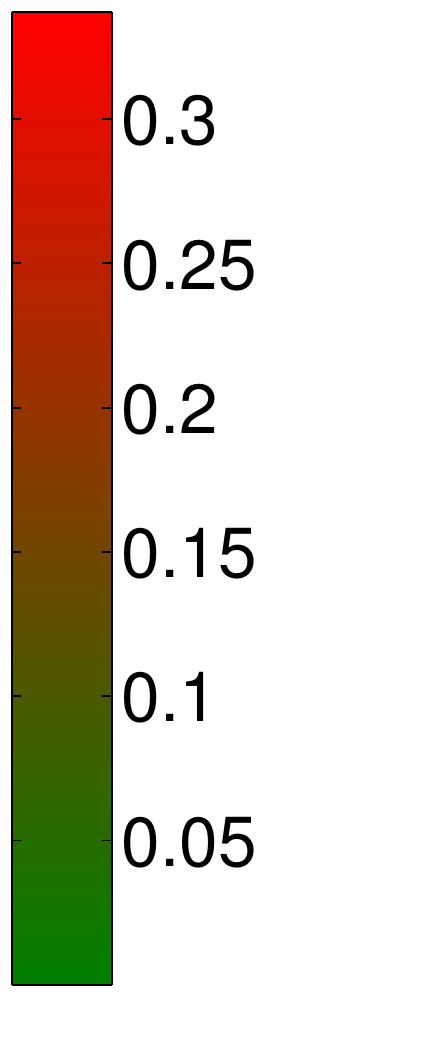}}
\end{subfigure}\\
\begin{subfigure}{0.020000\textwidth}
\centering
\rotcaption*{\num[group-separator={,}]{9800}\si{\kelvin}}
\end{subfigure} &
\begin{subfigure}{0.180000\textwidth}
\centering
\includegraphics[width=\textwidth]{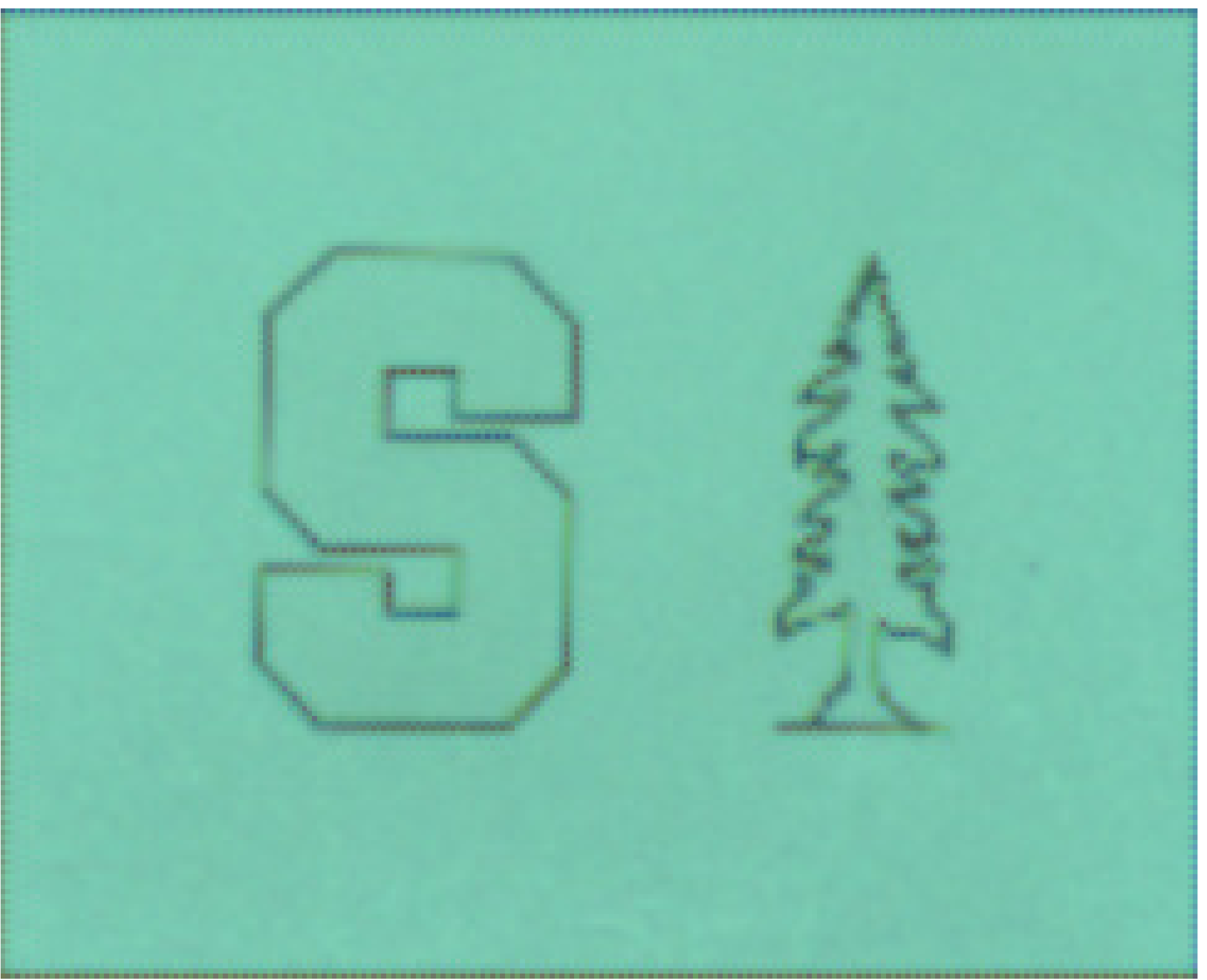}
\end{subfigure} &
\begin{subfigure}{0.180000\textwidth}
\centering
\includegraphics[width=\textwidth]{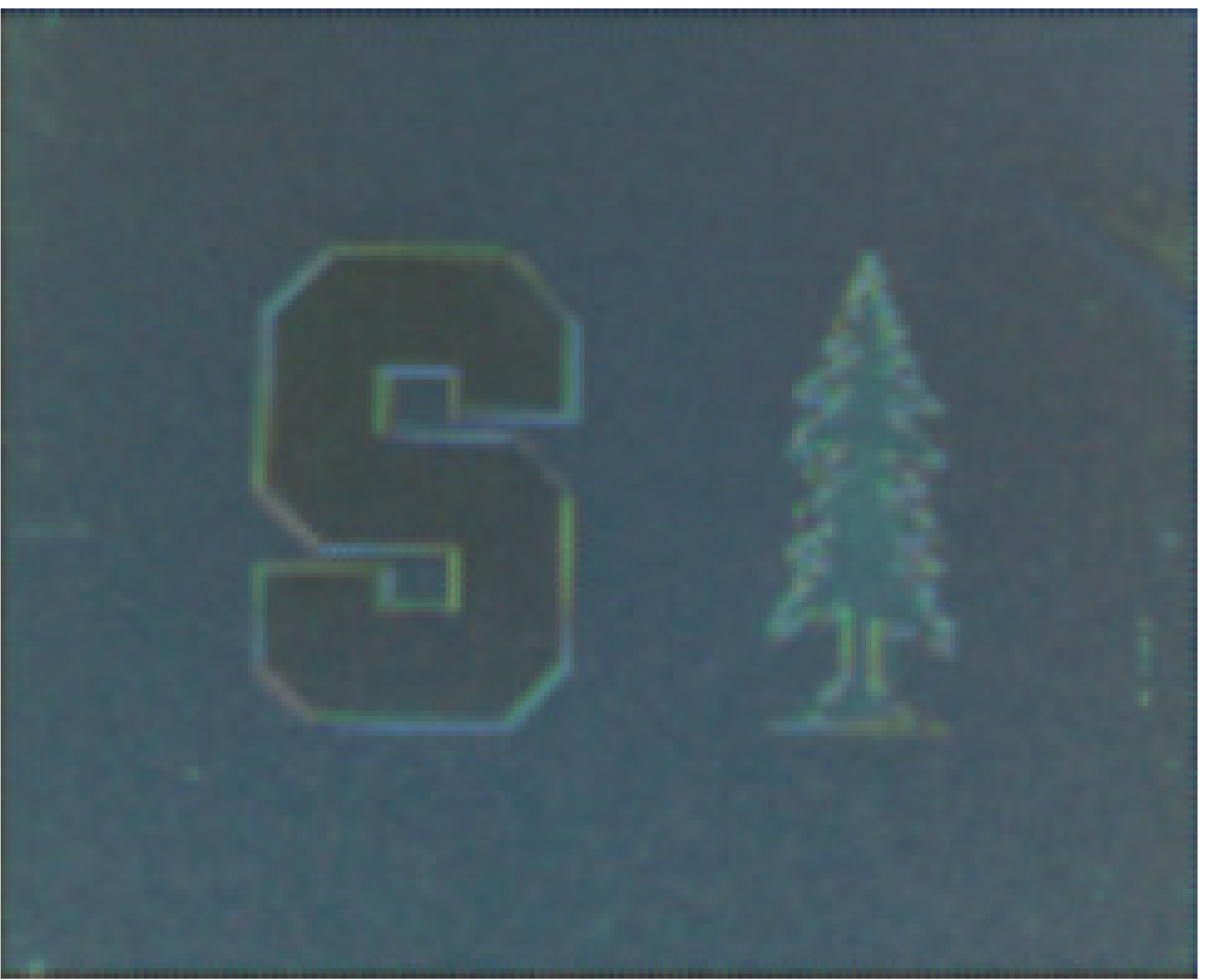}
\end{subfigure} &
\begin{subfigure}{0.180000\textwidth}
\centering
\includegraphics[width=\textwidth]{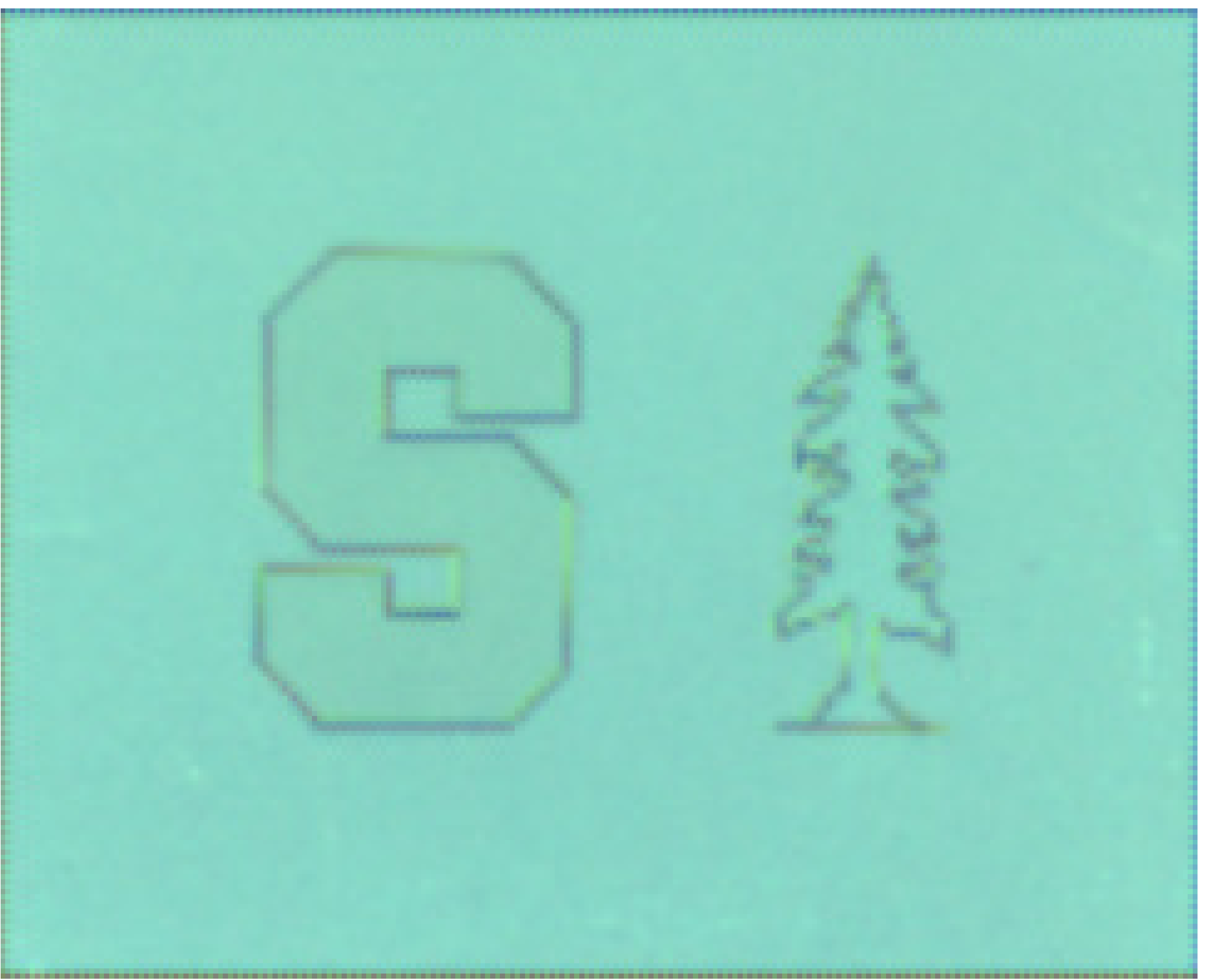}
\end{subfigure} &
\begin{subfigure}{0.180000\textwidth}
\centering
\includegraphics[width=\textwidth]{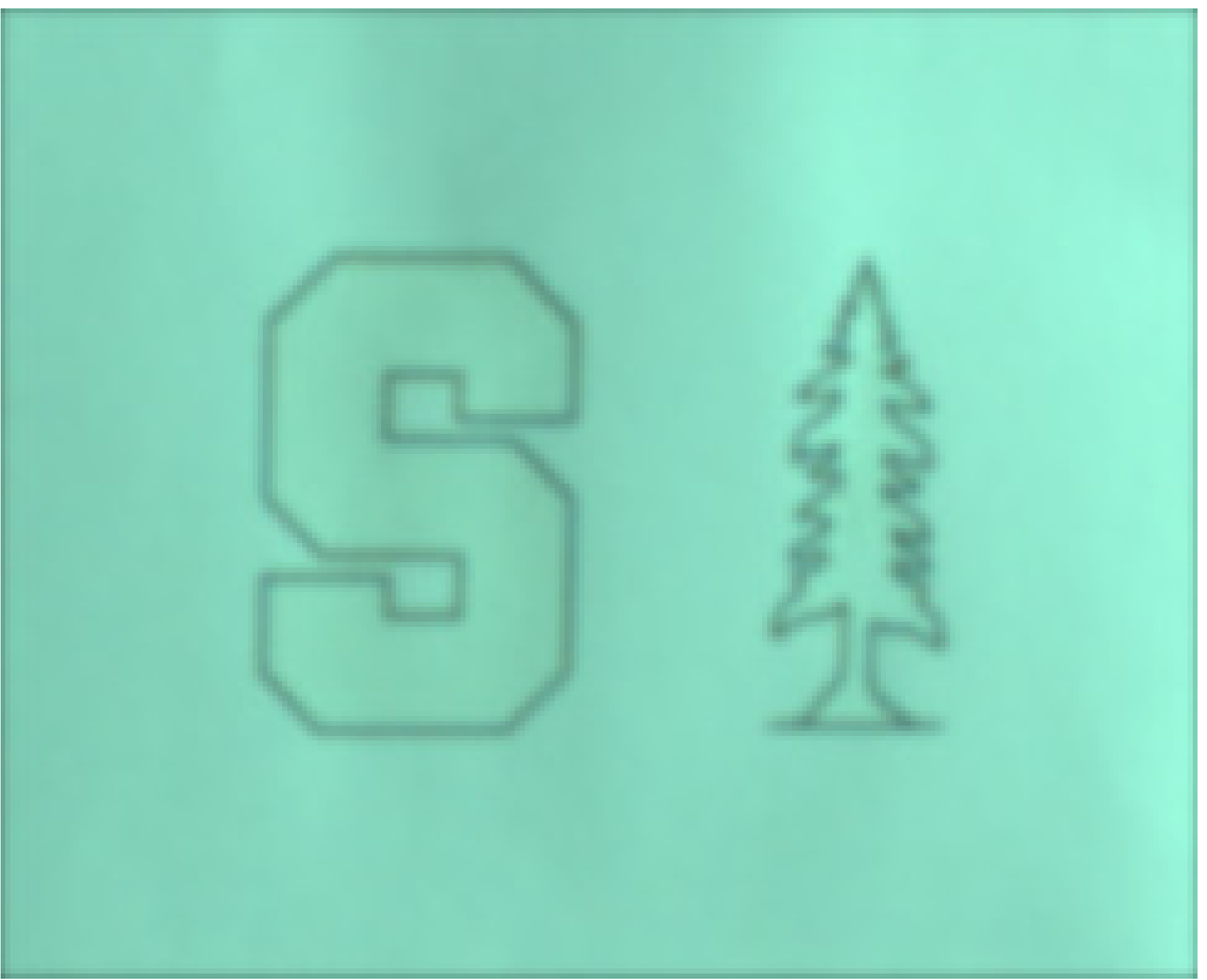}
\end{subfigure}& 
\begin{subfigure}{0.180000\textwidth}
\centering
\includegraphics[width=\textwidth]{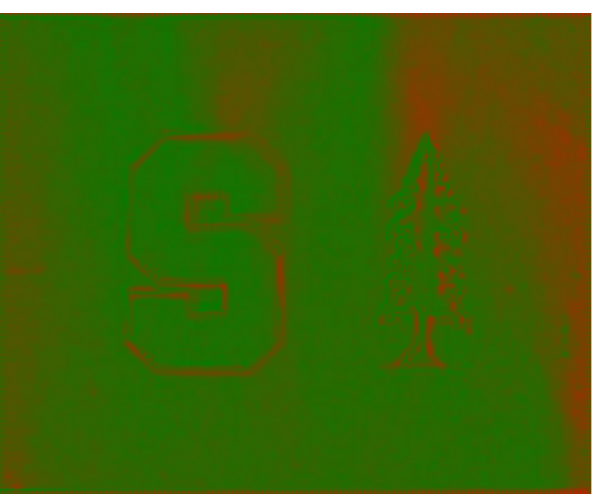}
\end{subfigure}&
\begin{subfigure}{0.030000\textwidth}
\centering
\raisebox{-0.5in}{\includegraphics[width=2.15\textwidth]{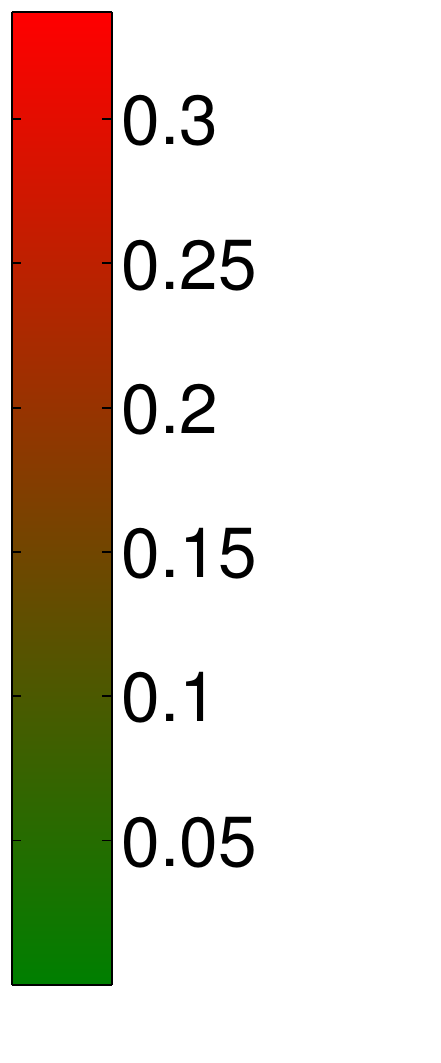}}
\end{subfigure}\\
\begin{subfigure}{0.020000\textwidth}
\centering
\rotcaption*{\num[group-separator={,}]{6500}\si{\kelvin}}
\end{subfigure}&
\begin{subfigure}{0.180000\textwidth}
\centering
\includegraphics[width=\textwidth]{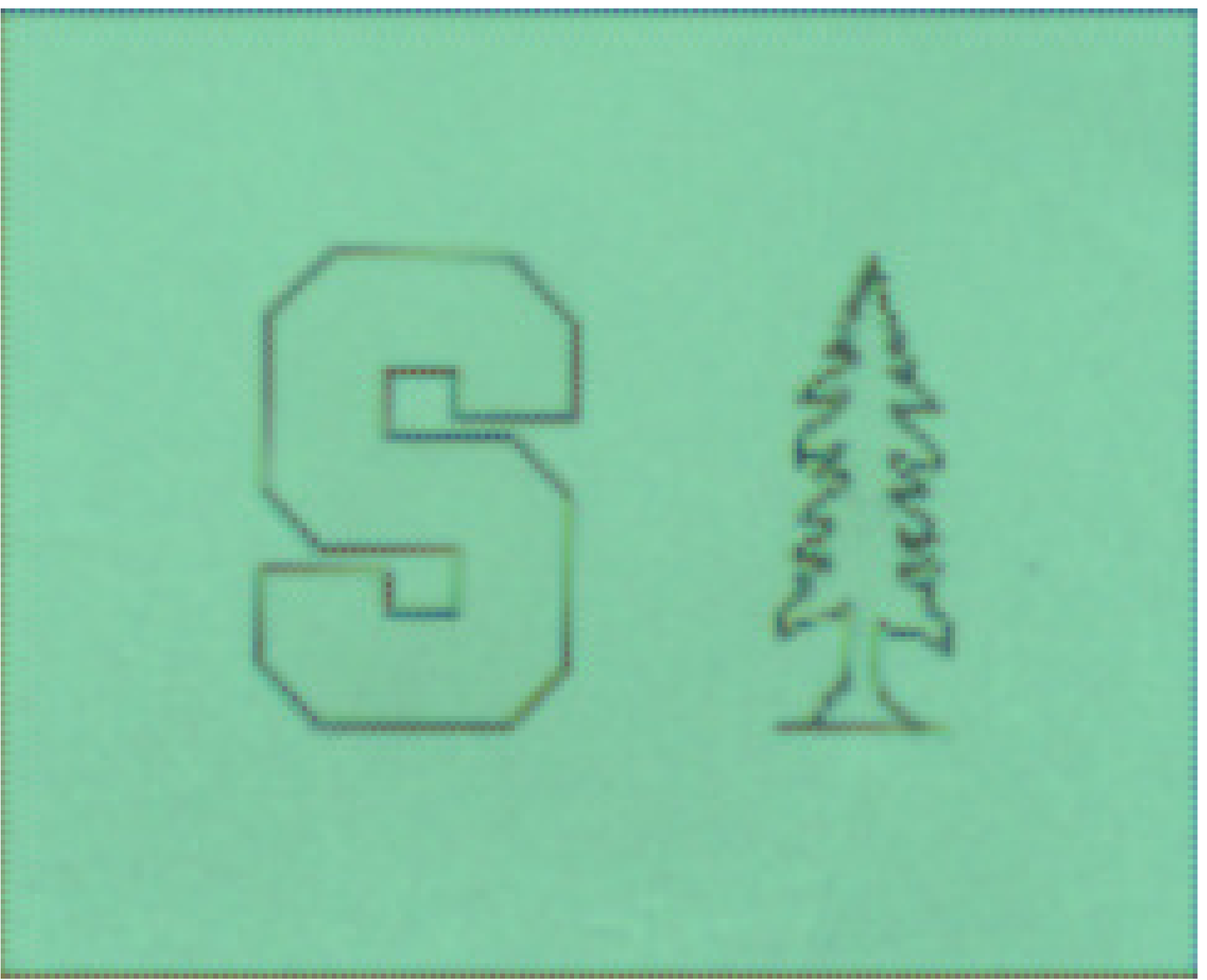}
\end{subfigure}&
\begin{subfigure}{0.180000\textwidth}
\centering
\includegraphics[width=\textwidth]{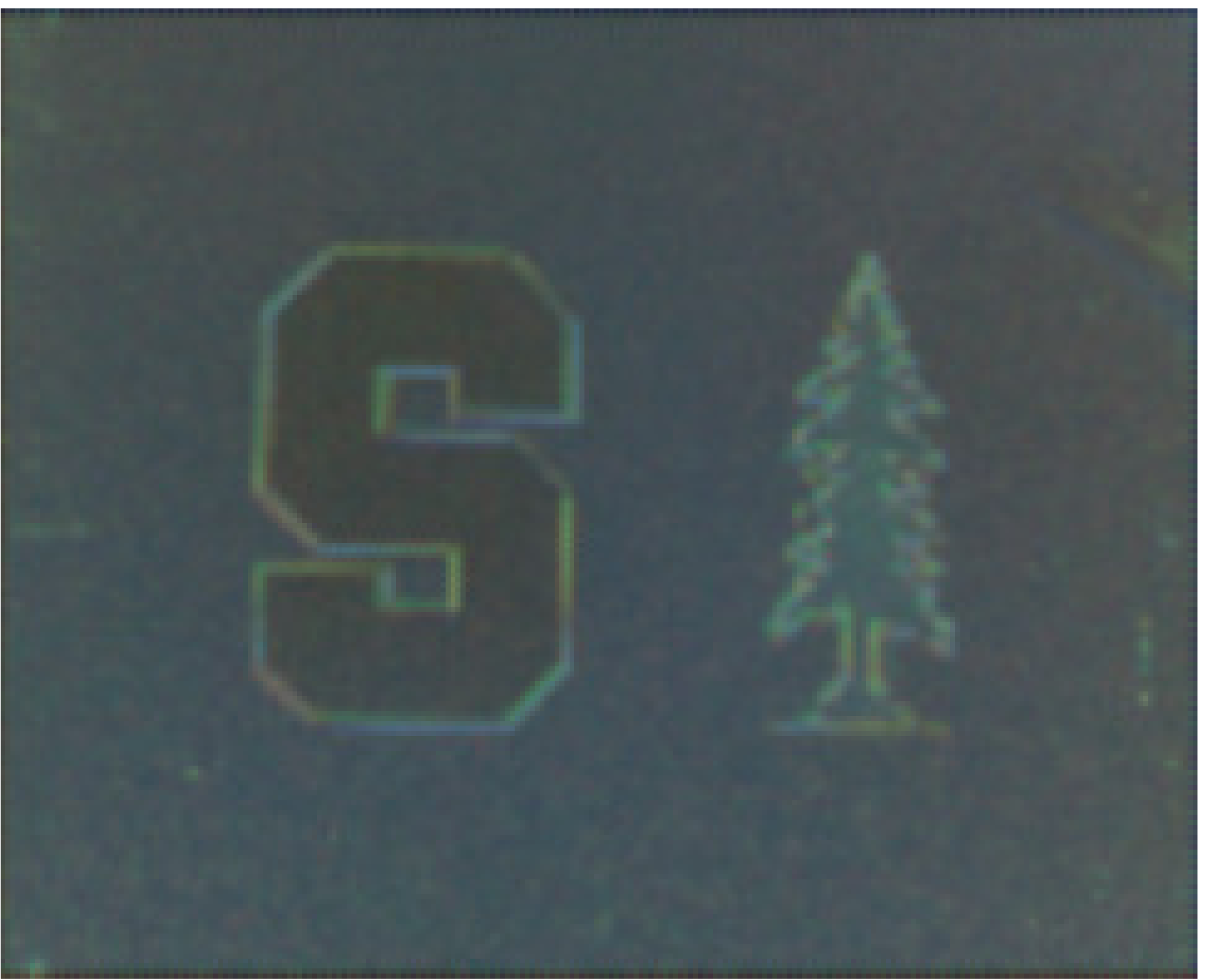}
\end{subfigure}&
\begin{subfigure}{0.180000\textwidth}
\centering
\includegraphics[width=\textwidth]{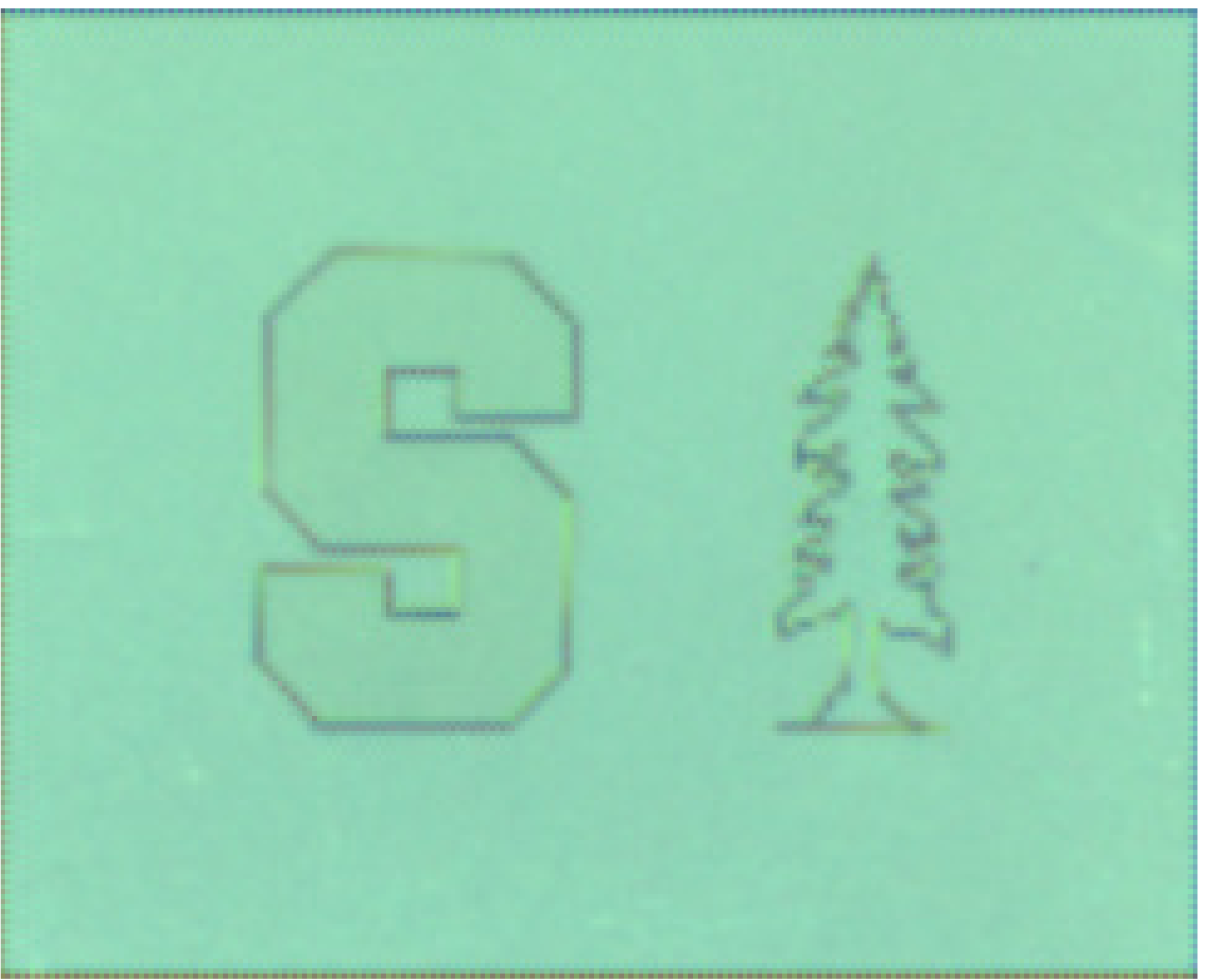}
\end{subfigure}&
\begin{subfigure}{0.180000\textwidth}
\centering
\includegraphics[width=\textwidth]{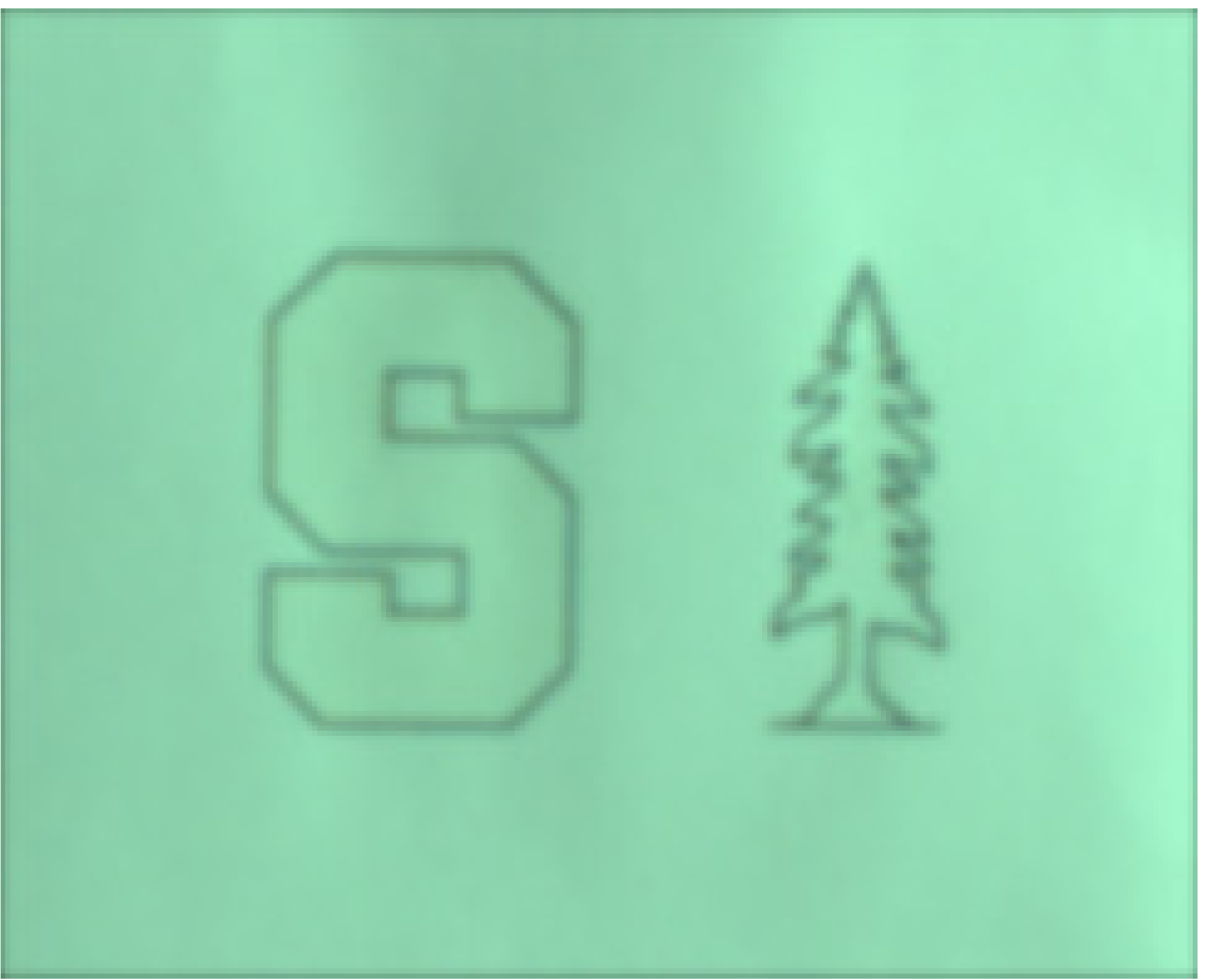}
\end{subfigure}&
\begin{subfigure}{0.180000\textwidth}
\centering
\includegraphics[width=\textwidth]{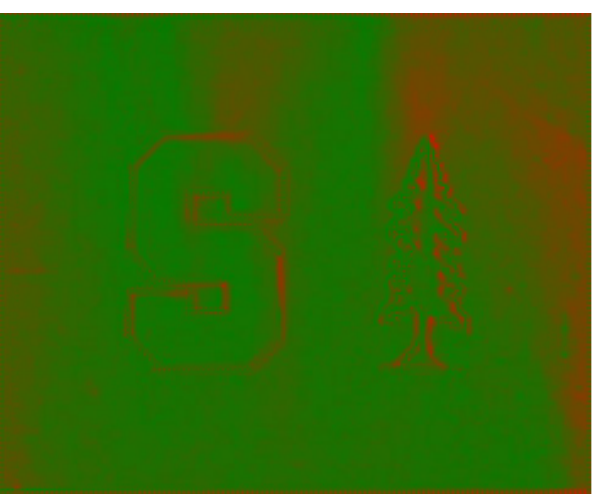}
\end{subfigure}&
\begin{subfigure}{0.030000\textwidth}
\centering
\raisebox{-0.5in}{\includegraphics[width=2.15\textwidth]{Errorbar}}
\end{subfigure}\\
\begin{subfigure}{0.020000\textwidth}
\centering
\rotcaption*{\num[group-separator={,}]{2000}\si{\kelvin}}
\end{subfigure}&
\begin{subfigure}{0.180000\textwidth}
\centering
\includegraphics[width=\textwidth]{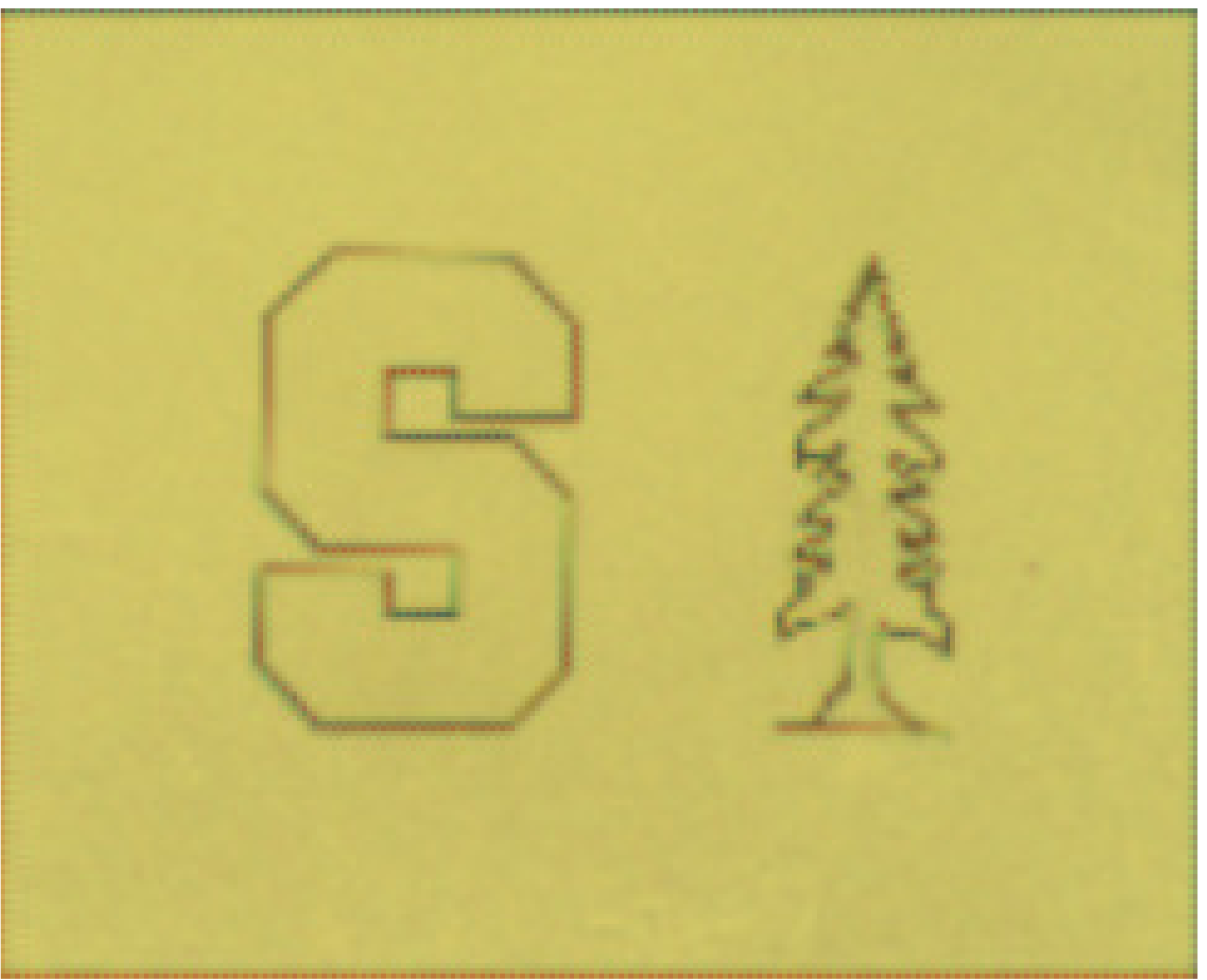}
\end{subfigure}&
\begin{subfigure}{0.180000\textwidth}
\centering
\includegraphics[width=\textwidth]{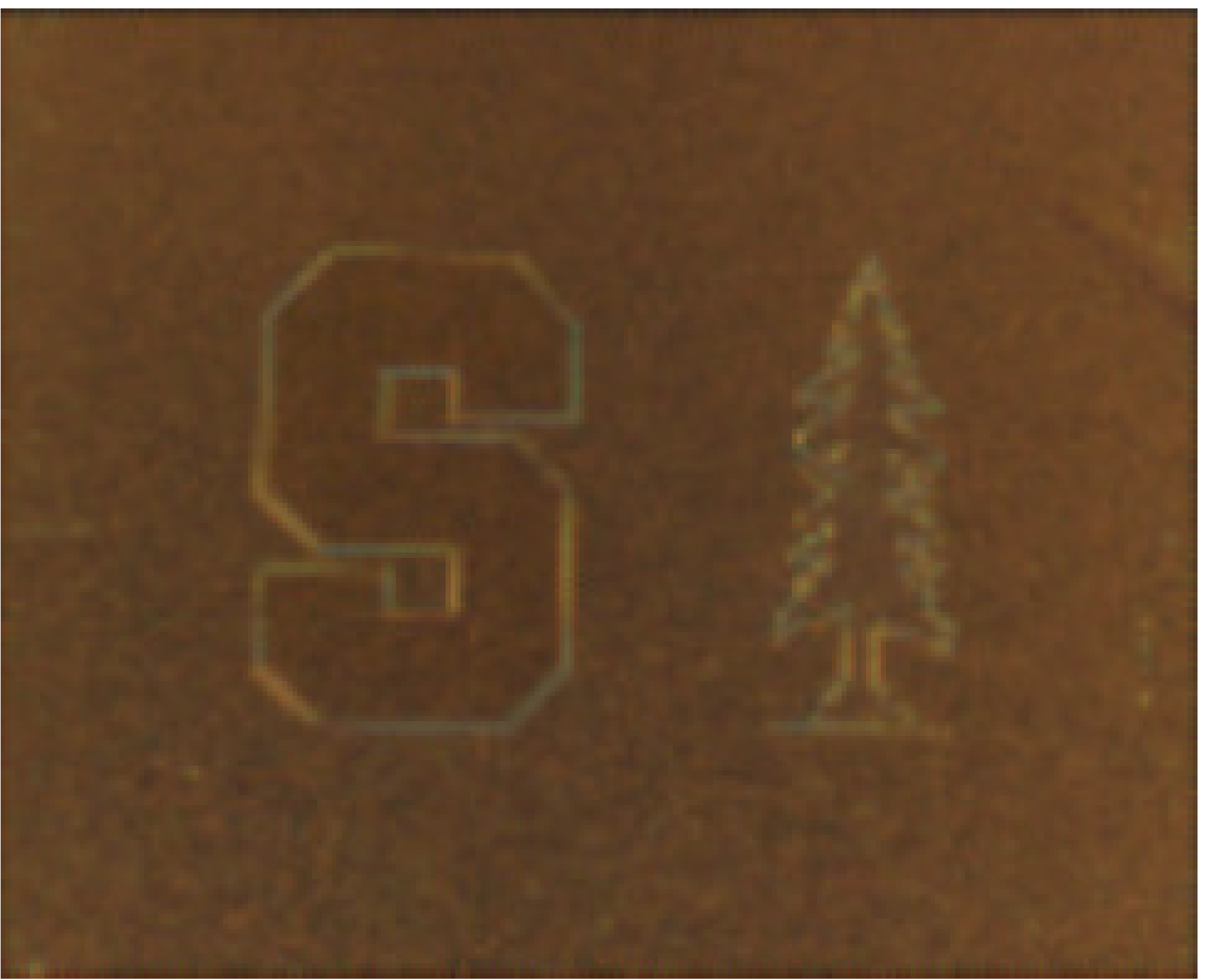}
\end{subfigure}&
\begin{subfigure}{0.180000\textwidth}
\centering
\includegraphics[width=\textwidth]{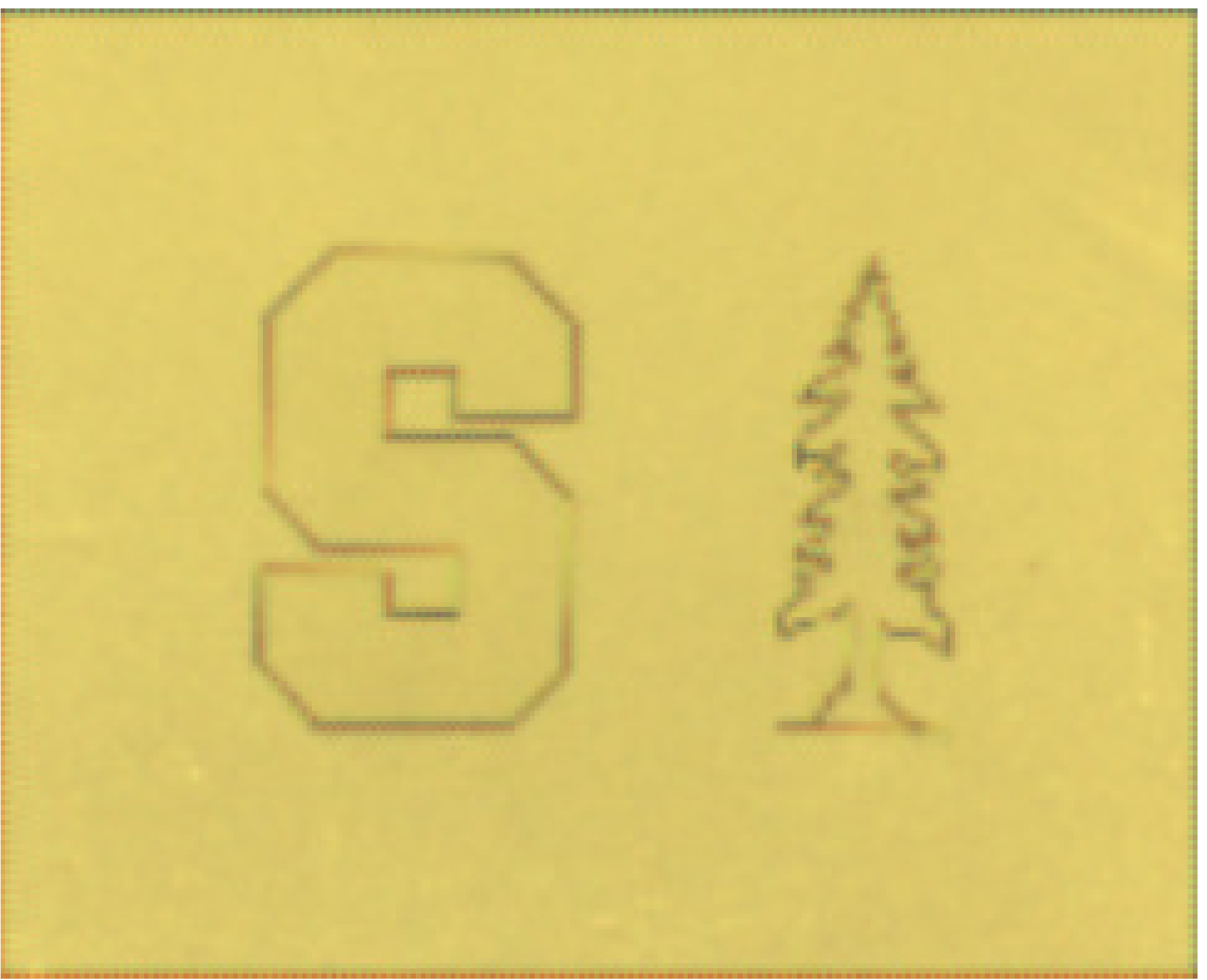}
\end{subfigure}&
\begin{subfigure}{0.180000\textwidth}
\centering
\includegraphics[width=\textwidth]{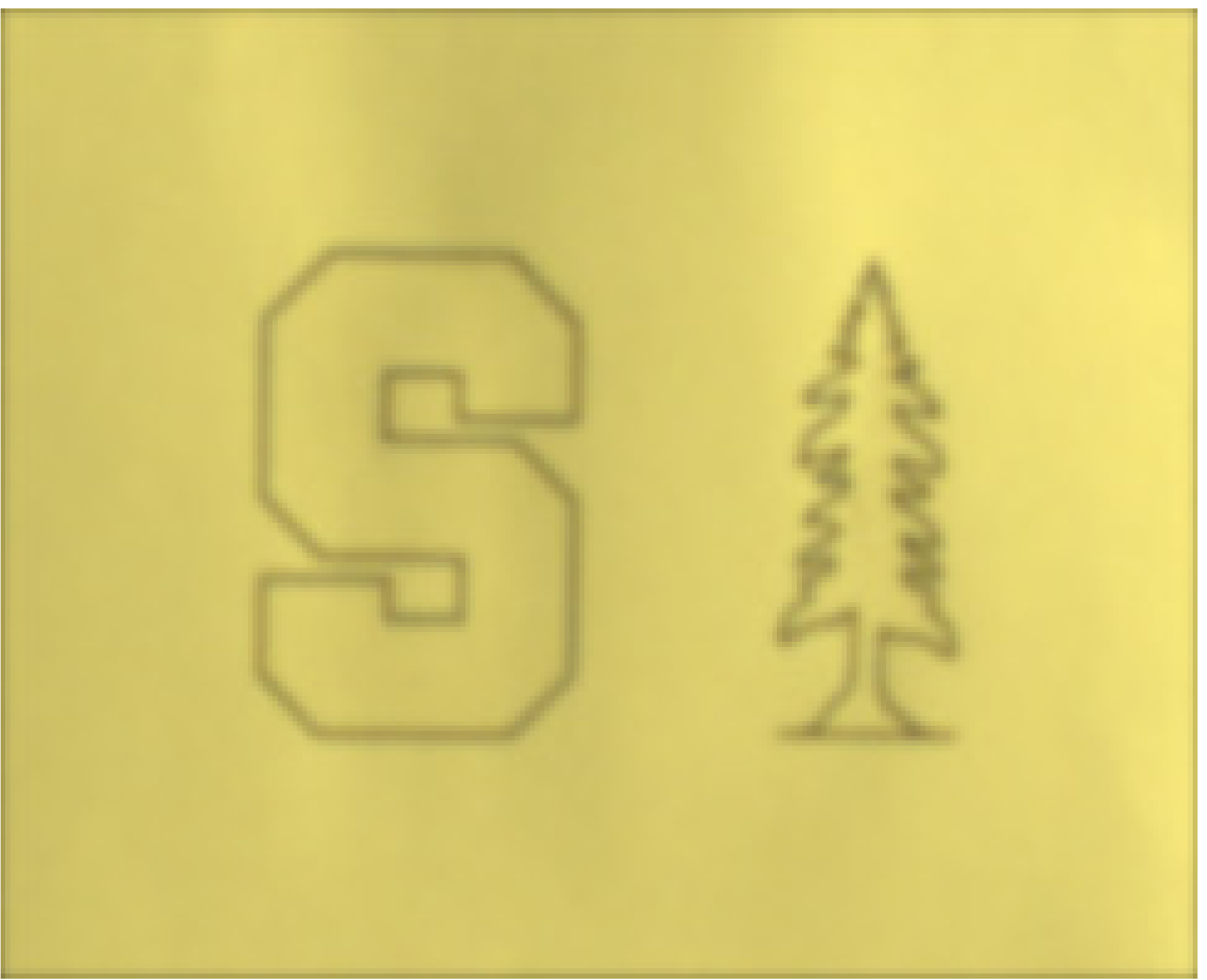}
\end{subfigure}&
\begin{subfigure}{0.180000\textwidth}
\centering
\includegraphics[width=\textwidth]{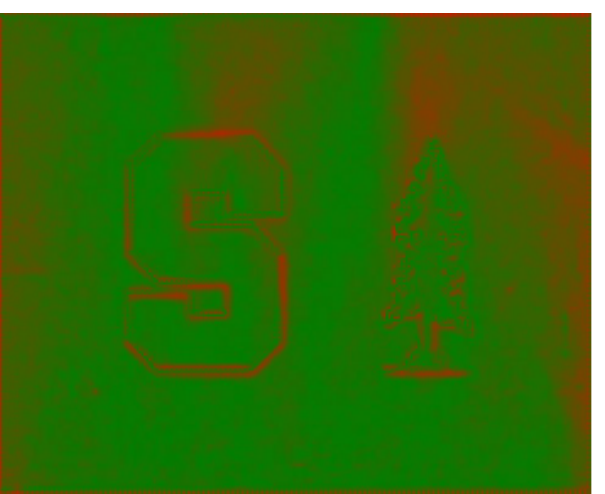}
\end{subfigure}&
\begin{subfigure}{0.030000\textwidth}
\centering
\raisebox{-0.5in}{\includegraphics[width=2.15\textwidth]{Errorbar}}
\end{subfigure}\\
\begin{subfigure}{0.020000\textwidth}
\centering
\rotcaption*{}
\end{subfigure}&
\begin{subfigure}{0.180000\textwidth}
\centering
\caption{Reflected}
\end{subfigure}&
\begin{subfigure}{0.180000\textwidth}
\centering
\caption{Fluoresced}
\end{subfigure}&
\begin{subfigure}{0.180000\textwidth}
\centering
\caption{Total}
\end{subfigure}&
\begin{subfigure}{0.180000\textwidth}
\centering
\caption{Captured}
\end{subfigure}&
\begin{subfigure}{0.180000\textwidth}
\centering
\captionsetup{justification=centering}
\caption{Camera RGB RMSE }
\end{subfigure}&
\begin{subfigure}{0.030000\textwidth}
\centering
\end{subfigure}\\
\end{tabular}
\caption{Fluoresced and reflected radiance separation and relighting. The images are synthesized from reflectance and fluorescence estimated with the multi-fluorophore algorithm. A realistic, multi-fluorophore target is composed of an 'S' and a 'tree' shapes printed on a sheet of office paper and colored with fluorescent dyes. The rows represent analyses under different narrowband and broadband illuminants at different color temperatures.  The first three columns represent the simulated reflected (a), fluoresced (b) and total (c) components rendered using a model of a consumer camera (Canon G7 X). We compare these simulated renderings visually (d), and qualitatively (e) to actual images captured with the same  camera.}
\label{fig:relighting}
\end{figure*}

%% file: singleFlAccuracy.tex
\begin{table*}
\renewcommand{\arraystretch}{1.3}
\centering
\caption{Comparison of single fluorophore estimation algorithms. The RMSE $\pm$ 1 sd are shown.There is a scaling ambiguity between the excitation and emission spectra; without loss of generality we assume that the peak emission spectrum is scaled to a value of 1 and that the fluorescence intensity is contained entirely within the scale of the excitation spectrum. Hence, we include the RMSE of the absolute and normalized excitation estimate, but there is no need to do so for the emission spectrum.}
\label{tab:singleAccExp}
\begin{tabular}{| l | c | c | c | c | c |}
\hline
\multirow{2}{*}{\diagbox{Algorithm}{Quantity}} & \multirow{2}{*}{Pixel values} & \multirow{2}{*}{Reflectance} & \multirow{2}{*}{Emission} & \multicolumn{2}{ c |}{Excitation} \\
\cline{5-6}
& &  &  & {Absolute $\times 10^{-1}$} & {Normalized} \\
\hline
\hline
Ours -- CIM & $\mathbf{0.02 \pm 0.00}$ & $0.05 \pm 0.03$ & $\mathbf{0.09 \pm 0.02}$ & -- & -- \\
Ours -- Single fluorophore & $\mathbf{0.02 \pm 0.00}$ & $\mathbf{0.04 \pm 0.02}$ & $0.14 \pm 0.05$ & $\mathbf{0.03 \pm 0.02}$ & $\mathbf{0.15 \pm 0.04}$ \\
Fu et al. \cite{Fu:14} & $0.07 \pm 0.02$ & $ 0.26 \pm 0.17$ & $0.28 \pm 0.13$ & $0.12 \pm 0.09$ & $ 0.30\pm 0.12$ \\
\hline
\end{tabular}
\end{table*}